\documentclass[a4paper,12pt,oneside]{report}

\usepackage[left=1.5in,right=1.5in,top=1.2in,bottom=1.2in]{geometry}

\usepackage{graphicx}
\usepackage{verbatim}
\usepackage{latexsym}
\usepackage{mathchars}
\usepackage{setspace}
\usepackage[toc,page]{appendix}
\usepackage{chngcntr}           
\counterwithin{figure}{chapter}
\counterwithin{table}{chapter}

\usepackage{float}
\usepackage{subcaption}

\usepackage{url}
\usepackage{ragged2e}
\usepackage{tabularx}
\usepackage{booktabs}
\usepackage{multirow}
\usepackage[normalem]{ulem}
\usepackage{pdfpages}
\usepackage[acronym]{glossaries}
\usepackage{enumitem}
\usepackage{amsmath}
\usepackage{changepage}
\usepackage{acro}
\usepackage{amsfonts}
\usepackage{pdflscape}

\usepackage{tocloft,calc}
\AtBeginDocument{\addtolength\cftchapnumwidth{\widthof{Chapter }}}

\usepackage{minitoc}
\mtcsetoffset{minitoc}{0pt}
\mtcsetdepth{minitoc}{4}
\mtcsetfont{minitoc}{*}{\small\rmfamily\upshape\mdseries}
\mtcsetfont{minitoc}{section}{\small\rmfamily\upshape\bfseries}
\usepackage{color}
\usepackage{listings}
\usepackage{spverbatim}

\definecolor{red}{rgb}{0.6,0,0}
\definecolor{blue}{rgb}{0,0,0}
\definecolor{BLUE}{rgb}{0,0,0}
\definecolor{BLACK}{rgb}{0,0,0}
\definecolor{green}{rgb}{0,0.8,0}
\definecolor{cyan}{rgb}{0.0,0.6,0.6}

\usepackage{fancyhdr}
\fancypagestyle{front}{
  \fancyhf{}
  
  \cfoot{\thepage}
}
\fancypagestyle{main}{
  \fancyhf{}
  \fancyhead[C]{\slshape \rightmark}
  \fancyfoot[C]{\thepage}
  
}
\makeatletter
  \newcommand\frontpagestyle{\cleardoublepage\pagestyle{front}\let\ps@plain\ps@front}
  \newcommand\mainpagestyle{\cleardoublepage\pagestyle{main}\let\ps@plain\ps@main}
\makeatother

\input{blocked.sty}
\input{boxit.sty}
\input{thesis.sty}

\newcommand{\NN}{{\sf I\kern-0.14emN}}   
\newcommand{\ZZ}{{\sf Z\kern-0.45emZ}}   
\newcommand{\QQQ}{{\sf C\kern-0.48emQ}}   
\newcommand{\RR}{{\sf I\kern-0.14emR}}   

\newcommand{\normallinespacing}{\renewcommand{\baselinestretch}{1.5} \normalsize}

\newcommand{\syncc}{~\stackrel{\textstyle \rhd\kern-0.57em\lhd}{\scriptstyle L}~}

\usepackage{fix-cm}
\usepackage{bm}
\usepackage{amssymb}
\usepackage{xcolor}
\usepackage{array}
\usepackage{graphicx}
\usepackage{tcolorbox}

\usepackage{booktabs}
\usepackage{threeparttable}
\usepackage{algorithm}
\usepackage{algpseudocode}
\usepackage{amsmath}
\usepackage{tabularx}
\usepackage{siunitx}
\usepackage{changepage}
\usepackage{adjustbox}
\usepackage{ragged2e}
\usepackage{pdflscape}
\usepackage{etoolbox}

\usepackage[tableposition=top]{caption}

\usepackage{amsthm}

\usepackage{tocloft}
\cftsetindents{subsection}{1.5em}{2.3em}
\cftsetindents{subsubsection}{3.8em}{3.2em}

\usepackage[explicit]{titlesec}
\usepackage{lipsum}

\usepackage{silence}
\usepackage[hidelinks]{hyperref}
\usepackage{xurl}

\hypersetup{
  colorlinks=false,
  linkcolor=black,
  filecolor=black,
  citecolor=black,
  urlcolor=black,
  bookmarksdepth=6
}

\usepackage{setspace}

\newcommand\revision[1]{{\color{black}#1}}

\def\sectionautorefname{Section}
\def\subsectionautorefname{Subsection}

\usepackage[numbers]{natbib}

\usepackage{doi}
\usepackage{notoccite}

\makeatletter
\let\NAT@orig@newblock\newblock

\renewcommand{\newblock}{%
  \NAT@orig@newblock
  \@ifnextchar U{\NAT@rm@URLprefix}{}%
}

\def\NAT@rm@URLprefix U R L{%
  \@ifnextchar:\NAT@rm@URLcolon\NAT@rm@URLspace
}

\def\NAT@rm@URLcolon:\space{}
\def\NAT@rm@URLspace\space{}

\makeatother

\newcommand{\chapterwithquote}[3]{%
  \begingroup
  \def\thequote{#1}%
  \chapter{#2}\label{#3}%
  \endgroup
}

\titleformat{\chapter}[block]
  {\normalfont\LARGE\bfseries}
  {}
  {0em}
  {%
    {\scriptsize\color{gray}\textit{\thequote}}\\
    \parbox[b]{\linewidth}{%
      \hfill
      \parbox[b]{5cm}{%
        \centering
        {\textcolor{gray}{\fontsize{14}{16}\selectfont\textbf{CHAPTER}}}\\[0.2ex]
        {\textcolor{gray}{\fontsize{120}{110}\selectfont\thechapter}}%
      }%
    }\\[2cm]%
    \parbox[b]{\linewidth}{%
      \raggedright
      {\fontsize{14}{14}\selectfont\MakeUppercase{#1}}%
    }%
  }

\titleformat{name=\chapter,numberless}
  {\normalfont\LARGE\bfseries\filleft}
  {}
  {0em}
  {\parbox[b]{\dimexpr\linewidth-2.5cm\relax}{#1}}

\titlespacing*{\chapter}
  {0pt}{0pt}{0pt}

\title{
    \vspace{-0.1\baselineskip}
    \fontsize{16pt}{18pt}\selectfont
    Automated Feature Engineering, AutoML, and Decision-Focused Learning for Improved Energy Consumption Forecasting
}

\author{Nasser Alkhulaifi}
\degree{Doctor of Philosophy}
\submitdate{\normalfont\selectfont September 2025}
\studentid{}

\supervisor{
Prof Isaac Triguero
\\ Prof Nicholas J. Watson
\\ Dr Alexander L. Bowler
\\ Dr Direnc Pekaslan
\\ Prof Dario Landa Silva
\vspace{2\baselineskip}\\
{\normalfont\fontsize{14pt}{14pt}\selectfont
School of Computer Science, Faculty of Science\\}
\vspace{1\baselineskip}
{\normalfont\fontsize{16pt}{14pt}\selectfont
University of Nottingham}
}

\begin{document}


\normallinespacing
\maketitle


\preface


\cleardoublepage
\newgeometry{left=1.5in,right=1.5in,top=1.05in,bottom=1.0in} 
\chapter*{Abstract}
\addstarredchapter{Abstract}
\chaptermark{Abstract}
\begin{singlespace}
\begin{justify} 

The rising cost and demand for energy, coupled with the need to meet environmental sustainability goals, create pressing challenges for energy management. Energy Consumption Forecasting (ECF) supports informed planning by predicting future consumption patterns, yet Machine Learning (ML) models for ECF remain highly dependent on domain expertise. \revision{This dependence is driven largely by manual, expert-driven Feature Engineering (FE), since raw energy data often require preprocessing and transformation before ML algorithms can learn effectively. Moreover, real-world datasets are often small due to data collection limitations, privacy issues, or resource constraints; in such cases, FE can partially compensate by extracting and selecting informative features to maximise the utility of available data and improve predictive performance. 

While state-of-the-art AutoML frameworks streamline model selection and hyperparameter tuning, they typically assume that data preparation and FE have already been completed. Existing automated FE (AFE) methods are largely domain-agnostic and often fail to capture energy-specific temporal patterns and exogenous effects. In addition, because ECF forecasts usually drive downstream operational decisions, optimising prediction accuracy alone can allow residual errors to propagate into suboptimal actions; Decision-Focused Learning (DFL) methods aim to address this by integrating prediction with the downstream optimisation objective, yet have seen limited real-world evaluation in automated ECF settings.}

\revision{This thesis, therefore, addresses these challenges through three key contributions. First, it establishes and evaluates a comprehensive FE pipeline for ECF, investigates domain-specific features, and provides the empirical foundation for subsequent FE automation efforts. Second, it introduces AutoEnergy, a domain-tailored AFE algorithm that generates interpretable features from timestamps and lagged consumption via rule-based transformations, and integrates with AutoML to enable end-to-end automated ECF modelling. Across eighteen diverse real-world energy datasets spanning residential, commercial, industrial, renewable, and grid power domains, AutoEnergy reduces forecasting error by 19.52\%--84.72\% relative to baseline AutoML and established AFE methods, while running 1.31--4.41 times faster, with performance gains varying by dataset. Third, it leverages AutoEnergy within a DFL framework for a Battery Energy Storage System (BESS) problem, jointly forecasting electricity prices and demand while optimising a charging and discharging strategy; on a real-world UK property dataset, this AFE--DFL approach reduces operating costs by 22.9\%--56.5\% compared with the same DFL models without AFE.}

\revision{While the empirical results in this thesis are dataset-dependent, they are drawn from multiple real-world datasets spanning diverse energy settings. This supports two general conclusions: (i) integrating AutoEnergy with AutoML can enable automated ECF modelling that reduces reliance on manual FE while improving forecasting accuracy; and (ii) DFL enhanced with AFE can translate predictive improvements into measurable operational benefits for ECF applications. These contributions have broader implications for energy management systems in settings with limited domain expertise and small datasets, and demonstrate potential to support the transition to automated, AI-based energy systems.}

\end{justify}  
\end{singlespace}


\clearpage
\thispagestyle{empty}
\null
\newpage


\thispagestyle{empty}

\vspace*{\fill}

{\fontsize{12pt}{14pt}\selectfont\bfseries
\noindent
The preceding page was deliberately left blank as a symbolic acknowledgement and profound recognition that human knowledge, for all its ambition, rigour, and achievement, is but a faint glimmer of light within a vast surrounding darkness of uncertainty and unknown unknowns; and that, however boldly we advance as a questioning species, vast immensities will remain beyond the utmost horizon of our knowing and comprehension.\par
}

\vspace*{\fill}

\begin{center}
{\fontsize{8pt}{9pt}\selectfont
Nasser Alkhulaifi, 01:55 am, Tuesday, 30 Sept 2025, Nottingham, England, United Kingdom}
\end{center}

\clearpage


\newenvironment{acknowledgements}{
    \cleardoublepage 
    \chapter*{Acknowledgements} 
    \addcontentsline{toc}{chapter}{Acknowledgements} 
    \chaptermark{Acknowledgements} 
    \normalsize 
    \setlength{\parskip}{1em} 
}{\cleardoublepage} 

\begin{acknowledgements}
\begin{singlespace}
\begin{justify}

Firstly, I would like to thank my supervision team: Professor Isaac Triguero, Professor Nicholas Watson, Dr Alexander Bowler, and Dr Direnc Pekaslan for their tremendous support. I truly appreciate the time they have given me, the challenging and fruitful discussions we have shared, and their guidance throughout my PhD journey. I'm especially grateful that our relationship has grown into friendships. I'm also thankful to Professor Dario Landa Silva for joining us for the final lap. It has been a real pleasure working with them all. \revision{Many thanks to my thesis examiners, Professor Christian Wagner and Professor Jaume Bacardit, for their time and constructive feedback.}

I would like to acknowledge the Engineering and Physical Sciences Research Council (EPSRC), the University of Nottingham, and the Horizon CDT for funding my PhD studentship. Special thanks go to Dr Andrea Haworth, Professor Steve Benford, Dr Adrian Hazzard, Laura Brian, and Monica Cano Gomez. Thanks to the academics and staff at the School of Computer Science and the Faculty of Engineering at the University of Nottingham for their support and good company. To my friends at the DaSCI Institute, thank you for making my research visit all the more memorable under the glorious Andalusian sunshine.

To my friends and colleagues, Gregor, Torran, Milly, Gabrielle, Sam, Callum, Yang, Emma, Carina, Matt, Ellie, Jenn, Rachel, Gokay, Ife, Tim, and Aigerim! Thank you for filling this journey with so many memories. I will always cherish our spontaneous walks with Greg (especially on those grey days), our chess games, and the silly, random chats we shared. I can't forget cross-campus walks and badminton games with Torran, though I never won a single match :) but I kept showing up because they were good times. I'm grateful to Milly for her support. I would also like to thank my family for their unwavering love and encouragement.

This has been a truly special chapter of my life, full of wonderful memories that will live with me forever, plenty of laughter, and yes, the occasional low. Thank you all for being part of it.

\end{justify}  
\end{singlespace}
\end{acknowledgements}



\frontpagestyle

    \setcounter{tocdepth}{6}
    \setcounter{minitocdepth}{6}
    
    \dominitoc 
    \dominilof 
    \dominilot
    
    \tableofcontents
    \pagestyle{plain}
    
    \cleardoublepage
    \chapter*{}
    \addstarredchapter{List of Tables}
    \chaptermark{List of Tables}
    \listoftables
    \pagestyle{plain}
    
    \cleardoublepage
    \chapter*{}
    \addstarredchapter{List of Figures}
    \chaptermark{List of Figures}
    \listoffigures
    \pagestyle{plain}
    
    \cleardoublepage
    \chapter*{List of Abbreviations}
    \addstarredchapter{List of Abbreviations}
    \chaptermark{List of Abbreviations}

\begin{description}[labelwidth=0.5cm, leftmargin=0.0cm, rightmargin=0.0cm, itemsep=0pt]
\footnotesize
    \item[AI] Artificial Intelligence
    \item[AFE] Automated Feature Engineering
    \item[ANN] Artificial Neural Network
    \item[ARIMA] Autoregressive Integrated Moving Average
    \item[ARIMAX] Autoregressive Integrated Moving Average with Explanatory Variable
    \item[AutoEnergy] Automated Feature Engineering Algorithm for Energy Consumption Forecasting
    \item[AutoGluon] Amazon's AutoML Framework
    \item[AutoML] Automated Machine Learning
    \item[BESS] Battery Energy Storage System
    \item[CAAFE] Context-Aware Automated Feature Engineering
    \item[CNN] Convolutional Neural Network
    \item[CO] Constrained Optimisation
    \item[CV] Coefficient of Variation
    \item[DBB] Differentiable Black-Box
    \item[DFL] Decision-Focused Learning
    \item[ECF] Energy Consumption Forecasting
    \item[FDCS] Food and Drink Cold Storage
    \item[FE] Feature Engineering
    \item[FFNN] Feedforward Neural Network
    \item[FLAML] Fast and Lightweight AutoML (Microsoft's AutoML Framework)
    \item[FS] Feature Selection
    \item[FT] FeatureTools
    \item[GRU] Gated Recurrent Unit
    \item[H2O] H2O AutoML Framework
    \item[KNN] K-Nearest Neighbours
    \item[LIME] Local Interpretable Model-agnostic Explanations
    \item[LLM] Large Language Model
    \item[LR] Linear Regression
    \item[LSTM] Long Short-Term Memory
    \item[MAE] Mean Absolute Error
    \item[MAPE] Mean Absolute Percentage Error
    \item[MILP] Mixed Integer Linear Programming
    \item[ML] Machine Learning
    \item[MSE] Mean Squared Error
    \item[MTL] Multi-Task Learning
    \item[nRMSE] Normalised Root Mean Squared Error
    \item[PCA] Principal Component Analysis
    \item[PTO] Predict-Then-Optimise
    \item[R\textsuperscript{2}] Coefficient of Determination
    \item[RF] Random Forest
    \item[RMSE] Root Mean Squared Error
    \item[RNN] Recurrent Neural Network
    \item[SHAP] SHapley Additive exPlanations
    \item[SoC] State of Charge
    \item[SPO+] Smart Predict-Then-Optimise
    \item[TabPFN] Tabular Prior-data Fitted Network
    \item[TPOT] Tree-Based Pipeline Optimisation Tool
    \item[TS] TSFresh
    \item[XGB] Extreme Gradient Boosting (XGBoost)

\end{description}  
    \pagestyle{plain}
    
    \cleardoublepage
    \pagenumbering{arabic}
    \doublespacing
    
    \adjustmtc


\mainpagestyle

\cleardoublepage
\chapterwithquote{"If we knew what we were doing, it would not be called research, would it?" - Albert Einstein}{Introduction}{ch:intro}

This chapter introduces the thesis and establishes the foundation upon which the subsequent research is built. \autoref{intro:Background} presents the research background and motivation. After that, \autoref{intro:Aim_objectives} outlines the aim and objectives that guided the research process, while \autoref{intro:Contribution} highlights the key contributions and novelties of the thesis. Finally, \autoref{intro:Structure} provides a roadmap of the thesis structure to guide the reader through the forthcoming chapters, while \autoref{intro:Publications} lists the publications arising from this thesis.
\newpage 

\section{Background and Motivation}
\label{intro:Background}

The growing urgency of achieving net-zero emissions targets has intensified the need for effective energy management strategies across all sectors, particularly as energy-related CO\textsubscript{2} emissions reached a record 37.4 Gt in 2023 while global energy demand and cost continue to increase at an accelerated pace \cite{IEA2024energy, alimohamadi2025energy, Ashraf2025, Growth_in_global24}. In response to this challenge, countries worldwide have committed to ambitious net-zero emissions goals by 2050 or earlier, following agreements made at COP29 and previous climate conferences, with these commitments further reinforced through the United Nations Sustainable Development Goals \cite{koch2024cop29, UNEP2024climate, UN_DESA_2024_A, UN_DESA_2024_B}. Central to achieving these targets is the recognition that energy efficiency improvements can deliver over 40\% of necessary emissions reductions by 2040, making the accelerated deployment of smart technologies and energy management systems essential for global climate goals \cite{IEA2021_40}. This imperative has driven significant policy developments, exemplified by the UK Government's comprehensive approach through the Net Zero Strategy, Smart Systems and Flexibility Plan, and Net Zero Growth Plan, which collectively emphasise the critical role of energy efficiency and smart technologies \cite{BEIS2021, DESNZ2021, DESNZ2023}. The strategic importance of integrating Artificial Intelligence (AI) with energy systems has been further demonstrated through recent policy initiatives, including the establishment of the AI Energy Council in 2025, the AI Opportunities Action Plan, and the Energy Digitalisation Strategy, highlighting the growing recognition of AI's potential to transform energy management \cite{GOVUK2025aicouncil, UKGov2025_AI_Plan, UKGovend2021_AI}.

\begin{figure}[!t]
  \centering
  \includegraphics[width=\textwidth]{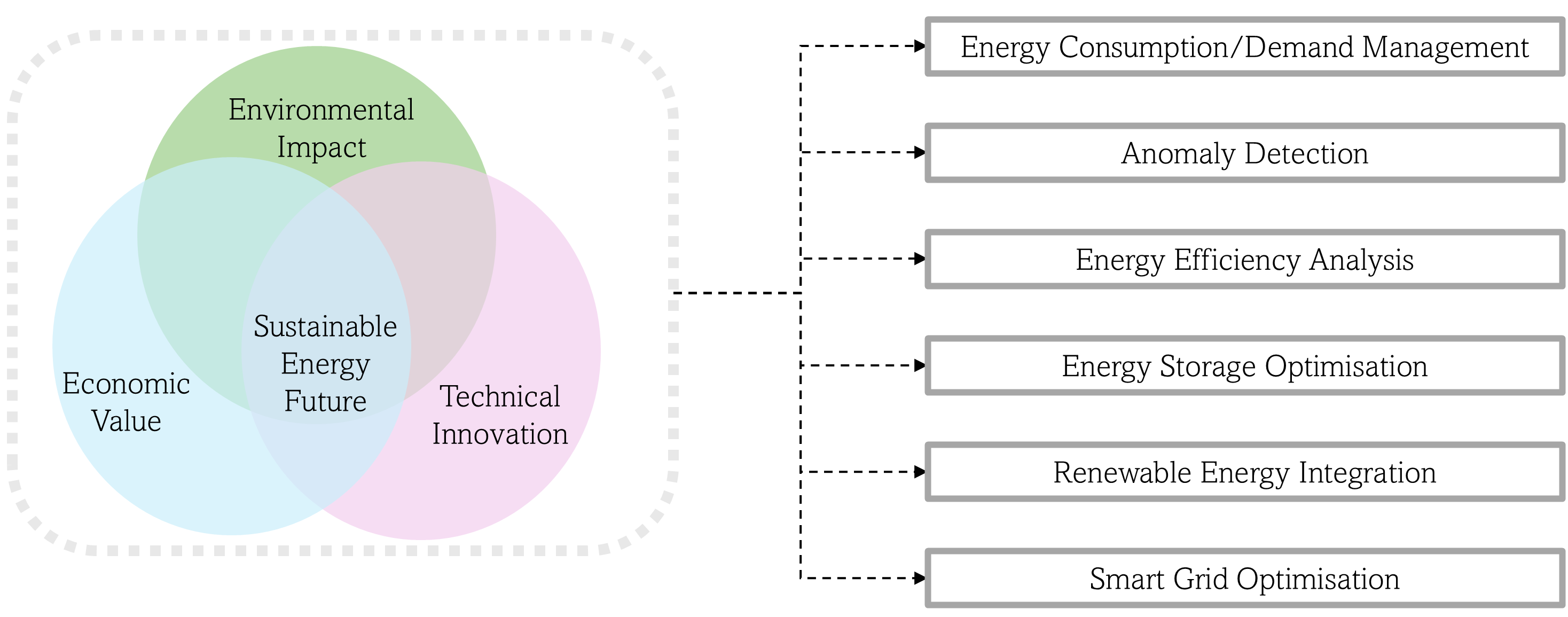}
  \caption{Different potential applications of ECF. These applications rely on accurate ECF to optimise resource allocation, reduce costs, enhance grid stability, and support the transition to sustainable energy systems.}
  \label{fig:enery_forecasting_matter}
\end{figure}

In this context, Energy Consumption Forecasting (ECF) emerges as a useful tool to aid in addressing these challenges, achieving these ambitious policy targets and supporting the transition to sustainable, smart energy systems. For example, ECF can enable more informed energy management by predicting future consumption patterns, allowing decision-makers to take pre-emptive action and implement effective planning strategies to reduce energy use and minimise waste and emissions \cite{deb2017review, amasyali2018review}. ECF can estimate future demand profiles, offering a baseline to help assess the potential impact of energy-saving interventions by comparing projected and actual consumption trends. \autoref{fig:enery_forecasting_matter} illustrates potential applications of ECF.


Three approaches to ECF modelling have been studied in the literature: white-box (physics-based), grey-box (hybrid), and black-box Machine Learning (ML) methods \cite{sun2020review}. ML methods have become the most widely studied approach in recent ECF literature due to their capacity to learn complex patterns directly from data without requiring explicit programming or physical model formulation \cite{Nti2020, Klyuev2022}, which distinguishes them from hybrid methods that still require some level of physics-based modelling expertise. Therefore, this thesis focuses specifically on ML techniques to model ECF. In general, ML methods can be categorised into four main types: supervised, unsupervised, semi-supervised, and reinforcement learning \cite{ayodele2010types}. This thesis uses supervised learning methods as ECF inherently involves historical input-output pairs where consumption values (output) are associated with corresponding features (inputs) such as time variables (e.g., hour of the day), weather conditions (e.g., outdoor temperature), and historical consumption data \cite{Ahmad2018}.

Developing ML models for ECF is a resource-intensive task requiring significant time, technical expertise and computational power \cite{zhang2021review}. Additionally, accurate ECF requires domain knowledge for Feature Engineering (FE), a crucial step in developing ML pipelines, to transform raw data into informative features, as this process significantly influences model performance \cite{Wang_2022, Mumuni_2024}. This is because raw data typically require preprocessing and transformation before algorithms can learn
effectively \cite{Verdonck2021, Liu2024}. Furthermore, real-world datasets are often small or limited due to data collection limitations, privacy issues, or resource constraints \cite{grinsztajn2022tree, hollmann2022tabpfn}. \revision{In this thesis, \textit{small dataset size} refers to a limited number of time-indexed observations (e.g., less than three months of data) available for model training and evaluation, relatively small compared to related studies (e.g., 6-12 months \cite{Fan_2017,Rahman2018}, two years \cite{kim2019Predicting,Wang2019Power}}). In such scenarios, FE could compensate for the limited data by extracting informative features. Moreover, FE may improve computational efficiency by eliminating noisy, redundant data \cite{Wang_2022} and supporting model explainability \cite{Gosiewska2021}.  However, manual FE for ECF remains a time-consuming process that is prone to human error while relying heavily on domain expertise and iterative experimentation \cite{Wang_2022}.

\revision{
In this thesis, \textit{domain knowledge} in FE refers to human judgement derived from subject-matter expertise that informs which variables, representations, and transformations are considered meaningful for a given modelling problem \cite{Duboue2020_art_FE, Bjrneld2023_FE}. In practice, ECF domain experts can identify plausible drivers of consumption and the kinds of patterns the system may exhibit, while data scientists validate these hypotheses empirically and operationalise them as model inputs. For example, domain experts are likely to know that energy demand often shows periodic behaviour, with recurring daily, weekly, or seasonal cycles. This motivates the construction of explicit temporal features (e.g., lagged values and rolling-window statistics) to encode time-dependent structure (i.e., representative features) in a form that learning algorithms can exploit. 

Accordingly, manual FE effort refers to the repeated, trial-and-error work of proposing, implementing, and tuning these features, and of iterating on them across datasets until performance becomes acceptable \cite{amasyali2018review, sun2020review, Wang_2022}. Within this framing, references in the thesis to \textit{reducing reliance on domain knowledge} should be understood as a reduction in the amount of bespoke, repeated, expert-driven feature design required each time the modelling pipeline is applied to a new ECF dataset. This interpretation is adopted consistently throughout the remainder of the thesis.
}

\revision{
Note that this thesis distinguishes between \textit{generalisable modelling knowledge} and \textit{context-specific knowledge}. Generalisable modelling knowledge refers to a widely applicable understanding of how energy consumption time series may behave and which representations are typically useful for ECF, such as diurnal and weekly cycles and short-term temporal dependence. In contrast, Context-specific knowledge refers to local factors that may be unique to a particular site or asset (e.g., occupancy schedules, equipment control policies, and maintenance) and that shape consumption in ways that may not be generalisable. The work proposed in this thesis, particularly in \autoref{ch:mainchapter5}, is designed to encode and automate generalisable modelling knowledge through potentially generalisable FE for ECF problems across different energy systems. It does not attempt to replace context-specific knowledge, which typically requires site-specific expertise to interpret, validate, and incorporate via additional signals. This type of knowledge is inherently difficult to generalise because each energy system operates under its own conditions and constraints.
}

While deep learning methods, a subset of ML, can automatically learn representations \cite{Fan2019Deep}, thereby minimising the need for expert-driven FE, they are generally less interpretable and typically require large datasets to excel, a condition that is often challenging in real-world scenarios \cite{grinsztajn2022tree, Wang2019}. These limitations have driven growing interest in Automated FE (AFE) methods to minimise reliance on expert input for each new dataset \cite{Wu2022,hollmann2024large}. Modern Automated ML (AutoML) frameworks \cite{hutter2019automated} such as AutoGluon \cite{AutoGluon2020}, H2O \cite{h2o}, and FLAML \cite{wang2021flaml} offer streamlined solutions for ML pipeline development through automated model selection and hyperparameter tuning. Nevertheless, these solutions often assume that data preparation and feature generation have been completed, and the data are ready for training. As a result, tasks such as FE and the integration of domain knowledge are largely left to human practitioners \cite{hollmann2024large}. While there have been growing attempts at AFE, such as TSfresh (TS) \cite{Christ2018} and FeatureTools (FT) \cite{kanter2015deep}, these approaches were not specifically designed for ECF problems and may not capture domain-specific temporal patterns and consumption characteristics inherent in energy data. Therefore, it is important to minimise dependency on domain expertise and enhance the performance of AutoML in ECF problems through AFE.

ECF ultimately serves downstream decision-making processes, especially in energy management tasks where optimising downstream tasks is important. However, while improving forecasting accuracy remains a key focus in ECF research, such improvements alone do not ensure better practical outcomes \cite{Wilder2019, donti2017task}. It is therefore important to study whether gains in predictive accuracy, particularly those achieved through AFE, lead to better operational decisions. \revision{This could be addressed by using downstream decision quality as the target: a \textit{better} decision may be understood as one that improves the downstream optimisation outcome (i.e., yields a lower operational cost/higher utility under the task’s objective), rather than merely reducing forecasting error.} This necessitates extending AFE beyond forecasting accuracy-focused approaches to optimise for downstream decision quality. Consequently, such methods should be evaluated within Predict-Then-Optimise (PTO) frameworks and the emerging paradigm of Decision-Focused Learning (DFL) \cite{Boettiger2022, Mandi2024}. This is to ensure that features generated not only enhance forecasting accuracy but also improve operational decision quality, thereby evaluating the real-world value of AFE in practical ECF applications.

\revision{
\begin{spacing}{0.95}
In response to these challenges, this thesis formulates the following hypothesis:
\begin{quote}
\textit{Integrating domain-specific AFE with AutoML and DFL can enhance ECF applications by minimising reliance on domain knowledge, improving forecasting accuracy, and optimising downstream tasks compared with conventional ML approaches.}
\end{quote}
Informed by this hypothesis, the overarching aim and objectives are articulated in \autoref{intro:Aim_objectives}.
\end{spacing}
}


\section{Research Aim and Objectives}
\label{intro:Aim_objectives}

\textbf{Aim:} This thesis aims to develop ML models for ECF that minimise reliance on domain knowledge and can be applied across diverse energy systems while maximising forecasting accuracy and optimising downstream tasks. \revision{As aforementioned, \textit{minimising reliance on domain knowledge} is scoped to the stages of the ML pipeline where expert input is typically heaviest, namely FE and its associated selection choices. In other words, to reduce the amount of manual, dataset-specific FE required before reliable models can be trained.} To achieve this aim, the following research objectives (ROs) were defined:

\begin{enumerate}
    \item \textbf{RO1: To establish baseline ML models for ECF and investigate the role of domain knowledge in FE for such forecasting problems}. In particular, to design a detailed ML pipeline for ECF problems. This objective includes:
    
        \begin{enumerate}
            \item Collecting historical electricity consumption data along with weather data to serve as a real-world case study for modelling ECF.
            
            \item Identifying and investigating domain-specific features for ECF.
    
            \item Evaluating different Feature Selection (FS) techniques, algorithm types, and hyperparameter tuning for ECF.
    
            \item Analysing feature importance in ECF problems and the impact of dataset size on model performance.  
        \end{enumerate}

\noindent Completing this objective will produce a comprehensive pipeline for ECF problems that can help other practitioners minimise the time needed to implement domain knowledge-based features for ECF applications.


    \item \textbf{RO2: To develop an AFE method for ECF problems, thereby streamlining ML model development by addressing the most time-consuming and expert-dependent task in the pipeline}. In particular, this objective addresses the following question: \textit{How can the domain knowledge required for FE be minimised through automation in ECF problems?} It further breaks down into two sub-questions:

        \begin{enumerate}
            \item Which features to generate?

            \item How to automate FS?
        \end{enumerate}
    
    \noindent \revision{Satisfying this objective will produce an AFE algorithm that researchers can use to automatically generate a set of features tailored for ECF problems. The generated features are human-readable and traceable to explicit, well-defined transformations of the input data, rather than an opaque learned representation, while minimising both time and expertise barriers to accurate ECF modelling and improving AutoML performance.}


    \item \textbf{RO3: To leverage the proposed AFE algorithm to optimise the downstream tasks beyond predictive performance}. Specifically, through three complementary sub-objectives:

     \begin{enumerate}
            \item Expanding evaluation of the proposed AFE beyond predictive performance to include downstream optimisation problems.

            \item Improving the nascent DFL methods.

            \item Evaluating performance using a novel real-world dataset and settings to examine DFL's practical viability.

        \end{enumerate}

    \noindent Fulfilling this objective will enhance the development of AFE that not only maximises forecasting accuracy but also is decision-aware, thus delivering tangible operational value in practical ECF applications.
    
\end{enumerate}


\section{Contribution and Novelty}
\label{intro:Contribution}

In this thesis, a series of studies has been conducted based on the identified ROs. The novelty contained within this thesis can be summarised as follows:

\begin{enumerate}

    \item \textbf{Development of a detailed ML pipeline for ECF (RO1).} This pipeline is (A) capable of predicting one week into the future at an hourly resolution (\revision{multi-step forecasting where the model predicts the next 168 hours}) to maximise usability, as short-term forecasts often lack practical value for planning, and (B) suitable for small dataset sizes (\revision{i.e., settings where only a limited number of time-indexed observations are available for model training}). This includes an extensive investigation into the role of domain knowledge in FE, a comparison of eight different FS methods, feature importance analysis, and dataset size implications. \revision{The contribution and novelty of this pipeline are detailed in \autoref{ch:mainchapter3}}

    \item \textbf{Development of AutoEnergy, an AFE algorithm tailored for ECF problems to minimise reliance on domain knowledge (RO2).} This algorithm was evaluated across eighteen datasets from diverse energy systems and environments, including residential, commercial, industrial, renewable, and grid power domains, thus demonstrating robust generalisation potential. AutoEnergy automatically generates interpretable features from timestamps and past consumption values without manual intervention, further streamlining ML development for ECF through seamless integration with AutoML. \revision{Although the amount of human-expert effort in manual FE is difficult to quantify directly, it is a key practical bottleneck in ECF pipelines \cite{amasyali2018review, sun2020review, Wang_2022}, AutoEnergy addresses this by replacing repeated, dataset-specific manual FE design with an automated, rule-based generation of ECF-relevant features}. It consistently outperforms general-purpose baseline AutoML. Furthermore, comprehensive benchmarking against existing FE methods shows that, on average, AutoEnergy achieves greater reductions in both forecasting errors and processing time. \revision{The contribution and novelty of AutoEnergy are detailed in \autoref{ch:mainchapter5}}

    \item \textbf{Development of a decision-aware, end-to-end ML forecasting and optimisation framework that jointly forecasts electricity demand and price over a 24-hour horizon while leveraging AutoEnergy to optimise operational decisions in ECF applications and enhance the nascent DFL (RO3).} This framework was evaluated on a Battery Energy Storage System (BESS) problem using real-world data and experimental settings, where PTO and DFL approaches assessed decision quality through regret metrics rather than conventional forecasting metrics. \revision{The contribution and novelty of this framework, including the evaluation and benchmarking methods, are detailed in \autoref{ch:mainchapter6}.}

\end{enumerate}

\section{Thesis Structure}
\label{intro:Structure}
This thesis is structured to address the ROs outlined in \autoref{intro:Aim_objectives}. Each chapter contributes to achieving these objectives through a series of interconnected studies, organised as follows:

    \begin{itemize}
    
        \item \autoref{ch:lit_review} reviews related work and delves into the ML pipeline for ECF, laying the groundwork for subsequent chapters by outlining the current state of the field.

        \item \autoref{ch:mainchapter3} addresses \textbf{RO1} by developing a robust ML pipeline for ECF. It highlights the role of domain knowledge in FE and evaluates multiple FS techniques, establishing a robust empirical foundation for FE automation in later chapters.

        \item \autoref{ch:mainchapter5} pursues \textbf{RO2} by introducing AutoEnergy, an AFE algorithm tailored for ECF. This chapter benchmarks AutoEnergy against well-established AFE methods across eighteen diverse energy datasets, demonstrating superior forecasting performance and computational efficiency.
        
        \item \autoref{ch:mainchapter6} fulfils \textbf{RO3} by evaluating the downstream impact of AutoEnergy within PTO and DFL frameworks. Using real-world BESS data, this chapter demonstrates how improvements in AFE influence operational decision quality, particularly under data scarcity, highlighting the decision-aware benefits of the proposed algorithm.

        \item \autoref{ch:conclusion} summarises the findings from all chapters and discusses the thesis’s contributions, reflects on limitations and outlines recommendations for future work.

    \end{itemize}

\revision{To further clarify how the thesis addresses different forms of domain knowledge, the chapters can be read as targeting complementary layers. \autoref{ch:mainchapter3} analyses domain knowledge in the practical, context-specific sense by demonstrating how manually engineered, dataset-dependent choices affect performance and robustness in a real-world setting (i.e., case study in two food storage facilities). \autoref{ch:mainchapter5} targets the generalisable layer by encoding reusable ECF feature categories and selection heuristics into a rule-based AFE method that can be applied across diverse energy systems and datasets. \autoref{ch:mainchapter6} extends this perspective to decision contexts by showing how forecasting representations affect downstream optimisation quality.}


\begin{landscape}
\begin{figure}[!t]
  \centering
  \includegraphics[width=\linewidth]{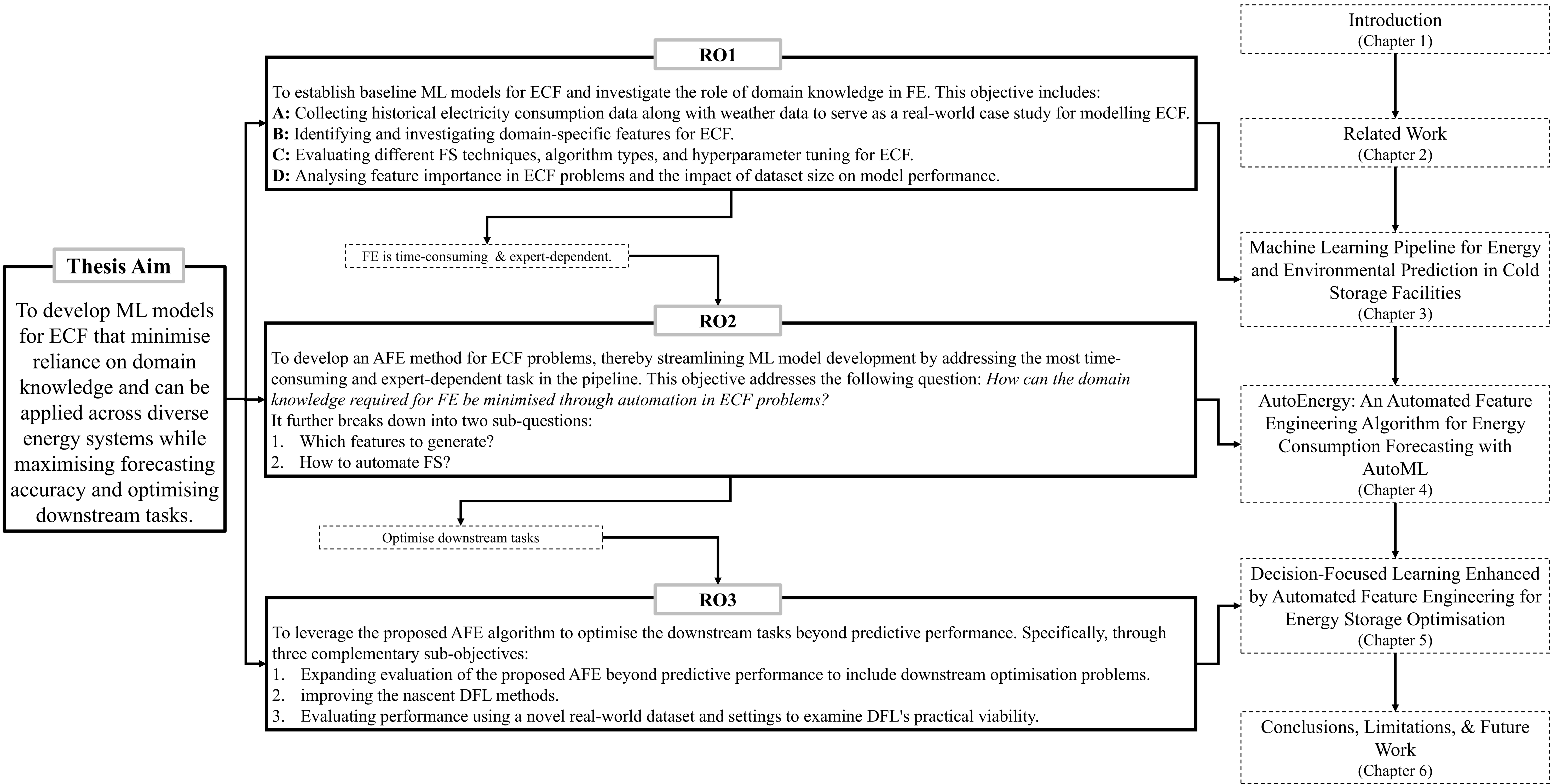}
  \caption{Chapter structure and alignment with the research aim and objectives of this thesis. The abbreviations used in the figure are as follows: RO (Research Objective), ML (Machine Learning), AutoML (Automated Machine Learning), ECF (Energy Consumption Forecasting), FE (Feature Engineering), AFE (Automated Feature Engineering), FS (Feature Selection), and DFL (Decision-Focused Learning).}
  \label{fig:thesis_outline}
\end{figure}
\end{landscape}


\section{List of Publications}
\label{intro:Publications}

First-author publications (included in this thesis):

\begin{itemize}

    \item \textbf{Alkhulaifi, N.}, Bowler, A.L., Pekaslan, D., Serdaroglu, G., Closs, S., Watson, N.J., and Triguero, I., 2024. Machine Learning Pipeline for Energy and Environmental Prediction in Cold Storage Facilities. \textit{IEEE Access}. \href{https://doi.org/10.1109/ACCESS.2024.3482572} {https://doi.org/10.1109/ACCESS.2024.3482572}. \\
    The content of this article is covered in \autoref{ch:mainchapter3} \cite{Alkhulaifi2024_pipeline}.

    \item \textbf{Alkhulaifi, N.}, Bowler, A.L., Pekaslan, D., Triguero, I., and Watson, N.J., 2024. Exploring Automated Feature Engineering for Energy Consumption Forecasting with AutoML. \textit{IEEE International Conference on Systems, Man, and Cybernetics (SMC) (pp. 2993-2998)}.     \href{https://doi.org/10.1109/SMC54092.2024.10831959}{https://doi.org/10.1109/SMC54092.2024.10831959}. \\
    The content of this article is presented in \hyperref[ch:mainchapter4]{Appendix A} \cite{Alkhulaifi2024}.

    \item \textbf{Alkhulaifi, N.}, Bowler, A.L., Pekaslan, D., Watson, N.J., and Triguero, I., 2025. AutoEnergy: An automated feature engineering algorithm for energy consumption forecasting with AutoML. \textit{Knowledge-Based Systems}, 329, 114300. \href{https://doi.org/10.1016/j.knosys.2025.114300}{https://doi.org/10.1016/j.knosys.2025.114300}.\\
    The content of this article is covered in \autoref{ch:mainchapter5} \cite{Alkhulaifi2025}.

    \item \textbf{Alkhulaifi, N.}, Dogan, I.G., Cargan, T.R., Bowler, A.L., Pekaslan, D., Watson, N.J., and Triguero, I., 2026. Decision-Focused Learning Enhanced by Automated Feature Engineering for Energy Storage Optimisation. Expert Systems with Applications, 302, 130554. \href{https://doi.org/10.1016/j.eswa.2025.130554}{https://doi.org/10.1016/j.eswa.2025.130554}\\
    The content of this article is covered in \autoref{ch:mainchapter6} \cite{alkhulaifi2025DFL}. \\
    
\end{itemize}


\noindent Co-authored publications (excluded from this thesis):

\begin{itemize}
    \item Alagoz, B.B., Keles, C., Ates, A., Özdemir, E. and \textbf{Alkhulaifi, N.}, 2025. Optimal deep neural network architecture design with improved generalization for data-driven cooling load estimation problem. \textit{Neural Computing and Applications}, 37(19), pp.13597-13616. \href{https://doi.org/10.1007/s00521-025-11212-7}{https://doi.org/10.1007/s00521-025-11212-7} \cite{Alagoz2025_co_authored}. 

    \item Canatan, M., \textbf{Alkhulaifi, N.}, Watson, N. and Boz, Z., 2025. Artificial Intelligence in Food Manufacturing: A Review of Current Work and Future Opportunities. \textit{Food Engineering Reviews}, 17(2), pp.189-219. \href{https://doi.org/10.1007/s12393-024-09395-1}{https://doi.org/10.1007/s12393-024-09395-1} \cite{Canatan2025_co_authored}. 

     \item Bowler, A.L., \textbf{Alkhulaifi, N.}, Bowler, S., Sier, J.H., Ferreira, C., Greetham, D., Pennells, J., Knoerzer, K. and Watson, N.J., 2026. Hybrid modelling and transfer learning for Bayesian optimisation of yeast protein production from food waste substrates. bioRxiv, 2026.07.24.740460. \href{https://doi.org/10.64898/2026.07.24.740460}{https://doi.org/10.64898/2026.07.24.740460} \cite{Bowler2026_co_authored}.

\end{itemize}

\cleardoublepage

\chapterwithquote{"Science is a way of thinking much more than it is a body of knowledge" - Carl Sagan}{Related Work}{ch:lit_review}

This chapter reviews the research landscape of ML-based ECF, highlighting current limitations and unexplored opportunities in the field. \autoref{ch2:Paper1_S1} begins by formulating ECF as a supervised learning problem, followed by a review of model inputs (\autoref{ch2:Paper1_S2}), FE approaches including feature extraction and selection (\autoref{ch2:Paper1_S3}), AFE methods (\autoref{ch2_Automated_Feature_Engineering}), and ML algorithms used for ECF (\autoref{ch2:Paper1_S4}). \autoref{ch2:Paper2} reviews AutoML approaches for streamlining the development of ML pipelines for ECF and outlines their current limitations. Finally, \autoref{ch2:Paper4} explores DFL approaches and their role in improving decision-making processes within ECF applications.
\newpage 
\section{\revision{Energy Consumption Forecasting using Supervised Machine Learning}}
\label{ch2:Paper1_S1}



\revision{The recent surge in data availability, fuelled by lower sensor costs and improved data processing capabilities, has ushered in numerous data-driven methods for modelling and predicting dynamic behaviours such as ECF \cite{fisher20, montáns19}. These approaches are particularly useful when the underlying physics of the system is not well understood or is difficult to model \cite{montáns19,zhao_12}. In a data-driven model, data gathered from regular operations or specific tests is analysed using algorithms such as statistical regression to understand the relationship between input and output variables \cite{mosavi2019energy,amasyali2018review}. While statistical methods such as Autoregressive Moving Average \cite{pappas08}, Autoregressive Integrated Moving Average (ARIMA) \cite{ho21,chen2009arima,sen2016application,nepal20}, and Autoregressive Integrated Moving Average with eXogenous variables (ARIMAX) \cite{newsham2010building,pereira15,cui15} have been used for forecasting energy consumption in buildings, they depend on restrictive assumptions such as linearity and stationary input data (constant statistical properties over time).  These limitations have led to the exploration of ML techniques, algorithms that learn from data to make predictions without explicit programming, thereby overcoming such constraints and demonstrating growing interest in this field \cite{deb2017review,robinson17,Seyedzadeh2018,hoang2021development}.}

\begin{figure}[!t]
\centering
\includegraphics[width=1\linewidth]{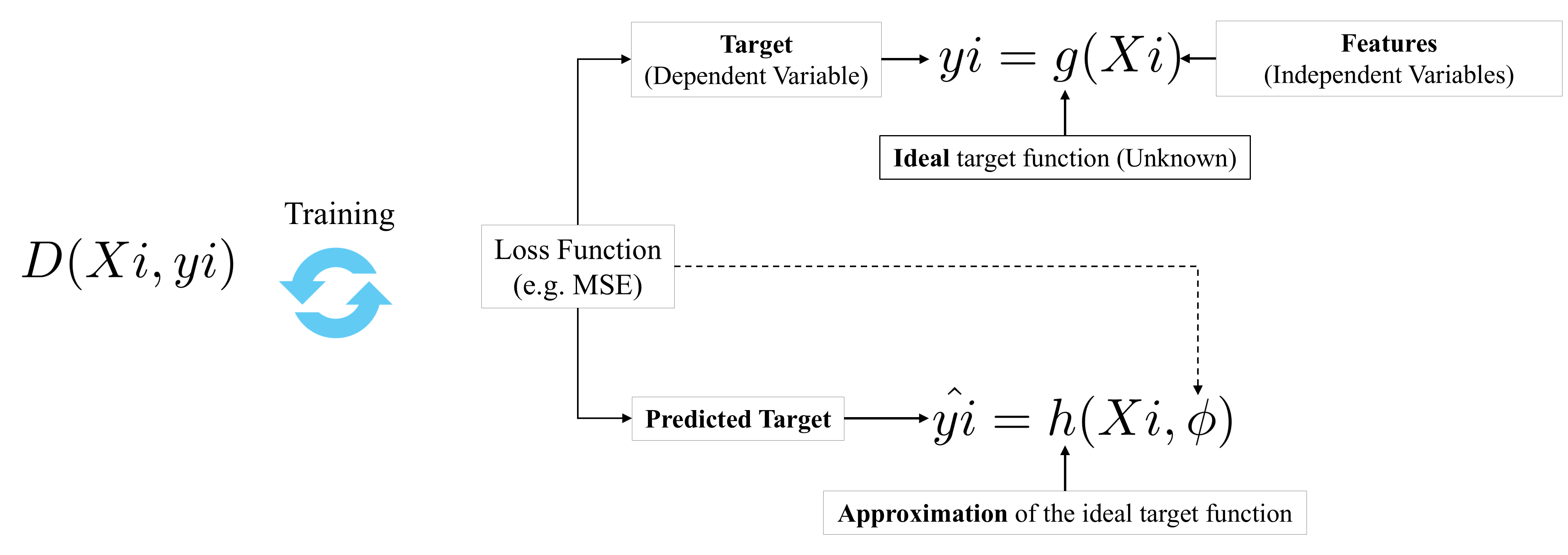}
\caption{Schematic representation of supervised ML adapted from \cite{joseph2022modern} with modifications for clarity. It illustrates the relationship between training data $D = \{(X_i, y_i)\}_{i=1}^N$ and the model learning process, where $X_i$ represents the features, $y_i$ represents the target, and $i$ denotes the $i$-th data point in the dataset. The ideal target function $g(X_i)$, which is unknown, maps features (i.e., inputs) to the target (i.e., output). During training, a parametric model produces predicted target $\hat{y}_i = h(X_i, \phi)$, where $h$ is the model function and $\phi$ represents the learnable parameters.}
\label{Supervised_ML_schematic}
\end{figure}

Supervised learning is a branch of ML where models learn to map inputs to outputs based on labelled examples. As illustrated in \autoref{Supervised_ML_schematic}, this involves training a parametric model to approximate the unknown ideal target function using available data and a suitable loss function. In this context, ECF can be formulated as a supervised ML problem where the model learns the relationship between the inputs (i.e., known as features, predictors or independent variables) and the output (known as target, label or dependent variable) \cite{zhang2021review,Bourdeau2019}. Formally, given a dataset $D = \{(X_i, y_i)\}_{i=1}^N$ with $N$ data points, where $X_i$ represents the feature vector (e.g., weather conditions, time features) and $y_i$ represents the target energy consumption value, the objective is to learn a function $h(X_i, \phi)$ that approximates the ideal target function $g(X_i)$, where $h$ is the model function and $\phi$ represents the learnable parameters. This is achieved by minimising a loss function (e.g., Mean Squared Error for regression) that quantifies the difference between predicted values $\hat{y}_i$ and actual values $y_i$ across the training dataset, guiding iterative parameter optimisation to approximate the ideal target function.

\subsection{\revision{Model Inputs for Energy Consumption Forecasting}}
\label{ch2:Paper1_S2}
Appropriate model inputs are essential for developing accurate ML energy usage prediction models \cite{Ding2017}. As weather heavily influences energy use in buildings through heating needs in cold climates and cooling needs in warm ones, numerous studies have utilised weather data such as temperature, dew point, humidity, precipitation, wind speed, air pressure, and solar radiation as input features for energy prediction models \cite{Xu2019Probabilistic,Cao2023,Rahman2018,Liu2023,Ding2017,Gao2021,Hong2022,Wei2019,Moon2020}. Historical energy data, capturing the complexities of actual consumption patterns influenced by various factors such as abnormal events and human activities, is also commonly used as inputs because it numerically indicates both the pattern and trend of the load profile \cite{Deb2016,Fan_2017,kim2019Predicting,Wang2020}. In addition to weather and historical data, very few studies have incorporated indoor features such as the number of occupants, zone air temperature, zone relative humidity \cite{Ding18Model}, indoor humidity, indoor temperature, and indoor carbon dioxide levels \cite{Wei2019}. However, obtaining such data often requires specialised sensors that may be unavailable due to privacy concerns, logistics, and cost \cite{Wang2020}. \autoref{tab:related_work} summarises the inputs, outputs, and prediction methods used in related works for ECF. 

\begin{landscape}
\begin{table}[t!]
\centering
\fontsize{9pt}{11pt}\selectfont
\setlength{\tabcolsep}{4.5pt}
\renewcommand{\arraystretch}{1.25}

\caption{\revision{Summary of different feature engineering approaches, inputs, and algorithms studied in the related work on energy consumption forecasting, along with the energy system/application contexts. See \autoref{tab:FS_methods} for additional details on feature selection methods}}

\label{tab:related_work}

{\color{black}
\begin{tabular}{
>{\centering\arraybackslash}m{2cm}
>{\centering\arraybackslash}m{2cm}
>{\raggedright\arraybackslash}p{5.5cm}
>{\raggedright\arraybackslash}p{4cm}
>{\raggedright\arraybackslash}p{4cm}}
\toprule
\multirow[c]{16}{2.5cm}[-4cm]{\raggedright\textbf{Feature Engineering}} &
\multirow{10}{2.5cm}[-2.5cm]{\raggedright\textbf{Feature Extraction}} &
\textbf{Feature Type} &
\textbf{Algorithm(s)} &
\textbf{Energy System Application Context} \\
\cline{1-5}

& &
Calendar features: e.g., hour, day-of-week, month, holiday \cite{Xu2019Probabilistic,Wang2020,Tan2020,Hong2022}. &
ANN; KNN, RF, XGB, SVM, stacking \cite{Wang2020}; MTL-SVM \cite{Tan2020}; KNN ensemble \cite{Hong2022}. &
Campus buildings (heating) \cite{Wang2020}; Industrial parks \cite{Tan2020}; Public/commercial facilities \cite{Hong2022}. \\
\cline{3-5}

& &
Sin/Cos transformations: cyclical encoding (e.g., hour of day, day of week) using sine/cosine \cite{Kim2019Recurrent,Moon2020}. &
Recurrent Inception CNN \cite{Kim2019Recurrent}; Stacking with PCR \cite{Moon2020}. &
Industrial complexes \cite{Kim2019Recurrent}; Office building (HQ) \cite{Moon2020}. \\
\cline{3-5}

& &
Lag-based (cyclic history): previous hour, day, week \cite{Wang2020,Deb2016,Gao2021,Liu2023}. &
KNN, RF, XGB; ANN \cite{Deb2016}; LSTM + attention \cite{Gao2021}; Deep MTL \cite{Liu2023}. &
Campus heating \cite{Wang2020}; Institutional cooling \cite{Deb2016}; Office electricity load \cite{Gao2021}; Commercial building \cite{Liu2023}. \\
\cline{3-5}

& &
Rolling-window statistics: e.g., moving min, max, mean, std \cite{Fan_2017}. &
Elastic Net, GB Trees, RF, SVR, DNN, Auto-encoder \cite{Fan_2017}. &
Educational building (chilled-water system) \cite{Fan_2017}. \\
\cline{3-5}

& &
System/Site-specific: E.g., occupant count, indoor air temperature, humidity, zone loads; building metadata; lighting, equipment electricity \cite{Massana2015,Ding18Model,Cao2023,OluAjayi2023}. &
Bagging + genetic selection \cite{Massana2015}; SVR, MLP with Wavelet/Correlation \cite{Ding18Model}; CNN-GRU + SHAP \cite{Cao2023}; 14 different ML methods \cite{OluAjayi2023}. &
Occupancy-aware office \cite{Massana2015}; Office heating \cite{Ding18Model}; Educational building (electricity) \cite{Cao2023}; UK residential (energy rating) \cite{OluAjayi2023}. \\
\cline{2-5}

& \multirow{5}{2.5cm}[-1.5cm]{\raggedright\textbf{Feature Selection}} &
\textbf{Selection Method} & & \\
\cline{3-3}

& &
Filter-based: E.g., Pearson correlation, ANOVA, Chi-square, mutual information \cite{Moon2020,OluAjayi2023}. &
Stacking ensemble \cite{Moon2020}; ML comparative study \cite{OluAjayi2023}. &
Office building \cite{Moon2020}; UK residential \cite{OluAjayi2023}. \\
\cline{3-5}

& &
Wrapper-based: E.g., genetic search for optimal feature subset \cite{Massana2015}. &
Bagging trees \cite{Massana2015}. &
Office building (occupancy-driven) \cite{Massana2015}. \\
\cline{3-5}

& &
Embedded-based: E.g., Feature ranking during model training (permutation) \cite{OluAjayi2023}. &
& UK residential buildings \cite{OluAjayi2023}. \\
\cline{3-5}

& &
Other: E.g., wavelet, ablation \cite{Ding18Model,Cao2023}. &
SVR, MLP \cite{Ding18Model}; CNN-GRU, stacking \cite{Cao2023}. &
Office heating \cite{Ding18Model}; Educational building \cite{Cao2023}. \\
\toprule
\end{tabular}
}
\end{table}
\end{landscape}


\subsection{\revision{Feature Engineering for Energy Consumption Forecasting}}
\label{ch2:Paper1_S3}
FE is the process of generating, extracting, and selecting features that transform raw data into informative inputs to enhance model performance when developing ML pipelines \cite{Domingos2012, Bengio2013}. This is because raw data often require preprocessing and transformation before algorithms can learn effectively \cite{Liu2024, Verdonck2021}. Furthermore, real-world datasets are often small or limited due to data collection limitations, privacy issues, or resource constraints \cite{grinsztajn2022tree, hollmann2022tabpfn}. In such scenarios, FE could compensate for the limited data by extracting informative features. Moreover, FE may improve computational efficiency by eliminating noisy, redundant, and irrelevant data \cite{Wang_2022, Mumuni_2024}. Finally, FE may improve not only predictive accuracy but also explainability \cite{Gosiewska2021}.

To better understand the FE process and its objectives in supervised learning contexts, it can be formally defined as follows: given a dataset \( D = \{(x_i, y_i)\}_{i=1}^N \) with \( N \) instances, the objective is to find a feature transformation function \( \phi : X \rightarrow X' \) that improves the performance of a learning algorithm \( A \) when trained on the transformed dataset \( D' = \{(\phi(x_i), y_i)\}_{i=1}^N \). Here, \( x_i \) represents the input features, \( y_i \) denotes the corresponding target values, \( X \) is the original feature space, and \( X' \) is the transformed feature space \cite{hollmann2024large}. 

\subsubsection{\revision{Feature Generation and Extraction}}
\revision{Generating and extracting representative features that capture relationships between inputs and outputs is central to ML \cite{zhang2021review,wahid2017statistical}. In ECF, raw measurements rarely expose all relevant structure directly; feature generation and Extraction, therefore, play a key role in presenting the model with informative signals (i.e., input features) that improve learning \cite{zhang2021review,sun2020review}. Because prior ECF studies describe this process using varied terminology and groupings, this thesis organises the most commonly used practices into four main groups, as follows:}

\paragraph{Calendar and Schedule Indicators}

\revision{Calendar and schedule features describe the operating context of the energy system through explicit time variables, enabling models to distinguish routine-driven consumption patterns from irregular fluctuations \cite{Fan_2017}. In ECF studies, this commonly includes timestamp-derived indicators (e.g., hour of day, weekday versus weekend, and month) \cite{Xu2019Probabilistic, kim2019Predicting, Wang2018Random}, and may be extended with context markers such as holidays, seasons, and time-of-use regimes (e.g., on-peak, off-peak) represented as binary variables \cite{Gao2021, Liu2023, Kim2019Recurrent}. These indicators are particularly valuable across heterogeneous energy systems because they encode operational schedules into predictive features that are robust to site-specific differences.}

\paragraph{Smooth Cyclic Time Encodings}

\revision{Smooth cyclic encodings represent periodic structure in energy consumption patterns in a continuous manner, avoiding artificial discontinuities that arise when cyclic variables are treated as ordinary integers (e.g., the boundary between the end and start of a day) \cite{Moon_2019}. A widely used approach is to apply sine and cosine transformations \cite{Verdonck2021} to cyclical time indicators (e..g, hour of the day), which preserve neighbourhood relationships on the cycle and capture daily regularities more faithfully. This improves the inductive bias of many ML models by making periodicity easier to learn, rather than forcing the model to infer wrap-around behaviour from data alone. As a result, cyclic encodings can improve predictive performance in ECF \cite{Deb2016}.}

\paragraph{Historical Dependence and Rolling-windows}

\revision{Historical dependence features capture temporal structure by extracting predictors from past observations of the target series, reflecting the fact that energy consumption often exhibits short-term inertia as well as longer-term trends \cite{MartnezComesaa2020}. Prior studies construct time-shifted predictors using varying history lengths, ranging from very recent windows (e.g., the most recent few hours) \cite{Deb2016} to longer horizons (e.g., up to the past 24 hours) \cite{Fan_2017}, depending on the forecasting setting and the dynamics of the underlying energy system. Complementing shifted histories, statistical summaries computed over recent history (e.g., rolling means and other windowed aggregates) are commonly used to represent local level and variability \cite{Gao2021}. Together, lagged values and rolling summaries expose autocorrelation and local regime changes to the ML model, supporting more accurate forecasts \cite{zhang2021review,sun2020review}.}


\begin{table}
\centering
\caption{\revision{Summary of different feature selection methods used in related work.}}
\fontsize{9pt}{11pt}\selectfont
\renewcommand{\arraystretch}{1.1}

{\color{black}
\begin{tabular}{
>{\centering\arraybackslash}p{1.8cm}
>{\centering\arraybackslash}p{3cm}
>{\justifying\arraybackslash}p{7cm}}
\hline
\textbf{Method} & \textbf{Specific technique} & \textbf{Description} \\
\hline

Filter & Filter-driven with SVR kernels &
\noindent Best features are chosen based on acquisition feasibility and performance scores using filter methods. These features are tested on datasets using support vector regression with radial basis and polynomial kernels \cite{Zhao2012Feature}. \\
\cline{1-3}

Embedded & Recursive feature elimination &
\noindent This backward selection technique involves training a model with all variables, ranking them for importance, and iteratively removing the least important ones until no further reduction is possible \cite{Fan2014}. \\
\cline{1-3}

Wrapper & Genetic search &
\noindent This method explores feature subsets using a genetic search strategy, focusing on predictive ability and minimising redundancy among features \cite{Massana2015}. \\
\cline{1-3}

Embedded & Data permutation-based importance &
\noindent Optimal features are identified by assessing their role in enhancing predictive performance, with a focus on the impact of introducing irrelevant or noisy information \cite{Wang2018}. \\
\cline{1-3}

Hybrid & Two-stage filter-wrapper approach &
\noindent Starts with a filter method to remove irrelevant features, reducing dimensionality without compromising accuracy. Then, a wrapper method conducts an exhaustive search for the most effective feature set, balancing accuracy and simplicity \cite{Zhang2019}. \\
\hline
\end{tabular}
}

\label{tab:FS_methods}
\end{table}


\subsubsection{Feature Selection}
\label{lit_review_FS}

After generating and extracting features, additional preprocessing techniques such as FS can be applied where appropriate, to identify the most relevant variable inputs for developing an ML model \cite{guyon2003introduction}. While this is not compulsory,  this preprocessing technique can be useful for ECF \cite{Jurado2015}. Neglecting to filter inputs can result in larger datasets and slower training speeds, potentially degrading model performance \cite{zhang2021review,sun2020review}. For ECF, FS is predominantly conducted manually, guided by domain knowledge as demonstrated in prior research \cite{Bagnasco2015,Li2018,Ding2020comparative}. Nevertheless, some studies, as presented in \autoref{tab:FS_methods}, have incorporated different FS methods such as \cite{Zhao2012Feature,Fan2014,Massana2015,Wang2018,Zhang2019}. These FS methods, encompassing filters, wrappers, and embedded techniques \cite{OluAjayi2023}. In brief, the filter technique, rooted in statistical procedures, assigns a value to each feature and ranks them, determining whether they should be retained or discarded; the wrap-per approach evaluates the predictive capabilities of models by assessing various subsets of potential features; and the embedded technique integrates FS directly into ML algorithms. Such studies indicate that the effectiveness of FS is somewhat context-dependent (e.g., data characteristics and model type), emphasising that no single method is universally superior. 

Despite these advantages, FE remains a time-consuming process that is prone to human error while relying heavily on domain expertise and iterative experimentation \cite{Wu2022}. It often requires both energy domain expertise and data science skills \cite{Wang_2022, Mumuni_2024}. Energy domain experts identify the key factors influencing energy consumption, while data scientists validate these insights through data analysis and extract the relevant features from the raw data. Consequently, there is a pressing need to automate FE in ECF problems to streamline the process, reduce human bias, and enhance model performance without requiring extensive domain-specific knowledge for new data sets. Deep learning algorithms \cite{Fan2019Deep, Dong2021}, while capable of automatically learning useful representations from raw data, lack interpretability and typically require large datasets to perform well, a condition that is often infeasible in real-world scenarios \cite{grinsztajn2022tree, wang2019review}. This has led to a growing interest in AFE methods \cite{hollmann2024large}.

\subsection{\revision{Automated Feature Engineering}}
\label{ch2_Automated_Feature_Engineering}

AFE encompasses a range of approaches aimed at reducing manual intervention in the feature generation and selection process by leveraging algorithmic solutions \cite{Wang2023}. The concept of \textit{automated} FE, as proposed by \cite{Mumuni_2024}, involve two key steps: first, generating a comprehensive search space of possible feature processing operations, and second, employing optimisation techniques to identify the most effective FE strategy. These operations, which include transformations such as aggregation functions and arithmetic operations, can be combined to create a set of new informative features. Although the literature lacks explicit and comprehensive categorisations of AFE methods relevant to ECF problems, we may broadly classify the current landscape into the following categories:

\begin{itemize}[leftmargin=15pt, itemsep=-2pt]

\item {Traditional Statistical Methods}: These methods apply statistical and mathematical transformations to create new features through dimensionality reduction or statistical computations. For example, Principal Component Analysis (PCA) \cite{Li_2015} and wavelet decomposition \cite{Peng_2022} are widely employed in ECF to improve performance and automate FE. However, these methods present significant challenges when applied to ECF problems. While PCA effectively reduces data dimensionality, it often obscures the intuitive relationship between the transformed principal components and the original data features, complicating interpretation. Similarly, wavelet decomposition generates coefficients that are more complex and less interpretable than the raw data, requiring specialised domain knowledge for proper interpretation. These interpretability challenges are particularly critical in ECF, where understanding the factors influencing energy use is essential for developing effective policies and solutions.

\item {Learning-Based Methods}: These methods primarily utilise artificial neural networks to learn and automatically generate new features from raw data. A prominent example is Autoencoders, which have been studied in ECF \cite{Fan2019Deep, Yu2022, Moon2024}. However, Autoencoders require large datasets and produce abstract features with unclear ties to the original data, hindering explainability. While post hoc methods such as LIME \cite{Ribeiro2016}, which is limited to local interpretability, and SHAP \cite{SHAP_paper} offer insights, their reliability is debated \cite{Rudin2019, Slack2020}. LIME’s explanations, for example, depend heavily on parameter choices and may exclude important features \cite{Garreau2020}, further emphasising the need for inherently interpretable features for ECF.

\item {Hybrid Semi-Automated Methods}: The CAAFE method \cite{hollmann2024large} is an example of a context-aware semi-AFE approach designed for tabular data. This method integrates human expertise (i.e., domain knowledge) with large language models (LLMs) to streamline the FE process. While some steps in this method are automated, human intervention is still necessary to guide the process and make critical decisions. Such methods face two key challenges: computational limitations when handling datasets with a large number of attributes, and the potential for LLM hallucinations during feature generation.

\item {Heuristic-Based Methods}: These search-based algorithms rely on predefined rules to generate new features. For example, TS \cite{Christ2018}, a widely used AFE method in the literature, applies various time series characterisation techniques to compute features without the need for manual intervention. This algorithm characterises time series data in terms of data point distribution, correlation properties, stationarity, entropy, and nonlinear time series analysis. However, this method can be computationally expensive and risk overfitting due to the large number of generated features \cite{Yang2020, Moghadam2023}. Another widely recognised search-based FE method is FT, which was developed based on the Deep Feature Synthesis algorithm \cite{kanter2015deep}. This algorithm applies a series of transformation functions to create new features. These functions include mathematical operations such as summing, averaging, and counting. Although it has shown promising results in domains such as education and e-commerce, particularly in predicting student dropout, project excitement, and repeat buyer behaviour, its application in domains such as ECF remains unexplored. 

\end{itemize}

\revision{In addition to the categories above, it is worth acknowledging that evolutionary feature synthesis approaches (e.g., genetic programming) can construct new features by searching over compositions of primitive operators (e.g., arithmetic and non-linear transforms) guided by a predictive objective \cite{Guo2005_Genetic_Programming}. However, they are typically computationally expensive and can overfit without careful regularisation and validation \cite{LaCava2020_Genetic_Programming_limitations, Gulati2025_Genetic_Programming_limitations}.} Nevertheless, the literature lacks clarity regarding what \textit{automated} truly means in the context of FE in supervised learning. That said, the work presented in \autoref{ch:mainchapter5} focuses on heuristic-based methods as benchmarks for comparison with an improved FE algorithm for the following four key reasons: (a) they represent established FE techniques directly applicable to ECF problems without additional training or human intervention, (b) the proposed method aligns with the heuristic-based approaches by employing rule-based criteria for FE, (c) these methods generally maintain traceable FE processes supporting interpretable energy consumption analysis and (d) they are potentially more readily integrated into AutoML frameworks, thus enabling more practical automated pipelines.

\section{\revision{Machine Learning Algorithms for Energy Consumption Forecasting}}
\label{ch2:Paper1_S4}

Different ML algorithms have been used for ECF. Traditional ML algorithms such as K-Nearest Neighbours (KNN) have been used for buildings’ ECF \cite{Hong2022,Wang2020}. Given their capabilities in addressing complex forecasting problems, there is a notable increasing trend in using neural network-based algorithms for ECF, such as using Feedforward neural networks (FFNN) for daily ECF of institutional buildings \cite{Deb2016}; Recurrent Neural Networks (RNN) for predicting 24-hour sequences of electric load \cite{Rahman2018}; a Multilayer Perceptron (MLP) for 1-hour-ahead prediction of office building \cite{Ding2018}; Long Short-Term Memory (LSTM) for office building ECF \cite{Gao2021}. Hybrid methods have also been studied, where models combine the feature extraction capabilities of Convolutional Neural Networks (CNNs) with the sequence modelling capabilities of LSTMs for predicting power consumption in large distribution complexes \cite{kim2019Predicting} and CNN-Gated Recurrent Units (CNN-GRU) for predicting hourly energy usage in educational buildings \cite{Cao2023}. 

Additionally, ensemble-based methods such as Extreme Gradient Boosting (XGB) and Random Forest (RF) have shown promising results when utilised for ECF \cite{Wang2020,Wang2018Random,li2021xgboost,Bassi2021,Seyedzadeh2019}. In contrast to these single-task methods, Multi-Task Learning (MTL) \cite{Caruana1997} leverages existing ML algorithms to simultaneously model multiple input-output relationships and task interdependencies, capitalising on these to improve prediction accuracy. For example, MTL was used to predict a building's electrical load and outdoor temperature simultaneously, leveraging outdoor temperature forecasting as a secondary task and employing a hyperparameter $c$, to balance the auxiliary task's weight \cite{Liu2023}. Utilising MTL in combination with a Temporal Convolutional Network (TCN) for short-term multi-energy load predictions has shown promising results \cite{Wang2022Multi}, and reduced training times were observed when MTL was combined with a Support Vector Machine (SVM) for similar predictions \cite{Tan2020}. The summary of previous studies, detailed in \autoref{tab:related_work}, shows algorithm selection is context-dependent, influenced by dataset challenges, computational efficiency, model interpretability, and potentially by the researchers' preference for methods with which they are most acquainted.

\section{Automated Machine Learning}
\label{ch2:Paper2}

Recently, AutoML frameworks \cite{hutter2019automated} such as H2O \cite{h2o}, Tree-Based Pipeline Optimisation Tool (TPOT) \cite{RN575}, Amazon AutoGluon \cite{AutoGluon2020}, and Microsoft FLAML \cite{wang2021flaml} have emerged as promising methods designed to streamline the ML pipeline, \revision{making them relevant approaches for addressing the time and expertise-intensive nature of ECF.} In brief, AutoML is designed to automate the process of applying ML methods, minimising the need for manual intervention to enable individuals to leverage such techniques without necessitating prior technical knowledge or domain-specific understanding \cite{hutter2019automated}. 

While these methods excel in automating \revision{\textit{some} preprocessing steps within the ML pipeline (e.g., handling missing values), as well as model selection and hyperparameter tuning}, they still lack the capacity to generate new, useful, and interpretable features, which is crucial for improving predictions in complex environments such as those observed in energy consumption data. In other words, these frameworks often assume that \revision{dataset-specific FE has already been performed and that the resulting feature set is ready for training}. As a result, tasks such as FE and the integration of domain knowledge are largely left to human practitioners \cite{hollmann2024large}.

Furthermore, many general-purpose AutoML systems are proposed for broad applicability across ML tasks, but their focus on generalisation often limits their effectiveness in specialised domains \cite{He2021}. Such limitations become particularly evident in domains that require interpretable features, such as ECF of time series data, where understanding the factors influencing energy consumption is important for well-informed decisions. This challenge may be attributed to the complex nature of power usage patterns, which can involve various linear and nonlinear relationships, fluctuating behaviours, and potential dependencies on temporal and environmental factors \cite{zhang2021review, manandhar2023current}. Therefore, domain-relevant FE could be a beneficial approach for AutoML to excel in modelling ECF problems. 

\revision{
It is worth emphasising that throughout this thesis, AutoML \cite{hutter2019automated} and AFE are treated as complementary rather than interchangeable. AutoML, as explained in \autoref{ch2:Paper2}, is primarily used to automate model selection and hyperparameter tuning once a learning-ready feature set is available, whereas AFE, as described in \autoref{ch2_Automated_Feature_Engineering}, targets the upstream bottleneck of constructing informative and interpretable features from raw data. This distinction matters because many AutoML systems still assume that feature generation has already been performed by a practitioner \cite{hollmann2024large}, meaning that \textit{automation} can be overstated if FE remains manual.
}

\section{Decision-Focused Learning}
\label{ch2:Paper4}
\revision{
\subsection{Concept and Rationale}}
Decision-making under uncertainty is common in real-world applications where unknown parameters significantly complicate the process \cite{Ibrahim2020, Reza2023}. For example, in residential energy systems with BESS, operators must make critical decisions about when to charge or discharge batteries to exploit time-varying tariffs (i.e., minimising electricity costs), and how much energy to store or release, while respecting physical and operational constraints of BESS \cite{Yang2022, Yu2023}. These decisions are made under uncertainty in both future electricity prices and household demand. In the literature, addressing these challenges typically involves a two-stage process: ML models forecast unknown variables, and then use these predictions as input parameters for Constrained Optimisation (CO) to determine optimal decisions within set boundaries \cite{Bergmeir2025}. This traditional sequential approach, also known as PTO, handles prediction and optimisation in isolation \cite{Vanderschueren2022}. The limitations of this approach manifest in two distinct ways: a) cascading errors arise from the sequential structure of PTO, where inaccuracies in the prediction stage may propagate and amplify through the optimisation stage, leading to suboptimal decisions \cite{Wilder2019, mandi2022_Through_the_lens}; and b) PTO suffers from objective misalignment in the utility of information extraction, as it focuses on deriving features and patterns from data solely to minimise prediction errors, without prioritising the information most relevant or useful for the downstream decision task \cite{donti2017task, Boettiger2022}.

To overcome this challenge, an emerging approach known as DFL integrates predictive modelling and optimisation directly into the learning process \cite{Wilder2019}. In this approach, forecasts are selected or assessed based on their impact on the actual downstream cost of the optimisation problem, rather than on standard error metrics (i.e., error-based loss function) such as Mean Squared Error (MSE). In order to achieve this, a task-aware loss function, such as regret, can be used \cite{Mandi2020}. By incorporating regret-based loss functions, which measure the difference between realised outcomes under uncertainty and optimal outcomes under perfect foresight, DFL methods have the potential to enhance decision quality by aligning training with the end-use task and prioritising it over minimising forecasting errors \cite{AnisLahoud2025}. \revision{Yet, DFL approaches remain largely untested in practical settings, with most evaluations conducted on artificial datasets and simplified scenarios \cite{kotary2021end, Mandi2024, geng2024benchmarking}. To establish their real-world effectiveness, DFL methods must be validated beyond basic synthetic benchmarks \cite{Mandi2020, Zhou2024} using actual operational data and realistic constraints.

\revision{Integrated learning methods highlight a fundamental shift in how forecast performance is evaluated for decision-making. In some cases, models trained with decision-aware losses intentionally sacrifice accuracy on traditional metrics to improve the downstream objective. For instance,\cite{Zhang2024_Value_Oriented_Prediction} found that DFL-trained policy (i.e., agent) yielded the lowest average system cost in a power system simulation, even though its forecast error was higher than that of standard predictors. This counterintuitive result underscores the idea that the “best” forecast in isolation may not necessarily yield the best overall decisions, particularly when downstream tasks are not accounted for in the learning process, a phenomenon known as the forecast trap \cite{Boettiger2022}.}

Recent DFL applications have used extensive datasets spanning multiple years \cite{Bergmeir2025, Paredes2025, Wang2025_AI_Optimized, Sang2022_6_years}, yet many practical implementations face data scarcity due to collection constraints, privacy limitations, or resource restrictions \cite{grinsztajn2022tree, hollmann2022tabpfn, Alkhulaifi2024_pipeline}. This creates a critical research gap where DFL performance remains largely unvalidated under constrained, small-scale real-world conditions typical of operational energy systems. FE can maximise information extraction from limited datasets \cite{Wang_2022}, but FE remains a manual, expert-dependent task \cite{Wang_2022, Wu2022}. While Automated FE (AFE) shows promise for improving forecasting metrics \cite{hollmann2024large, Alkhulaifi2025}, its effectiveness for enhancing DFL in data-constrained environments remains unexplored.

These gaps in DFL validation under data-constrained real-world conditions and the unexplored potential of AFE for enhancing DFL while reducing reliance on domain expertise motivated this work. First, we propose a decision-aware, end-to-end ML framework that jointly forecasts electricity demand and prices while optimising BESS operations using regret-based objectives, specifically designed for small dataset sizes. Second, we enhance the nascent DFL approach by integrating domain-specific AFE to extract richer data representations without requiring extensive domain expertise, thereby streamlining DFL pipeline development for energy applications. Third, we validate this framework using novel real-world data from a UK-based property, providing a comprehensive comparative analysis between traditional PTO and DFL approaches to demonstrate the practical viability of DFL methods in operational BESS systems.
}

\revision{
\subsection{Key Decision-Focused Learning Methods}}
\label{ch2:Paper4_dfl_methods} 
In gradient-based DFL literature, Smart ``Predict, then Optimise'' (SPO$^{+}$) \cite{Elmachtoub2022} and Differentiable Black-Box (DBB) \cite{Pogančić2020Differentiation} are two seminal methods. In brief, the SPO$^{+}$ method employs a convex surrogate loss function, derived via duality theory, that upper bounds the SPO loss, which measures the decision error (suboptimality gap) induced by predicted cost vectors in a linear, convex, or integer optimisation problem, enabling efficient gradient-based training tailored to optimise decision quality. The DBB method implements an efficient backward pass for blackbox combinatorial solvers with linear objective functions by constructing a continuous interpolation function, whose gradient is computed using a single solver call on perturbed inputs, enabling the integration of combinatorial algorithms into neural network architectures. These methods have been applied to various classical DFL problems, such as the travelling salesman and shortest path problems \cite{Pogančić2020Differentiation}. However, despite promising theoretical advances and growing interest in DFL research, these methods have predominantly been evaluated on synthetic benchmark problems, with a lack of real-world applications \cite{kotary2021end, Mandi2024, geng2024benchmarking}. Therefore, DFL methods need to be explored beyond small-scale synthetic problems (referred to as \textit{toy-level} problems in \cite{Mandi2020, Zhou2024}) to demonstrate their practical viability using real-world data and constraints.

As the integration of renewables accelerates, prediction and optimisation approaches have attracted growing attention. For instance, as part of the IEEE-CIS Technical Challenge \cite{Bergmeir2025}, participants used different methods to forecast 15-minute power demand and solar production for six buildings and six solar arrays, while the optimisation task was to generate a schedule for a set of activities that minimised electricity costs across the buildings. Building on this, \cite{abolghasemi2022predict} reported strong positive Pearson correlations (0.81-0.9) between forecasting accuracy and optimisation cost across overforecast and underforecast scenarios (i.e., perturbed). Yet, they found that this correlation is asymmetric, meaning unequal effects between overforecasting and underforecasting, and that forecast‑accuracy metrics may be sub‑optimal for minimising complex optimisation costs.

\revision{A growing number of studies have started applying DFL frameworks to broader energy scheduling problems, including grid-level dispatch, renewable trading, and microgrid operations. These works report that aligning predictive models with operational objectives can yield measurable performance gains even in complex power systems \cite{Zhang2025_Uncertainty}. For example, \cite{Stratigakos2022_trees} employed a prescriptive model tree to integrate forecasting and trading decisions for renewable energy, resulting in modest profit increases (approximately 3.8\% on average) in the French electricity market. \cite{DiasGarcia2025_Closed_Loop} demonstrated in large-scale grid scheduling cases (up to a 3000-bus system) that decision-focused training improved economic outcomes by approximately 11–13\% compared to traditional forecasts. In an IEEE 39-bus test system, \cite{Lu2022_Economic_Dispatch} reported that a task-tailored learning approach reduced the additional costs caused by forecast errors by about 5.5\%, while dramatically accelerating training convergence (by several orders of magnitude) relative to a standard indirect method. These studies underscore that even in multi-resource or network-level scheduling, optimising predictions for decision quality can translate to tangible cost savings or profit gains.}

Similarly, the need for aligning optimisation with forecasting is particularly important in domains such as BESS operations \cite{Yang2022, Yu2023}, where such problems serve as a testing ground for evaluating the interplay between forecasting accuracy and downstream decision quality. In such energy scheduling problems, where future energy demand and costs are unknown, prediction errors can significantly affect downstream decisions \cite{Mandi2020Interior}, highlighting the limitations of traditional PTO methods and underscoring the potential of DFL to yield more practically valuable outcomes. In various contexts, literature on BESSs has studied PTO methods \cite{Hannan2021, Song2024}, while only recently have a few studies investigated DFL-based approaches for reserve-market participation \cite{Paredes2025} and for developing bidding strategies for microgrids in day-ahead electricity markets \cite{Alrasheedi2024}. \revision{\cite{Sang2022_6_years} applied a decision-focused strategy to price forecasting for battery arbitrage in an energy storage system, and achieved nearly 47\% higher profits compared to a conventional error-minimising predictor, while the regret (suboptimality) of decisions dropped by over 90\%. These domain-specific applications confirm that DFL can materially enhance the economic performance of storage scheduling and bidding tasks by directly tying predictions to operational outcomes, rather than purely to forecast accuracy.}

Nonetheless, several critical limitations persist that challenge effective implementation in practical energy systems. Traditional PTO approaches suffer from cascading errors arising from their sequential structure, where prediction inaccuracies propagate and amplify through the optimisation stage, and from objective misalignment, as they focus on minimising prediction errors rather than optimising information most relevant for downstream decision tasks \cite{mandi2022_Through_the_lens, donti2017task, Boettiger2022}. 

\revision{Fundamentally, and within the work conducted in this thesis, the prediction and optimisation stages are not independent in terms of decision quality, because the optimiser uses the forecasts as inputs and therefore forecast errors directly perturb the objective and constraints that determine the chosen action \cite{Mandi2024}. In contrast, a conventional sequential PTO approach treats prediction and optimisation as independent \textit{during training}, since the forecaster is fitted to a statistical error metric without accounting for how its errors translate into downstream cost or constraint violations \cite{Wilder2019}. This decoupling can be reasonable when the optimiser is relatively insensitive to forecast errors; however, in our setting (BESS scheduling), small forecast shifts can change charging and discharging decisions (i.e., when to charge/discharge and by how much), amplifying prediction error into decision regret. DFL attempts to address this dependence by placing the predictor inside the optimisation loop, so the learning signal reflects the downstream decision objective (i.e., regret metric, see \autoref{Evaluation_metrics}) rather than forecast accuracy alone.}

Whilst there are recent attempts to use DFL in real-world settings with different dataset sizes (e.g., multi-year \cite{Bergmeir2025}, one year \cite{Paredes2025}, two years \cite{Wang2025_AI_Optimized}, and six years \cite{Sang2022_6_years}), these datasets are significantly larger than those found in many practical applications where data scarcity is common due to collection limitations, privacy concerns, or resource constraints \cite{grinsztajn2022tree, hollmann2022tabpfn, Alkhulaifi2024_pipeline}. This highlights a critical gap in evaluating DFL's performance under constrained, small-scale real-world conditions, such as the 55-day dataset used in this work. In such scenarios, FE can compensate for limited data by extracting informative features, thus maximising the utility of the available data as well as enhancing computational efficiency by eliminating noisy or irrelevant data \cite{Wang_2022}. However, manual FE for energy forecasting remains a time-consuming process that is prone to human error while relying heavily on domain expertise and iterative experimentation \cite{Wang_2022, Wu2022}. As a result, tasks such as FE and the integration of domain knowledge are largely left to human practitioners, which in turn has led to growing interest in AFE methods \cite{hollmann2024large}. While AFE traditionally focuses on improving energy forecasting error metrics \cite{Alkhulaifi2025}, it remains unclear whether such improvements yield better operational outcomes when integrated with DFL approaches.
\cleardoublepage

\chapterwithquote{"Knowledge is not power until it is applied" - Dale Carnegie}{Machine Learning Pipeline for Energy and Environmental Prediction in Cold Storage Facilities}{ch:mainchapter3}


This chapter addresses \textbf{RO1} by developing a comprehensive ML pipeline for ECF that establishes baseline performance while investigating the critical role of domain knowledge in FE. Specifically, the chapter (A) utilises two novel real-world electricity consumption and weather datasets to serve as case studies for ECF modelling, (B) identifies and investigates domain-specific features for ECF applications, (C) systematically evaluates different FS techniques, algorithm types, and hyperparameter tuning, and (D) analyses feature importance and dataset size implications on model performance. This comprehensive approach ultimately produces a practical pipeline that guides practitioners in efficiently implementing domain knowledge-based features for ECF applications and establishes the empirical foundation necessary for subsequent FE automation efforts.
\newpage
\revision{
\section{\revision{Introduction}}
}
With escalating energy demands and costs, the UK's food and beverage industry is increasingly motivated to enhance energy efficiency, driven by the need to reduce operational costs and meet consumer expectations for sustainability. Food and Drink Cold Storage (FDCS) rooms (\revision{as illustrated in \autoref{fig3}}), as a component of food systems, play a crucial role in the food supply chain by preserving a wide array of perishable goods, including dairy, meat, and fresh produce, and ensuring their safety by maintaining acceptable temperature and humidity levels. For example, the maintenance of temperature-sensitive beverages in cellar rooms demands a consistent temperature, and abrupt changes can either hasten their ageing process in warmer conditions or inhibit their development in cooler environments \cite{gaspar19}. Lower humidity can lead to dehydration, particularly of natural elements such as cork, resulting in air infiltration in bottles, while conversely, excessively high humidity can encourage mould development and damage bottle labels \cite{reynolds10}. Accurate forecasts of electricity consumption, indoor temperature and humidity (driven by cooling and humidity control systems) in FDCSs can enhance operations and scheduling (e.g. managing door opening frequency to minimise energy loss, aligning maintenance with lower demand periods, determining the optimal times for restocking) leading to reduced energy consumption and ensuring food product preservation and quality \cite{gaspar19, walther21}. The energy consumption in FDCS is influenced by weather, temporal, operator activity, and less measurable factors such as the type, size, quantity, and packaging of food \cite{hoang15, evans14}.

Current studies on ECF using ML methods have predominantly concentrated on various types of buildings, such as institutional and educational buildings \cite{Deb2016,Fan_2017,Xu2019Probabilistic,Cao2023}, commercial and residential buildings \cite{Rahman2018,Ding2018,kim2019Predicting,Liu2023}, office and governmental buildings \cite{Ding2017,Gao2021}, community buildings \cite{Hong2022,Li2022}, factory building \cite{Moon_2019}, and industrial distribution complexes \cite{Kim2019Recurrent}. However, there remains a notable lack of research specifically targeting ECF in FDCS. While there are parallels between predicting energy consumption in buildings and FDCS, FDCS presents unique challenges, particularly the stringent requirements for maintaining temperature and humidity levels to ensure food safety and quality. This work addresses this gap in the literature by trialling ML techniques, commonly used for building ECF, to determine a tailored pipeline to the specific characteristics of FDCS. It focuses on the prediction of electricity usage, indoor humidity, and indoor temperature within FDCS environments, whereas most previous studies on buildings have only focused on predicting energy consumption.

\begin{landscape}
\begin{figure}[!t]
\centering
\includegraphics[width=.9\linewidth]{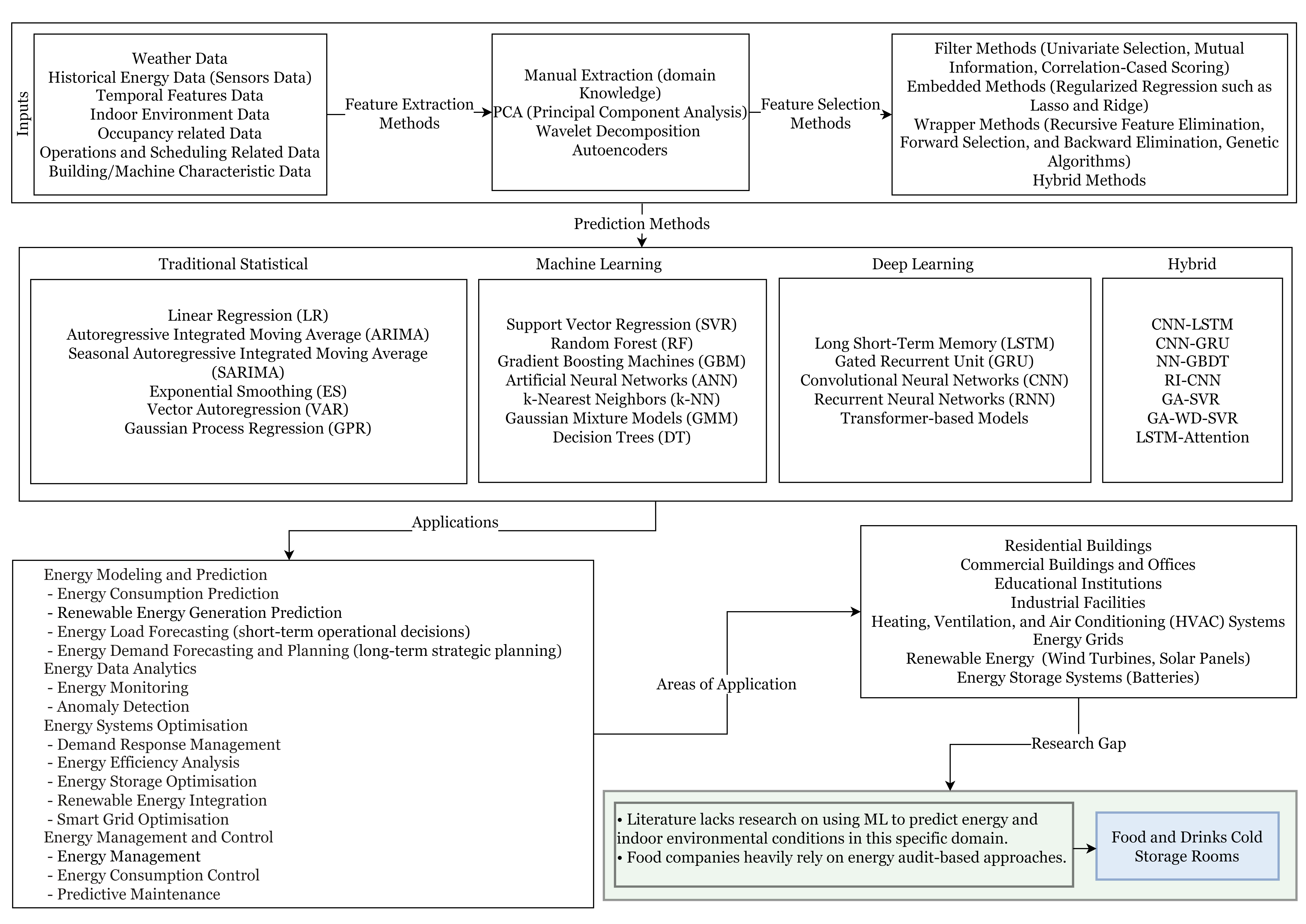}
\caption{Overview of literature on predicting energy consumption, including input features, FE, prediction methods, applications, areas of application, and identified research gaps for this work.}
\label{fig1}
\end{figure}
\end{landscape}


The novelty and contributions of this work lie in: 
\begin{itemize}
    \item Using ML methods for forecasting essential parameters (electricity consumption, indoor temperature, and humidity) in the unique context of FDCS. This specificity is critical because, although previous studies have explored forecasting energy using ML methods in various domains such as commercial and residential buildings, Heating, Ventilation, and Air Conditioning (HVAC) systems, smart grids, wind turbines, and solar panels, FDCS presents its own set of unique challenges, particularly their stringent requirements for maintaining temperature and humidity levels to ensure food safety and quality.
    
    \item Proposing a detailed pipeline for ML techniques in forecasting one week (hourly) into the future of electricity consumption, temperature, and humidity specific to FDCS environments and suitable for small dataset sizes. The proposed pipeline was validated using two newly collected datasets from different FDCS rooms located in Nottingham, UK. The inclusion of these datasets enables a comparative analysis, facilitating a more robust evaluation of ML methods in the specific context of FDCS.
    
    \item Emphasising multi-variable forecasting, in contrast to existing studies that often focus solely on energy consumption. This work underlines the importance of also forecasting indoor temperature and humidity, factors crucial for maintaining the quality and lifespan of stored items in FDCS.
    
    \item Investigating the often-overlooked aspect of FS methods. This involves examining the impact of eight different methods, including filter-based, embedded, wrapper-based, and hybrid methods, on different ML regression algorithms in FDCS settings, providing insights for future research in a similar context, even though such an investigation can be considered data-dependent.
    
    \item Trialling of different dataset sizes to demonstrate the impact of dataset volume on model accuracy in FDCS environments. Such environments often face limitations in data collection due to time, cost, or operational challenges. By evaluating model performance across varying dataset volumes, the research not only highlights the influence of dataset size on accuracy but also offers an estimation of the minimum dataset size needed for forecasting in FDCSs.
\end{itemize}


\section{Methodology}
\label{sec:Methodology}
\begin{landscape}
\begin{figure}[!t]
\centering
\includegraphics[width=.9\linewidth]{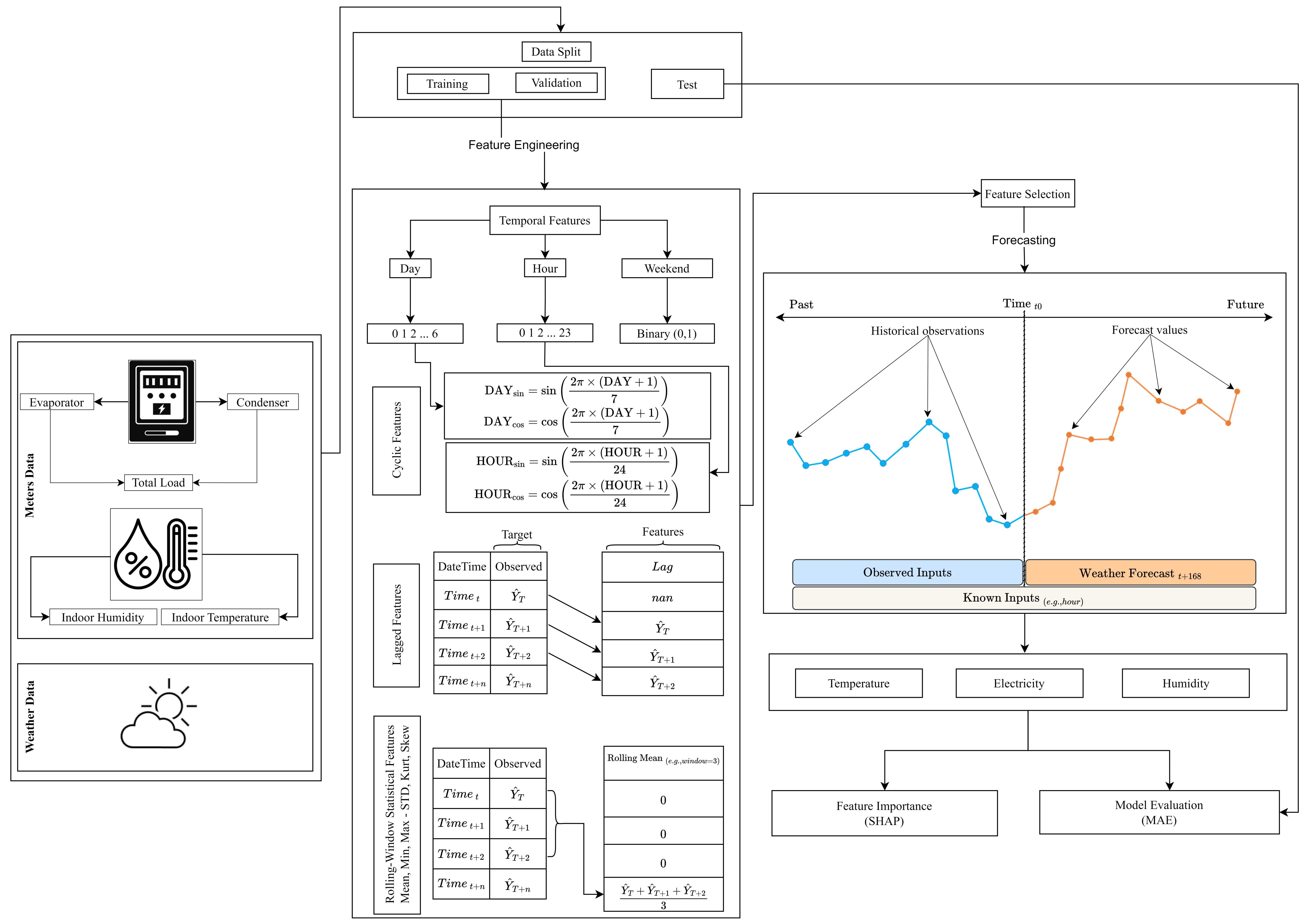}
\caption{Proposed ML pipeline for predicting electricity consumption, indoor temperature, and indoor humidity one week into the future in FDCS settings using weather data.}
\label{fig2}
\end{figure}
\end{landscape}

This section explains the methodology used in this work and the proposed pipeline to predict electricity consumption, temperature, and humidity one week (hourly) into the future, as shown in \autoref{fig2}. The methodology includes collecting electricity consumption data and indoor environment conditions through metering systems and weather. Different FE methods were examined, owing to their effectiveness in previous related works and to assess their influence within the specific context of FDCS. Various ML methods were explored, focusing on those that have shown superior performance in comparable settings, as explained in the preceding related work section.


\subsection{Data Collection}
\begin{landscape}
\begin{figure}[!t]
\centering
\includegraphics[width=.8\linewidth]{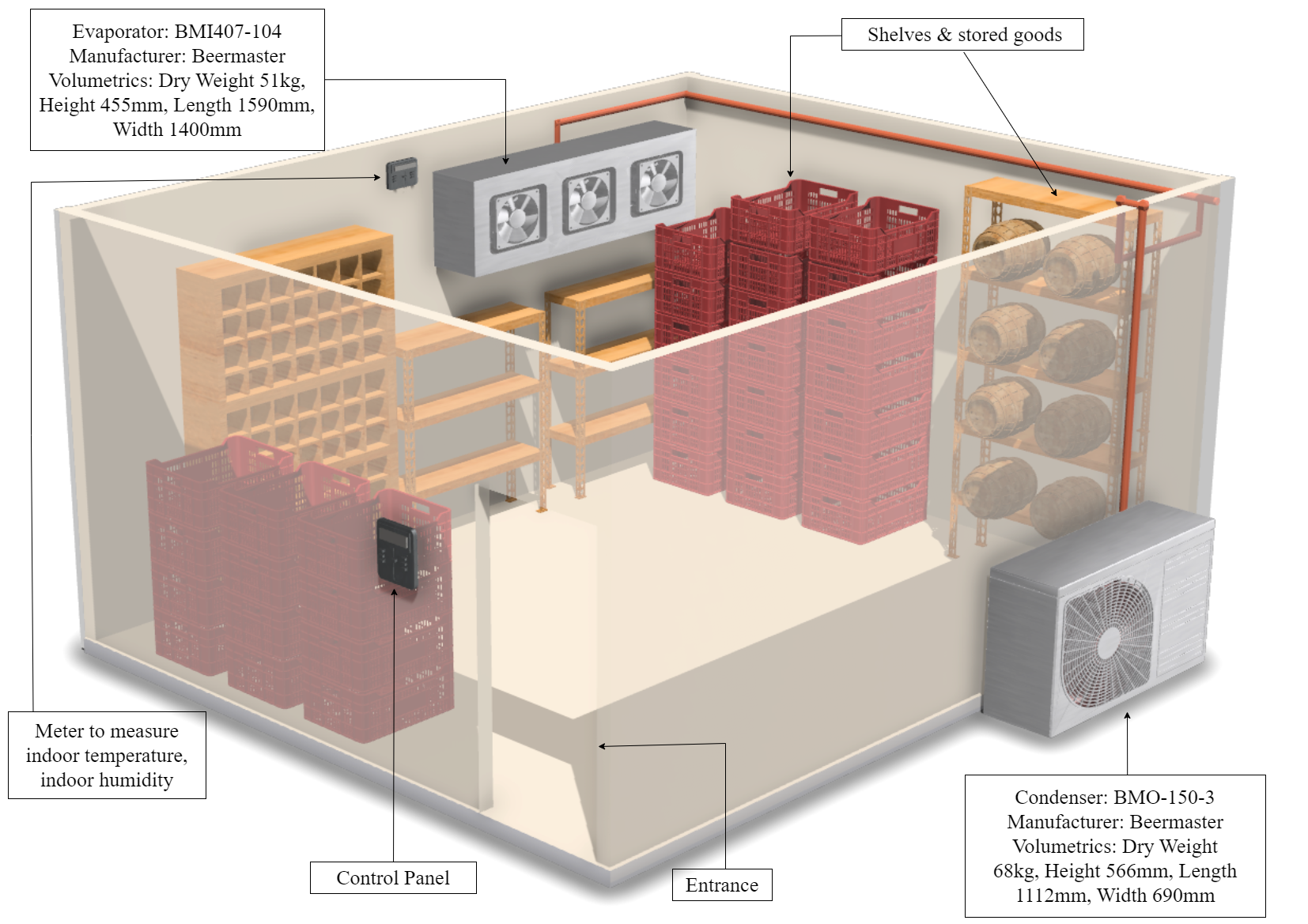}
\caption{3D schematic representation of the investigated FDCSs. Key components indicated include (i) evaporator unit, responsible for absorbing heat from the storage space and maintaining low temperatures; (ii) condenser unit, essential for releasing the absorbed heat outside the storage room and condensing the refrigerant back into a liquid; (iii) an integrated meter collecting indoor temperature and humidity values; and (iv) a control panel that facilitates monitoring and adjusting the storage condition.}
\label{fig3}
\end{figure}
\end{landscape}

\begin{figure}[!t]
\centering
\includegraphics[width=1\linewidth]{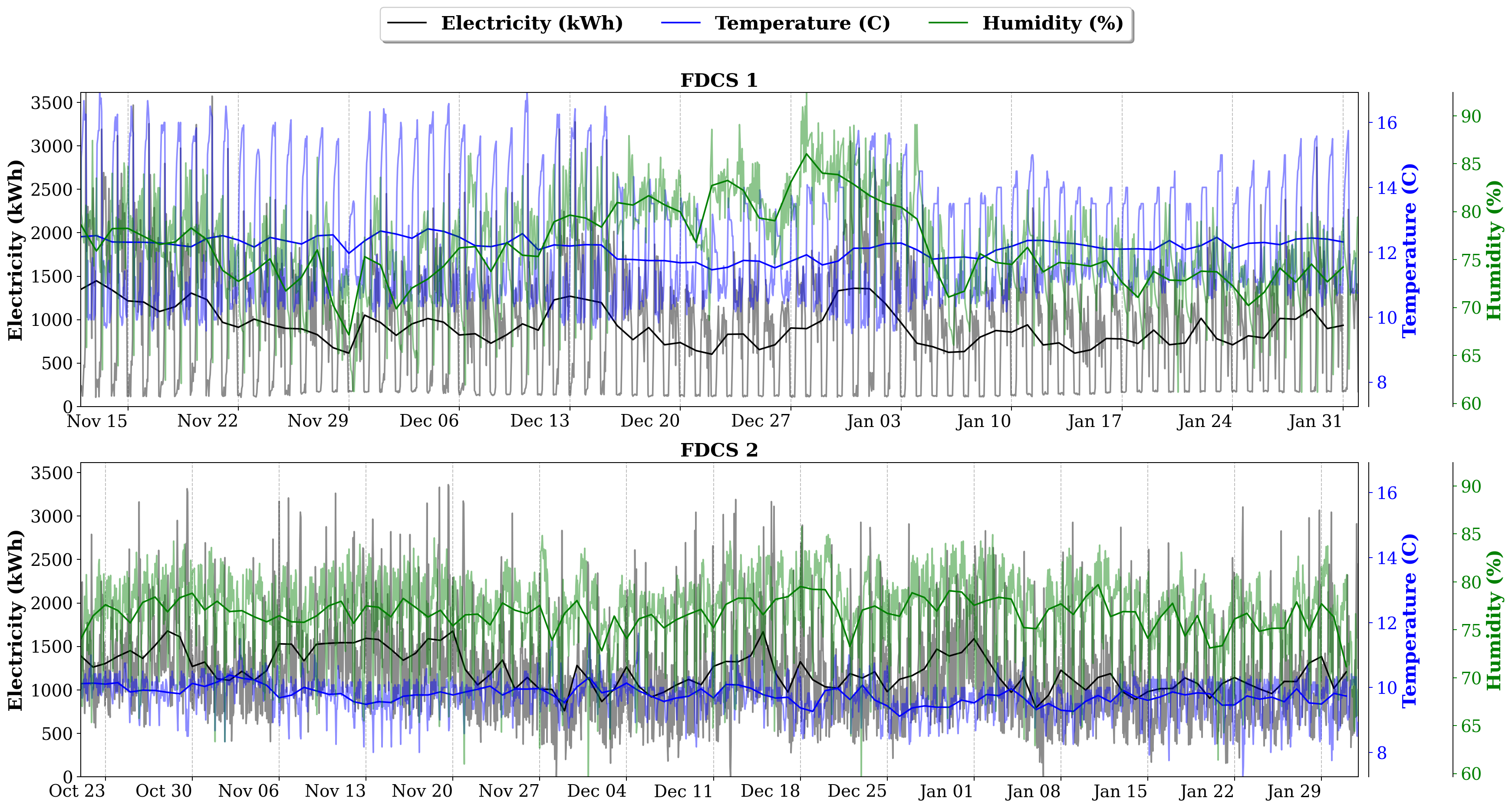}
\caption{Collected data from the two FDCS rooms plotted as a time series. Raw data points are represented by faint lines, while mean values are shown as solid lines. The mean is calculated by resampling the data into 24-hour blocks and taking the average value within each block, which smooths out short-term fluctuations and helps identify trends and anomalies.}
\label{fig4}
\end{figure}

\begin{figure}[!t]
\centering
\includegraphics[width=1\linewidth]{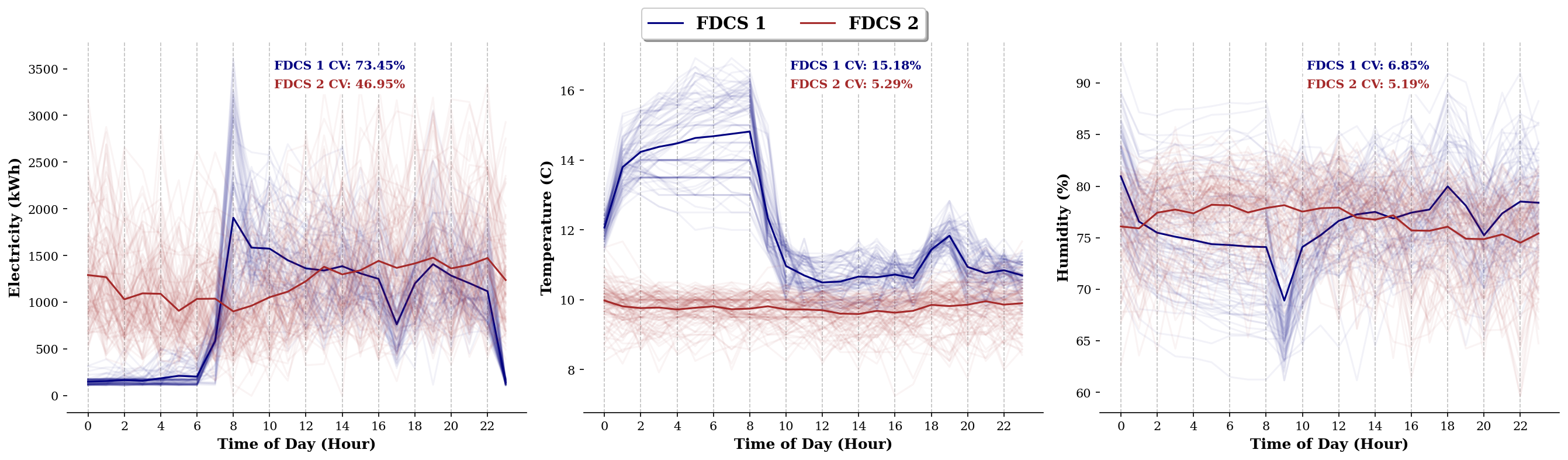}
\caption{Hourly electricity, temperature, and humidity patterns for the two FDCSs. Each hourly reading is shown with reduced opacity, while mean hourly values are depicted in darker lines for clarity. The Coefficient of Variation (CV) quantifies relative variability. FDCS 1 shows higher electricity, temperature, and humidity variability (higher CV) than FDCS 2, indicating greater sensitivity to internal and/or external factors in FDCS 1 versus a more regulated environment in FDCS 2.}
\label{fig5}
\end{figure}

In this work, electricity consumption data for the condenser and evaporator units, as well as internal ambient conditions (temperature and humidity), were collected from two FDCSs based in Nottingham, United Kingdom, as shown in \autoref{fig3} and \autoref{tab:Summarystats}. Electricity consumption data were collected at 4-second intervals for FDCS 1 from 12 November 2021 up to 31 January 2022, and for FDCS 2 from 21 October 2021 up to 31 January 2022. The internal temperature and internal humidity were recorded at 10-minute intervals within the storage room and close to the evaporator unit. \autoref{fig4} and \autoref{fig5} present the collected datasets of the three variables from the FDCS datasets over time, resampled to a 1-hour resolution. The hourly weather data utilised in this work was sourced from the NASA Langley Research Centre’s POWER Project, a repository of solar and weather data sets produced by NASA to support renewable energy and building energy efficiency research \cite{nasaNASAPOWER}. In this work, the exact locations of each FDCS location were identified for retrieving weather data using geographical coordinates (latitude and longitude). 

These observations are historical instead of forecasts to reduce uncertainty from forecast errors and better understand the impact of various input features.  The data were partitioned into three subsets: 70\% for training the ML models, 15\% for validation (to fine-tune hyperparameters and monitor performance), and 15\% for testing to evaluate the models' performance on previously unseen data. \revision{It is worth mentioning that the data were not shuffled, as energy consumption data exhibit temporal patterns; therefore, a chronological training, validation, and test split was used to preserve the temporal ordering of the time stamps.} A 5-fold cross-validation approach, based on trial and error, was also employed, and for the final model training, the training and validation sets were combined to maximise the data utilised. This approach supports the development of a robust ML model and aligns with the methodologies adopted in related studies \cite{Liu2023, Wang2022Multi, SendraArranz2020, Somu2020}.

\revision{
Note that evaluating on two real-world FDCS datasets, specifically collected for this work, is appropriate to satisfy \textbf{RO1} (see \autoref{intro:Aim_objectives}). First, it enables a detailed investigation of the role that domain knowledge plays in FE for ECF problems, and allows us to assess whether the conducted FE approach and the main patterns observed (e.g., which feature families/categories are repeatedly selected or emerge as important) are consistent across two distinct real-world datasets rather than being an artefact of a single system. Second, ECF does not have a single traditional evaluation setup, as prior related work (reviewed in \autoref{ch:lit_review} and summarised in \autoref{tab:related_work} varies substantially in available inputs (e.g., weather, occupancy, indoor variables), temporal granularity, forecasting horizons, and reported metrics, making direct cross-paper benchmarking unreliable. 

Accordingly, the two datasets provide a purposeful robustness assessment because they exhibit different operating regimes (see the consumption patterns in \autoref{fig5}). Nonetheless, it is worth acknowledging that two sites cannot span all possible ECF configurations; however, they provide an appropriate empirical basis for establishing a realistic and reproducible baseline ECF pipeline, with limitations discussed later in \autoref{con:Research_Limitations}. Finally, while this chapter uses two FDCS datasets to investigate domain knowledge in FE in depth, \autoref{ch:mainchapter5} extends the evaluation to eighteen real-world datasets spanning multiple energy systems (e.g., residential and commercial buildings, wind turbines, industrial settings, and grid power consumption) to provide a broader test of generality for automated FE.}

\begin{table}
    \centering
    \caption{Summary statistics of the collected data for electricity consumption, indoor humidity, temperature, and weather in both FDCSs.}
    \footnotesize
    \renewcommand{\arraystretch}{1.1}  
    \begin{tabular}{p{3.5cm}  p{.9cm} p{.9cm} p{.9cm} p{.9cm} p{.9cm} p{.9cm} p{.9cm} p{.9cm}}
        \toprule
        \multirow{2}{*}{\textbf{Variable}} & \multicolumn{7}{c}{\textbf{Statistics}} \\
        \cmidrule(l){2-8}
        & {Mean} & {Std Dev} & {Min} & {25th Perc} & {50th Perc} & {75th Perc} & {Max} \\
        \midrule
        \multicolumn{8}{l}{\textbf{FDCS 1}} \\
        Electricity Consumption (kWh) & 922.22 & 677.51 & 106.01 & 178.35 & 958.15 & 1385.56 & 3613.5 \\
        Humidity Closer to Evaporator (\%) & 76.21 & 5.22 & 61.17 & 72.67 & 76.25 & 79.75 & 92.42 \\
        Temperature Closer to Evaporator (\si{\celsius}) & 12.16 & 1.84 & 9.5 & 10.67 & 11.42 & 13.5 & 16.92 \\
        Outdoor Temperature (\si{\celsius}) & 4.62 & 3.44 & -3.30 & 2.04 & 4.44 & 6.80 & 13.76 \\
        Outdoor Dew Point (\si{\celsius}) & 3.96 & 3.34 & -4.15 & 1.55 & 3.48 & 6.39 & 12.33 \\
        Outdoor Wet Bulb Temperature (\si{\celsius}) & 4.29 & 3.35 & -3.69 & 1.87 & 3.94 & 6.48 & 12.87 \\
        Specific Humidity (g/kg) & 5.17 & 1.25 & 2.81 & 4.21 & 4.88 & 5.98 & 8.97 \\
        Outdoor Relative Humidity (\%) & 95.03 & 5.69 & 60.12 & 93.38 & 96.75 & 98.69 & 100.0 \\
        Precipitation (mm/hour) & 0.06 & 0.18 & 0.0 & 0.0 & 0.01 & 0.04 & 2.09 \\
        Surface Pressure (kPa) & 100.32 & 1.50 & 96.13 & 99.14 & 100.44 & 101.58 & 102.81 \\
        Wind Speed (m/s) & 5.14 & 2.61 & 0.28 & 3.13 & 4.76 & 6.70 & 14.88 \\
        Wind Direction (Degrees) & 238.70 & 74.11 & 0.25 & 212.86 & 249.14 & 292.39 & 359.66 \\
        \midrule
        \multicolumn{8}{l}{\textbf{FDCS 2}} \\
        Electricity Consumption (kWh) & 1218.32 & 572.11 & 0 & 781.78 & 1134.07 & 1537.62 & 3361.45 \\
        Humidity Closer to Evaporator (\%) & 76.71 & 3.98 & 59.67 & 74.65 & 77.33 & 79.5 & 85.75 \\
        Temperature Closer to Evaporator (\si{\celsius}) & 9.77 & 0.52 & 7.25 & 9.5 & 9.83 & 10.08 & 11.67 \\
        Outdoor Temperature (\si{\celsius}) & 5.53 & 3.78 & -3.23 & 2.56 & 5.44 & 8.26 & 16.47 \\
        Outdoor Dew Point (\si{\celsius}) & 4.71 & 4.71 & -4.15 & 2.02 & 4.81 & 7.28 & 12.99 \\
        Outdoor Wet Bulb Temperature (\si{\celsius}) & 5.12 & 3.60 & -3.61 & 2.28 & 5.12 & 7.83 & 14.68 \\
        Specific Humidity (g/kg) & 5.47 & 1.37 & 2.81 & 4.39 & 5.34 & 6.39 & 9.36 \\
        Outdoor Relative Humidity (\%) & 94.16 & 6.68 & 47.07 & 92.08 & 96.36 & 98.48 & 100.0 \\
        Precipitation (mm/hour) & 0.07 & 0.22 & 0.0 & 0.0 & 0.01 & 0.04 & 4.30 \\
        Surface Pressure (kPa) & 100.19 & 1.45 & 96.14 & 99.11 & 100.30 & 101.25 & 102.81 \\
        Wind Speed (m/s) & 5.35 & 2.64 & 0.38 & 3.29 & 4.99 & 7.02 & 14.55 \\
        Wind Direction (Degrees) & 238.78 & 68.62 & 0.32 & 209.72 & 245.02 & 286.44 & 357.56 \\
        \bottomrule
    \end{tabular}
    \label{tab:Summarystats}
\end{table}

\begin{table}
\centering
\caption{Summary of input features used in this work.}
\footnotesize
\renewcommand{\arraystretch}{1.1}  
\begin{tabular}{>{\raggedright\arraybackslash}p{2cm}p{1.5cm}p{8.5cm}}
\toprule
\textbf{Features} & \textbf{Type} & \textbf{Description} \\
\midrule
Weather data & Continuous & Outdoor Temperature (\si{\celsius}), Outdoor Dew Point (\si{\celsius}), Outdoor Wet Bulb Temperature (\si{\celsius}), Specific Humidity (\si{g/kg}), Outdoor Relative Humidity (\%), Precipitation (mm/hour), Surface Pressure (kPa), Wind Speed (\si{m/s}), Wind Direction (Degrees). \\
\addlinespace
Temporal features & Integer Value & Hour of the day (0-23), day of the week (0-6), day of the month (1-31), the month of the year (1-12), weekday vs weekend (0-1) \\
\addlinespace
Operation hours & Binary & The 'Is\_Open' feature serves as a binary indicator: 1 indicates that the storage is operating within working hours, while 0 indicates otherwise. \\
\addlinespace
Cyclical features & Continuous & Sine and cosine transformation of temporal features (hour of day and day of week) \\
\addlinespace
Time-lag features & Continuous & Previous values of the given target (e.g., electricity consumption from preceding hours/days) \\
\addlinespace
Rolling-window statistical features & Continuous & Summary statistics (maximum, minimum, mean, kurtosis, skewness, and standard deviation) of a given target computed over a fixed-size window \\
\bottomrule
\end{tabular}
\label{tab:inputs}
\end{table}


\subsection{Feature Extraction}
\label{ch3_Feature Extraction}
Given that no previous work has identified the optimal features to extract in the context of FDCSs, and considering their relevance and potential impact, this work uses a comprehensive set of extracted features. These include indicator variables accounting for categorical events such as the hour of the day and weekends, cyclical features that capture temporal patterns, time-lag features, and rolling-window statistical features. The following subsections describe the feature extraction process used in this work, and \autoref{tab:inputs} presents a summary of all the input variables employed.

\subsubsection{Indicator Variables and Operation Hours}
Indicator variables and temporal features are crucial in time series analysis to account for the impact of categorical events, such as weekdays and weekends \cite{sun2020review, Khalil2022}. For example, in predicting the electricity consumption of FDCS, these encoded indicator variables enhance the model's ability to capture the effects of these events on energy use. A 'weekend' variable, for instance, can indicate a potential increase or decrease in demand for FDCS systems on weekends, which may lead to lower or raised electricity consumption. Additionally, variables that represent the daily operation hours (working hours) of the two FDCSs have been employed. Such integration could be important as different operation hours can significantly influence the patterns of electricity consumption and other metrics being predicted.

\subsubsection{Cyclical Features}
Although time-based features such as the hour of day and day of the week provide temporal information, these features may not always be effective in representing time-based patterns. To overcome this, sine and cosine transformations can be applied to these features to capture the temporal patterns in the data \cite{Moon_2019}. This enhances the model's ability to capture the cyclical and periodic patterns in the data, leading to improved prediction accuracy. Equations \eqref{eq:daysin} and \eqref{eq:daycos} calculate the sinusoidal and cosinusoidal transformation of the day of the week respectively, allowing for cyclical pattern representation. Similarly, Equations \eqref{eq:hoursin} and \eqref{eq:hourcos} perform the sinusoidal and cosinusoidal transformation of the hour of the day respectively. By adjusting the input by +1 and normalising by the period (7 for days and 24 for hours), these transformations capture cyclical temporal patterns in data.

\begin{equation}
    DAYsin = \sin\left(\frac{2\pi \times (DAY + 1)}{7}\right)
    \label{eq:daysin}
\end{equation}

\begin{equation}
    DAYcos = \cos\left(\frac{2\pi \times (DAY + 1)}{7}\right)
    \label{eq:daycos}
\end{equation}

\begin{equation}
    HOURsin = \sin\left(\frac{2\pi \times (HOUR + 1)}{24}\right)
    \label{eq:hoursin}
\end{equation}

\begin{equation}
    HOURcos = \cos\left(\frac{2\pi \times (HOUR + 1)}{24}\right)
    \label{eq:hourcos}
\end{equation}

\subsubsection{Time-Lag Features}
\begin{figure}[!t]
\centering
\includegraphics[width=1\linewidth]{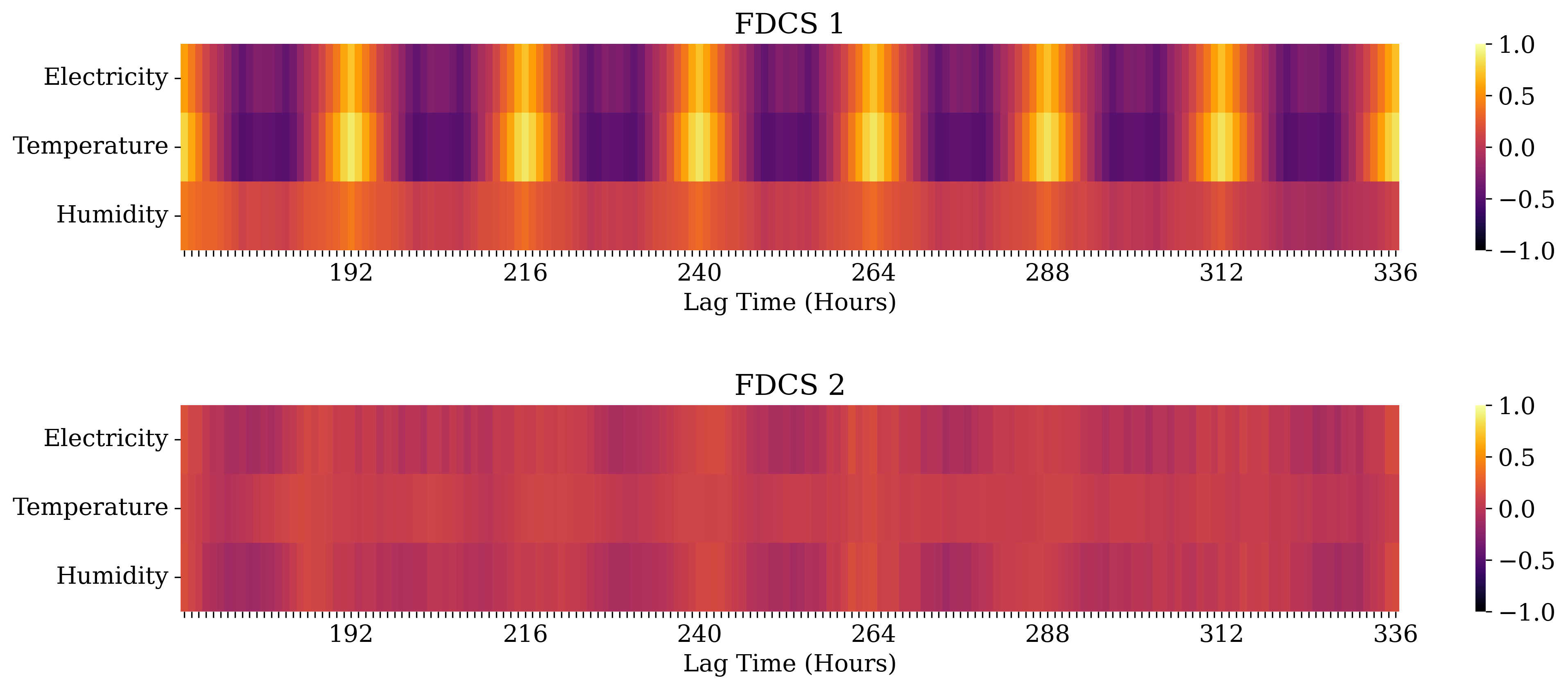}
\caption{Correlation heatmaps of target variables against their time lags over one week for FDCS 1 and 2. The colour scale represents the Pearson correlation coefficients}
\label{fig6}
\end{figure}

Time-lag features in time series data are values from previous time points. They are created by shifting the target variable back in time $t$ by a certain number of steps $k$. The primary purpose of these features is to capture the relationship between the current value of the target variable and its past values. For example, in an FDCS system, the energy consumption at time $t$ might be influenced by consumption levels at $t - k$ reflecting the inertia of cooling systems. Similarly, the indoor temperature or humidity at a given moment could be a result of conditions from previous hours. By incorporating such features, models can better account for historical influences, capture temporal dependencies and trends, and ultimately improve their predictive accuracy. In this work, as the forecasting horizon is 168 hours into the future (one week), the top 10 lags showing the highest correlation with the respective target variable were employed, selected from the last 168-336 lags, \autoref{fig6}, to ensure the use of only available data at the time of forecasting.

\subsubsection{Rolling-Window Statistical Features}

The extraction of statistical features from the target variables in this work employed a rolling-window technique applied to historical data. This approach involved segmenting the time series data into smaller windows, allowing for the capture of evolving data patterns and trends over time. To ensure that only past information was used for feature computation, and to avoid look-ahead bias, rolling statistics were assigned to a timestamp after the window and from the last 168-336 hours as the forecast is 168 hours into the future. These features included the rolling window of the mean, variance, skewness (a measure of asymmetry), and kurtosis (a measure of the distribution's tail heaviness). The mean can be computed using equation \eqref{eq:mean}, while equation \eqref{eq:variance} can be utilised to calculate the variance, equation \eqref{eq:skewness} enables the computation of the skewness, and equation \eqref{eq:kurtosis} is employed to derive the kurtosis, where $X_i$ represents the target variable (e.g., electricity consumption) during the $i^{th}$ hour of the day, with $i$ ranging from 0 to 23. The total number of hours is denoted by $N$. The symbols $M (\mu)$, $V$, $S$, and $K$ represent the mean, variance, skewness, and kurtosis, respectively.

\begin{equation}
M = \frac{1}{N} \sum_{i=1}^N X_i
\label{eq:mean}
\end{equation}

\begin{equation}
V = \frac{1}{N} \sum_{i=1}^N (X_i - \mu)^2
\label{eq:variance}
\end{equation}

\begin{equation}
S = \frac{1}{N} \sum_{i=1}^N (X_i - \mu)^3
\label{eq:skewness}
\end{equation}

\begin{equation}
K = \frac{1}{N} \sum_{i=1}^N (X_i - \mu)^4
\label{eq:kurtosis}
\end{equation}


\revision{
The input features in this chapter and this experiment, summarised in \autoref{tab:inputs}, are chosen to support the core objective of \textbf{RO1}, namely to investigate the role of domain knowledge in FE for ECF and to compare feature categories (see \autoref{ch3_Feature Extraction}) under a reproducible pipeline, such that their contributions can be examined systematically. This approach is more informative because it examines which features repeatedly matter (i.e., useful predictors) for ECF problems and which are redundant under different modelling choices. These input features also align with the wider ECF literature (see \autoref{ch2:Paper1_S3}).}

\subsection{Feature Selection Methods}
\label{Ch3_FS_Methods}
To examine the impact of FS methods on predicting electricity use, temperature, and humidity in FDCS environments, this work employed eight distinct FS methods, as shown in \autoref{tab:fs_methods_2}, including filter-based, embedded, wrapper-based and hybrid. These methods were selected for their effectiveness in related work and potential suitability for FDCS challenges. Utilising diverse FS techniques allowed for a thorough examination, leveraging each method's strengths to comprehensively assess feature relevance. \revision{In this experiment, FS is treated as a modular pre-processing stage that is intentionally separated from the downstream regression model, because the primary objective is to compare FS strategies under a consistent ECF pipeline rather than to tailor a different FS routine to each learner. Many FS approaches are, by design, model-agnostic filters or embedded rankings, whereas wrapper methods such as RFE require a specific estimator to define feature importance and guide elimination \cite{OluAjayi2023} (reviewed in  \autoref{lit_review_FS} and summarised in \autoref{tab:FS_methods}). We therefore use Linear Regression (LR) as the ranking estimator for RFE (and for Sequential Forward Selection) to provide a simple, stable and computationally tractable selection signal that can be applied consistently across all subsequent forecasting algorithms  (see \autoref{tab:fs_methods_2}). This design avoids confounding the comparison by making the selected subset depend on the particular regressor used later in the pipeline, which would otherwise complicate the interpretation of whether performance differences are due to the FS method or the forecasting model itself}. An analysis and comparison of these methods is presented in \autoref{Ch3_Results_and_Discussion}.

\begin{table}
\centering
\caption{Overview of FS methods used in the work (k = number of features to select).}
\footnotesize
\renewcommand{\arraystretch}{1.1}  
\begin{tabularx}{\textwidth}{>{\raggedright\arraybackslash}p{1.5cm}p{1.5cm}X>{\centering\arraybackslash}p{2.85cm}}
\toprule
\textbf{Method} & \textbf{Type} & \textbf{Description} & \textbf{Hyperparameters} \\
\midrule
Correlation (F-Test) & Filter & Select top-k features using univariate LR tests based on F-statistics. & k = 20 \\
\addlinespace
Mutual Information & Filter & Chooses top-k features based on mutual information with the target, capturing non-linear relationships. & k = 20 \\
\addlinespace
Lasso Regularisation & Embedded & Uses L1 regularisation in Lasso regression to eliminate less important features by driving their coefficients to zero. & alpha = 0.01 \\
\addlinespace
Tree Importance (Extra Trees) & Embedded & Employs an ensemble of decision trees to rank features by importance, selecting those with higher importance. & n\_estimators = 300 \\
\addlinespace
ElasticNet Regularisation & Embedded & Combines L1 and L2 regularisation, eliminating features with coefficients that shrink to zero. & alpha = 0.01 \\
\addlinespace
Recursive Feature Elimination (RFE) & Wrapper & Uses RFE with LR to recursively eliminate the least important features. & k = 20 \\
\addlinespace
Sequential Forward Selection & Wrapper & Starts with no features, and iteratively adds important features based on negative mean squared error evaluated by LR. & k = 20 \\
\addlinespace
Filter + Embedded & Hybrid & Uses Correlation F-Test as a filter to select top-k features, followed by an embedded method with a Random Forest Regressor for final selection. & Filter: k = 20, n\_estimators = 300 \\
\bottomrule
\end{tabularx}
\label{tab:fs_methods_2}
\end{table}

\subsection{Machine Learning Algorithms}
In this work, electricity consumption, temperature, and humidity within two FDCSs were predicted using weather data and extracted features as input variables. The ML algorithms used in this work, as shown in \autoref{tab:ml_alg}, were chosen not only based on their established efficacy in predicting the energy consumption of buildings, as supported by the existing literature but also to maximise the strengths of each method, mitigate method-specific limitations, and enhance prediction accuracy and reliability. More precisely; KNN was employed for its simplicity and fast training speed \cite{Hong2022,Wang2020}; RF was chosen due to its robustness against overfitting, along with its embedded ability to provide insights into feature importance \cite{Wang2018}; XGB was included for its rapid performance and high efficiency \cite{Bassi2021,Seyedzadeh2019}; MLP was selected for its high ability to capture complex non-linear relationships, necessary for modelling interactions within FDCS environments \cite{Ding18Model}; LSTM was incorporated due to its proficiency in handling sequential data and capturing time dependencies \cite{Gao2021,Wang2019Power,Somu2020,Zhou2020Using}; and MTL was utilised for its potential to learn multiple related tasks simultaneously, aiming to enhance generalisation despite its inherent complexity and the potential for task interference \cite{Liu2023,Wang2022Multi,Tan2020}. Hyperparameter tuning, which involves setting model configurations before training, is crucial for optimising model performance. To fine-tune each model, a grid search of hyperparameter combinations was conducted using training and validation sets to identify the optimal values. The tuned models were then evaluated on the test set. \autoref{tab:hyperparameters} shows the results of the grid search for hyperparameter tuning across all models.

\begin{table*}
\centering
\caption{ML algorithms used in this work.}
\footnotesize
\begin{tabularx}{\textwidth}{>{\raggedright\arraybackslash}p{3cm}Xc}
\toprule
\textbf{Model} & \textbf{Description} & \textbf{Ref.} \\
\midrule
KNN Regression (KNR) & A non-parametric algorithm that predicts the target based on the average of the K closest training examples in the input feature space. & \cite{valgaev2017building} \\
\addlinespace
RF Regression (RFR) & An ensemble method that fits multiple decision trees on randomly sampled subsets of the data and combines their predictions. & \cite{breiman2001random} \\
\addlinespace
XGB Regression (XGBR) & An ensemble method that trains decision trees sequentially, each time fitting the residual errors of the previous tree. & \cite{Chen2016XGBoost} \\
\addlinespace
MLP Regressor & A feedforward ANN with multiple layers of nodes between input and output. Uses backpropagation to train the network weights and biases. & \cite{Afzal2023} \\
\addlinespace
LSTM & A class of RNNs that can learn long-term dependencies by using a memory cell and three gating mechanisms. & \cite{Hochreiter1997} \\
\addlinespace
MTL & A learning method where multiple tasks are handled concurrently, using a shared representation. By leveraging the similarities and variations across tasks, what is learnt for one task can aid in the learning of other tasks. & \cite{Caruana1997} \\
\bottomrule
\end{tabularx}
\label{tab:ml_alg}
\end{table*}


\begin{table}[b]
\centering
\caption{Grid search results for hyperparameter tuning of models predicting electricity consumption. Training used a 4-core Intel CPU with 32GB RAM, Python 3.11, Scikit-learn, and TensorFlow frameworks.}
\footnotesize
\begin{tabularx}{\textwidth}{@{}llX@{}}
\toprule
& \textbf{Model} & \textbf{Best hyperparameters} \\
\midrule
\multirow[c]{6}{*}{\textbf{FDCS 1}} 
& KNNR & 'leaf\_size': 10, 'n\_neighbours': 10, 'weights': 'distance' \\
& RFR & 'max\_depth': 10, 'min\_samples\_leaf': 2, 'min\_samples\_split': 10, 'n\_estimators': 500 \\
& XGBR & 'learning\_rate': 0.1, 'max\_depth': 10, 'min\_child\_weight': 2, 'n\_estimators': 500, 'subsample': 0.8 \\
& MLP & 'activation': 'relu', 'alpha': 0.0001, 'batch\_size': 16, 'hidden\_layer\_sizes': (50, 100), 'learning\_rate\_init': 0.01, 'max\_iter': 500, 'solver': 'adam' \\
& LSTM & LSTM\_Layers: [20,10], Optimizer: Adam, Learning\_Rate: 0.01, Epochs: 100, batch\_size: 32 \\
& MTL & Base\_Layers: 10, 40; Activation: ReLU; Optimizer: Adam; Learning\_Rate: 0.005; Loss\_Function: MSE; Epochs: 100; Batch\_Size: 32; Loss\_Weights: 1 \\
\midrule
\multirow[c]{6}{*}{\textbf{FDCS 2}}
& KNNR & 'leaf\_size': 10, 'n\_neighbours': 20, 'weights': 'distance' \\
& RFR & 'max\_depth': 10, 'min\_samples\_leaf': 10, 'min\_samples\_split': 2, 'n\_estimators': 500 \\
& XGBR & 'learning\_rate': 0.01, 'max\_depth': 20, 'min\_child\_weight': 10, 'n\_estimators': 500, 'subsample': 0.8 \\
& MLP & 'activation': 'relu', 'alpha': 0.001, 'batch\_size': 32, 'hidden\_layer\_sizes': (50, 100), 'learning\_rate\_init': 0.001, 'max\_iter': 500, 'solver': 'sgd' \\
& LSTM & LSTM\_Layers: [20,10], Optimizer: Adam, Learning\_Rate: 0.01, Epochs: 100, batch\_size: 32 \\
& MTL & Base\_Layers: 30, 30; Activation: ReLU; Optimizer: Adam; Learning\_Rate: 0.01; Loss\_Function: MSE; Epochs: 100; Batch\_Size: 32; Loss\_Weights: 1 \\
\bottomrule
\end{tabularx}
\label{tab:hyperparameters}
\end{table}


\subsection{Model Performance Assessment Metrics}
The performance of the ML regression models was evaluated using the Mean Absolute Error (MAE) \cite{chai2014root}. MAE, which measures average absolute differences between predicted and actual values in the original units, was selected for its interpretability. The equation of MAE \eqref{eq:mae} is shown below where $M$ is the number of samples in the studied dataset, $predicted_i$ is the predicted $i^{th}$ value, $observed_i$ is the true $i^{th}$ value, and $m\_observed$ is the mean of the true values.

\begin{equation}
    MAE = \frac{1}{M} \sum_{i=1}^m |predicted_i - observed_i|
    \label{eq:mae}
\end{equation}

\revision{MAE is used as the primary metric to compare the impact of FE and FS across the large set of experiments conducted in this chapter (six algorithms, eight FS methods, and additional analyses such as dataset size effects). Using this single, unit-consistent primary metric keeps comparisons readable and prevents conclusions from being driven by metric-specific scaling artefacts when summarising many runs.}

\begin{landscape}
\begin{figure}
\centering
\includegraphics[width=.9\linewidth]{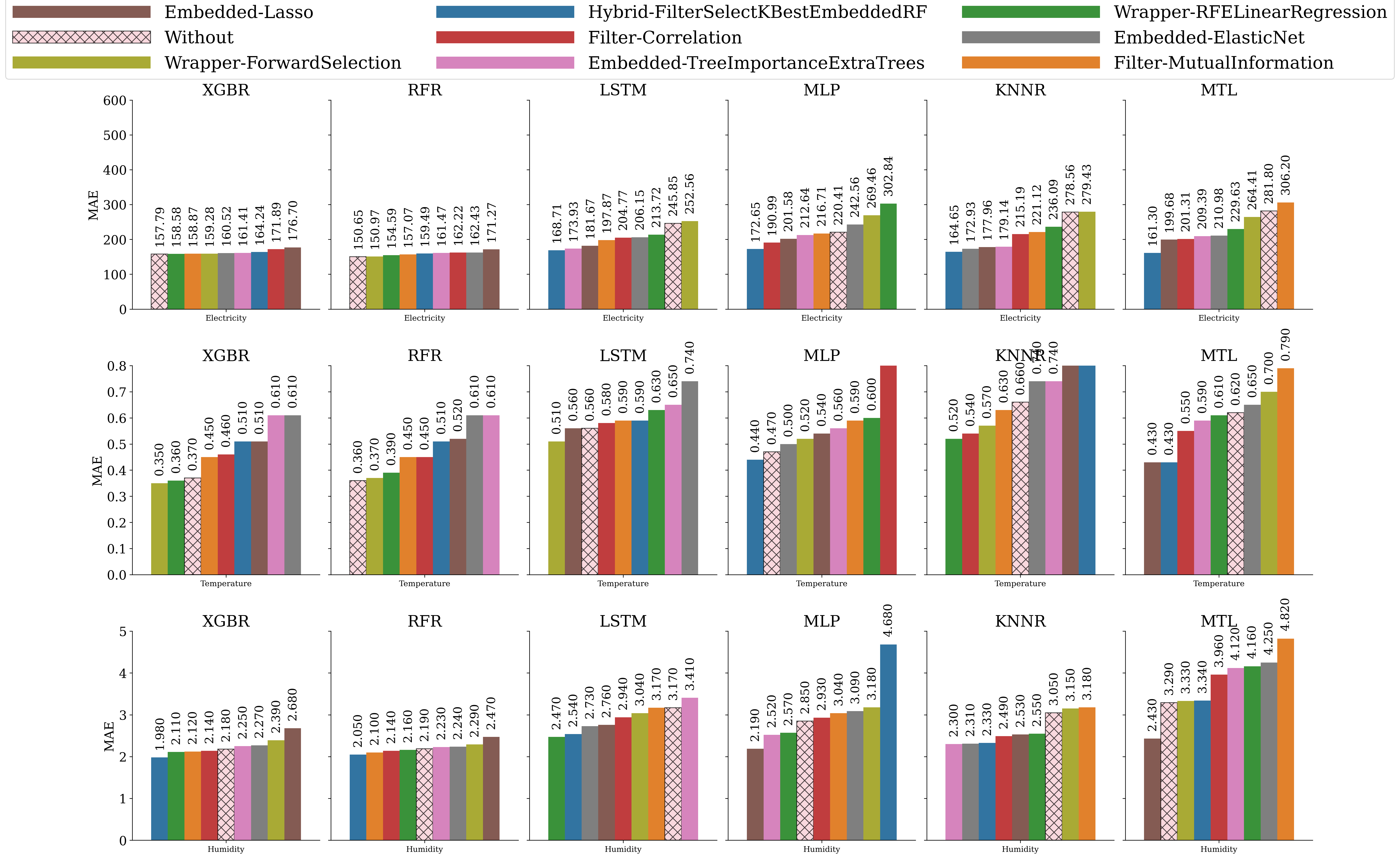}
\caption{Model performance evaluation using MAE of the test set of different FS methods in FDCS 1.}
\label{fig7}
\end{figure}
\end{landscape}

\begin{landscape}
\begin{figure}
\centering
\includegraphics[width=.9\linewidth]{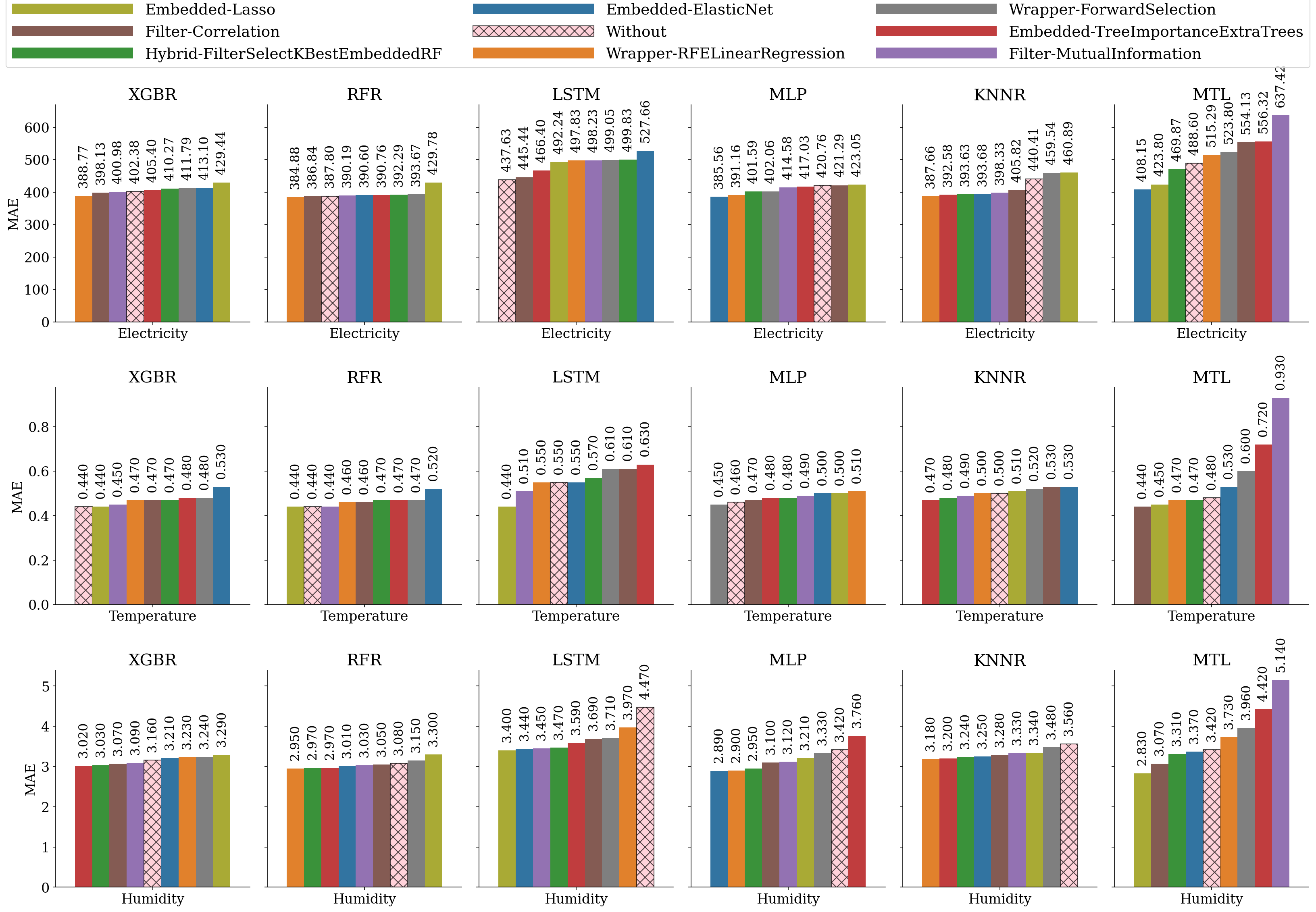}
\caption{Model performance evaluation using MAE of the test set of different FS methods in FDCS 2.}
\label{fig8}
\end{figure}
\end{landscape}
\begin{landscape}
\begin{figure}
\centering
\includegraphics[width=.9\linewidth]{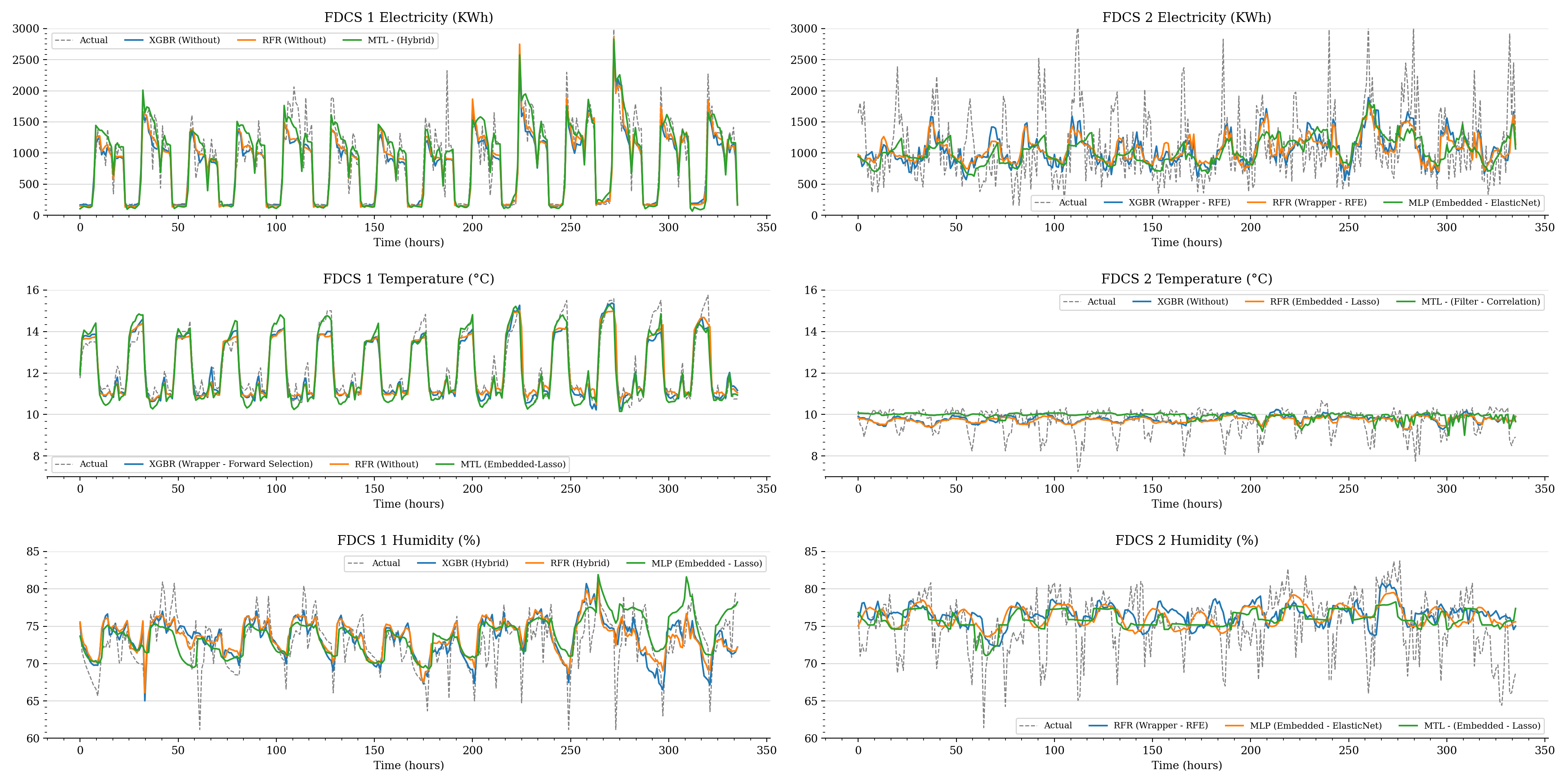}
\caption{Comparison of the three lowest model errors in predicting electricity, temperature, and humidity in FDCS 1 and 2. Plots show actual versus predicted values on the test set.}
\label{fig9}
\end{figure}
\end{landscape}

\section{Results and Discussion}
\label{Ch3_Results_and_Discussion}
This section presents an analysis and discussion of the experimental results, including the performance of ML models in prediction tasks within the FDCS settings on the test set, the influence of eight different FS methods, feature importance, and finally, the implications of dataset size.

\subsection{Performance Evaluation}
\label{subsec:Performance_evaluation}
The results of the experimental evaluation, as summarised in \autoref{fig7}, \autoref{fig8}, and \autoref{fig9}, demonstrate that ensemble methods (XGBR and RFR) can achieve more accurate predictions for forecasting electricity consumption and internal environmental conditions one week into the future in FDCS settings when compared to other algorithms. For electricity predictions of FDCS 1, these two methods outperformed others, even without applying FS methods (i.e., using all features), achieving the lowest errors in the test set with MAEs of 150.75 and 157.79, respectively. The promising performance of the XGBR algorithm observed in this work aligns with findings from other studies, such as \cite{Bassi2021,Seyedzadeh2019}. However, while comparing these findings with existing literature is important, such a comparison may not be entirely appropriate due to the unique context of FDCSs, which differs significantly from other domains. Notably, the hybrid (filter + embedded) was the FS method that most improved the performance of LSTM, MLP, KNNR, and MTL. Similar patterns were observed as the ensemble-based methods produced the lowest errors in predicting both indoor temperature and humidity in FDCS 1.

In FDCS 2, the MLP model, combined with the hybrid (filter + embedded) FS method, produced prediction errors almost matching those of ensemble-based methods for predicting electricity consumption, yet the prediction errors were noticeably high. For predicting indoor temperature in this storage, LSTM and MTL, alongside RFR and XGBR, produced the lowest prediction errors when the embedded-lasso FS method was applied. In understanding the most significant findings across the two different FDCSs, it's important to recognise that these storage systems differ in layout, size, and operations. Consequently, a direct comparison may not be entirely appropriate. Nevertheless, some significant patterns emerge. First, the ensemble methods XGBR and RFR consistently achieve top-tier performance with the lowest errors compared to other models in both environments. The success of these two algorithms likely stems from their advanced FS and ensemble techniques - bagging for RFR and boosting for XGBR. Secondly, the neural network-based models demonstrate considerable variability in their performance metrics across different FS methods. This fluctuation could be attributed to the models' sensitivity to specific features and/or the relatively small dataset sizes used for training.

\revision{A limitation of the FS modular design in the conducted experiments (see \autoref{Ch3_FS_Methods} for details) is that wrapper-based selection is inherently model-dependent: different estimators can yield different rankings because learners exploit feature interactions differently \cite{OluAjayi2023}. Consequently, using LR to guide RFE may not produce the optimal subset for every downstream regressor, particularly for non-linear models that can benefit from interaction effects that LR does not capture. However, coupling RFE separately to each forecasting model would multiply the experimental space and make cross-model comparisons harder to interpret, because the feature subset would change with the learner as well as with the FS method. We therefore prioritise a consistent evaluation of FS behaviour within a single pipeline, and treat algorithm-specific wrapper selection as a potential extension rather than a requirement for the conclusions drawn in this experiment.}

Additionally, these results shed light on the predictability of energy consumption and indoor variables in FDCSs. The models consistently demonstrate the lowest errors in FDCS 1, suggesting that its energy consumption and indoor variables are more predictable than those in FDCS 2. This observation aligns with the consistent daily trends observed in electricity consumption, indoor temperature, and humidity depicted in \autoref{fig5} for FDCS 1 compared to FDCS 2. While high predictability can facilitate planning and management, thereby boosting operational efficiency and cost savings, it should not be equated with efficiency. For example, an FDCS with high but predictable energy consumption may not be as efficient as one with less predictable but lower energy consumption. Therefore, these findings should be integrated into a broader strategy for energy efficiency understanding and improvement in FDCS. \revision{In the context of the overall thesis, this reinforces that forecasting predictability/accuracy (established here as a baseline under \textbf{RO1}) is informative for operational planning but is not, by itself, evidence of energy efficiency, thereby motivating the later work on automating ECF-relevant FE (\textbf{RO2}) and evaluating whether forecasting gains translate into improved downstream decision quality under PTO and DFL scenarios (\textbf{RO3}).}

\subsection{Evaluating Feature Importance Using SHAP}
\begin{landscape}
\begin{figure}
\centering
\includegraphics[width=1\linewidth]{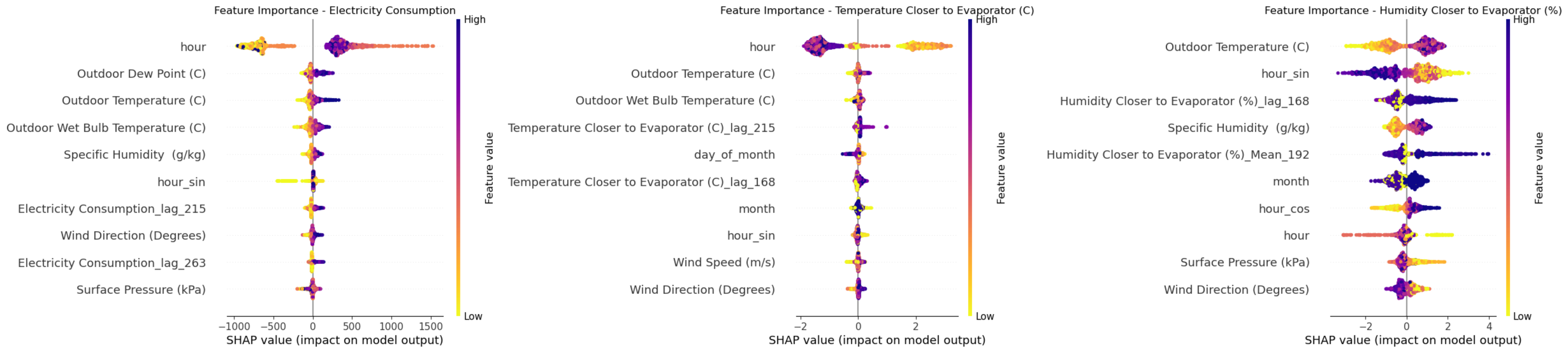}
\caption{Analysis of the top ten features in terms of their impact on model output, as explained by SHAP for electricity, temperature, and humidity variables in FDCS 1.}
\label{fig10}
\end{figure}
\end{landscape}

\begin{landscape}
\begin{figure}
\centering
\includegraphics[width=1\linewidth]{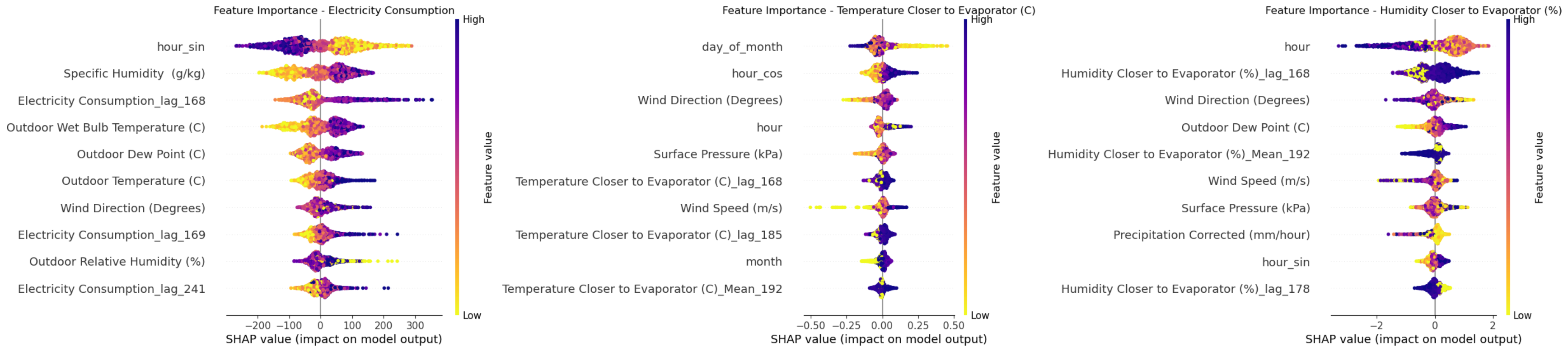}
\caption{Analysis of the top ten features in terms of their impact on model output, as explained by SHAP for electricity, temperature, and humidity variables in FDCS 2.}
\label{fig11}
\end{figure}
\end{landscape}

Building on the insights gained from the FS analysis in \autoref{subsec:Performance_evaluation}, this section aims to delve deeper into understanding feature importance. This work employs Shapley Additive Explanations (SHAP) \cite{SHAP_paper} for feature importance evaluation, as illustrated in \autoref{fig10} and \autoref{fig11}. SHAP was selected due to its model-agnostic properties, local accuracy, and game-theoretic foundation, which ensures a fair and consistent distribution of predictive power across features. This method ranks features by their impact on the model's predictions, with the top feature being the most influential and data points spread along a horizontal axis showing the direction and magnitude (negative or positive) of their impact. For this experiment, XGBR was chosen over RFR due to its faster training speed.

For FDCS 1, the most critical feature for predicting electricity consumption, and indoor temperature was the hour of the day, indicating a daily cyclical pattern, potentially influenced by operational routines in such an environment. Notably, in FDCS 1, among the top ten features for predicting electricity, four were extracted features, while the remaining six were weather-related features. Similarly, for predicting indoor temperature and humidity, some features were extracted features, highlighting the effectiveness of these methods in capturing complex patterns and trends in FDCS environments, thus improving the model's predictive accuracy.

Likewise, in FDCS 2, key features for predicting the same variables include time-lags, cyclical, and rolling-window statistical features, alongside weather-related features. These results across both systems underscore the importance of feature extraction in capturing FDCS complexities, thereby enhancing the model's prediction capability. The noticeable variance in SHAP values for electricity in FDCS 2, indicated by the more widely dispersed dots, implies that the features affecting its model predictions demonstrate greater variability. This could be due to its irregular usage patterns, aligning with earlier findings discussed in \autoref{subsec:Performance_evaluation} and \autoref{fig5}, unlike the predictable trends seen in FDCS 1. The weather impact on both FDCSs is noticeable, particularly the outdoor temperature, which directly influences the cooling demand, thereby affecting energy consumption, underscoring the considerable influence of weather-related features in energy forecasting strategies for such systems. \revision{This matters to the thesis because it identifies the key, practically meaningful feature families (e.g., temporal, lagged/rolling statistics) that drive ECF performance in a real FDCS setting, providing the empirical basis for encoding and automating these feature categories in AutoEnergy (\textbf{RO2}) and later investigating whether such representation improvements translate into better downstream decision quality (\textbf{RO3}).}


\begin{landscape}
\begin{figure}
\centering
\includegraphics[width=1\linewidth]{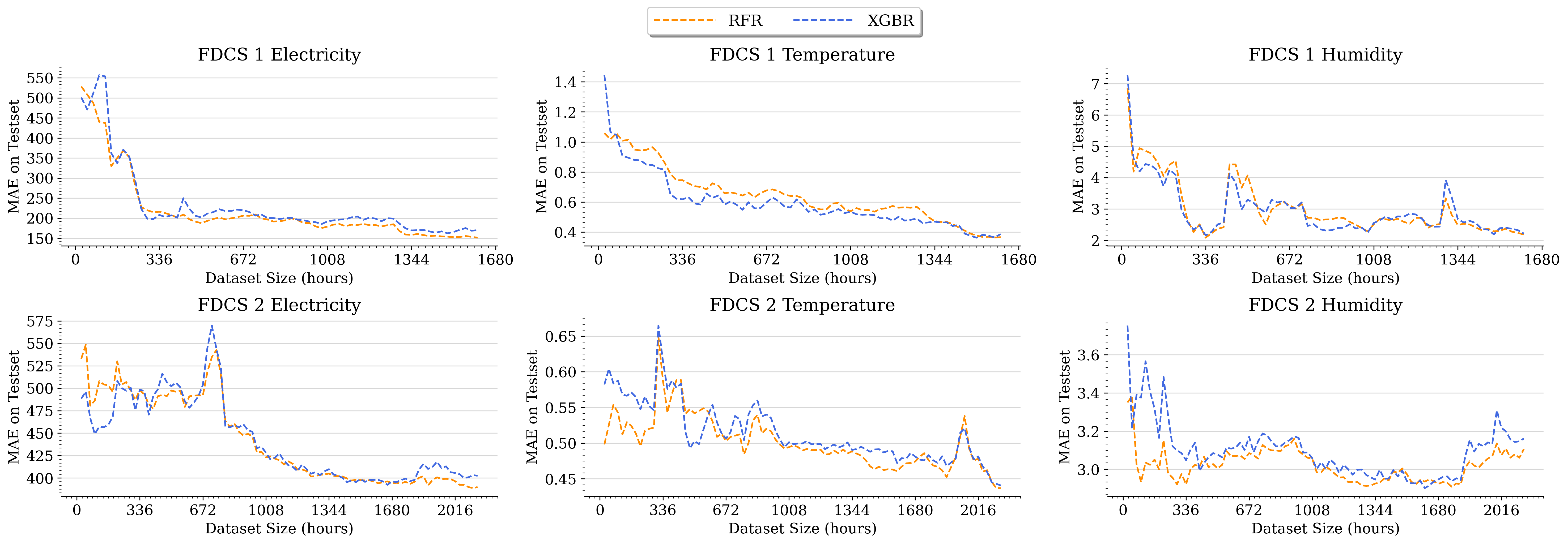}
\caption{Variations of MAE in the test set occur when predicting electricity, temperature, and humidity in both FDCSs, as a function of increasing the training dataset size for the XGBR and RFR models.}
\label{fig12}
\end{figure}
\end{landscape}


\begin{table*}
\centering
\caption{Summary of key recommendations for ML applications in FDCS systems, offering guidelines and insights to enhance the performance and efficiency of forecasting models in real-world applications.}
\footnotesize
\begin{tabularx}{\textwidth}{>{\raggedright\arraybackslash}p{3cm}X}
\toprule
\textbf{Focus Area} & \textbf{Recommendations} \\
\midrule
Enhancement of input features & 
Use feature extraction (e.g., the hour of the day, and cyclical features). \\
\addlinespace
FS techniques & 
Hybrid FS methods generally enhance model performance. Avoid wrapper methods if computational resources are limited. \\
\addlinespace
Algorithms selection & 
Ensemble-based learning methods (XGBR, RFR) were superior to traditional ML, NN-based, and deep learning models in FDCS predictions. XGBR is particularly recommended for its computational efficiency. \\
\addlinespace
Dataset collection & 
Contrary to popular belief, larger datasets were not necessary for accurate prediction in this work. Smaller datasets from real-world FDCS facilities may yield reliable predictions when enhanced with robust FE. \\
\bottomrule
\end{tabularx}
\end{table*}
\subsection{Dataset Size Implications}
\label{Ch3_results_Dataset_Size_Implications}
As demonstrated in \autoref{fig12}, the evaluation of how the volume of the training data affects forecasting performance was conducted using XGBR and RFR, as they had the most consistently superior performance as shown in previous analyses. In this experiment, the models were trained starting with a baseline of a single day's worth of hourly data. From this baseline, the dataset was expanded in one-day increments, each comprising 24 hourly data points, to systematically assess the impact of the train dataset size on prediction performance in the test set. While the best-performing models can provide valuable insights into the implications of dataset size, it is important to note that this approach is computationally expensive, as it involves iterative retraining of the models on an ever-expanding data corpus, a constraint that precluded a comprehensive investigation of dataset size implications across all examined algorithms.

In predicting electricity consumption, the XGBR and RFR models showed fluctuating yet overall declining MAE in the test set as the training dataset size increased in both FDCSs. The most notable improvements (i.e., reduction in prediction errors) for both models occurred at 1344 hours (56 days' worth of data) in FDCS 1 and at 1680 hours (70 days) in FDCS 2. After those levels, and as the dataset size continued to grow, both models showed a trend towards stability with minor MAE fluctuations, signalling performance plateaus. For temperature, both models showed signs of stabilisation around 1560 hours' worth of data, with minor MAE changes in FDCS 1 compared to FDCS 2. In analysing the dataset size impact on humidity prediction, MAE decreased as the data grew; however, there were more noticeable fluctuations in prediction errors.

\revision{This analysis provides empirical, albeit limited, evidence to aid in answering the practical question of how much historical data is needed to obtain reliable ECF performance in FDCS settings, thereby supporting the thesis aim of developing pipelines that remain effective under small dataset sizes by indicating where performance improves and where it plateaus as training data volume increases.} However, it is important to acknowledge that these conclusions are drawn from the available data, and additional research is needed to confirm these findings in broader applications. Yet, such observations could be valuable in scenarios where acquiring extensive historical data is impractical due to time, cost, or data availability limitations.

\section{Summary}
This work proposes an ML pipeline tailored for predicting electricity consumption, indoor temperature, and humidity one week (hourly) into the future in FDCS settings, addressing the unique challenges of such environments compared to previous building-focused ML studies and also suitable for small dataset sizes. Two real-world datasets of FDCSs have been employed for training, validation, and testing of the developed models. The results show that ensemble-based methods (RFR and XGBR) outperformed other models in both examined FDCS datasets, evidenced by the lowest MAE values while neural network-based and deep learning models showed varied performance. Eight FS methods have been investigated, and the results from the datasets used in this work indicate that hybrid methods generally enhance model performance, while wrapper methods are computationally expensive. The conducted feature importance analysis underscores the importance of feature extraction, given that the extracted features have a noticeable impact on model outputs, as evidenced by SHAP analysis. The implication of dataset size on model accuracy was analysed, providing some insights into estimating the minimum dataset size needed for forecasting electricity in FDCS (1344 hours’ worth of data for FDCS 1 and 1680 hours for FDCS 2). Nevertheless, these conclusions are drawn from the available datasets, and further research is needed. \revision{Nevertheless, these conclusions are drawn from the available FDCS datasets; for the thesis, they serve as an empirically grounded baseline and motivation for the following chapter’s broader multi-dataset evaluation of automated FE, while acknowledging that generality beyond these sites requires further evidence.} \revision{The limitations of this work are discussed in \autoref{con:Research_Limitations}.}

\cleardoublepage

\chapterwithquote{"If I have seen further, it is by standing on the shoulders of giants" - Isaac Newton}{AutoEnergy: An Automated Feature Engineering Algorithm for Energy Consumption Forecasting with AutoML}{ch:mainchapter5}


\revision{Building on the baseline ECF pipeline and empirical insights from \autoref{ch:mainchapter3} (notably which features consistently contribute to improved ECF performance under data limitations), this chapter transitions from analysis to automation by encoding these useful predictors into an AFE method that can be applied across diverse energy systems.} In particular, this chapter fulfils \textbf{RO2} by introducing AutoEnergy, a novel domain-aware AFE algorithm specifically tailored for ECF problems. 

AutoEnergy automatically generates interpretable features from timestamps and past consumption values through rule-based transformations, enabling fully automated ECF modelling when integrated with AutoML. \revision{It is rule-driven because the feature space is produced by a fixed set of FE functions based on ECF domain knowledge, with explicitly defined behaviour, rather than by a learned feature generator (see \autoref{ch:Method}). In other words, AutoEnergy generates features by applying an explicit, predefined set of transformation rules, rather than learning latent representations through an additional predictive model. This aligns with the heuristic-based AFE perspective discussed earlier in \autoref{ch2_Automated_Feature_Engineering}, where FE is governed by auditable rules and criteria rather than opaque learned embeddings. Concretely, AutoEnergy’s rules specify both (i) which feature families can be generated (temporal and cyclical encodings, lags, and rolling-window statistics) and (ii) which candidates are retained via statistical selection. The key point is that every generated feature can be traced back to a named rule and a parameter choice, which keeps the FE process reproducible and inspectable.}

AutoEnergy is validated across eighteen diverse energy datasets spanning residential, commercial, industrial, renewable, and grid power domains, including steel manufacturing environments and FDCS facilities, demonstrating robust generalisation potential. Through comprehensive benchmarking against existing AFE methods, AutoEnergy achieves superior reductions in forecasting errors while maintaining greater computational efficiency. The chapter further examines AutoEnergy's integration with the state-of-the-art Tabular Prior-Fitted Networks (TabPFN), resulting in significant forecasting error reductions across test sets.

\section{\revision{Introduction}}
\label{ch:Intro}

FE is a crucial step in developing ML pipelines, as it transforms raw data into informative features that enhance model performance \cite{Domingos2012, Bengio2013}. This is because raw data often require preprocessing and transformation before algorithms can learn effectively \cite{Liu2024, Verdonck2021}. Furthermore, real-world datasets are often small or limited due to data collection limitations, privacy issues, or resource constraints \cite{grinsztajn2022tree, hollmann2022tabpfn}. In such scenarios, FE could compensate for the limited data by extracting informative features. Moreover, FE may improve computational efficiency by eliminating noisy, redundant, and irrelevant data \cite{Wang_2022, Mumuni_2024}. Finally, FE may improve not only predictive accuracy but also explainability \cite{Gosiewska2021}. 

Despite these advantages, FE remains a time-consuming process that is prone to human error while relying heavily on domain expertise and iterative experimentation \cite{Wu2022, Wang_2022}. Deep learning algorithms \cite{Fan2019Deep, Dong2021}, while capable of automatically learning useful representations from raw data, lack interpretability and typically require large datasets to perform well, a condition that is often infeasible in real-world scenarios \cite{grinsztajn2022tree, wang2019review}. This has led to a growing interest in AFE methods \cite{hollmann2024large}.

Modern AutoML frameworks \cite{hutter2019automated}, such as AutoGluon \cite{AutoGluon2020}, H2O \cite{h2o}, and FLAML \cite{wang2021flaml} offer streamlined solutions for ML pipeline development through automated model selection and hyperparameter tuning. Nevertheless, these solutions often assume that data preparation and feature generation have been completed and the data are ready for training. As a result, tasks such as FE and the integration of domain knowledge are largely left to human practitioners \cite{hollmann2024large}.

Furthermore, many general-purpose AutoML systems are proposed for broad applicability across ML tasks, but their focus on generalisation often limits their effectiveness in specialised domains \cite{He2021}. Such limitations become particularly evident in domains that require interpretable features, such as ECF of time series data, where understanding the factors influencing energy consumption is important for well-informed decisions. This challenge may be attributed to the complex nature of power usage patterns, which can involve various linear and nonlinear relationships, fluctuating behaviours, and potential dependencies on temporal and environmental factors \cite{zhang2021review, manandhar2023current}. Therefore, domain-relevant FE could be a beneficial approach for AutoML to excel in modelling ECF problems.

\revision{It is worth clarifying further that the AutoML methods \cite{hutter2019automated} used in this work are used to reduce reliance on \textit{modelling expertise} that would otherwise be required for manual \textit{model selection} and \textit{hyperparameter tuning} across many datasets and experimental conditions (see \autoref{ch2:Paper2} for more details). Such methods are specifically designed to automate these choices by searching over algorithms and their hyperparameters in a data-driven manner, returning the best-performing configuration \cite{AutoGluon2020, h2o, wang2021flaml}. This motivation is complementary to, and distinct from, \textit{domain knowledge} in ECF problems. In particular, AutoML reduces the effort required to configure the learner, whereas domain knowledge in FE (i.e., human practitioners) is typically needed to decide which input features are useful \cite{Wang_2022, hollmann2024large}. Therefore, in our work, AutoML sits alongside domain knowledge rather than replacing it. Put differently, the proposed FE algorithm in this chapter reduces the manual FE burden by generating an automated domain-aware input feature set, while AutoML automated model selection and tuning, thereby supporting fully automated ML pipelines for ECF applications.}

Given the aforementioned limitations and as discussed in \autoref{ch2_Automated_Feature_Engineering}, this work proposes \texttt{AutoEnergy}, a domain-aware AFE method that constitutes a novel synthesis to: (a) improve AutoML performance through domain-specific FE optimised for ECF; (b) minimise the human intervention and domain expertise required by automating the time-consuming, manual FE process; and (c) maintain interpretability through a heuristic search design, generating human-readable and traceable features. \revision{It is worth clarifying that the work presented in this thesis, and particularly in this chapter, does not claim that time saved relative to manual FE by human experts can be quantified directly. Doing so would require domain experts to engineer features across datasets and report comparable person-hours, which is neither practical nor reproducible. Instead, the evaluation operationalises \textit{reduced manual effort} in terms of the degree of automation and reproducibility of the pipeline, and it reports empirical outcomes such as forecasting performance and computational overhead (e.g., FE runtime). Accordingly, claims in this chapter should be interpreted as reducing manual intervention and iteration, rather than measuring absolute time saved.}

The key novelty of the proposed FE algorithm lies not in the individual components (i.e., feature types), but in its fully automated, expert-free, domain-aware feature extraction, optimised selection, and strategic integration tailored to ECF problems, enabling end-to-end automated ECF modelling when integrated with AutoML. The method automatically generates features by applying a series of FE functions to the dataset's timestamps and target variables (i.e. past consumption values). To assess the proposed method on AutoML performance, a comprehensive set of eighteen real-world datasets, representing various energy consumption patterns in different domains (e.g., residential and commercial buildings, wind turbines, industrial settings, food storage facilities, and grid power consumption), was utilised \footnote{All results in this work are reproducible as the code and datasets used are publicly available on GitHub. See \protect\url{https://github.com/Nasser-Alkhulaifi/AutoEnergy}}. Additionally, the proposed method was systematically compared with well-established FE methods in the literature, namely TS and FT, assessing both (a) predictive accuracy and (b) computational efficiency. This thoroughly evaluated the method's overall effectiveness and practical applicability in real-world ECF tasks. While our previous exploratory study \cite{Alkhulaifi2024} laid the foundation for this article, the novelty and contributions of the current work are as follows:

\begin{itemize}[leftmargin=15pt, itemsep=-2pt]

\item An improved AFE method tailored for ECF is introduced with a higher degree of automation in feature extraction and optimised selection compared to our previous work, as introduced and explained in \autoref{ch:Method}. It considerably improves AutoML performance, as demonstrated by comparisons with and without the proposed FE method, as shown in \autoref{Impact_on_AutoML}. Additionally, the impact of the proposed FE method on the state-of-the-art TabPFN algorithm \cite{Hollmann2025} was also examined. \footnote{TabPFN is a foundation model that leverages transformer-based meta-learning, designed specifically for tabular data problems and recognised as a state-of-the-art AutoML framework.}

\item This work includes comprehensive benchmarking against existing FE methods. The proposed FE method achieves a superior reduction in forecasting errors and computational efficiency for ECF problems compared to the benchmarking methods, as shown in \autoref{Comparison_benchmarking_methods}. 

\item This work incorporates a wider range of energy consumption datasets, including new energy systems and settings such as steel manufacturing environments and food and drinks cold facilities. This allows for a more extensive evaluation of the proposed method's applicability and effectiveness across different settings and environments.

\end{itemize}

The structure of the remaining sections of this chapter is as follows: \autoref{ch:Method} outlines the proposed method for automating FE for ECF problems, \autoref{ch:Experimental_design} provides details of the experimental design, including the datasets used, benchmarking methods, and evaluation criteria. Analysis and discussion of findings are presented in \autoref{ch:Results_Discussion}. To conclude, \autoref{ch5:Conclusion} summarises key insights.

\section{AutoEnergy: An Automated End-to-End Feature Engineering Algorithm}
\label{ch:Method}

This section outlines the problem under investigation in \autoref{sec:problem_definition} and subsequently details the proposed method to address it in \autoref{sec:Proposed_Method}.


\subsection{Problem Definition}
\label{sec:problem_definition}
Given a dataset $\mathcal{D}$ consisting of $N$ instances, where each instance is represented by a tuple $(t_i, y_i)$, with $t_i$ denoting a timestamp and $y_i$ the corresponding target variable, we define the dataset as follows:
\begin{equation}
\mathcal{D} = \left\{ (t_1, y_1), (t_2, y_2), \ldots, (t_N, y_N) \right\}
\end{equation}

\noindent The aim is to develop an algorithm that implements an AFE process to enhance the performance of a predictive model $\mathcal{M}$ within an AutoML framework.

\subsection{The Proposed Feature Engineering Method}
\label{sec:Proposed_Method}

\begin{landscape}
\begin{figure}
\centering
\includegraphics[width=1\linewidth]{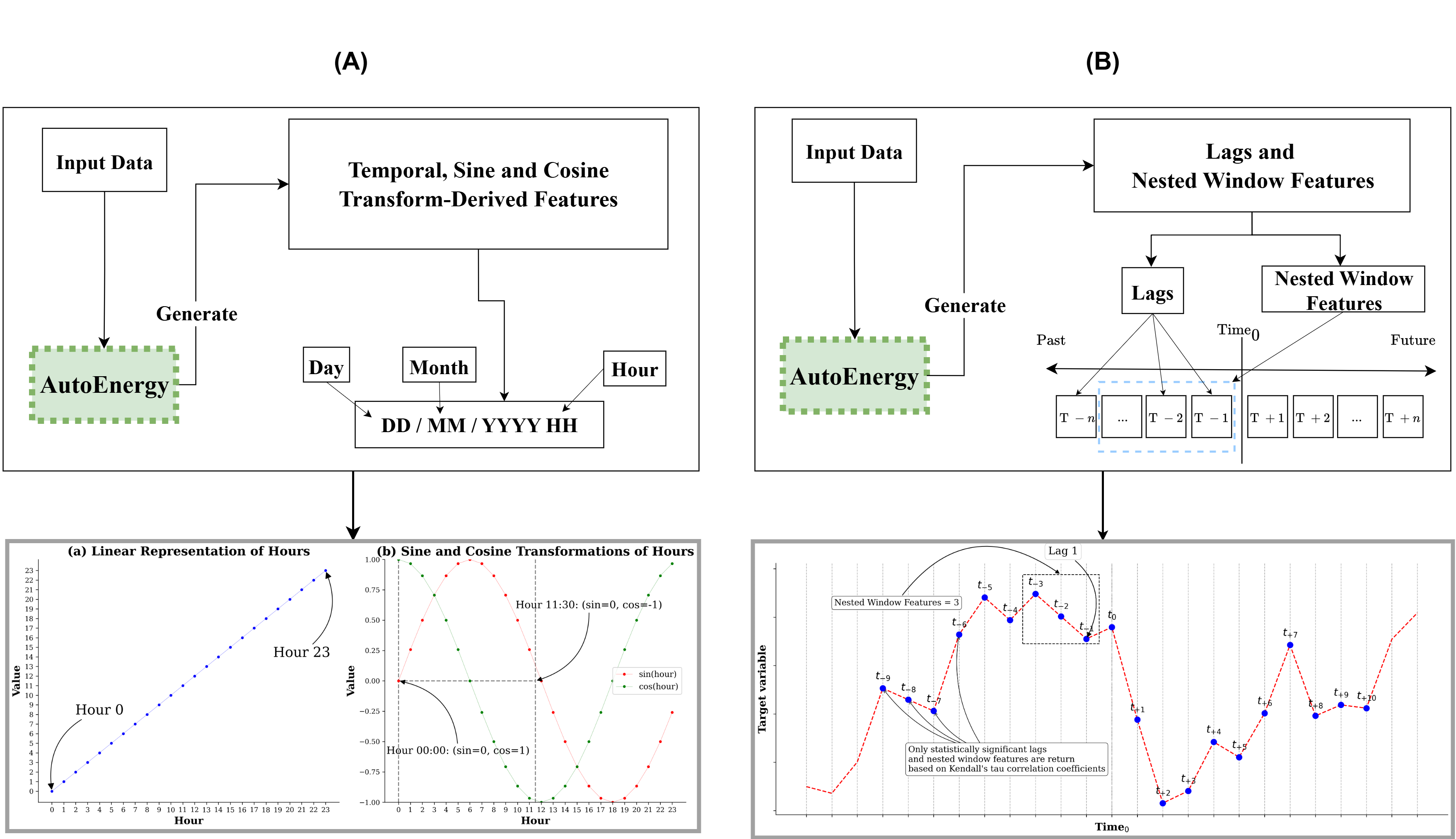}
\caption{The proposed \hyperref[pseudocode_AutoEnergy_Part1]{Algorithm 1} for generating temporal, sine, and cosine transform-derived features is illustrated in subfigure~(A), while the proposed \hyperref[pseudocode_AutoEnergy_Part2]{Algorithm 2} for generating lag and nested window features is illustrated in subfigure~(B).}
\label{AutoEnergy_Figure}
\end{figure}
\end{landscape}

The proposed FE method, as illustrated in \autoref{AutoEnergy_Figure}, applies a set of FE functions $\{\text{GeneratedFeatures}_j\}_{j=1}^{M}$ which process the timestamp $t_i$ and \revision{the historical target values available up to $t_i$ (denoted $\mathbf{y}_{1:i-1}$)} in $\mathcal{D}$ to generate a series of feature subsets $\mathbf{F}'_{i,j}$. The complete feature vector $\mathbf{F}'_{i}$ for each instance $i$ is then created by concatenating these feature subsets:

\revision{
\begin{equation}
\mathbf{F}'_{i} = \bigoplus_{j=1}^{M} \text{GeneratedFeatures}_j\!\left(t_i,\mathbf{y}_{1:i-1}\right), \quad \forall i \in \{1, \ldots, N\}
\end{equation}
}

\revision{\noindent where $\mathbf{y}_{1:i-1} = (y_1,\ldots,y_{i-1})$ denotes the historical target values available up to (but not including) time $t_i$, so that lag and rolling-window features at $t_i$ are computed from past values (e.g., $y_{i-k}$ and $\mathbf{y}_{i-w:i-1}$) rather than from $y_i$.} 
where also $\bigoplus$ denotes the concatenation operation, combining all feature subsets $\mathbf{F}'_{i,j}$ generated by the functions into a single feature vector for each instance.

Following this, the predictive model $\mathcal{M}$ is trained using these complete feature vectors $\mathbf{F}'_{i}$ along with their corresponding target variable $y_i$, in an AutoML framework:
\begin{equation}
\mathcal{M} = \text{AutoML}\left(\{(\mathbf{F}'_{i}, y_i)\}_{i=1}^{N}\right)
\end{equation}

\noindent Although many of these features exist in the literature, the proposed algorithm is fully automated, expert-free, domain-aware feature extraction, optimised selection, and strategic integration tailored to ECF problems, enabling end-to-end automated ECF modelling when integrated with AutoML. In particular, for lag and rolling-window statistical features, as described in the following subsections: 


\label{Pseudocode_AutoEnergy_Part1}
\begin{algorithm}[!t]
\caption{Temporal, Sine and Cosine Transform-Derived Features.}
    \label{pseudocode_AutoEnergy_Part1}
    \footnotesize
    \begin{algorithmic}[2]
        \Require DataFrame $\mathcal{D}$ with $t$ (timestamps) and $y$ (target variable)
        \Ensure DataFrame with temporal and cyclical features
        \State $\mathcal{D}' \gets \mathcal{D}$
        \State $Features_{\text{time}} \gets F_{\text{time}}(\mathcal{D}, t)$
        \State $Features_{\text{cyclical}} \gets F_{\text{cyclical}}(Features_{\text{time}})$
        \State $\mathcal{D}' \gets \mathcal{D}'$ append $Features_{\text{time}}$, $Features_{\text{cyclical}}$
        \Function{$F_{\text{time}}$}{$\mathcal{D}, t$}
            \State $Features_{\text{time}} \gets$ Extract time-based features from $t$
            \State \Return $Features_{\text{time}}$
        \EndFunction
        \Function{$F_{\text{cyclical}}$}{$Features_{\text{time}}$}
            \State $Features_{\text{cyclical}} \gets$ empty list
            \For{feature $f$ in $Features_{\text{time}}$}
                \If{$f$ is 'hour of day'}
                    \State $f_{\sin} \gets \sin\left(\dfrac{2\pi f}{24}\right)$
                    \State $f_{\cos} \gets \cos\left(\dfrac{2\pi f}{24}\right)$
                \ElsIf{$f$ is 'day of week'}
                    \State $f_{\sin} \gets \sin\left(\dfrac{2\pi f}{7}\right)$
                    \State $f_{\cos} \gets \cos\left(\dfrac{2\pi f}{7}\right)$
                \EndIf
                \State Append $f_{\sin}, f_{\cos}$ to $Features_{\text{cyclical}}$
            \EndFor
            \State \Return $Features_{\text{cyclical}}$
        \EndFunction
    \end{algorithmic}
\end{algorithm}


\subsubsection{Temporal, Sine and Cosine Transform-Derived Features}
\label{ch4_sine_cosine_section}
\begin{landscape}
\begin{figure}[!t]
\centering
\includegraphics[width=.8\linewidth]{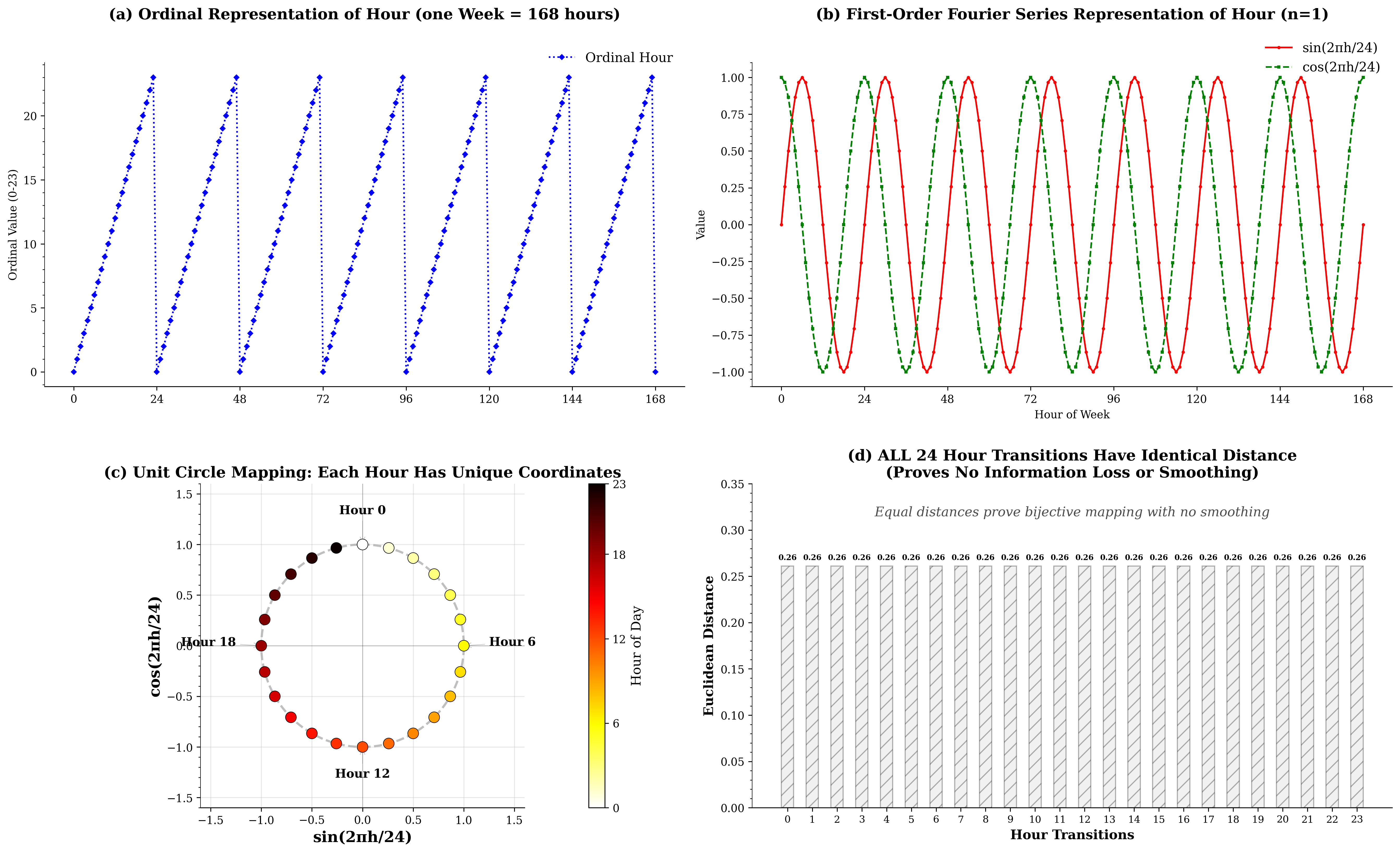}
\caption{\revision{\small Ordinal versus Fourier-based cyclical encoding of time features. (a) Ordinal hours (0-23) over one week show a discontinuity between 23 and 0. (b) First-order Fourier encoding, $\sin(2\pi h/24)$ and $\cos(2\pi h/24)$, yields continuous cyclical features. (c) Unit-circle mapping assigns each hour unique coordinates (bijective, no information loss). (d) Distances between consecutive hours are constant (0.26), confirming the encoding preserves temporal resolution without smoothing, see \autoref{ch4_sine_cosine_section} for more details.}}

\label{ordinal_fourier_representation}
\end{figure}
\end{landscape}

These features exploit temporal data to identify patterns influenced by time, such as distinguishing between on-peak vs off-peak hours and weekdays vs weekends, as shown in \autoref{AutoEnergy_Figure}. In the first function, namely $F_{\text{TIME}}$, of \hyperref[pseudocode_AutoEnergy_Part1]{Algorithm 1}, time-based features are extracted from the timestamp $t_i$ such as the hour of the day as outlined in steps 5-8. These time-related features provide insights into temporal patterns and trends in the data. The second function, namely $F_{\text{CYCLICAL}}$, of \hyperref[pseudocode_AutoEnergy_Part1]{Algorithm 1}, applies Fourier-based transformations, which have been used in time series feature encoding \cite{Verdonck2021}, to temporal variables through steps 9-22, where 'hour of day' ranges from 0 to 23, and 'day of week' ranges from 0 (Monday) to 6 (Sunday) to represent the daily and weekly seasonal cycle. This transformation is  justified by the following key principles:
\begin{itemize}
    \item The Fourier series theorem establishes that periodic patterns, such as energy consumption's daily and weekly seasonality, can be approximated through a combination of sinusoidal components. This mathematical foundation serves as an approach for capturing cyclical consumption patterns across various temporal frequencies.

    \item Sine/cosine transformations map temporal features onto a unit circle (where hour 0 and hour 23 become adjacent points), thus preserving the natural circular topology of time, see \autoref{ordinal_fourier_representation}. This mathematical property ensures that every time point has a unique, continuous representation while maintaining the cyclical relationship between adjacent time periods. More importantly, this transformation is bijective and information-preserving: each hour maps to distinct coordinates with no averaging or smoothing of features, as demonstrated by the uniform Euclidean distances (0.26) between all adjacent hours in \autoref{ordinal_fourier_representation} (d). The continuous appearance of the sine/cosine functions reflects their inherent mathematical properties rather than any loss of temporal resolution; each of the 24 hours retains its unique identity in the transformed space. Unlike linear encoding of hours (0-23), which may create a misleading maximum distance between hour 23 and hour 0, the circular transformation ensures these temporally adjacent hours maintain their true neighbouring relationship. This continuous representation aligns with the physical reality of energy usage patterns, where consumption often changes gradually due to thermal inertia and operational behaviours unless disrupted by sudden events.
\end{itemize}

\subsubsection{Lags and Nested Window Features}
\label{Lags_windows}
The proposed algorithm utilises an automated approach to identify statistically significant lags and computes rolling statistics across nested window sizes using Kendall's tau correlation. These features capture temporal dependencies and multi-scale statistical characteristics in the energy time series data, as shown in \hyperref[pseudocode_AutoEnergy_Part2]{Algorithm 2}, and \autoref{AutoEnergy_Figure}, as follows: 

\label{Pseudocode_AutoEnergy_Part2}

\begin{algorithm}[t]
    \caption{Lags and Nested Window Features}
    \label{pseudocode_AutoEnergy_Part2}
    \footnotesize
    \begin{algorithmic}[3]
        \Require DataFrame $\mathcal{D}$ with target variable $y$
        \Ensure DataFrame with lag and window features
        \State $\mathcal{D}' \gets \mathcal{D}$
        \State $top\_lags \gets F_{\text{lags}}(y)$ \Comment{Find significant lags}
        \State $Features_{\text{lags}} \gets$ Create features using $top\_lags$
        \State $top\_windows \gets F_{\text{stats}}(y)$ \Comment{Find optimal window sizes}
        \State $Features_{\text{stats}} \gets$ empty list
        \For{$window\_size$ in $top\_windows$}
            \State Compute rolling statistics of $y$ over $window\_size$
            \State Add computed statistics to $Features_{\text{stats}}$
        \EndFor
        \Function{$F_{\text{lags}}$}{$y$}
            \State Compute statistical significance for each lag
            \State \Return top 10 significant lags
        \EndFunction
        \Function{$F_{\text{stats}}$}{$y$}
            \State $max\_window\_size \gets \left\lfloor \text{length}(y)/3 \right\rfloor$ 
            \State Evaluate rolling windows up to $max\_window\_size$
            \State \Return top 10 significant window sizes
        \EndFunction
    \end{algorithmic}
\end{algorithm}


\noindent\textbf{A:} In the first function, namely $F_{\text{LAGS}}(y)$ of \hyperref[pseudocode_AutoEnergy_Part2]{Algorithm 2}, and as outlined in steps 10-13,  the algorithm employs Kendall's tau correlation to automatically identify statistically significant lags of  $y_i$ (i.e. the target variable). This process is inspired by the FRESH (Feature Extraction and Scalable Hypothesis Testing) algorithm \cite{Christ2016}, ensuring that only useful lags are retained. Kendall's tau is a rank-based correlation statistic to assess the strength and direction of association between two variables, where its p-value tests the null hypothesis that there is no association between the current and lag values. It is defined as:

\begin{equation}
\tau = \frac{(C - D)}{\sqrt{(C + D + T) \cdot (C + D + U)}}
\end{equation}

where \( C \) is the number of concordant pairs, \( D \) is the number of discordant pairs, \( T \) is the number of ties only in the first variable, and \( U \) is the number of ties only in the second variable. \revision{This approach is justified by the following:}

\begin{itemize}
    \item Non-parametric robustness: Kendall's tau assumes no specific distribution, which is suitable for energy data that may not follow normal distributions due to irregular consumption patterns. It demonstrates superior robustness to outliers compared to Pearson correlation  \cite{Croux2010}, where the latter is sensitive to such anomalies and consumption spikes, which is critical for energy data that frequently contain these irregularities.

    \item Statistical significance: hypothesis testing on Kendall's tau coefficients with a p-value threshold of 0.05 ensures that only lags exhibiting statistically significant correlations with the target variable $y_i$ are retained. Lags with ($p < 0.05$) indicate less than a 5\% probability that the observed correlation is due to chance. This statistical filter ensures that the selected lags preserve genuine temporal dependencies and provides an additional layer of validation for the chosen lags.

    \item {Computational efficiency}: to handle large datasets effectively and avoid highly correlated lags (i.e., reduce multicollinearity), only the top ten lags with the lowest p-values among the significant lags are generated.

\end{itemize}

\noindent\textbf{B:} In the second function, namely $F_{\text{STATS}}(y)$ of \hyperref[pseudocode_AutoEnergy_Part2]{Algorithm 2}, and as outlined in steps 14-18, the algorithm incorporates a nested rolling window approach to compute statistical features at multiple time scales. It evaluates a range of window sizes up to one-third of the series length. However, it is worth acknowledging that this cap is empirical and justified by the following:

\begin{itemize}
    \item Sensitivity analysis: an experiment was conducted across multiple window size thresholds (one-quarter, one-third, one-half, and the full sequence length) to systematically evaluate the impact of the maximum nested-window length on model performance and FE processing time. Although this approach is not theoretically optimal, it remains empirically grounded, as it preserves the automated element of FE while offering a practical trade-off balance between forecasting accuracy and processing efficiency.
    
    \item Computational efficiency and multi-scale pattern capture: limiting the maximum window to one-third of the series length attempts to balance capturing short-term fluctuations, medium-term variations, and longer-term trends while avoiding redundant historical information and containing computational complexity. It prevents the creation of excessively large windows that a) produce substantial overlap between consecutive rolling statistics, leading to multicollinearity; b) increase memory requirements and computational overhead without proportional gains in predictive value; and c) risk over-fitting by incorporating overly broad temporal contexts that may not reflect underlying energy-consumption patterns.

    \item Statistical significance: the algorithm computes Kendall’s tau correlations for the rolling-window statistics and retains window sizes with $p < 0.05$. This statistical filter further ensures that, irrespective of the maximum window-size threshold, only relationships that are statistically significant are preserved, providing an additional layer of validation for the selected window sizes.

\end{itemize}

For the retained window sizes, the algorithm computes the following rolling statistics:

{\small
\setlist[itemize]{leftmargin=*,align=left,label=}
\begin{itemize}
\item Rolling Mean:
\begin{equation}
\bar{x} = \frac{1}{n} \sum_{i=1}^{n} x_i
\end{equation}

\item Rolling Standard Deviation:
\begin{equation}
s = \sqrt{\frac{1}{n-1} \sum_{i=1}^{n} (x_i - \bar{x})^2}
\end{equation}

\item Rolling Maximum:
\begin{equation}
\max = \max(x_1, x_2, ..., x_n)
\end{equation}

\item Rolling Minimum:
\begin{equation}
\min = \min(x_1, x_2, ..., x_n)
\end{equation}

\item Rolling Kurtosis:
\begin{equation}
\text{Kurt} = \frac{\frac{1}{n}\sum_{i=1}^{n}(x_i-\bar{x})^4}{(\frac{1}{n}\sum_{i=1}^{n}(x_i-\bar{x})^2)^2} - 3
\end{equation}

\item Rolling Skewness:
\begin{equation}
\text{Skew} = \frac{\frac{1}{n}\sum_{i=1}^{n}(x_i-\bar{x})^3}{(\frac{1}{n}\sum_{i=1}^{n}(x_i-\bar{x})^2)^{3/2}}
\end{equation}
\end{itemize}}

where $n$ is the size of the rolling window, and $x_i$ are the values within the current window. By calculating these statistics for each retained window size, the algorithm captures different aspects of temporal dynamics in the energy data.\\

\begin{table}
\centering
\footnotesize
\caption{Notations and definitions for all mathematical symbols, variables, and notation used throughout \autoref{ch:Method}.}
\begin{tabular}{cl}
\toprule
\textbf{Symbol} & \textbf{Definition} \\
\midrule
$\mathcal{D}$ & Dataset composed of $N$ instances \\
$N$ & Total number of instances in the dataset \\
$M$ & Number of feature engineering functions \\
$t_i$ & Timestamp associated with the $i$-th instance \\
$y_i$ & Target variable associated with the $i$-th instance \\
\revision{$\mathbf{y}_{1:i-1}$} & \revision{Historical target values available up to time $t_i$} \\
\revision{$\mathbf{F}'_{i,j}$} & \revision{Feature subset generated at time $t_i$ using the $j$-th FE function from $(t_i,\mathbf{y}_{1:i-1})$} \\
$\mathbf{F}'_{i}$ & Complete feature vector for the $i$-th instance \\
$\bigoplus$ & Concatenation operation \\
$\mathcal{M}$ & Predictive model trained within an AutoML framework \\
$F_{\text{TIME}}$ & Temporal feature extraction function \\
$F_{\text{CYCLICAL}}$ & Cyclical feature extraction function \\
$F_{\text{LAGS}}$ & Lag feature extraction function \\
$F_{\text{STATS}}$ & Rolling-window statistical feature extraction function \\
$\tau$ & Kendall's tau correlation coefficient \\
$C, D, T, U$ & Concordant pairs, discordant pairs, ties in first/second variable \\
$n$ & Size of the rolling window \\
$x_i$ & Values within the current rolling window \\
$\bar{x}$ & Rolling mean \\
$s$ & Rolling standard deviation \\
\bottomrule
\end{tabular}

\label{tab:symbol_table}
\end{table}

\subsection{Considerations}
The design of the proposed FE method adheres to the following considerations:

\begin{itemize}
    \item {Computational Efficiency}: the algorithm handles large datasets by limiting the number of generated lags and window sizes based on statistical significance.

    \item {Avoidance of Data Leakage}: features at time $t_i$ are computed using only data available up to $t_i$, preventing look-ahead bias.

    \item {Interpretability}: the generated features have the potential to aid in model explainability. For instance, lag features capture direct historical dependencies, while rolling statistics quantify concepts such as trend (mean), volatility (standard deviation), extremes (max/min), and distribution shape (skewness/kurtosis) over specific time windows, enabling domain experts to understand how each feature contributes to the model's predictions. In other words, rolling means can reveal baseline consumption patterns, standard deviations can identify periods of irregular usage, and lag features can capture recurring behaviours such as daily routines or equipment cycling patterns, allowing energy engineers to understand the factors driving consumption in their systems. \autoref{tab:symbol_table} shows the notations and deﬁnitions for all mathematical symbols, variables, and notations used. \revision{For readability and consistency, notation is defined where first introduced, and \autoref{tab:symbol_table} provides a single consolidated reference for the symbols used in the AutoEnergy method (the most notation-heavy part of the thesis).}

\end{itemize}


\section{Experimental Design}
\label{ch:Experimental_design}
This section outlines the experimental design adopted in this work, including the dataset used in \autoref{datasets_subsection}, benchmarking methods in \autoref{Benchmarking_methods_subsection}, detailed experimental procedure in \autoref{Experimental_procedure}, and lastly the evaluation criteria and statistical tests in \autoref{Evaluation_statistical_tests}.

\subsection{Datasets}
\label{datasets_subsection}
\begin{landscape}
\begin{figure}[!t]
\centering
\includegraphics[width=\linewidth]{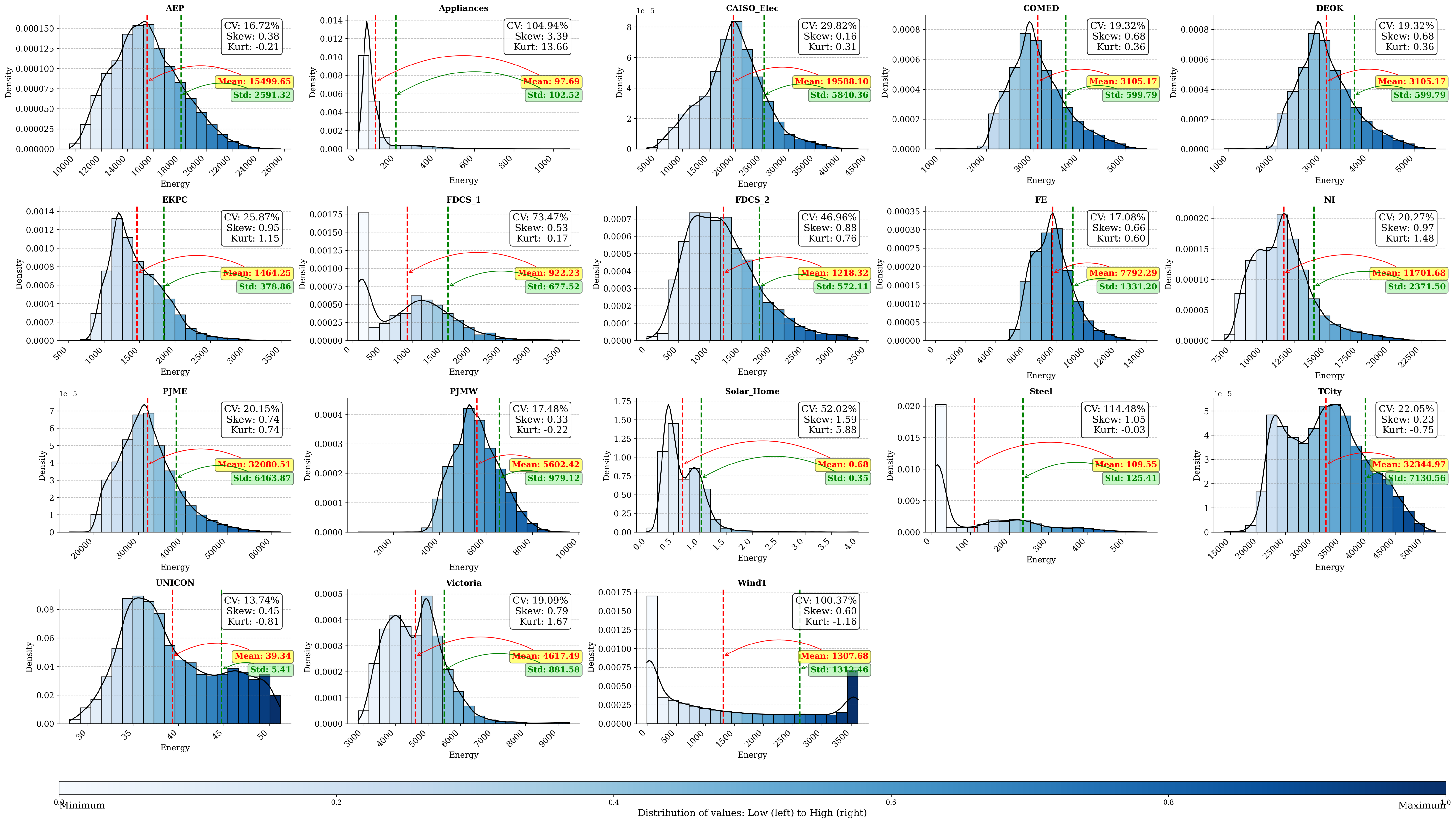}
\caption{The energy datasets used in this work are depicted in histograms of the target variable with colour-coded bars representing normalised bin positions. Red and green dashed lines indicate the mean and standard deviation, respectively. Annotated statistics include coefficient of variation (CV: relative variability), skewness (distribution asymmetry), and kurtosis (tailedness). The colour gradient in the histogram bars represents the distribution of values from low (left) to high (right).}
\label{datasets}
\end{figure}
\end{landscape}

Due to the limited availability of standardised benchmark datasets for ECF using AutoML, this work evaluated the proposed FE method using real-world datasets from related research and repositories. Eighteen energy datasets were employed, spanning a range of energy domains, including residential buildings (e.g., home appliances), industrial and manufacturing facilities (e.g., steel factory), food and drink cold storage facilities, urban or regional energy use, and renewable energy sources (e.g., wind turbines). \autoref{datasets_table} and \autoref{datasets}  provide further details about the datasets used. These datasets provide a comprehensive representation of different energy systems, encompassing both univariate and multivariate data with varying sample sizes and temporal resolutions.


\begin{table}[!t]
\centering
\scriptsize
\caption{Overview of the eighteen energy datasets used in this work. 
Column \textit{Dataset} lists the dataset name. 
Column \textit{Description} summarises the energy system and domain. 
Column \textit{Type} indicates whether the dataset is univariate (Uni) or multivariate (Multi$(k)$, where $k$ is the number of additional features, e.g., weather variables). 
Column \textit{N} gives the total number of samples. 
Column \textit{Resolution} shows the sampling interval (m = minutes, h = hours). 
Column \textit{Total Duration (days)} gives the total time span covered by the dataset in whole days. 
Column \textit{Ref.} cites the source.}

\label{datasets_table}
\renewcommand{\arraystretch}{1.1} 
\begin{tabular} {p{1.5cm} | p{3.7cm} | >{\centering\arraybackslash}  p{.6cm}| >{\centering\arraybackslash}  p{.8cm}| >{\centering\arraybackslash}  p{1.4cm}| >{\centering\arraybackslash}  p{2.1cm}| >{\centering\arraybackslash}  p{.5cm}} 
\toprule
\textbf{Dataset} & \textbf{Description} & \textbf{Type} & \textbf{N} & \textbf{Resolution} & \textbf{Total Duration
(days)} & \textbf{Ref.} \\

\midrule
AEP & Power consumption by American Electric Power & Uni & 121,269 & 1h & 5,053 & \cite{MULLA18} \\
Appliances & Energy consumption data from home appliances & Multi (27) & 19,735 & 10m & 138 & \cite{RN559} \\
CAISO\_Elec & Electricity load by California ISO & Uni & 26,304 & 1h & 1,096 & \cite{Mayes_2024} \\
COMED & Energy usage by Commonwealth Edison & Uni & 57,735 & 1h & 2,406 & \cite{MULLA18} \\
DEOK & Energy consumption by Duke Energy OH/KY & Uni & 57,735 & 1h & 2,406 & \cite{MULLA18} \\
EKPC & Energy usage by East Kentucky Power & Uni & 45,330 & 1h & 1,889 & \cite{MULLA18} \\
FDCS\_1 & Energy usage by Food and Drinks Cold Storage & Multi (9) & 1,944 & 1h & 81 & \cite{Alkhulaifi2024_pipeline} \\
FDCS\_2 & Energy usage by Food and Drinks Cold Storage & Multi (9) & 2,472 & 1h & 103 & \cite{Alkhulaifi2024_pipeline} \\
FE & Power consumption by FirstEnergy & Uni & 62,870 & 1h & 2,620 & \cite{MULLA18} \\
NI & Northern Illinois Hub energy usage & Uni & 58,450 & 1h & 2,436 & \cite{MULLA18} \\
PJME & Energy use in PJM East Region & Uni & 145,362 & 1h & 6,057 & \cite{MULLA18} \\
PJMW & Power consumption in PJM West & Uni & 143,202 & 1h & 5,967 & \cite{MULLA18} \\
Solar\_Home & Ausgrid solar home electricity & Uni & 17,568 & 30m & 366 & \cite{ausgrid2022solar} \\
Steel & Energy usage by AEWOO Steel Co. & Multi (8) & 8,760 & 1h & 365 & \cite{V_E_2020} \\
TCity & Power usage in Tetouan & Multi (3) & 52,416 & 10m & 364 & \cite{RN604} \\
UNICON & La Trobe University electricity consumption & Multi (4) & 8,663 & 1h & 361 & \cite{Moraliyage_2022} \\
Victoria & Electricity demand of 5 AU states & Multi (1) & 20,352 & 1h & 848 & \cite{o2021tsibbledata} \\
WindT & Wind turbine SCADA systems & Multi (2) & 50,530 & 10m & 351 & \cite{RN602} \\
\bottomrule
\end{tabular}
\end{table}


\begin{figure}[htbp]
\centering
\includegraphics[width=1\linewidth]{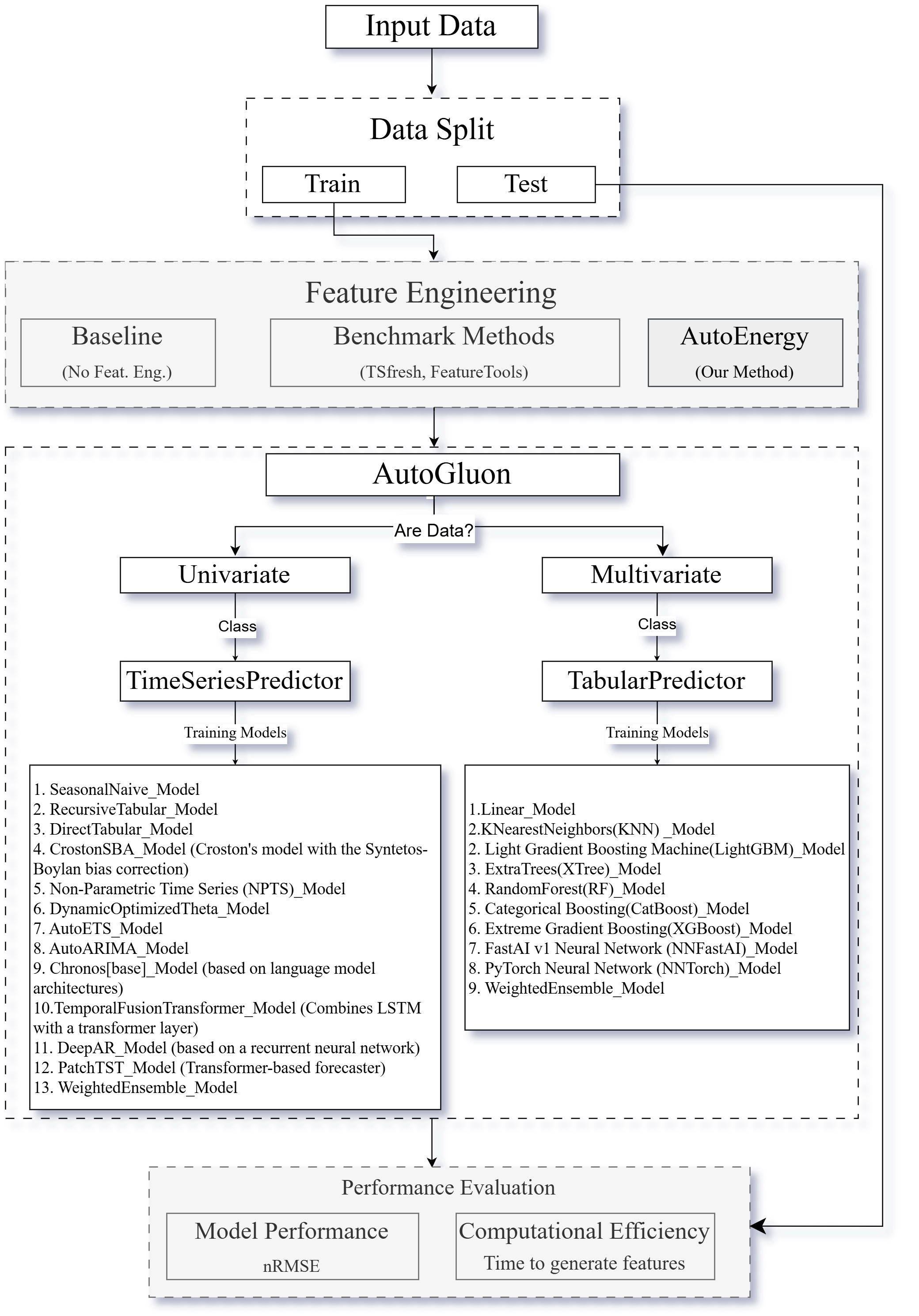}
\caption{Experimental design. See \autoref{Experimental_procedure} for a detailed explanation of the experimental setup.}
\label{experimental_design}
\end{figure}

\begin{figure}
\centering
\includegraphics[width=1\linewidth]{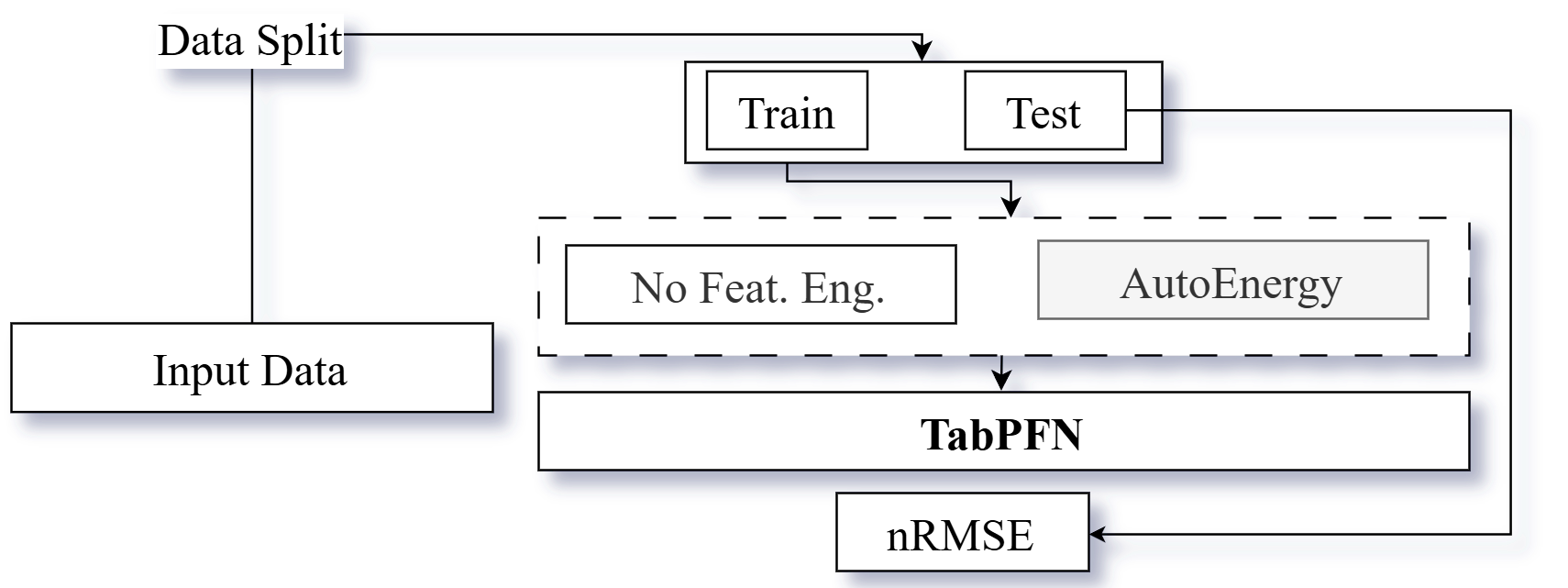}
\caption{Experimental design with TabPFN. See "Special Case TabPFN" in \autoref{Experimental_procedure} for a detailed explanation of this particular case.}
\label{experimental_design_tabPFN}
\end{figure}


\subsection{Benchmarking Feature Engineering Methods}
\label{Benchmarking_methods_subsection}
To demonstrate the effectiveness of the proposed FE method, this work compares key FE methods from the literature, namely TS and FT, as discussed in \autoref{ch2_Automated_Feature_Engineering}, with the proposed method.  These FE methods were selected as they represent leading open-source solutions that have been successfully applied to a variety of forecasting problems, thus providing a robust benchmark across different FE approaches. While energy-specific FE approaches exist in the literature \cite{Zhang2018}, these are predominantly manual, expert-driven methodologies lacking automated elements to minimise domain knowledge reliance, streamline ML model development, and improve AutoML performance. 

Therefore, TS and FT serve as the most comparable AFE baselines available for systematic comparison, as they share AutoEnergy's automated nature while offering extensive documentation and accessible implementations that ensure reproducible and methodologically sound comparative evaluation. This combination of automation capabilities and implementation accessibility makes them the most appropriate benchmarks for rigorous evaluation against the proposed FE method. It is noteworthy that the TS method offers three primary configurations for FE: a) MinimalFCParameters (TSMin), which includes a limited number of features suitable for quick tests; b) EfficientFCParameters (TSEff), which comprises all features generated by the TS method except those marked with the \textit{"high\_comp\_cost"} attributes\footnote{\scriptsize{Features with high computational costs, see: \protect\url{https://tsfresh.readthedocs.io/en/latest/api/tsfresh.feature_extraction.html\#tsfresh.feature_extraction.settings.ComprehensiveFCParameters}.}}; and c) ComprehensiveFCParameters (TSComp), which includes all generated features by the TS method. 

In this work, AutoGluon \cite{AutoGluon2020} was selected as the AutoML framework due to its superior performance in previous research  \cite{Alkhulaifi2024}, where it outperformed other AutoML methods. This selection is justified by: A) logical scientific progression that builds upon previously established findings and maintains focus alignment on comprehensive evaluation of different FE methods against the proposed FE method within the same AutoML framework; B) methodological complexity arising from using multiple AutoML frameworks with different architectures, optimisation strategies, and ensemble approaches, which may introduce confounding variables that obscure the true impact of AFE contributions (i.e., results performance variations may reflect differences between the AutoML frameworks rather than the contribution of the FE methods); and C) computational feasibility challenges associated with evaluating multiple AutoML frameworks across eighteen diverse datasets with multiple FE pipelines. Nevertheless, TabPFN \cite{Hollmann2025}, a state-of-the-art AutoML framework, was incorporated as an additional experiment within the work design to further assess the robustness and potential generalisability of the proposed FE method. In this experiment, all FE methods and models were trained using Python 3.11.\footnote{\scriptsize{All computational experiments were performed on a system featuring an x86\_64 architecture, 64 GB of RAM, and dual Quadro RTX 5000 GPUs.}}

\subsection{Experimental Procedure}
\label{Experimental_procedure}
The experimental procedure and steps conducted in this work, as shown in \autoref{experimental_design}, are as follows:
\begin{enumerate}[label=]
\item \textbf{Stage One}: Partitioning each dataset into two segments: 80\% was allocated to training the AutoGluon models (i.e., train dataset), while the remaining 20\% was reserved for assessing model performance on newly, previously unseen data (i.e., test dataset). This train-test validation method was selected due to its simplicity and computational efficiency, particularly given the extensive experimental design involving 18 datasets, each subjected to five FE methods. Moreover, in AutoGluon, cross-validation is inherently integrated, eliminating the need for a separate validation set, as models are trained on multiple folds of data, with each instance evaluated against the hold-out fold that was not used during training to generate out-of-fold predictions, which are then used to calculate the final cross-validation score \cite{AutoGluon2020}. It is also important to note that the data were not shuffled, as the energy consumption data exhibit temporal patterns, making it essential to preserve the chronological order of the timestamps. 

\item \textbf{Stage Two}: This stage involves two steps. In the first step, AutoGluon models were trained without FE (No.Feat.). This serves as a baseline scenario to assess AutoGluon's forecasting capabilities with minimal input, utilising the \textit{TimeSeriesPredictor}. This class, as shown in \autoref{experimental_design}, includes different forecasting methods, including statistical-based, neural network-based, hybrid, and ensemble methods. This deliberately simplified configuration is designed to replicate AutoML's performance under conditions that simulate: (a) a worst-case scenario (i.e., the absence of domain-specific FE knowledge), and (b) an initial testing phase where the model learns with limited data. In the second step, both benchmarking FE methods, as described in \autoref{Benchmarking_methods_subsection}, and the proposed FE method as described in \autoref{sec:Proposed_Method}, were applied to all datasets as a preprocessing step before training the AutoGluon models with the \textit{TabularPredictor} class.

\item \textbf{Stage Three}: In this final stage of the experimental procedure, the performance of all trained AutoGluon models was evaluated using the test sets and the evaluation metrics outlined in \autoref{Evaluation_statistical_tests}.

\textbf{Special Case TabPFN}: While TabPFN is highly effective \cite{Hollmann2025}, it operates within specific architectural constraints, exhibiting superior performance exclusively on datasets containing up to 10,000 samples and 500 features. Given these limitations, the experimental procedure (see \autoref{experimental_design_tabPFN} comparing TabPFN with versus without the proposed FE algorithm was confined to eight datasets that satisfied these specifications, as shown in \autoref{tab:TabPFN_AutoEnergy_Improvement}.

\end{enumerate}

\subsection{Evaluation Metrics and Statistical Tests}
\label{Evaluation_statistical_tests}
In this work, the nRMSE is used to evaluate and compare the proposed FE method versus the benchmarking methods described in \autoref{Benchmarking_methods_subsection}. This quantitative metric, as depicted in \autoref{eq:nRMSE}, offers a standardised measure of error magnitude. Its normalised nature facilitates quantitative performance comparisons across diverse datasets, regardless of the varying scales of their respective $y_i$ values.
\begin{equation}
\label{eq:nRMSE}
nRMSE = \frac{\sqrt{\frac{1}{n}\sum_{i=1}^{n}(y_i - \hat{y}_i)^2}}{y_{\max} - y_{\min}}
\end{equation}
where $y_i$ is the observed or actual values in the dataset; $\hat{y}_i$ is the values predicted by the model; $n$: the total number of observations in the dataset, $y_{\max}$ and $y_{\min}$ are The highest and lowest observed values in the dataset, respectively.

In addition to nRMSE, the work evaluates the computational efficiency of each FE method by measuring its processing times. This assessment is crucial for understanding the practical applicability of these methods in real-world scenarios, where computational resources may be limited, rapid processing is critical, or when dealing with large datasets. By comparing the execution times across different FE methods, the study provides valuable insights into their scalability and performance trade-offs. In other words, this evaluation helps identify methods that achieve low forecasting errors while maintaining computational cost, thereby offering a comprehensive evaluation of each method's overall effectiveness and efficiency.

To support the experimental findings statistically, non-parametric hypothesis tests were employed to identify significant differences between the methods \cite{Sheskin2003}. The Friedman Aligned-ranks test \cite{Garca2010} is employed to examine the presence of statistically significant differences among the FE methods, with the significance level set at \(\alpha = 0.05\). Following this, the Bonferroni post hoc procedure is applied to determine which specific FE methods have significant differences in the one-versus-many ($1 \ast n$) comparisons performed, where one method (i.e., AutoEnergy) is compared against $n$ other FE methods. Furthermore, the Wilcoxon Signed-Rank test \cite{demvsar2006statistical, garcia2008extension} was employed with $\alpha = 0.05$ to further investigate potential differences between pairs of FE methods that were not identified as significant by the previous tests, ensuring a comprehensive statistical analysis.\footnote{For more information on Machine Learning statistical tests, visit SCI2S's website on Statistical Inference in Computational Intelligence and Data Mining \protect\url{https://sci2s.ugr.es/sicidm}}
\section{Results and Discussion}
\label{ch:Results_Discussion}

This section presents an analysis and discussion of the experimental results. \autoref{Impact_on_AutoML} discusses the overall impact of AutoEnergy on the performance of AutoGluon and TabPFN. Subsequently, \autoref{Comparison_benchmarking_methods} provides a comparative evaluation of the proposed FE method against benchmark approaches, considering both forecasting errors and processing time. This is followed by \autoref{computational_efficiency_sensitivity_analysis}, which discusses computational efficiency and sensitivity analysis, and finally, \autoref{Analysis_Feature_Statistical_Testing} presents the feature importance analysis and statistical testing.

\subsection{Impact of AutoEnergy on AutoML Predictivity}
\label{Impact_on_AutoML}

\begin{landscape}
\begin{figure}
\centering
\includegraphics[width=.9\linewidth]{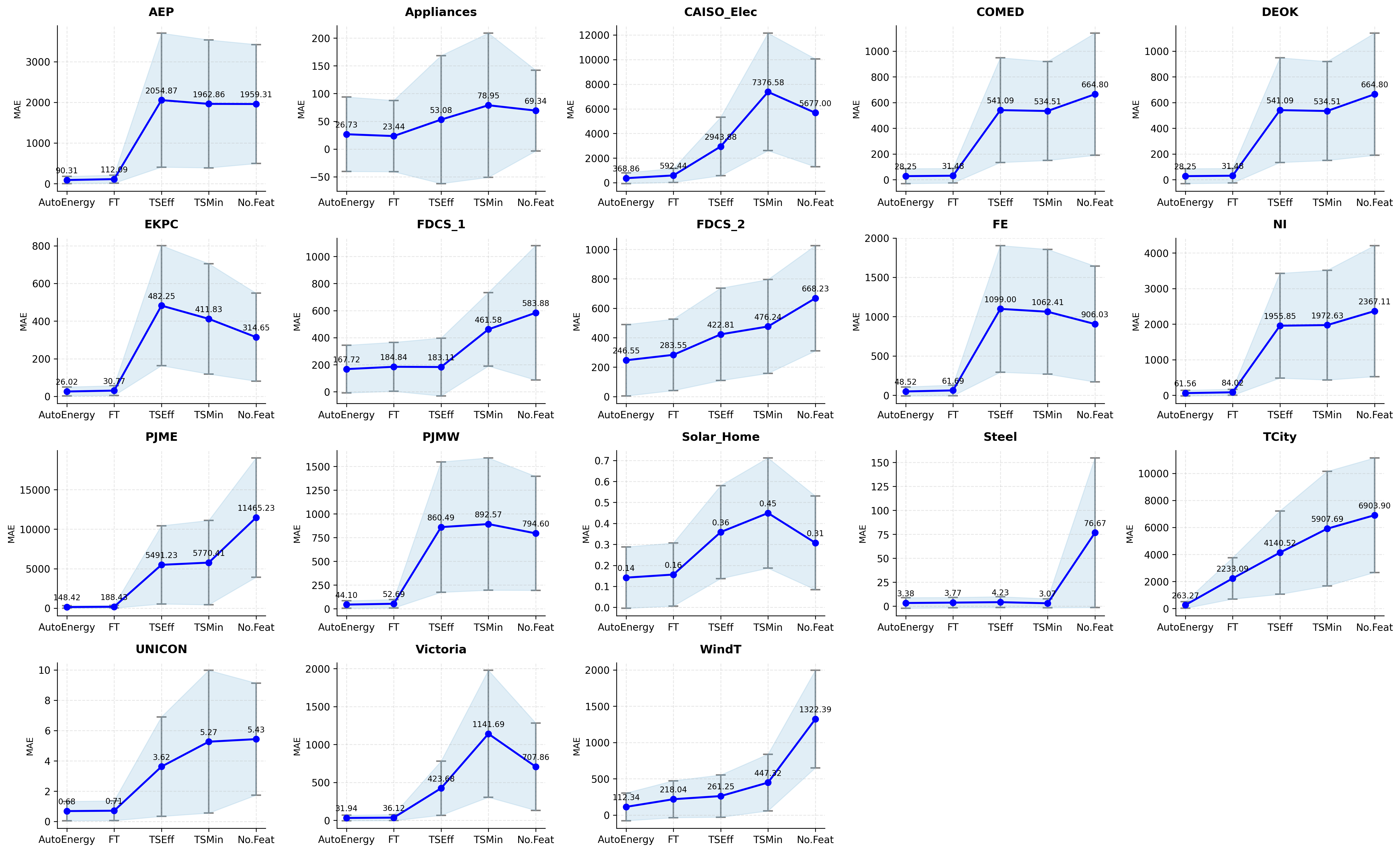}
\caption{Error analysis across test datasets. Each subplot shows the mean absolute error (MAE), computed as \(\text{MAE}=\frac{1}{n}\sum_{i=1}^{n}|y_i-\hat{y}_i|\), between true and predicted values. The black line represents the MAE (lower is better), while the shaded areas and error bars indicate error variability. MAE provides an average magnitude of prediction error, treating all deviations equally and offering insight into prediction consistency that complements the scale-sensitive nRMSE in \autoref{tab:nRMSE_Time}.}
\label{error_bars}
\end{figure}
\end{landscape}

The performance comparison between AutoEnergy and benchmark FE methods across eighteen test sets is presented in \autoref{tab:nRMSE_Time} and \autoref{error_bars}. Despite the diversity in dataset characteristics across different energy settings and environments (detailed in \autoref{datasets_table} and \autoref{datasets}), AutoEnergy consistently enhanced AutoGluon's performance compared to the baseline (i.e., without FE). The results show an average nRMSE reduction of 83.22\%, with improvements (i.e., reduction in forecasting errors) ranging from 28.22\% for the Appliances dataset to 98.39\% for the PJME dataset compared to the baseline. 

This improvement is complemented by a 53.69\% enhancement in prediction stability, as indicated by the lower standard deviation (0.0358 vs 0.0773). This shows the contribution of the proposed FE method to improving AutoGluon's predictive accuracy.  Moreover, the results in \autoref{tab:TabPFN_AutoEnergy_Improvement} indicate that AFE can further enhance the performance of state-of-the-art AutoML methods such as TabPFN \cite{Hollmann2025}. In six of the eight datasets that satisfy TabPFN’s constraints, integrating AutoEnergy with TabPFN yielded the lowest forecasting errors, outperforming FT, TSEff, and TSMin FE methods. The only exceptions were the Steel and Victoria datasets, where TSMin and FT achieved lower errors, respectively. When considered alongside the AutoGluon results in \autoref{tab:nRMSE_Time} and \autoref{error_bars}, these findings indicate that AutoEnergy can potentially generalise across diverse AutoML frameworks, which differ in architecture, model-selection mechanisms, hyper-parameter optimisation strategies, and ensemble approaches. \revision{It is worth clarifying that integration with TabPFN means that AutoEnergy is used purely as a preprocessing step to generate a TabPFN-compatible tabular feature matrix, after which TabPFN is applied unchanged as the predictive learner (see \autoref{Experimental_procedure} for more details on the experimental setup with TabPFN). Accordingly, \textit{successful integration} refers both to practical compatibility (i.e., AutoEnergy outputs satisfy TabPFN’s input constraints) and to empirical benefit (i.e., reduced forecasting error on test sets).}


\begin{landscape}
\begin{table}
\centering
\caption{\footnotesize Results obtained on all the test sets (N=18). nRMSE values and processing times in seconds (lower is better) for the proposed AutoEnergy method compared to other FE methods. Bold values highlight the lowest nRMSE and fastest processing time for each dataset. nRMSE gives more weight to larger errors due to the squaring operation, making it more sensitive to outliers. It is normalised by the range (max–min) of the target variable, allowing for fair comparison of forecasting error across datasets with different scales. \autoref{Appendix_A_AutGluon} in the appendices provides additional results of RMSE, MAE, MAPE and R$^2$. Note that the TScomp configuration is not reported, as it proved computationally infeasible in our experiments (runs crashed after several hours of running due to excessive memory use). We include this note to highlight the practical computational limits of exhaustive FE approaches (e.g., TSComp) in real-world energy forecasting and to underscore the importance of considering computational feasibility and practicality in future work.}
\fontsize{9pt}{7pt}\selectfont
\renewcommand{\arraystretch}{2} 

\begin{tabular}{p{1.8cm}|p{1.1cm}p{1.1cm}|p{1.1cm}p{1.1cm}|p{1.1cm}p{1.1cm}|p{1.1cm}p{1.1cm}|p{1cm}p{1.05cm}}
\toprule
& \multicolumn{10}{c}{\textbf{FE Methods}} \\
\cmidrule{2-11}
\textbf{Dataset} & \multicolumn{2}{c|}{\textbf{AutoEnergy}} & \multicolumn{2}{c|}{\textbf{FT}} & \multicolumn{2}{c|}{\textbf{TSMin}} & \multicolumn{2}{c|}{\textbf{TSEff}} & \multicolumn{2}{c}{\textbf{No.Feat.}} \\
& nRMSE & Time & nRMSE & Time & nRMSE & Time & nRMSE & Time & nRMSE & Time \\
\midrule
AEP & \textbf{0.0096} & 1585.90 & 0.0113 & 1576.84 & 0.1659 & \textbf{636.50} & 0.1737 & 4280.70 & 0.1856 & 0.00 \\
Appliances & 0.0870 & \textbf{44.33} & \textbf{0.0824} & 382.34 & 0.1831 & 177.06 & 0.1529 & 1411.01 & 0.1212 & 0.00 \\
CAISO\_Elec & \textbf{0.0163} & \textbf{88.30} & 0.0232 & 211.91 & 0.2495 & 211.06 & 0.1074 & 1468.03 & 0.2036 & 0.00 \\
COMED & \textbf{0.0150} & 372.41 & 0.0151 & 558.55 & 0.1535 & \textbf{352.04} & 0.1579 & 2506.55 & 0.1905 & 0.00 \\
DEOK & \textbf{0.0150} & \textbf{360.22} & 0.0151 & 561.21 & 0.1535 & 361.33 & 0.1579 & 2502.54 & 0.1905 & 0.00 \\
EKPC & \textbf{0.0135} & \textbf{226.76} & 0.0155 & 416.53 & 0.1931 & 295.27 & 0.2208 & 2089.80 & 0.1499 & 0.00 \\
FDCS\_1 & \textbf{0.0861} & \textbf{2.03} & 0.0914 & 14.65 & 0.1893 & 22.80 & 0.0996 & 51.06 & 0.2711 & 0.00 \\
FDCS\_2 & \textbf{0.1173} & \textbf{2.79} & 0.1266 & 18.62 & 0.1944 & 25.21 & 0.1786 & 66.12 & 0.2571 & 0.00 \\
FE & \textbf{0.0097} & 427.77 & 0.0119 & 625.56 & 0.1719 & \textbf{262.22} & 0.1766 & 2035.33 & 0.1513 & 0.00 \\
NI & \textbf{0.0070} & \textbf{376.61} & 0.0083 & 564.64 & 0.1732 & 425.00 & 0.1693 & 3111.96 & 0.2074 & 0.00 \\
PJME & \textbf{0.0059} & 2061.00 & 0.0068 & 2492.46 & 0.2104 & \textbf{666.59} & 0.1979 & 4663.42 & 0.3675 & 0.00 \\
PJMW & \textbf{0.0100} & 2205.06 & 0.0117 & 2321.66 & 0.1834 & \textbf{676.22} & 0.1784 & 4304.59 & 0.1682 & 0.00 \\
Solar\_Home & \textbf{0.0808} & \textbf{39.66} & 0.0860 & 133.49 & 0.2065 & 131.43 & 0.1673 & 853.15 & 0.1507 & 0.00 \\
Steel & \textbf{0.0104} & \textbf{13.95} & 0.0122 & 62.22 & 0.0122 & 49.94 & 0.0132 & 266.38 & 0.2040 & 0.00 \\
TCity & \textbf{0.0123} & \textbf{320.00} & 0.0922 & 506.37 & 0.2478 & 439.00 & 0.1759 & 3276.16 & 0.2760 & 0.00 \\
UNICON & \textbf{0.0408} & \textbf{13.23} & 0.0421 & 61.29 & 0.3090 & 50.78 & 0.2137 & 263.87 & 0.2875 & 0.00 \\
Victoria & \textbf{0.0100} & \textbf{55.47} & 0.0110 & 158.34 & 0.2911 & 170.75 & 0.1138 & 1088.43 & 0.1877 & 0.00 \\
WindT & \textbf{0.0614} & \textbf{299.46} & 0.0930 & 469.73 & 0.1647 & 431.06 & 0.1084 & 3197.56 & 0.4115 & 0.00 \\
\midrule
\textbf{Mean} & \textbf{0.0338} & 471.94 & 0.0420 & 618.68 & 0.1918 & \textbf{299.12} & 0.1535 & 2079.81 & 0.2212 & 0.0 \\
\textbf{Std. Dev.} & \textbf{0.0358} & 705.75 & 0.0404 & 744.73 & 0.0633 & \textbf{216.15} & 0.0495 & 1507.69 & 0.0773 & 0.0 \\
\bottomrule
\end{tabular}
\label{tab:nRMSE_Time}
\end{table}
\end{landscape}


 The consistent improvements across diverse energy systems may suggest that underlying consumption patterns share common characteristics that AutoEnergy's proposed functions, explained in \autoref{sec:Proposed_Method}, can effectively capture and transform raw energy data into useful predictive features for training AutoML models. Such findings demonstrate AutoEnergy's effectiveness in automating the FE process for ECF problems without requiring domain expertise. Meanwhile, the limited effectiveness of AutoML frameworks in specialised domains such as ECF can be attributed to their general-purpose design, which prioritises broad applicability across ML tasks at the expense of domain-specific optimisations that FE methods, such as AutoEnergy, can provide.

\begin{table}[!t]
\centering
\caption{nRMSE for TabPFN without FE versus with FE methods across datasets that satisfy the TabPFN constraints ($N \le 10{,}000$, $D \le 500$). Lower values are better. Boldface indicates the best results. \autoref{Appendix_A_TabPFN} in the appendices provides additional results of RMSE, MAE, MAPE and R$^2$.}
\footnotesize
\begin{tabular}{p{2.5cm}|c|c|c|c|c}
\toprule
\textbf{Dataset} & \textbf{TabPFN} & \textbf{AutoEnergy} & \textbf{FT} & \textbf{TSEff} & \textbf{TSMin} \\
\midrule
Appliances & 0.1412 & \textbf{0.0710} & 0.0712 & 0.1153 & 0.1497 \\
FDCS\_1    & 0.1549 & \textbf{0.0718} & 0.0778 & 0.0984 & 0.1822 \\
FDCS\_2    & 0.1504 & \textbf{0.1128} & 0.1278 & 0.1514 & 0.1620 \\
Steel      & 0.0080 & 0.0081 & 0.0082 & 0.0135 & \textbf{0.0076} \\
TCity      & 0.2691 & \textbf{0.0184} & 0.0212 & 0.0701 & 0.2299 \\
UNICON     & 0.2226 & \textbf{0.0391} & 0.0422 & 0.2026 & 0.1963 \\
Victoria   & 0.1496 & 0.0158 & \textbf{0.0146} & 0.0994 & 0.1453 \\
WindT      & 0.1792 & \textbf{0.0841} & 0.0974 & 0.1595 & 0.1269 \\
\bottomrule
\end{tabular}

\label{tab:TabPFN_AutoEnergy_Improvement}
\end{table}

\subsection{Comparison with Benchmark Methods}
\label{Comparison_benchmarking_methods}

The proposed AutoEnergy FE method demonstrated an overall superior performance across test sets compared to the benchmarking methods. This is evident by achieving a mean nRMSE of 0.0338 compared to FT (0.042), TSMin (0.1918), and TSEff (0.1535), as presented in \autoref{tab:nRMSE_Time}. This represents average reductions in forecasting errors of 19.52\%, 82.38\%, and 77.98\%, respectively, over these methods. In terms of computational efficiency (i.e., time needed for FE), AutoEnergy achieved a mean processing time of 471.94 seconds, which was 1.31x faster (23.72\% improvement) than FT (618.68s) and 4.41x faster (77.31\% improvement) than TSEff (2079.81s), though 1.58x slower (57.78\% slower) than TSMin (299.12s). While the latter showed faster processing time, AutoEnergy reduced forecasting errors by 82.38\% on average compared to TSMin. \revision{Note that the comparison is intended to evaluate predictive performance and computational overhead against other automated FE baselines, not to quantify person-hours saved relative to manual FE.}

\begin{figure}
\centering
\includegraphics[width=1\linewidth]{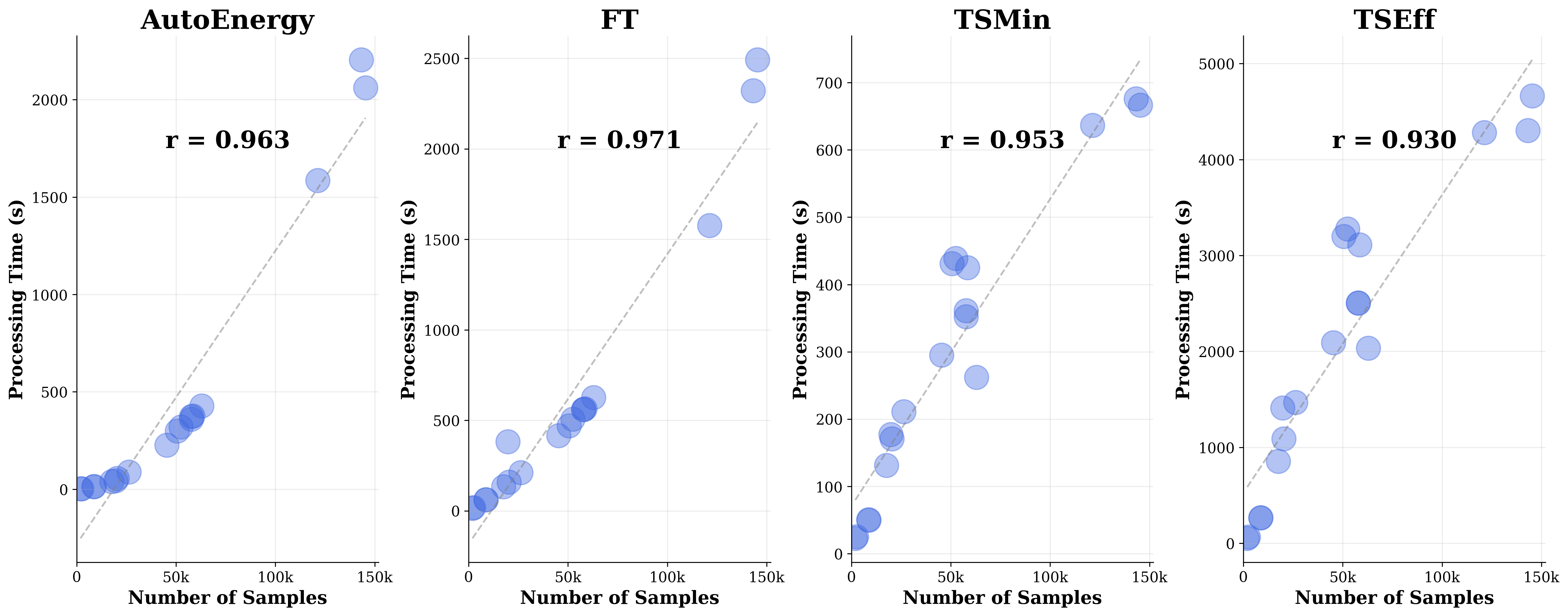}
\caption{Pearson correlation between dataset size (i.e., number of samples) and processing time for each method across all eighteen energy datasets. Markers represent individual datasets.}
\label{Fig:Pearson_correlation}
\end{figure}

As shown in \autoref{tab:nRMSE_Time}, processing time varies substantially across datasets, primarily driven by dataset size (i.e., the number of samples). \autoref{datasets_table} provides an overview of the characteristics of the eighteen energy datasets, including their sample sizes. Notably, larger datasets require significantly more intensive computations during AFE. This relationship is evident in specific examples: high-sample datasets such as PJME (N = 145,362; AutoEnergy = 2,061s; FT = 2,492.4s; TSEff = 4,663.4s) exhibit considerably longer processing times due to the computational scaling of AFE operations, whereas smaller datasets such as FDCS\_1 (N = 1,944; AutoEnergy = 2s; FT = 14.6s; TSEff = 2s) are processed much faster. Importantly, this pattern is not unique to AutoEnergy but is consistently observed across all benchmarking methods. This observation is further supported by \autoref{Fig:Pearson_correlation}, which shows that all FE methods display strong positive Pearson correlations ($r = 0.93$--$0.97$) between dataset size and processing time. This systematic relationship confirms that the computational overhead is intrinsic to AFE, rather than a limitation specific to AutoEnergy.

AutoEnergy's performance advantage is less pronounced on datasets with skewed distributions that also include exogenous features (Appliances with 27 features and Steel with 8 features, see \autoref{datasets_table}). These additional features, such as weather-related, already capture useful predictive information, thereby reducing the relative improvement that AutoEnergy's temporal FE can provide. Furthermore, the skewed distributions in these datasets may suggest that consumption patterns are dominated by extreme values or specific operational modes, where exogenous variables are likely stronger predictors than temporal patterns. When datasets already contain informative variables that directly influence consumption, the marginal benefit of AutoEnergy is likely to be diminished, though AutoEnergy maintains competitive performance, achieving the lowest nRMSE for the Steel dataset while remaining competitive for the Appliances dataset, where FT performs best.


\begin{table}[!t]
\centering
\caption{Sensitivity analysis results (nRMSE and processing time) for the time series fraction constraint defining the candidate range for nested window statistical feature generation, where only features with statistically significant Kendall's tau correlation within this range are retained to generate the final model input features. The examined lengths are: quarter (1/4), one-third (1/3), half (1/2), and full sequence length (1). This experiment evaluates the impact of varying maximum rolling window fractions on predictive model performance and efficiency, using solely the nested window features as inputs to isolate their standalone contributions and inform optimal parameter selection. See \autoref{Lags_windows} for more details.}
\fontsize{7pt}{7pt}\selectfont
\renewcommand{\arraystretch}{1.6}
\begin{tabular}{p{1.45cm}|rr|rr|rr|rr}
\toprule
\textbf{Dataset} & \multicolumn{8}{c}{\textbf{Maximum Length}} \\
\cmidrule{2-9}
 & \multicolumn{2}{c|}{\textbf{1/4}} & \multicolumn{2}{c|}{\textbf{1/3}} & \multicolumn{2}{c|}{\textbf{1/2}} & \multicolumn{2}{c}{\textbf{1}} \\
 & nRMSE & Time & nRMSE & Time & nRMSE & Time & nRMSE & Time \\
\midrule
AEP & 0.2241 & 1498.01 & 0.2229 & 1634.39 & \textbf{0.2227} & 1816.42 & 0.3124 & 2110.72 \\
Appliances & 0.2492 & 40.28 & 0.2447 & 44.35 & 0.2966 & 50.82 & \textbf{0.2429} & 63.48 \\
CAISO\_Elec & 0.2002 & 85.50 & 0.1776 & 92.99 & \textbf{0.1669} & 104.12 & 0.1849 & 125.12 \\
COMED & 0.1519 & 367.54 & 0.1513 & 388.01 & \textbf{0.1498} & 439.48 & 0.1826 & 517.67 \\
DEOK & 0.1508 & 359.59 & 0.1501 & 384.05 & \textbf{0.1465} & 432.18 & 0.1811 & 511.69 \\
EKPC & 0.1625 & 224.61 & \textbf{0.1560} & 243.71 & 0.1735 & 279.53 & 0.2243 & 326.94 \\
FDCS\_1 & 0.1723 & 1.73 & \textbf{0.1626} & 1.85 & 0.1729 & 2.10 & 0.1690 & 2.68 \\
FDCS\_2 & 0.1558 & 2.32 & 0.1545 & 2.46 & \textbf{0.1507} & 2.79 & 0.1509 & 3.55 \\
FE & 0.2009 & 432.82 & 0.2180 & 460.15 & \textbf{0.1674} & 524.25 & 0.1708 & 617.90 \\
NI & 0.1906 & 378.06 & 0.1977 & 412.28 & 0.2182 & 464.61 & \textbf{0.1812} & 537.66 \\
PJME & 0.2289 & 2030.82 & 0.2289 & 2173.77 & 0.2289 & 2427.23 & \textbf{0.1883} & 2808.38 \\
PJMW & 0.1876 & 1826.87 & \textbf{0.1863} & 1987.15 & 0.1871 & 2219.01 & 0.1887 & 2570.72 \\
Solar\_Home & 0.1766 & 38.03 & 0.1892 & 41.23 & 0.2003 & 46.85 & \textbf{0.1660} & 57.38 \\
Steel & 0.0117 & 13.54 & \textbf{0.0112} & 14.60 & 0.0121 & 16.54 & 0.0122 & 20.39 \\
TCity & 0.2574 & 275.11 & \textbf{0.2078} & 297.74 & 0.2349 & 332.94 & 0.2088 & 389.78 \\
UNICON & 0.2458 & 12.90 & 0.2458 & 13.90 & \textbf{0.2133} & 15.73 & 0.2594 & 19.56 \\
Victoria & \textbf{0.1377} & 53.73 & 0.1395 & 57.58 & 0.1711 & 65.52 & 0.1680 & 79.12 \\
WindT & 0.1294 & 247.24 & \textbf{0.1217} & 267.58 & 0.1348 & 300.13 & 0.1483 & 355.55 \\
\bottomrule
\end{tabular}

\label{tab:maxlen_results}
\end{table}

\autoref{tab:maxlen_results} reveals that capping the nested window length at one-third of each series offers a favourable accuracy–cost trade-off for ECF across the eighteen datasets, as evidenced by its lowest average nRMSE of 0.1758 and second-fastest processing time overall. It achieves the minimum error in 6 of 18 datasets, versus 1, 4, and 7 for the quarter, full-length and half caps, respectively, yet the latter was considerably slower in processing time compared to the one-third cap. Although such findings remain data-dependent, with some datasets showing sub-optimal results under the one-third constraint, this configuration consistently delivers near-optimal performance across diverse energy datasets representing different energy systems, offering a balance between forecasting accuracy and computational efficiency for AFE in ECF tasks; it thus minimises the need for expert input and manual intervention, enabling fully automated, end-to-end ECF systems.


\subsection{Computational Efficiency and Sensitivity Analysis}
\label{computational_efficiency_sensitivity_analysis}

While computational efficiency claims are often dataset-dependent, AutoEnergy shows promising results in scalability relative to other existing well-established FE methods.  For instance, for the two datasets exceeding 140,000 samples (PJME = 145,362; PJMW = 143,202 – see \autoref{datasets_table} for dataset characteristics), AutoEnergy maintained computational feasibility across those cases. More precisely,  it achieved faster processing times (2,061s and 2,205s, respectively) compared to FT (2,492s and 2,322s) and significantly outperformed TSEff (4,663s and 4,305s). However, more rigorous scalability evaluation on very large datasets (e.g., millions of samples) and streaming data scenarios is therefore recommended for future work to fully establish the method's scalability characteristics. It is worth noting, though, that AFE methods such as AutoEnergy are particularly beneficial for smaller datasets, where raw data often lacks sufficient informative patterns, whereas very large datasets may inherently contain rich relationships that could reduce the relative necessity for extensive FE.

As real-world datasets are often small due to cost and privacy concerns \cite{grinsztajn2022tree, hollmann2022tabpfn}, AutoEnergy's performance on small datasets is particularly noteworthy. For the three smallest datasets (FDCS\_1: 1,944 samples, FDCS\_2: 2,472 samples, and UNICON: 8,663 samples), AutoEnergy reduces nRMSE error to an average of 0.0814, showing significant improvements compared to baseline (0.2719, 70.06\% lower error), TSMin (0.2309, 64.75\% lower error), TSEff (0.1640, 50.36\% lower error), and FT (0.0867, 6.11\% lower error). Regarding the processing time for these datasets, AutoEnergy generates features in 6.02 seconds on average, which is 21.1x faster than TSEff (127.02s), 5.5x faster than TSMin (32.93s), and 5.2x faster than FT (31.52s). Although this observation is data-dependent, AutoEnergy shows better performance on small datasets, improving both forecasting accuracy and computational efficiency compared to the benchmarking methods. This superior performance could be attributed to AutoEnergy being specifically designed for ECF problems, compared to the benchmarking FE methods.



\begin{figure}[!t]
\centering
\includegraphics[width=1\linewidth]{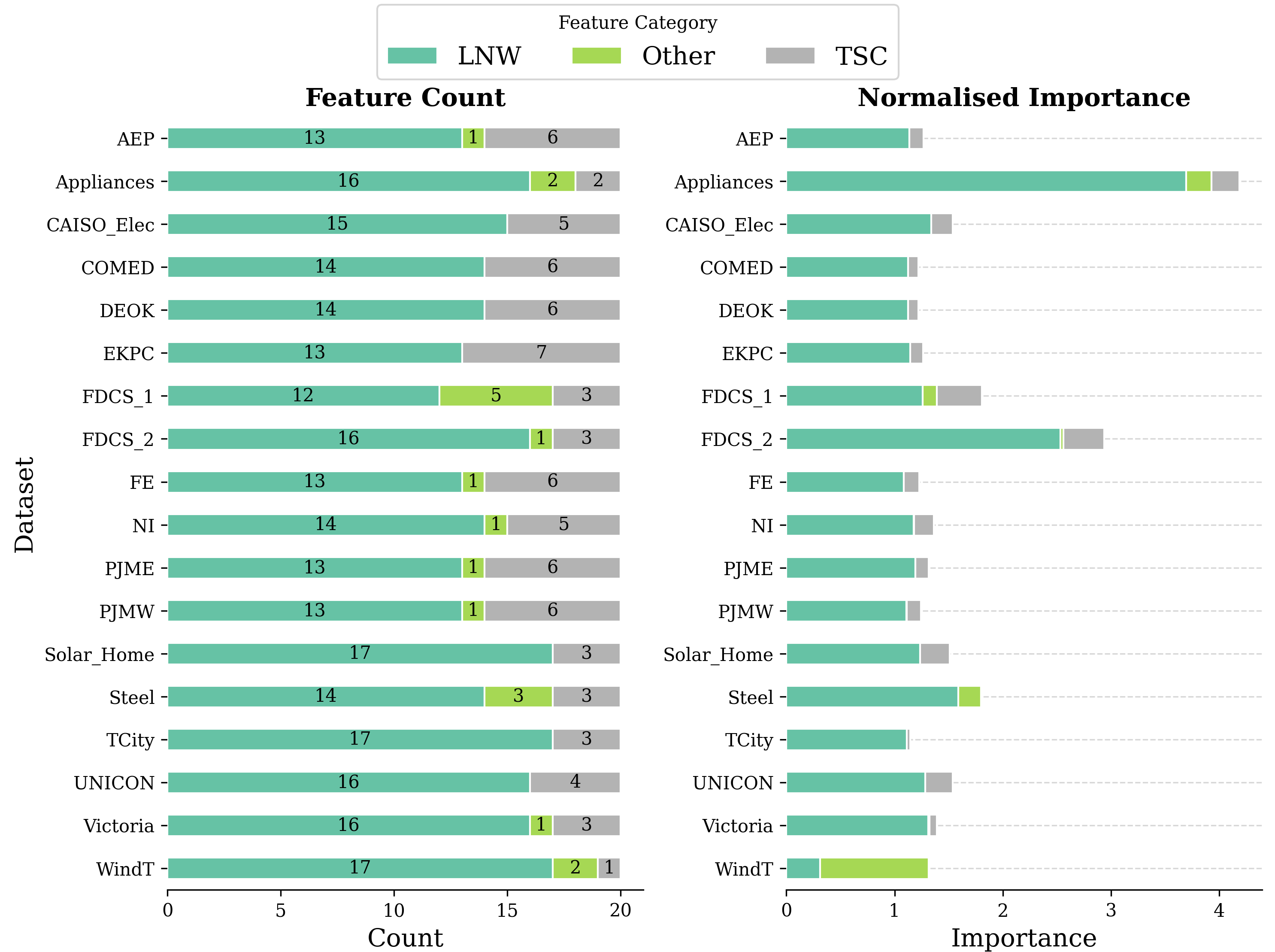}
\caption{\revision{\small Comparison of the top 20 most important features across datasets, grouped into three categories: Temporal, Sine and Cosine Transform-Derived Features (TSC; \hyperref[pseudocode_AutoEnergy_Part1]{Algorithm 1}), Lags and Nested Window Features (LNW; \hyperref[pseudocode_AutoEnergy_Part2]{Algorithm 2}), and Other (e.g., weather). The left panel shows feature counts, and the right panel shows cumulative normalised importance (proportion of total importance). LNW features contribute most, consistent with their ability to capture temporal dependencies in energy consumption, including lagged effects such as thermal inertia \cite{MartnezComesaa2020} and multi-scale behaviour via nested windows. More analysis is provided in \autoref{Analysis_Feature_Statistical_Testing}, while the full feature-importance scores are provided in the supplementary material (see \autoref{Appendix_B} in the appendices).}}
\label{Feature_importnace_AutoEnergy}
\end{figure}



\subsection{Analysis of Feature Importance and Statistical Testing}
\label{Analysis_Feature_Statistical_Testing}

The importance and distribution of the top 20 features across the examined energy datasets, computed by AutoGluon, are shown in \autoref{Feature_importnace_AutoEnergy}, where feature importance is calculated using permutation importance \cite{AutoGluon2020}. It measures the decrease in model performance when the values of a specific feature are randomly shuffled, thereby quantifying each feature's contribution to predictive accuracy in a model-agnostic manner \footnote{\begin{minipage}[t]{\linewidth}\raggedright
For more information, see 
\url{https://auto.gluon.ai/dev/api/autogluon.tabular.TabularPredictor.feature_importance.html} and 
\url{https://auto.gluon.ai/dev/api/autogluon.timeseries.TimeSeriesPredictor.feature_importance.html}.
\end{minipage}}. The results reveal that features generated by AutoEnergy are the most predictive, even though eight of the eighteen datasets have additional features such as weather attributes. In particular, statistically significant lags and nested window features, as explained in \autoref{Lags_windows}, appear to be the dominant predictors contributing to model performance. This may suggest that lag features effectively capture historical influences, such as thermal inertia in buildings \cite{MartnezComesaa2020}, where past consumption directly affects immediate future demand due to gradual changes in heating or cooling systems, revealing that energy behaviours may exhibit persistence and path-dependency rather than abrupt shifts. Meanwhile, the effectiveness of nested window features may indicate that energy consumption operates across multiple temporal scales (i.e., short-term windows capture immediate operational fluctuations while longer-term windows encode baseline trends and seasonal adjustments). Because these temporally explicit predictors may map directly to operational concepts such as “daily average demand,” practitioners can trace a model’s output back to concrete, intuitive drivers, thereby markedly enhancing interpretability. In spite of that, feature importance is often dataset-dependent, so this observation may vary across different datasets and settings. Further feature-importance scores and details, including eighteen tables (one for each dataset), are provided in the supplementary material (see \autoref{Appendix_B} ).

Beyond feature-importance rankings, the engineering of temporal, cyclical, lag, and nested features by AutoEnergy, as explained in \autoref{sec:Proposed_Method}, enhances interpretability and transparency in several useful ways. First, sinusoidal time encodings (sine/cosine of hour-of-day and day-of-week) map directly to familiar operational cycles; their effects, therefore, can be read as “daily/weekly seasonality” rather than opaque latent factors. Second, lagged and windowed statistics (e.g., load at $t-24h$ or the previous 24-hour mean) preserve units and time scales, making the direction and magnitude of their influence predictable (higher recent demand plausibly raises near-term forecasts), which may support counterfactual “what-if” reasoning and analysis. Third, AutoEnergy’s design is a rule-based (heuristic) search algorithm that applies explicit, auditable transformations (temporal shifts, rolling aggregates, and trigonometric projections), keeping provenance clear and making each generated feature human-readable, traceable, and reproducible, thereby strengthening the interpretability advantage of the method. Finally, because the generated features are emitted as standard tabular columns with fixed names and units, they can be integrated directly into AutoML frameworks that provide built-in feature-importance tools (e.g., AutoGluon), enabling non-experts to see how engineered features influence forecasts and thereby preserving interpretability in fully automated, end-to-end ECF workflows.

\revision{It is worth clarifying that the interpretability reported in this experiment is limited by the choice of AutoML method (i.e., AutoGluon) rather than by AutoEnergy itself. Since AutoEnergy is evaluated inside AutoGluon’s AutoML pipeline, the final predictor may be a stacked ensemble of heterogeneous learners, and AutoGluon therefore exposes a uniform, model-agnostic explanation interface via permutation-based feature importance \cite{AutoGluon2020}. This provides a comparable global ranking of engineered features across datasets; however, it does not yield a single, consistent set of richer model-specific or instance-level explanations across all candidate models and ensembles considered during the AutoML search. Accordingly, the feature-importance results in this section (see \autoref{Appendix_B} in the appendices for detailed feature-importance scores) should be interpreted as a practical diagnostic of which engineered feature families matter most under the chosen AutoML framework, rather than as a complete explanation of the learned forecasting function.}

\begin{figure}[!t]
\centering
\includegraphics[width=1\linewidth]{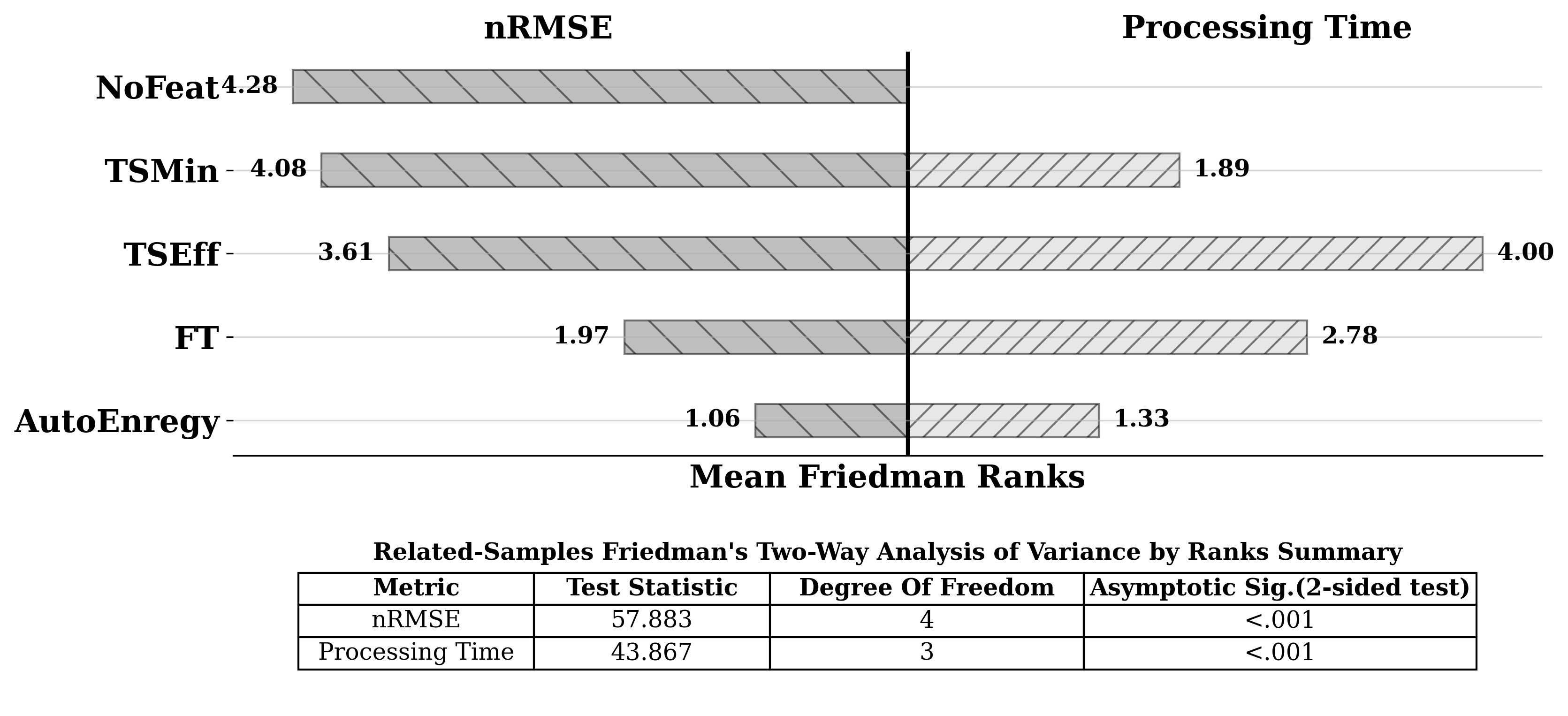}
\caption{Comparisons between the proposed FE algorithm and other FE methods using the Friedman Average Ranking (lower is better).}
\label{Friedman_Ranking}
\end{figure}


To evaluate the statistical significance of performance differences, both the Friedman and the Wilcoxon tests were conducted as detailed in \autoref{Evaluation_statistical_tests}. Friedman’s test revealed significant differences among the methods ($p < .001$), with AutoEnergy achieving the best mean rank for nRMSE (1.06), followed by FT (1.97), TSEff (3.61), TSMin (4.08), and No.Feat. (4.28), as presented in \autoref{Friedman_Ranking}. Post hoc analysis using Bonferroni correction demonstrated that AutoEnergy significantly outperformed TSEff, TSMin, and No.Feat. in terms of nRMSE (all $p < .001$), as shown in \autoref{tab:Friedman_test}. While the Friedman test with Bonferroni adjustment showed no significant difference between AutoEnergy and FT ($p = .820$), the pairwise Wilcoxon test, which specifically examines the direct comparison between these two methods, revealed a statistically significant improvement ($p = .001$) as depicted in \autoref{tab:Wilcoxon_Test}. The discrepancy may stem from the Friedman test’s omnibus design and conservative Bonferroni adjustment: the original AutoEnergy-FT comparison was non-significant pre-adjustment ($p=0.082$), and the correction further attenuated this effect ($p=0.820$), whereas the Wilcoxon test, focused solely on this pairwise comparison without multiple-testing penalties, thereby detected a statistically significant difference ($p=0.001$) that the more conservative post hoc analysis may have overlooked.


\begin{table}[!b]
\caption{Friedman test results for nRMSE and processing time comparisons.}
\label{tab:Friedman_test}
\footnotesize
\begin{tabular}{lp{1.65cm}p{1cm}p{1.65cm}p{1cm}p{1cm}}
\toprule
\textbf{Comparison} & \textbf{Test Statistic} & \textbf{Std. Error} & \textbf{Std. Test Statistic} & \textbf{Sig.} & \textbf{Adj. Sig.$^a$} \\
\midrule
\multicolumn{6}{l}{\textbf{nRMSE}} \\
AutoEnergy-FT & -0.917 & 0.527 & -1.739 & 0.082 & 0.820 \\
AutoEnergy-TSEff & -2.556 & 0.527 & -4.849 & $<$0.001 & 0.000 \\
AutoEnergy-TSMin & -3.028 & 0.527 & -5.745 & $<$0.001 & 0.000 \\
AutoEnergy-No.Feat. & -3.222 & 0.527 & -6.114 & $<$0.001 & 0.000 \\
\midrule
\multicolumn{6}{l}{\textbf{Processing Time}} \\
AutoEnergy-TSMin & -0.556 & 0.430 & -1.291 & 0.197 & 1.000 \\
AutoEnergy-FT & -1.444 & 0.430 & -3.357 & $<$0.001 & 0.005 \\
AutoEnergy-TSEff & -2.667 & 0.430 & -6.197 & $<$0.001 & 0.000 \\
\bottomrule
\multicolumn{6}{p{0.95\textwidth}}{Note: Each row tests the null hypothesis that distributions are the same.} \\
\multicolumn{6}{l}{$^a$Bonferroni-adjusted significance values.} \\
\end{tabular}
\end{table}

\begin{table}[!t]
\centering
\caption{Wilcoxon Signed Ranks test results comparing AutoEnergy with other FE methods for nRMSE and processing time.}
\footnotesize
\begin{tabular}{llcc}
\toprule
\textbf{Metric} & \textbf{Comparison} & \textbf{Z} & \textbf{Asymp. Sig. (2-tailed)} \\
\midrule
\multicolumn{4}{l}{\textbf{nRMSE}} \\
& FT - AutoEnergy & -3.202\textsuperscript{b} & 0.001 \\
& TSEff - AutoEnergy & -3.724\textsuperscript{b} & \textless{}0.001 \\
& TSMin - AutoEnergy & -3.724\textsuperscript{b} & \textless{}0.001 \\
& No.Feat. - AutoEnergy & -3.724\textsuperscript{b} & \textless{}0.001 \\
\midrule
\multicolumn{4}{l}{\textbf{Processing Time}} \\
& FT - AutoEnergy & -3.680\textsuperscript{b} & \textless{}0.001 \\
& TSEff - AutoEnergy & -3.724\textsuperscript{b} & \textless{}0.001 \\
& TSMin - AutoEnergy & -0.762\textsuperscript{b} & 0.446 \\
\bottomrule
\multicolumn{4}{l}{\textsuperscript{b} Based on negative ranks.} \\
\end{tabular}
\label{tab:Wilcoxon_Test}
\end{table}


In terms of computational efficiency, as presented in \autoref{Friedman_Ranking}, Friedman's test ranked AutoEnergy first (1.33), with post hoc tests showing significant improvements over FT ($p = .005$) and TSEff ($p < .001$), while the difference with TSMin was not statistically significant ($p = 1.000$), a finding also confirmed by the Wilcoxon test ($p = .446$). Although TSMin achieved better processing times, AutoEnergy’s superiority in accuracy, demonstrated earlier through its substantial reduction in forecasting errors, remained noticeable. These results provided statistical evidence supporting AutoEnergy's favourable balance between accuracy and computational efficiency for ECF problems compared to the benchmarking methods.

\section{Summary}
\label{ch5:Conclusion}

This work proposed AutoEnergy, an AFE method designed specifically for ECF to improve AutoML performance. This algorithm automatically generates interpretable features from timestamps and historical energy consumption values while reducing reliance on domain expertise for FE. Through comprehensive evaluation across eighteen diverse real-world energy datasets, encompassing residential buildings, wind turbines, industrial settings, and grid power consumption, AutoEnergy demonstrated significant enhancement of AutoML's predictive performance compared to existing FE methods.

The experimental results revealed that AutoEnergy achieved superior forecasting accuracy with a mean nRMSE of 0.0338, representing substantial reductions in forecasting errors of 19.52\%, 82.38\%, 77.98\%, and 84.72\% compared to FT, TSMin, TSEff, and baseline approaches, respectively. These improvements were statistically validated through both the Friedman and the Wilcoxon tests. In terms of computational efficiency, AutoEnergy demonstrated better processing times, averaging 471.94 seconds across the test sets, performing 1.31x and 4.41x faster than FT and TSEff, respectively. Notably, the method exhibited exceptional performance on small datasets, which is particularly relevant given the common constraints of data availability in real-world energy applications. 

The superior performance of AutoEnergy can be attributed to its domain-specific design, where the proposed functions detailed in \autoref{ch:Method} generate interpretable features that effectively capture underlying consumption patterns across diverse energy systems, leading to improved AutoGluon performance. This was further validated through its integration with the state-of-the-art TabPFN algorithm, where AutoEnergy achieved forecasting error reductions ranging from 2.1\% to 92.8\% across different datasets compared to using TabPFN without AutoEnergy. These findings highlight that while general-purpose AutoML frameworks prioritise broad applicability, domain-specific FE methods can significantly enhance performance for specialised tasks such as ECF, offering an effective balance between accuracy and computational efficiency. \revision{The limitations of this work are discussed in \autoref{con:Research_Limitations}.}

\cleardoublepage

\chapterwithquote{"All models are wrong, but some are useful" - George Box}{Decision-Focused Learning Enhanced by Automated Feature Engineering for Energy Storage Optimisation}{ch:mainchapter6}


\revision{Having established in \autoref{ch:mainchapter5} that AutoEnergy can automate FE for ECF and improve forecasting performance across diverse datasets, this chapter closes the loop by investigating whether those representation gains translate into better downstream operational decisions in a realistic optimisation setting.} In particular, this chapter fulfils \textbf{RO3} by leveraging the proposed AFE algorithm presented in \autoref{ch:mainchapter5}  to enhance downstream tasks beyond predictive performance. Specifically, the chapter addresses three complementary sub-objectives: (A) expanding evaluation of the proposed AFE beyond predictive performance to include downstream optimisation problems, (B) improving the nascent DFL methods, and (C) evaluating performance using a novel real-world dataset and settings to examine DFL's practical viability. This comprehensive approach ultimately produces an end-to-end ML and optimisation framework that jointly forecasts electricity demand and price over a 24-hour horizon. By assessing decision quality through regret metrics rather than conventional forecasting metrics, the work demonstrates that domain-specific AFE enhances DFL and reduces reliance on domain expertise for BESS optimisation, delivering tangible operational value in practical ECF applications with broader implications for energy management systems facing similar challenges.

\section{\revision{Introduction}}
\label{ch6:Intro}

\revision{While DFL methods, as discussed in \autoref{ch2:Paper4},  address the limitations of traditional PTO approaches by integrating predictive modelling and optimisation directly into the learning process \cite{Wilder2019}}, they have been predominantly evaluated on synthetic benchmark problems with limited real-world applications \cite{kotary2021end, Mandi2024, geng2024benchmarking}, highlighting the need to assess their practical viability beyond small-scale synthetic problems \cite{Mandi2020, Zhou2024}. Furthermore, real-world datasets often exhibit greater variability and data scarcity due to collection limitations, privacy concerns, or resource constraints \cite{grinsztajn2022tree, hollmann2022tabpfn, Alkhulaifi2024_pipeline}, creating a critical gap in evaluating DFL's performance under constrained conditions typical of practical applications. Although AFE can compensate for limited data by extracting informative features \cite{Wang_2022}, manual feature engineering for energy forecasting remains time-consuming, error-prone, and heavily dependent on domain expertise \cite{Wang_2022, Wu2022}, while it remains unclear whether AFE improvements in forecasting metrics translate to better operational outcomes when integrated with DFL approaches \cite{Alkhulaifi2025}.

\revision{
In this work, \textit{translating to better operational outcomes} means that improvements in the predictive model are reflected in the quality of the downstream optimisation decision, not merely in a lower forecasting error. In the BESS setting, even small prediction errors can have asymmetric cost effects, because mistakes during high-price or high-demand periods may lead to disproportionately expensive charge or discharge schedules. For that reason, the evaluation emphasises decision quality metrics (i.e., regret) that directly measure how much performance is lost relative to the true optimal schedule, rather than relying only on conventional forecasting metrics. This definition is used consistently when comparing PTO and DFL approaches in the remainder of this Chapter.
}

Therefore, this work, addresses gaps that challenge effective BESS optimisation, stemming from: A) DFL methods promise to align prediction with downstream objectives \cite{Wilder2019, Mandi2020}; however, they are relatively new and have been tested primarily on synthetic datasets or small-scale problems (i.e., simplified benchmarks) \cite{mandi2023towards, Zhou2024}, highlighting the need to assess their practical viability in real-world applications such as BESS problems; and B) real-world datasets often exhibit greater variability and data scarcity due to practical constraints \cite{grinsztajn2022tree, hollmann2022tabpfn} which can compromise DFL performance and necessitate enhanced feature representations to extract richer information from limited data without the need for domain expertise. The novelty and contributions of this work are summarised as follows\footnote{\scriptsize{To support transparency and reproducibility, the historical electricity price and weather data, along with the code used for the experiments in this work, are publicly available on GitHub. See \protect\url{https://github.com/Nasser-Alkhulaifi/DFL}}}:

\begin{itemize}[leftmargin=15pt, itemsep=-2pt]

    \item Proposing a decision-aware, end-to-end ML forecasting and optimisation framework for BESS problems and is suitable for small dataset sizes. Rather than treating forecasting and optimisation as separate tasks, the framework uses DFL to jointly forecast electricity demand and prices while optimising BESS operations using a regret-based objective.
    
    \item Improving the nascent approach of DFL by leveraging domain-specific AFE to extract richer representations without reliance on domain expertise, as presented in our previous work \cite{Alkhulaifi2025}, thereby streamlining the development of DFL pipelines for BESS problems.
    
    \item Using novel real-world data collected from a UK-based property to evaluate the proposed framework with a comprehensive comparative analysis of PTO versus DFL approaches, with and without AFE, thereby demonstrating the practical viability of DFL methods in real-world BESS applications.

\end{itemize}

The structure of the remaining sections of this chapter is as follows: \autoref{ch6:Method} outlines the methods used for the BESS problem in this work, \autoref{ch6:Experimental_design} provides details of the experimental design, including the datasets used and evaluation criteria. Analysis and discussion of findings are presented in \autoref{ch:Results_and_Discussion}. To conclude, \autoref{ch6:Conclusion} summarises key insights.

\section{Methodology}
\label{ch6:Method}

This section outlines the problem under investigation in \autoref{ch6_sec:problem_definition}, presents the mathematical model for the BESS optimisation problem in \autoref{Mathematical_model}, and describes the prediction methods in \autoref{Pred_model}.


\subsection{Problem Definition}
\label{ch6_sec:problem_definition}

\begin{landscape}
\begin{figure}[!t]
\centering
\includegraphics[width=.8\linewidth]{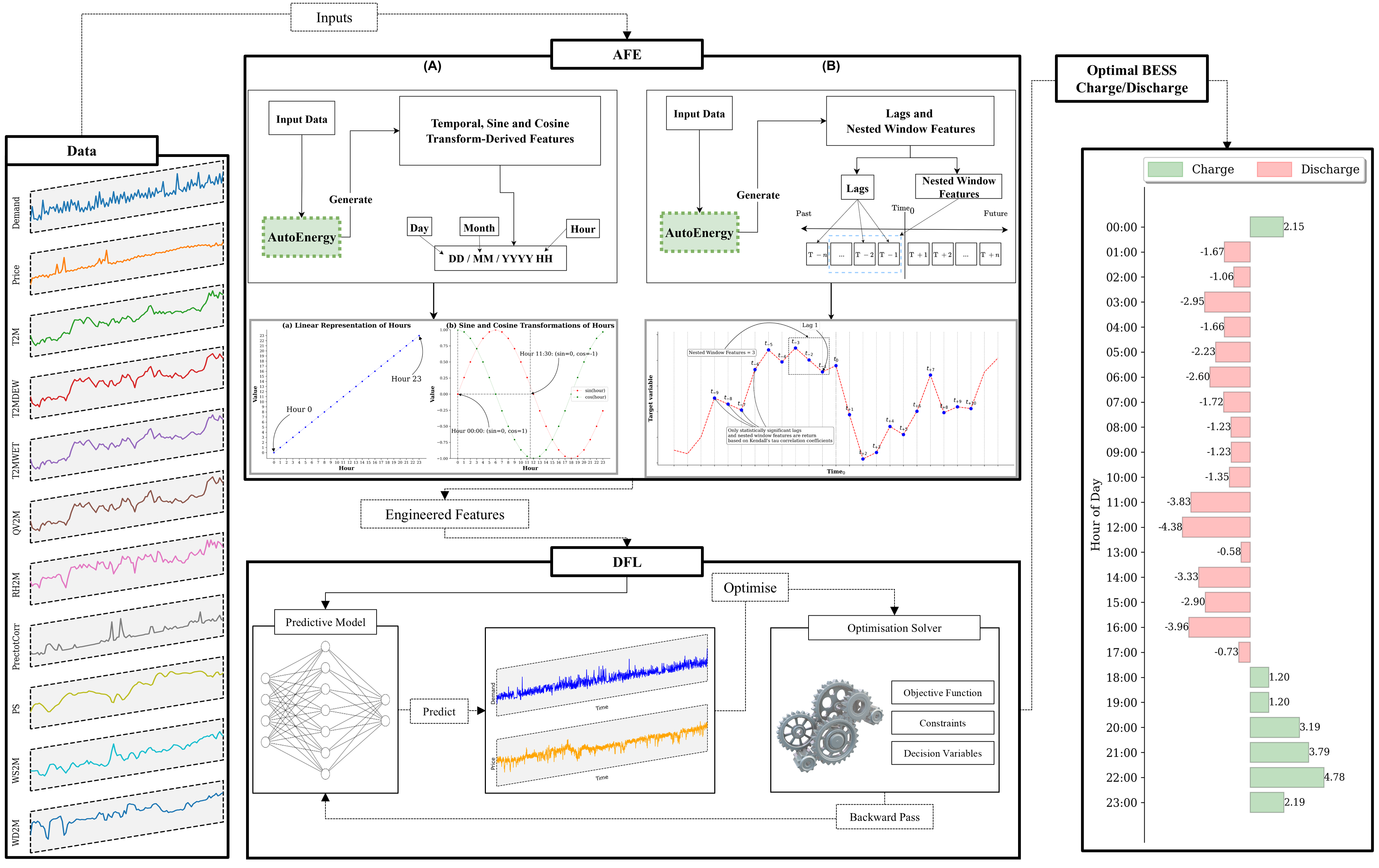}
\caption{\revision{\small The proposed framework jointly forecasts next-day electricity prices and demand while optimising the downstream BESS scheduling task. It uses automated FE to enrich the dataset with domain-specific features via \textit{AutoEnergy} \cite{Alkhulaifi2025} (details in \autoref{Auto_FE}). The goal is to reduce domain-knowledge requirements and improve decision quality under data scarcity. While the problem formulation (see \autoref{Mathematical_model}) is BESS-specific, the overall methodology has the potential to be adapted to other similar energy storage systems by revising the constraints, variables, and objectives.}}
\label{Framework}
\end{figure}
\end{landscape}

The BESS problem entails forecasting unknown parameters (electricity prices and household demand), which serve as inputs to an optimisation model that computes the optimal charge/discharge schedule for the battery over the planning horizon while minimising electricity costs and satisfying energy demand and operational constraints. As illustrated in \autoref{Framework}, the proposed framework leverages AFE to enrich the dataset with domain-specific features \cite{Alkhulaifi2025} while reducing domain knowledge requirements, thereby enhancing decision quality and improving the nascent approach of DFL for BESS scheduling problems.

\subsubsection{Multiple Unknowns, Decision-Making, and Objectives}
In this work, the BESS problem involves predicting two key unknowns (electricity demand and price) over a 24-hour planning horizon, mimicking real-world residential energy management where household BESS uses day-ahead predictions to optimise charging and discharging schedules. These uncertainties can be framed as: \textbf{(A)} What will the electricity prices be over the next 24 hours? and \textbf{(B)} What levels of electricity demand are expected? Given these forecasts, the core decision problem is to determine the optimal battery operation strategy across the 24-hour horizon, which can be formulated through the following operational questions: \textbf{(C)} At what times (i.e., hour of the day) should the battery be charged or discharged? and \textbf{(D)} By how much should it be charged or discharged? Successfully addressing these four prediction and operational questions enables the achievement of the following objectives:
\begin{itemize}
    \item Exploit price differentials through optimal battery charging and discharging schedules to reduce overall electricity costs (i.e., cost minimisation).
    \item Ensure household energy requirements are consistently met through appropriate combinations of grid power and battery discharge (i.e., demand satisfaction).
    \item Generate a battery operation strategy that respects all battery physical constraints, including power limits, energy capacity, and state-of-charge boundaries (i.e., constraint compliance).
\end{itemize}


\subsection{\revision{Mathematical Model of the Battery Energy Storage System Optimisation Problem}}
\label{Mathematical_model}

In this work, the optimisation problem of the BESS is formulated as a Mixed Integer Linear Programming (MILP) problem. MILP is widely adopted for BESS optimisation because it can efficiently handle both continuous variables (e.g., power flows and state of charge) and discrete decisions (e.g., on/off states or charge/discharge modes), enabling accurate modelling of operational constraints and system logic with computationally tractable and optimal solutions \cite{Yang2022, Yu2023}. The parameters are defined in \autoref{tab:parameters} and decision variables presented in \autoref{tab:Decision_Variables}, followed by the objective function \autoref{objective_func} subject to the listed constraints.

\revision{
The BESS optimisation problem focuses on minimising the total electricity cost over the entire planning horizon. The objective is to determine, for each time interval, the amount of energy to be drawn from the battery or the grid to satisfy the predicted property demand, as well as the amount of energy to be charged into the battery. Due to physical constraints, the battery cannot be charged and discharged simultaneously. Moreover, to ensure reliable operation and maintain long-term battery health, a minimum state-of-charge level must be preserved at all times. The problem also incorporates limitations on the maximum charging and discharging rates of the battery. It is assumed that the initial state-of-charge, the unit price of electricity (per kWh), and the energy demand for each interval are known in advance. For more details on the studied system, see \autoref{BESS_configuration}.
}

Let \( T = \{1, \ldots, T\} \) denote the set of all time intervals. For each interval \( t \in T \), \( p_t \) and \( d_t \) represent the unit price of electricity (i.e., the cost of 1 kWh) and the energy demand, respectively. The parameters \( \mathit{max\_charge} \) and \( \mathit{max\_discharge} \) denote the maximum amount of energy that can be charged to or discharged from the battery in a given interval. The parameter \( e_0 \) indicates the initial battery level (in kWh). The decision variable \( e_t \) denotes the battery level at the end of interval \( t \). The decision variables \( g_t \), \( b_t \), and \( c_t \) represent the amount of energy supplied from the grid to the property, from the battery to the property, and from the grid to the battery, respectively. The binary variable \( z_t \) takes the value 1 if energy is supplied from the battery to the property in interval \( t \), and 0 otherwise. 

\begin{table}[!t]
	\begin{center}
        \footnotesize
		\caption{Notations}
		{\renewcommand{\arraystretch}{1.3}
		\label{tab:parameters}
		\begin{tabular}{l l l} 
			\textbf{Set} & \textbf{Definition} & \textbf{ }  \\
			\hline	
			$T$ & Set of (time) intervals  &  \\
			$max\_charge$ & Maximum energy that can be added to the battery in an interval & \\
			$max\_discharge$ & Maximum energy that can be drained from the battery in an interval & \\ 
			$p_t$ & Price of 1 kWh energy at interval $t$ & $\forall t \in T $ \\
			$d_t$ & Energy demand at interval $t$  & $\forall t \in T $ \\
			$e_0$ & Initial battery level (kWh)  & \\
                $E_{\max}$ & Usable battery capacity (kWh) & \\
                $SoC_{\min}$ & Minimum allowable state-of-charge (fraction of $E_{\max}$) & \\
			\hline
		\end{tabular}}
	\end{center}
\end{table}

\begin{table}[!t]
	\begin{center}
        \footnotesize
		\caption{Decision Variables}
		{\renewcommand{\arraystretch}{1.3}
		\label{tab:Decision_Variables}
		\begin{tabular}{l l l} 
			\textbf{Dec. Var.} & \textbf{Definition} & \textbf{ }  \\
			\hline	
			$e_t$  & Amount of energy (kWh) in the battery at the end of interval $t$ & \\
			$g_t$  & Amount of energy (kWh) provided from grid to the property at interval $t$   & \\
			$b_t$  & Amount of energy (kWh) provided from battery to the property at interval $t$   & \\
			$c_t$  & Amount of energy (kWh) charged to battery at interval $t$   & \\
			$z_t$  & 1, if energy is being sent from battery to the property at interval $t$ & \\
			& 0, otherwise & \\

			\hline
		\end{tabular}}
	\end{center}
\end{table}


\begingroup
\allowdisplaybreaks
\setlength{\jot}{10pt}
\begin{align}
	& \text{Min} \quad \sum_{t \in T} p_t \cdot (g_t + c_t) && \label{objective_func}\\
	& \text{Subject to} && \notag \\
	& g_t + b_t = d_t && \forall t \in T \label{demand_satisfaction}\\
	& e_t = e_{t-1} + c_t - b_t && \forall t \in T \label{energy_balance}\\
	& b_t \leq e_{t-1} && \forall t \in T \label{available_energy_use}\\
        & b_t \leq M * z_t && \forall t \in T \label{if_charge_then_z}\\
	& c_t \leq M * (1 - z_t) && \forall t \in T \label{if_discharge_then_no_charge}\\
	& c_t \leq max\_charge && \forall t \in T \label{max_charge}\\
	& b_t \leq max\_discharge && \forall t \in T \label{max_discharge}\\
        & e_t \leq E_{\max} && \forall t \in T \label{capacity_limit}\\
        & e_t \geq SoC_{\min}\,E_{\max} && \forall t \in T \label{min_soc}\\
        & g_t, b_t, e_t, c_t \in \mathbb{R}^+ && \forall t \in T  \label{real_ranges} \\
	& z_t \in \{0,1\} && \forall t \in T \label{binary_ranges}
\end{align}
\endgroup

The objective function \eqref{objective_func} minimises the total electricity cost. Constraint \eqref{demand_satisfaction} ensures that the energy demand is satisfied for all intervals \( t \in T \). Constraint \eqref{energy_balance} enforces energy flow conservation (i.e., updating the battery’s energy level based on charging and discharging). Constraint \eqref{available_energy_use} ensures that the energy drawn from the battery never exceeds the available energy at the beginning of the interval. Constraints \eqref{if_charge_then_z} and \eqref{if_discharge_then_no_charge} jointly ensure that the battery cannot be charged and discharged simultaneously (i.e., using a large positive constant \( M \) to enforce the binary logic of \( z_t \)). Constraints \eqref{max_charge} and \eqref{max_discharge} impose upper bounds on charging and discharging rates. Constraint \eqref{capacity_limit} limits the state of charge so that it never exceeds the battery’s usable capacity \(E_{\max}\). Constraint \eqref{min_soc} enforces a floor of \(SoC_{\min}\,E_{\max}\) that prevents deep cycling, thereby reducing degradation and safeguarding long-term operational longevity. Constraints \eqref{real_ranges} and \eqref{binary_ranges} define the domains and the ranges of the decision variables.

\revision{Although the decision variables (see \autoref{tab:Decision_Variables}) are indexed by interval $t$, the MILP is solved jointly over the full horizon \( T = \{1, \ldots,t., \ldots, T\} \). The solver find the optimum values for the entire sequence $\{g_t,b_t,c_t,e_t,z_t\}_{t \in T}$ while minimising the objective \eqref{objective_func} subject to constraints \eqref{demand_satisfaction} to \eqref{binary_ranges}. The intertemporal energy-balance \eqref{energy_balance}, the capacity limit \eqref{capacity_limit}, the minimum state-of-charge bound \eqref{min_soc}, and the charge and discharge limits \eqref{max_charge} and \eqref{max_discharge} couple decisions across time. In other words, the optimisation considers all decision variables for all intervals at once to minimise total electricity cost over the whole horizon while satisfying all constraints.}

\subsection{\revision{Prediction Methods of the Battery Energy Storage System Forecasting Problem}}
\label{Pred_model}

Three ANN-based methods (PTO, SPO$^{+}$, DBB) are used for the BESS forecasting problem in this work, based on the following methodological rationale. First, the conventional PTO method serves as the established baseline, representing the predominant approach in BESSs where prediction and optimisation are handled separately \cite{Vanderschueren2022}. Second, two DFL methods, namely SPO$^{+}$ \cite{Elmachtoub2022} and DBB \cite{Pogančić2020Differentiation}, as introduced in \autoref{ch6:Intro}, are selected as the two seminal and most widely adopted DFL methods in gradient-based DFL literature. This selection enables a comprehensive comparison between the conventional PTO approach and the two main DFL methods, thereby addressing the research gap regarding DFL's practical viability in real-world BESS applications.
 
The DFL-based methods are centred on training predictive models in an end-to-end approach using decision loss derived from the associated optimisation task, and therefore, an optimisation problem needs to be solved in each training iteration. In short, the SPO$^{+}$ method uses a convex surrogate loss to upper-bound decision regret in downstream optimisation tasks, while the DBB method enables end-to-end training by approximating solver outputs for gradient computation. Further details on loss computation for the SPO$^{+}$ and DBB methods can be found in \cite{Elmachtoub2022} and \cite{Pogančić2020Differentiation}, respectively, and an explanation of the Python implementation is provided in \cite{Tang2024}. In contrast, the PTO method trains the predictive model with MSE loss, which measures the average squared difference between the model’s forecasts and the actual values, as shown in \autoref{MSE}. This approach, therefore, focuses solely on minimising forecasting errors during training, without considering the downstream optimisation task. \revision{See \autoref{ch2:Paper4_dfl_methods} for more details on the fundamental difference between combining prediction and optimisation sequentially for independent processes.}
\begin{equation}
\operatorname{MSE} = \frac{1}{N}\sum_{i=1}^{N}\bigl(\hat{y}_{i} - y_{i}\bigr)^{2}
\label{MSE}
\end{equation}
where \(\hat{y}_{i}\) is the model's prediction for the \(i^{\text{th}}\) observation, \(y_{i}\) denotes the true value for that observation, and \(N\) is the total number of observations. 

\subsection{Automated Feature Engineering}
\label{Auto_FE}
\revision{The experiment conducted in this chapter leverages AutoEnergy for AFE, which was introduced in \autoref{ch:mainchapter3}.} This method aims to streamline ML pipeline development by automatically generating features from timestamps and historical energy consumption data, thereby minimising reliance on domain expertise for FE. It generates two primary categories of features: A) Temporal features: these features extract time-based patterns from timestamps such as hour of the day, day of the week, weekdays and weekends. To enhance the representation of cyclical patterns, these temporal features undergo sine and cosine transformations. Specifically, Fourier-based transformations are applied to capture periodic patterns that represent daily and weekly cycles inherent in energy data. B) Lag and nested window features: these features capture temporal dependencies and multi-scale statistical characteristics within the energy time series data. The method computes statistically significant lags and rolling statistics (e.g., mean and standard deviation) using Kendall's tau correlation coefficient. This approach ensures that only meaningful temporal relationships are incorporated into the feature set. Additional details of this method are explained in.

\section{Experimental Design}
\label{ch6:Experimental_design}
This section outlines the experimental design used in this work, including the datasets used in \autoref{ch6_datasets_subsection}, BESS configuration in \autoref{BESS_configuration}, experimental procedure in \autoref{ch6_Experimental_procedure}, and lastly the evaluation criteria in \autoref{Evaluation_metrics}. 

 \begin{landscape}
\begin{figure}[!t]
\centering
\includegraphics[width=.6\linewidth]{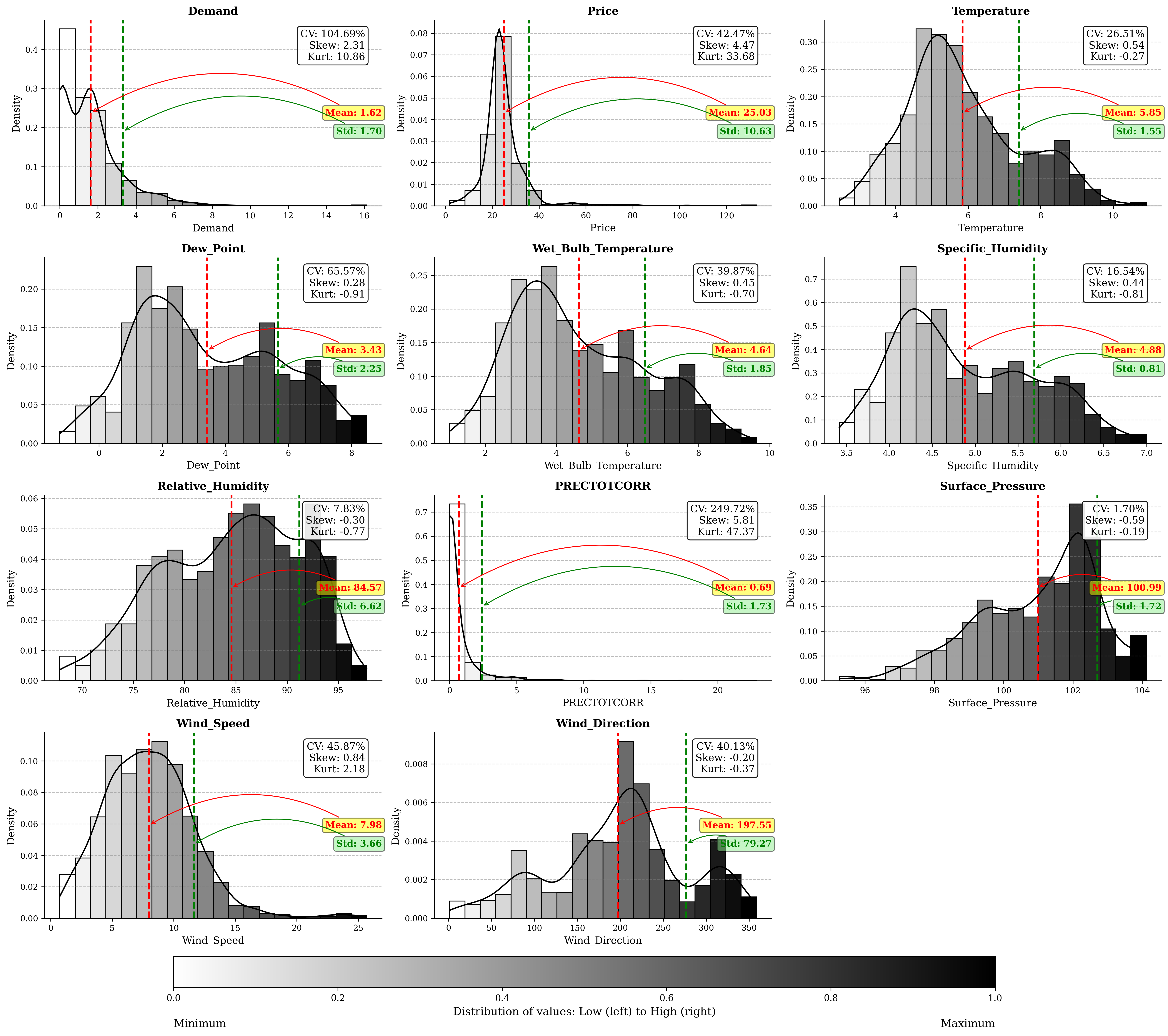}
\caption{\small The dataset used in this work is depicted in histograms with colour-coded bars representing normalised bin positions. Red and green dashed lines indicate the mean and standard deviation, respectively. Annotated statistics include coefficient of variation (CV: relative variability), skewness (distribution asymmetry), and kurtosis (tailedness). The colour gradient in the histogram bars represents the distribution of values from low (left) to high (right).}
\label{fig:stats}
\end{figure}
 \end{landscape}

\subsection{Datasets}
\label{ch6_datasets_subsection}
\begin{figure}[!t]
\centering
\includegraphics[width=1\linewidth]{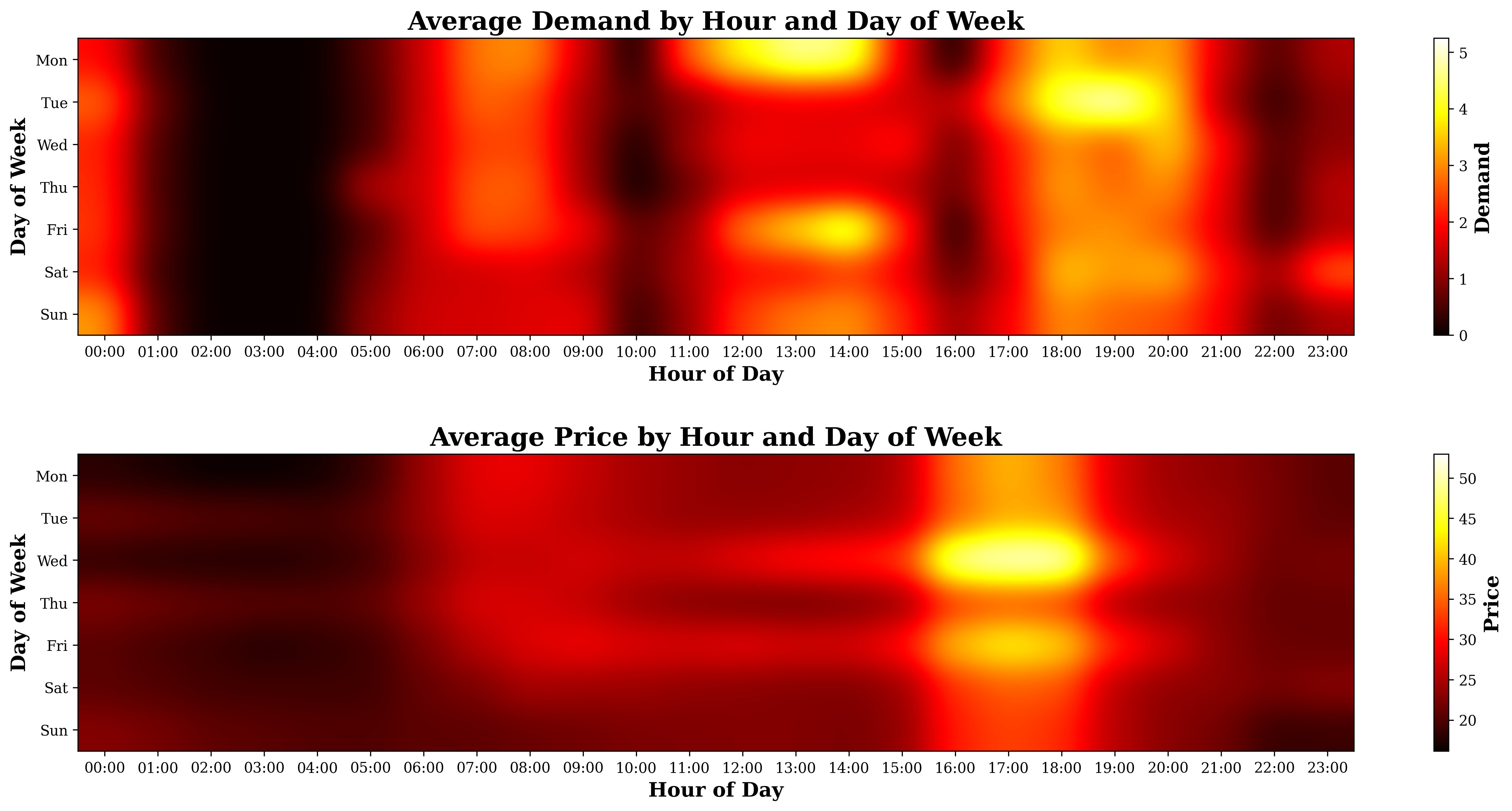}
\caption{Heatmaps of average hourly household electricity demand (top) and price (bottom) by day of week. Demand displays consistent peaks in the early morning (around breakfast), midday (lunch), and most prominently between 17:30 and 20:00 (dinner time), reflecting typical residential consumption patterns. Price peaks are concentrated between 16:00 and 19:30. Both demand and price are lowest during the early morning hours (01:00 to 05:00), reflecting reduced residential activity and system load during overnight periods.}
\label{fig:Demand_Price_Heatmaps}
\end{figure}

In this work, a dataset spanning 1 January 2025 to 24 February 2025 (55 days) is used. This scale is substantially smaller than in related works (e.g., multi-year datasets \cite{Bergmeir2025}, one year \cite{Paredes2025}, two years \cite{Wang2025_AI_Optimized}, and six years \cite{Sang2022_6_years}), thereby distinguishing the study’s contribution further by rigorously evaluating DFL, enhanced by AFE, under limited real-world data. This hourly historical real-world electricity demand and BESS data were provided by the Intelligent Plant platform \cite{IntelligentPlant}. Historical electricity prices were sourced via the Octopus Energy API \cite{OctopusEnergy}. Weather data was obtained from the NASA Langley Research Center's POWER Project \cite{nasaNASAPOWER}, a repository of solar and meteorological datasets developed by NASA to support renewable energy and building energy efficiency research. \autoref{fig:stats} presents summary statistics of the collected data, while \autoref{fig:Demand_Price_Heatmaps} displays patterns of average hourly electricity demand and price by day of the week.


\subsection{\revision{Battery Energy Storage System Configuration}}
\label{BESS_configuration}
\begin{figure}[t!]
\centering
\includegraphics[width=1\linewidth]{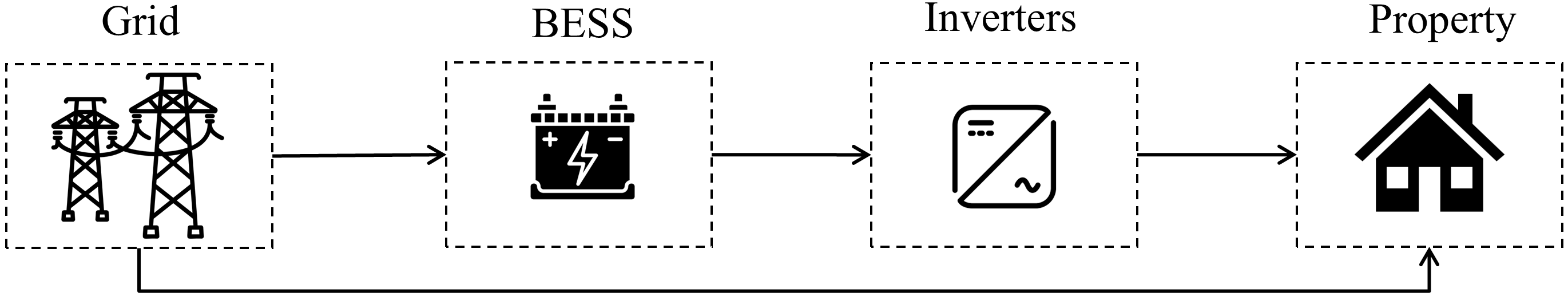}
\caption{Power flow diagram, where: A) power flows from the grid to charge the battery, which then supplies the property via inverters performing DC–AC conversion to meet energy demand; and B) power flows directly from the grid to the property when the battery capacity is insufficient.}
\label{fig:Power_flow}
\end{figure}

In this experiment, the maximum charging rate was set to 5 kWh per interval to comply with operational battery constraints, while the maximum discharging rate was limited to 4.5 kWh per interval to account for round-trip efficiency losses and system resistive losses inherent in BESSs. The minimum state of charge was maintained at 10\% to prevent battery degradation and ensure operational longevity. The capacity was set to 50 kWh, imposed by physical battery constraints. It is worth noting that the current BESS configuration operates without any solar panels. \revision{ \autoref{tab:fixed_params} provides a summary of the parameters and their values used in the experiment}. In this system, as shown in \autoref{fig:Power_flow}, electrical power follows two primary pathways: first, power flows from the grid to charge the battery, which then supplies the property through inverters that perform DC-AC conversion to meet energy demand; second, power flows directly from the grid to the property when battery capacity is insufficient. This dual-path configuration ensures a continuous power supply through direct grid connection while enabling BESS optimisation for energy arbitrage (i.e., storing electricity during low-cost periods and supplying the property during high-cost periods), thereby reducing energy cost and contributing to grid stability.


\begin{table}[t!]
\centering
\footnotesize
\revision{
\caption{\revision{Summary of parameters and values used in the experiment. See \autoref{BESS_configuration} for BESS configuration details and \autoref{ch6_Experimental_procedure} for the experimental procedure.}}
\label{tab:fixed_params}
\begin{tabular}{ll}
\toprule
\textbf{Parameter} & \textbf{Value} \\
\midrule
Planning horizon $T$ [intervals] & $24$ \\
Interval duration $\Delta t$ [hours] & $1$ \\
Usable capacity $E_{\max}$ [kWh] & $50$ \\
Minimum SoC fraction $\mathrm{SoC}_{\min}$ & $0.10$ \\
Max charge per interval $max\_charge$ [kWh] & $5.0$ \\
Max discharge per interval $max\_discharge$ [kWh] & $4.5$ \\
Initial energy $e_0$ [kWh] & $10$ \\
Electricity price $p_t$ [£/kWh] & Variable by interval \\
Energy demand $d_t$ [kWh] & Variable by interval \\
\bottomrule
\end{tabular}}
\end{table}


\subsection{Experimental Procedure}
\label{ch6_Experimental_procedure}
In this work, the dataset is partitioned into three subsets: \(\sim\)50\% (27 days) allocated for initial training, \(\sim\)25\% (14 days) for validation to fine-tune hyperparameters, and the remaining \(\sim\)25\% (14 days) reserved for assessing model performance on previously unseen data (i.e., test dataset). ANN-based methods, as explained in \autoref{Pred_model}, were used to predict the price and demand for the next 24 hours (i.e., the following day) based on input features such as weather conditions (e.g., outdoor temperature) and the engineered features described in \autoref{Auto_FE}, while accounting for the operational and physical constraints of the BESS explained in \autoref{Mathematical_model}. This approach mimics real-world scenarios where BESS schedules charging and discharging operations daily to optimise cost savings by taking advantage of low electricity prices. It is important to note that, to better capture the interrelated dynamics between electricity price and demand, a multitask learning approach was used \cite{Wang2022Multi}, in which the same input features are processed through shared ANN layers to predict both outputs simultaneously, thereby leveraging shared information across tasks.

\begin{figure}[!t]
\centering
\includegraphics[width=1\linewidth]{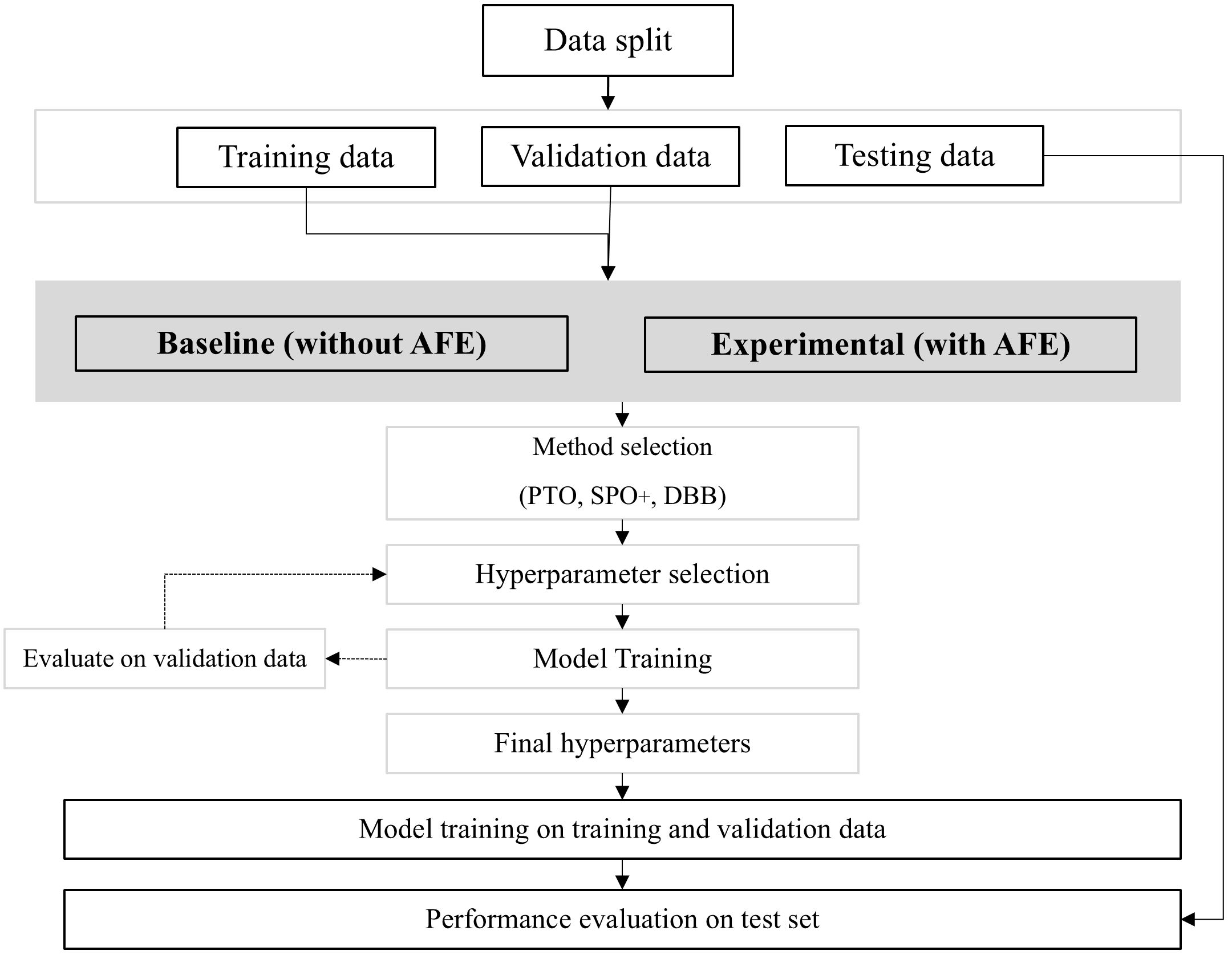}
\caption{Experimental design. See \autoref{ch6_Experimental_procedure} for a detailed explanation of the experimental procedure.}
\label{fig:experimental_design}
\end{figure}


Five-fold cross-validation and grid search optimisation were employed to tune the ANN architecture and learning hyperparameters of all three methods (PTO, DBB, and SPO$^{+}$). During the hyperparameter tuning phase, many model configurations are trained exclusively on the training set and evaluated on the validation set until optimal hyperparameters are identified, after which the training and validation sets are combined (totalling 75\%, 41 days of the original dataset) to train the final model using the previously optimised hyperparameters, thereby maximising the utilisation of available data for final model training. \autoref{fig:experimental_design} shows the experimental design and \autoref{tab:hyperparameters_space} shows the hyperparameter search space. Finally, the performance of the PTO, DBB, and SPO$^{+}$ methods was evaluated on the test set using the evaluation metrics described in \autoref{Evaluation_metrics}.  

\revision{The grid search includes architectures with one or two hidden layers; the best models use 256 and 128 neurons in the first and second hidden layers, respectively. The grid search explores 12 configurations with 5-fold cross-validation, yielding 60 candidate models per method and 180 ANNs in total (PTO, SPO\textsuperscript{+}, DBB); see \autoref{Pred_model} for details. For each training window, a 24-interval MILP (\autoref{Mathematical_model}) is solved using Gurobi \cite{gurobi} to optimise the battery schedule. During training, the solver is called once per sample (i.e., one day) per epoch; therefore, the optimisation layer dominates the total run time. This procedure is repeated twice (see \autoref{fig:experimental_design}), once with AFE and once without AFE, as explained in \autoref{ch6_Experimental_procedure}. Unlike \cite{Paredes2025}, which tunes core regressor hyperparameters on PTO and transfers them to DFL, we perform method-specific tuning: for PTO method, the models were trained with MSE loss on predictions; for DFL methods (SPO$^{+}$ and DBB), training integrates optimisation results via their respective decision-focused losses (see \cite{Mandi2020, Tang2024} for more details). The total running time, including hyperparameter tuning, is approximately 39.97 hours.}

To ensure the robustness and reliability of the experimental results, models were trained and evaluated across ten independent runs with different random seeds. This practice mitigates standard deviation introduced by stochastic training procedures (e.g., such as random weight initialisation) and enables statistically meaningful comparisons by reporting mean performance metrics together with their standard deviations. The results of all ten independent runs are reported in the supplementary material. \revision{It is also important to note that the data were not shuffled, as the energy data exhibit temporal patterns, making it essential to preserve the chronological order of the timestamps.}


\begin{table}[!t] 
\centering 
\begin{threeparttable}

\footnotesize
\caption{Hyperparameter search space and the corresponding best hyperparameters for the ANN models, determined through grid search on the validation dataset. To avoid over-parameterisation, the search space is constrained to the most critical parameters identified in related work \cite{Paredes2025}. \revision{However, it is worth mentioning that this may have restricted the discovery of optimal hyperparameter combinations that could further enhance model performance}.}
\setlength{\extrarowheight}{1pt} 
\begin{tabularx}{\textwidth}{
  @{}
  >{\centering\arraybackslash}m{3cm}
  >{\centering\arraybackslash}m{2.5cm}
  | >{\scriptsize\centering\arraybackslash}X
  | >{\scriptsize\centering\arraybackslash}X
  | >{\scriptsize\centering\arraybackslash}X
  | >{\scriptsize\centering\arraybackslash}X
  | >{\scriptsize\centering\arraybackslash}X
  | >{\scriptsize\centering\arraybackslash}X
  @{}
}

\toprule
\multirow{4}{*}{Hyperparameter} & \multirow{4}{*}{Search Space} & \multicolumn{6}{c}{Best Hyperparameter} \\ 
\cmidrule(lr){3-8} 
& & \multicolumn{2}{c!{\vrule width 0.8pt}}{PTO} & \multicolumn{2}{c!{\vrule width 0.8pt}}{SPO$^{+}$} & \multicolumn{2}{c}{DBB} \\ 
\cmidrule(lr){3-4} \cmidrule(lr){5-6} \cmidrule(lr){7-8} 
& & No AFE & AFE & No AFE & AFE & No AFE & AFE \\ 
\midrule 
Number of Layers & $\in \{1,\,2\}$ & 2 & 2 & 2 & 2 & 2 & 1 \\ 
Epochs & $\in \{10,\,20,\,30\}$ & 30 & 30 & 30 & 30 & 30 & 30 \\ 
Learning Rate & $\in \{10^{-3}, 10^{-5}\}$ & $10^{-3}$ & $10^{-3}$ & $10^{-5}$ & $10^{-5}$ & $10^{-3}$ & $10^{-3}$ \\ 
\bottomrule 
\end{tabularx} 
\label{tab:hyperparameters_space} 

\begin{tablenotes}
\scriptsize
\item In this experiment, models were trained using Python 3.11, Gurobi API \cite{gurobi}, and PyEPO API \cite{Tang2024}. Computational experiments were performed on an Ubuntu system featuring an \texttt{x86\_64} architecture with 16 physical CPU cores (32 logical cores), 64GB of RAM.
\end{tablenotes}

\end{threeparttable}
\end{table}

To evaluate the impact of AFE on decision quality, two conditions were established: A) a baseline scenario without AFE and B) an experimental scenario incorporating AFE. The baseline models serve to assess forecasting and optimisation capabilities with minimal inputs. This deliberately simplified configuration replicates model performance under conditions that simulate both a worst-case scenario (i.e., absence of domain-specific FE knowledge) and an initial testing phase where the model learns with limited data enhancement. Conversely, the experimental scenario applies the AFE method explained in \autoref{Auto_FE}, thereby enabling the assessment of the AFE impact on the downstream optimisation task (i.e., decision-making quality).

\revision{
It is worth mentioning that the experimental setup with two well-established DFL methods and the traditional PTO approach (see \autoref{Pred_model}) and systematic AFE evaluation (with vs. without; see \autoref{Auto_FE}), yielding a comprehensive experiment with six method conditions, is justified by: A) alignment with the stated research objective of demonstrating DFL's practical viability under data scarcity conditions rather than providing exhaustive benchmarking, where the 55-day dataset deliberately reflects real-world BESS deployment constraints faced by practitioners during initial operation (i.e., expanding to additional methods or datasets may obscure the core research questions); B) methodological clarity requirements, where incorporating more DFL methods may introduce confounding variables that complicate attribution of performance differences (i.e., results variations may reflect differences between DFL approaches rather than the contribution of AFE); and C) computational resource constraints associated with training additional DFL methods, where each training iteration necessitates solving an MILP optimisation problem for every sample in every epoch, such that additional methods would require multiplicative increases in computational demands, potentially exceeding resources available in practical deployment scenarios.
}


\subsection{Evaluation Metrics and Statistical Tests}
\label{Evaluation_metrics}

In this work, model performance is evaluated using the normalised regret metric \cite{Tang2024}. The notion of regret is used to measure the error in decision-making. It is characterised as the difference in the objective value between the true optimal solution and the optimal solution obtained by utilising the predicted coefficients. For minimisation problems, the normalised regret is defined as:

\begin{equation}
\label{eq:normalized_regret}
\frac{\sum_{i=1}^{n_{\text{test}}} \mathcal{L}_{\text{Regret}}(\hat{c}^i, c^i)}{\sum_{i=1}^{n_{\text{test}}} \lvert z^*(c^i) \rvert}
\end{equation}

where $\mathcal{L}_{\text{Regret}}(\hat{c}^i, c^i) = c^{i\top} w^*(\hat{c}^i) - z^*(c^i)$ denotes the regret for instance $i$. Here, $w^*(\hat{c}^i)$ represents the optimal solution obtained using the predicted cost vector $\hat{c}^i$, $c^{i\top} w^*(\hat{c}^i)$ is the objective value achieved by this solution when evaluated under the true cost vector $c^i$, and $z^*(c^i)$ is the true optimal objective value under the actual cost vector $c^i$. The regret thus measures the difference between the objective value of the solution derived from predicted costs (evaluated under true costs) and the true optimal objective value. The denominator normalises the aggregated regret by the sum of absolute values of true optimal objectives across the test set, providing a scale-invariant measure of decision quality degradation \cite{Tang2024}.

Non-parametric hypothesis tests were used to identify significant differences between the DFL and PTO methods, as well as between AFE versus without AFE and to support the experimental findings statistically \cite{Sheskin2003}. The Friedman Aligned-ranks test \cite{Garca2010} first evaluated overall differences among the methods with the significance threshold set at \(\alpha = 0.05\). Additionally, for pairwise comparisons, the Wilcoxon Signed-Rank test \cite{demvsar2006statistical, garcia2008extension} was applied at \(\alpha = 0.05\) to explore potential differences between method pairs that the preceding test did not flag as statistically significant, thereby ensuring a comprehensive statistical analysis.

\section{Results and Discussion}
\label{ch:Results_and_Discussion}

This section presents an analysis and discussion of the results.  
\autoref{Performance_overview} discusses the overall performance of the PTO and DFL methods for the investigated BESS problem. Subsequently, \autoref{FE_impact} provides a comparative analysis of the impact of AFE on the downstream optimisation task (i.e. decision quality) supported by feature importance analysis. Finally, \autoref{Statistical_Significance} evaluates the statistical significance of the observed performance differences between the methods, to verify that the reported results are unlikely to be due to chance. \autoref{6_Appendix_A} in the appendices presents the detailed results of each of the ten runs across the test days, while \autoref{6_Appendix_B} in the appendices illustrates the daily BESS optimisation schedules for the best run of each day using the best-performing method, showing how predicted prices and demand, subject to operational constraints, shaped the BESS scheduling decisions.

\subsection{\revision{Performance Overview of Predict-Then-Optimise Compared to Decision-Focused Learning}}
\label{Performance_overview}

\begin{figure}[!t]
\centering
\includegraphics[width=1\linewidth]{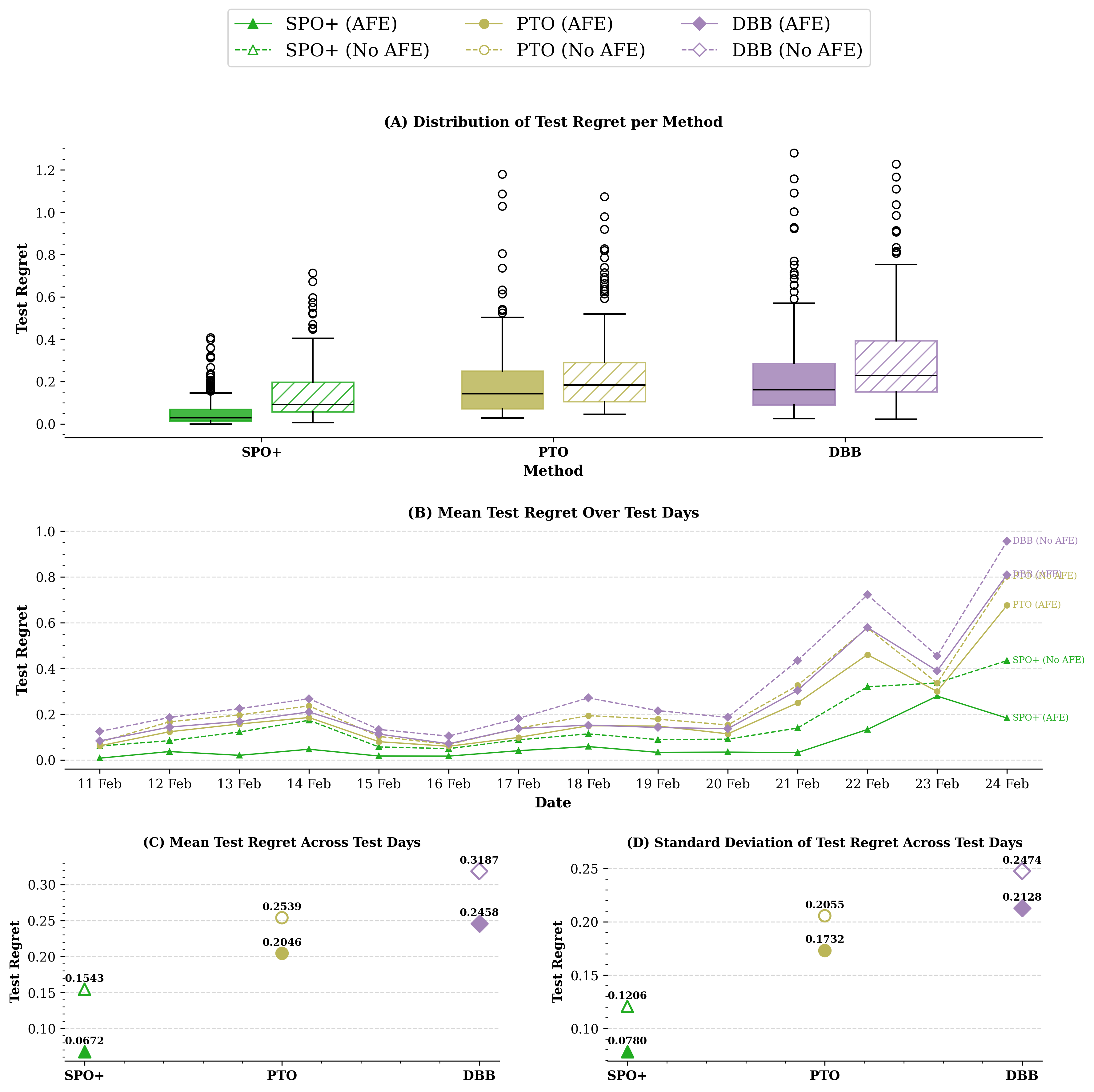}
\caption{\revision{ Test regret across methods with and without AFE. (A) Box plots of test regret across all experimental runs and test days for PTO and DFL methods (SPO$^{+}$ and DBB). The distributions reflect performance variability across multiple independent experimental runs and test days, providing a comprehensive view of method robustness and central tendency. (B) Daily mean test regret over the fourteen test days. (C) Mean test regret across test days for each method variant. (D) Standard deviation of test regret across test days. \revision{These results support DFL being superior \emph{in this experimental setting} when combined with AFE (as measured by regret), but they do not imply that DFL is universally superior across all datasets, objectives, or optimisation problems.}}}

\label{Box_plots}
\end{figure}

The performance comparison results between the PTO and DFL methods across the test set are presented in \autoref{Box_plots} and \autoref{tab:test_regret}.
Across the fourteen‑day test horizon, the SPO$^{+}$ method outperformed both the traditional PTO  and DBB methods. On average, the PTO method with AFE incurred a regret of 0.2046, while the SPO$^{+}$ method achieved a remarkably low regret of 0.0672. This improvement corresponds to a substantial reduction of approximately 67.16\%, underscoring the significant benefits of integrating the optimisation layer during training with the SPO$^{+}$. In other words, when the model is trained to predict electricity price and property demand while minimising downstream cost using the SPO$^{+}$ loss, rather than a generic pointwise error metric (i.e., error-minimising forecasts), it yields cheaper battery schedules. 

\revision{Interestingly, the DBB method, which is considered a DFL approach, achieved the highest average regret (i.e., the highest cost in BESS scheduling). This observation is broadly in line with recent related work  \cite{Wang2025_AI_Optimized, zharmagambetov2023landscape}, where DBB underperforms SPO$^{+}$, despite differences in datasets, experimental settings and problem formulations. Although DBB can differentiate through black-box combinatorial solvers, the findings of this work indicate that this method is less suitable for BESS optimisation tasks under data scarcity, as evidenced by the higher regrets compared with other methods, potentially due to its original design for problems with linear objectives \cite{Mandi2024}. Additionally, the poor performance of DBB compared to SPO$^{+}$ and PTO methods may indicate that DBB requires more training data to learn effective gradient approximations, making it particularly vulnerable in data-scarce environments such as the 55-day dataset examined in this work.}

An examination of day‑to‑day performance reveals that SPO$^{+}$ with AFE delivers notably consistent superior performance, demonstrating reliability on a per-instance basis. This consistency is reflected in lower standard deviation (0.0780 vs 0.1732 for PTO and 0.2128 for DBB), as alternative approaches exhibited greater sensitivity to volatile market conditions such as those observed on February 22nd and 24th. The stability demonstrated by SPO$^{+}$ suggests that training with regret-based loss may produce models that are more robust to diverse price-demand patterns compared to both traditional PTO and DBB methods. Such findings, while data-dependent, indicate that SPO$^{+}$ could offer practical advantages for BESS strategies by potentially reducing exposure to costly high-regret scheduling decisions, though broader validation across different operational contexts would strengthen these conclusions. \revision{The consistently higher variance observed in DBB's performance compared to SPO$^{+}$ further indicates potential instability in DBB's learning process under data constraints. This instability likely stems from the inherent challenges of approximating gradients through black-box solvers when training data is limited, resulting in less reliable gradient signals during optimisation.}


\begin{table}[!t]
\centering
\footnotesize
\caption{Daily test regrets (lower is better) averaged across ten independent runs to improve the robustness and reliability of the results (see supplementary materials for detailed results of each of the ten runs across the test set days). The lowest mean and standard deviation values are bolded to indicate the best performance. The comparison includes PTO and DFL methods (SPO$^{+}$, DBB), each evaluated with and without AFE (see \autoref{ch6_Experimental_procedure} for more details on the experimental procedure). Appendix \ref{Appendix_A} provides detailed daily BESS optimisation schedules for the best run of each day using the best-performing method.}
\setlength{\extrarowheight}{1pt}
\begin{tabularx}{\textwidth}{ @{} >{\centering\arraybackslash}m{2.8cm} >{\centering\arraybackslash}X >{\centering\arraybackslash}X !{\vrule width 0.8pt} >{\centering\arraybackslash}X >{\centering\arraybackslash}X !{\vrule width 0.8pt} >{\centering\arraybackslash}X >{\centering\arraybackslash}X @{} }
\toprule
\multirow{4}{*}{Date} & \multicolumn{6}{c}{Test Set Regrets} \\
\cmidrule(lr){2-7}
 & \multicolumn{2}{c!{\vrule width 0.8pt}}{PTO} & \multicolumn{2}{c!{\vrule width 0.8pt}}{SPO$^{+}$} & \multicolumn{2}{c}{DBB} \\
\cmidrule(lr){2-3} \cmidrule(lr){4-5} \cmidrule(lr){6-7}
 & AFE & No AFE & AFE & No AFE & AFE & No AFE \\
\midrule
11 Feb & 0.0620 & 0.0780 & 0.0079 & 0.0611 & 0.0832 & 0.1247 \\
12 Feb & 0.1234 & 0.1669 & 0.0369 & 0.0845 & 0.1440 & 0.1858 \\
13 Feb & 0.1573 & 0.1968 & 0.0205 & 0.1222 & 0.1685 & 0.2243 \\
14 Feb & 0.1856 & 0.2363 & 0.0469 & 0.1736 & 0.2101 & 0.2678 \\
15 Feb & 0.0800 & 0.1030 & 0.0173 & 0.0571 & 0.1133 & 0.1335 \\
16 Feb & 0.0599 & 0.0681 & 0.0169 & 0.0496 & 0.0708 & 0.1042 \\
17 Feb & 0.0985 & 0.1384 & 0.0406 & 0.0880 & 0.1377 & 0.1818 \\
18 Feb & 0.1495 & 0.1939 & 0.0586 & 0.1142 & 0.1525 & 0.2711 \\
19 Feb & 0.1483 & 0.1784 & 0.0334 & 0.0891 & 0.1420 & 0.2156 \\
20 Feb & 0.1144 & 0.1530 & 0.0344 & 0.0909 & 0.1366 & 0.1862 \\
21 Feb & 0.2495 & 0.3261 & 0.0322 & 0.1395 & 0.3041 & 0.4344 \\
22 Feb & 0.4604 & 0.5771 & 0.1334 & 0.3199 & 0.5789 & 0.7216 \\
23 Feb & 0.2994 & 0.3363 & 0.2791 & 0.3367 & 0.3896 & 0.4549 \\
24 Feb & 0.6756 & 0.8021 & 0.1832 & 0.4343 & 0.8098 & 0.9551 \\
\midrule
Mean & 0.2046 & 0.2539 & \textbf{0.0672} & 0.1543 & 0.2458 & 0.3187 \\
Std & 0.1732 & 0.2055 & \textbf{0.0780} & 0.1206 & 0.2128 & 0.2474 \\
\bottomrule
\end{tabularx}

\label{tab:test_regret}
\end{table}



\subsection{\revision{Impact of Automated Feature Engineering on Predict-Then-Optimise and Decision-Focused Learning Methods with Feature Importance Analysis}}
\label{FE_impact}

The addition of AFE significantly enhanced the performance for all methods, evidenced by the results presented in \autoref{Box_plots} and \autoref{tab:test_regret}, with particularly pronounced benefits for the DFL approaches. Without AFE, SPO$^{+}$ outperformed the other methods with a mean regret of 0.1543, compared to the PTO approach (0.2539) and DBB (0.3187). This initial superiority suggests that the SPO$^{+}$ formulation is inherently effective at learning the mapping between inputs and optimal decisions for the investigated problem, even with less feature representations. However, the introduction of AFE amplified these advantages, with SPO$^{+}$ achieving 56.48\% improvement in mean regret, while the PTO improved by 19.42\% and DBB by 22.87\%. The overall reductions in regrets indicate that the AFE, while minimising the need for expert-driven FE, successfully generates useful features not only for maximising forecasting accuracy (i.e., the PTO approach) but also for optimising downstream decision-making tasks within DFL frameworks (i.e., SPO$^{+}$ and DBB). 

Moreover, the notable responsiveness of SPO$^{+}$ to enhanced feature representations may suggest the synergistic effect between this method and AFE in BESS optimisation problems. It implies that AFE provides SPO$^{+}$ with the contextual information necessary to better approximate the discrete optimisation problem via its convex surrogate, leading to more cost-effective battery scheduling decisions. The enriched feature representations could enable the convex surrogate loss to better capture the relationship between prediction errors and their downstream decision costs. The differential response to AFE across methods provides additional insight into DBB's limitations. \revision{ While SPO$^{+}$ achieved a higher improvement with AFE, DBB's improvement was noticeably lower, suggesting that DBB's gradient approximation mechanism may be less capable of effectively leveraging enhanced feature representations. This reduced feature sensitivity could stem from DBB's reliance on perturbation-based gradient estimation, which may not capture the complex relationships between engineered features and optimal decisions as effectively as the convex surrogate approach of SPO$^{+}$.}

\revision{It is important to note that the regret values in \autoref{tab:test_regret} are measured in monetary units (GBP) and represent the additional electricity cost incurred due to imperfect predictions when making battery scheduling decisions, where a regret of zero indicates optimal decision-making. Because regret equals additional electricity cost, the differences in mean regret quantify cost savings. Averaged over the 14-day test horizon, PTO (AFE) incurred a regret of 0.2046, whereas SPO$^{+}$ (AFE) achieved 0.0672, which is an absolute reduction of 0.1374 and a 67.16\% relative reduction. Within each method, AFE further reduced regret: SPO$^{+}$ from 0.1543 to 0.0672 (56.48\%), PTO from 0.2539 to 0.2046 (19.42\%), and DBB from 0.3187 to 0.2458 (22.87\%). These results indicate that training the predictor with a decision-focused loss yields materially cheaper day-ahead battery schedules and that AFE consistently improves decision quality. While absolute regret values may appear small in monetary terms because the BESS is sized for a single household, they correspond to sizeable percentage reductions, accumulate over time, and scale with asset size; using the worst and best mean regrets among the compared methods (DBB without AFE: £0.3187; SPO$^{+}$ with AFE: £0.0672) gives a daily difference of £0.2515, which is approximately £91.80 per year if sustained.}

 \begin{landscape}
\begin{figure}[!t]
\centering
\includegraphics[width=.7\linewidth]{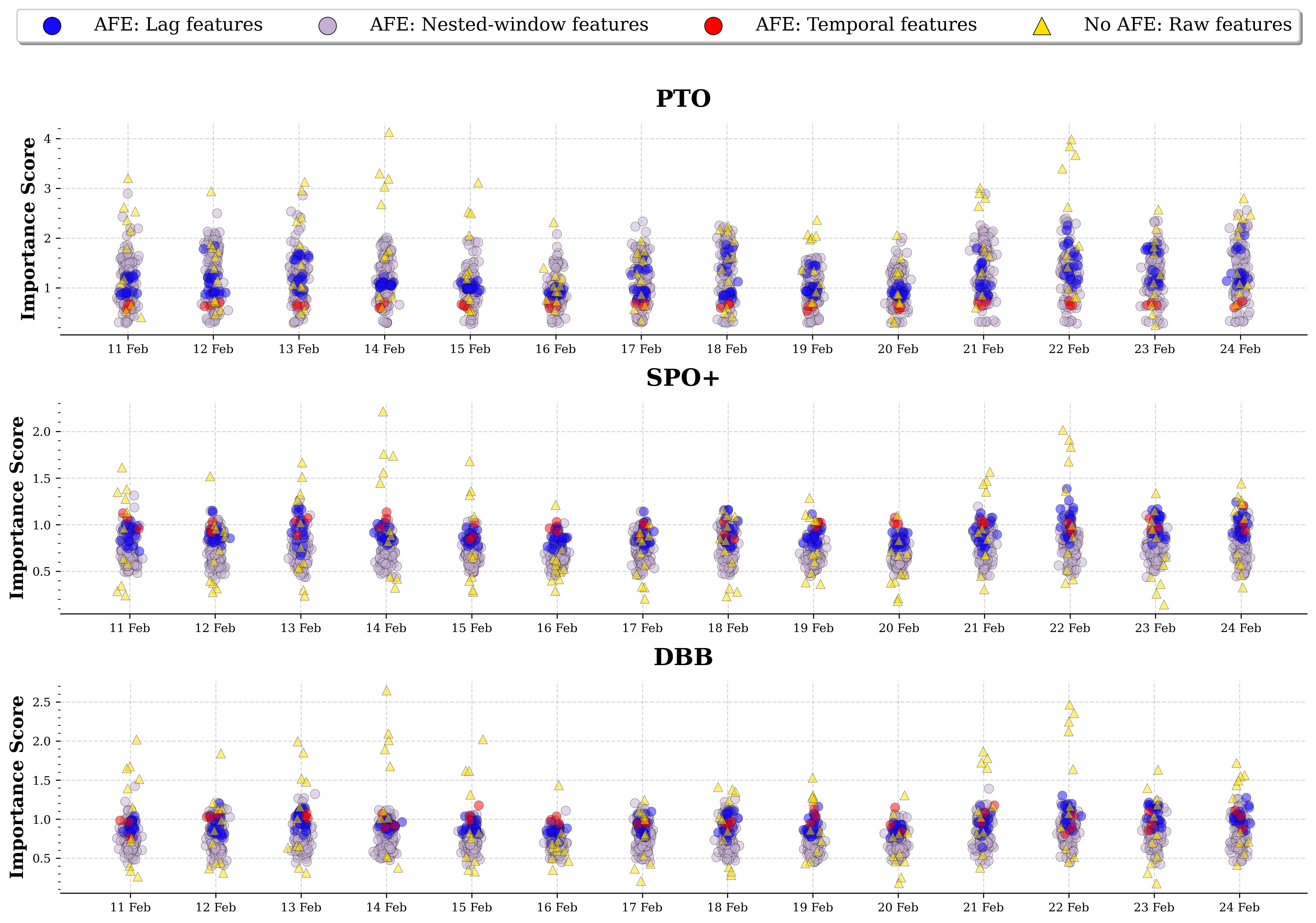}
\caption{\revision{\small
Feature importance across all experimental runs and test days. Each subplot corresponds to one method and reports the mean SHAP importance per feature. Raw inputs (“No AFE”, e.g., weather) are compared with engineered features (“AFE”, e.g., lags, rolling-window statistics, and calendar indicators). Markers (colour and shape) denote feature categories; each point is a feature, with x-axis jitter to reduce overlap and show within-category dispersion. More detailed analysis is provided in \autoref{shap_analysis}}.}
\label{fig_SHAP_by_method}
\end{figure}
 \end{landscape}

Nevertheless, the magnitude of AFE's influence varied across test set days, yet no instances of AFE-induced performance degradation were observed across any method. For example, adding AFE to the SPO$^{+}$ method improved its performance, leading to regret reductions of 83.22\% and 62.2\% on February 13 and 20, respectively. However, on some days, the improvements were less pronounced. For instance, on 23 February, SPO$^{+}$ improved by 17.1\% (from 0.3367 to 0.2791), indicating that while AFE consistently enhances performance (i.e., yielding lower regrets than without AFE), its effectiveness may vary depending on the relevance of the features generated to the underlying data patterns.

Beyond average improvements, AFE also enhanced the consistency and reliability of performance across all methods. The standard deviations of regrets decreased for all approaches with AFE: from 0.2055 to 0.1732 for PTO (15.72\% reduction), from 0.1206 to 0.0780 for SPO$^{+}$ (35.32\% reduction), and from 0.2474 to 0.2128 for DBB (13.9\% reduction). Notably, SPO$^{+}$ achieved the largest proportional standard deviation reduction from AFE, indicating potential suitability for BESS optimisation problems where data scarcity heightens the importance of methods that respond strongly to AFE.

\revision{It is worth noting that, although the optimisation layer ultimately outputs a single day-ahead control action (i.e., one feasible BESS schedule; see \autoref{ch6:Experimental_design} for more details on the experimental design), the analysis does not treat that action as uncertainty-free. Decision robustness is quantified empirically by computing regret (explained in \autoref{Evaluation_metrics} per test day and summarising performance across repeated runs (reporting both mean and standard deviation), which provides a distributional view of realised decision quality rather than a single deterministic claim. Nonetheless, the current framework is limited to point forecasts for uncertain inputs; therefore, it does not incorporate probabilistic forecasts (e.g., prediction intervals or quantile forecasts) nor propagate this uncertainty into risk-aware optimisation, which may be important in safety-critical or risk-averse deployment settings.}

\revision{It is worth mentioning that, a practical consequence of integrating forecasting with the downstream optimisation task end-to-end is that not all regression algorithms considered in earlier chapters are directly usable in this DFL setting. In particular, end-to-end training requires the learning signal to propagate through the downstream optimisation layer, which favours predictors and training pipelines that support gradient-based optimisation (discussed in \autoref{ch2:Paper4}). As a result, some promising regression models used in \autoref{ch:mainchapter3} cannot be easily incorporated in the same end-to-end manner without additional approximation or surrogate modelling, even if they achieve high standalone forecasting accuracy. This highlights an explicit trade-off in the work presented in this chapter. Specifically, the methods used in this chapter are constrained by end-to-end decision training, and the comparisons here therefore focus on methods that are compatible with the integrated PTO objective rather than on the full space of best-performing regressors from earlier chapters.}

\revision{Note that, in principle, a less tight integration alternative is possible that still connects forecasting to the downstream task while allowing the use of any high-performing regressor. Potentially, one can adopt a sequential PTO design in which a forecaster is trained independently (using the best-performing methods from earlier chapters), and its point or probabilistic forecasts are then fed into the optimiser, optionally selecting the forecaster based on downstream cost on a validation set rather than on forecasting error alone. This relaxation may improve model-class flexibility discussed above, but it might sacrifice the end-to-end DFL coupling, meaning the forecaster is no longer trained directly against the optimisation objective and the pipeline becomes more exposed to objective misalignment and error propagation. That said, the work presented in this chapter focuses on end-to-end integration to test whether learning with a decision-aligned signal can outperform purely sequential designs under realistic constraints, while recognising that hybrid designs are a viable practical option when maximising standalone forecasting accuracy is the priority.}

\subsection{\revision{Feature Importance Analysis Across Predict-Then-Optimise and Decision-Focused Learning Methods}}
\label{shap_analysis}

\revision{
SHAP values (SHapley Additive exPlanations) were used to interpret the relationships learned by the PTO and DFL methods for predicting electricity price and property demand simultaneously. Across all feature combinations, SHAP computes each input's average marginal effect on model outputs (i.e., predictions), thereby quantifying both the magnitude and the direction (positive or negative) of feature influence \cite{lundberg2017unified}. As shown in \autoref{fig_SHAP_by_method}, features generated through AFE demonstrate considerable temporal variation in their impact on model outputs across different test days. This day-to-day variation in feature importance likely reflects the dynamic nature of energy consumption patterns and market conditions during the test period. For instance, while engineered features dominated the importance rankings on certain days (e.g., 17 and 18 February), raw features such as weather data maintained higher importance scores on others (e.g., 14 and 22 February). This temporal variability suggests that the relative value of different information sources fluctuates based on underlying consumption patterns, weather conditions, and market dynamics, highlighting the importance of adaptive AFE in operational BESS systems.

Within the AFE categories generated by the AutoEnergy algorithm (see \autoref{Auto_FE}), statistical nested rolling-window and lag features consistently demonstrate higher impact scores compared to temporal features (e.g., hour of day and its sine/cosine transformations) across all methods. This dominance of historical pattern-based features may reflect strong autocorrelation in electricity demand time series, particularly due to building thermal inertia \cite{MartnezComesaa2020}, where past usage influences near-future demand through gradual thermal changes, making historical patterns more predictive than calendar-based patterns for next-day forecasting. The consistent importance of lag and rolling-window features across different methods (PTO, SPO$^{+}$, and DBB) suggests that these feature types capture fundamental underlying relationships in energy data that remain valuable regardless of the learning approach employed. 

While feature importance analysis is inherently dataset-dependent and these observations may vary across different datasets and environments, a key practical insight emerges in real-world BESS optimisation scenarios where additional exogenous features are unavailable (e.g., when weather data are not accessible), prediction performance can still be substantially improved by generating richer inputs from timestamp and historical data alone through AFE. This finding has potential implications for deployment in data-constrained environments where external data sources may be unreliable or unavailable.
}


\begin{figure}[!t]
\centering
\includegraphics[width=1\linewidth]{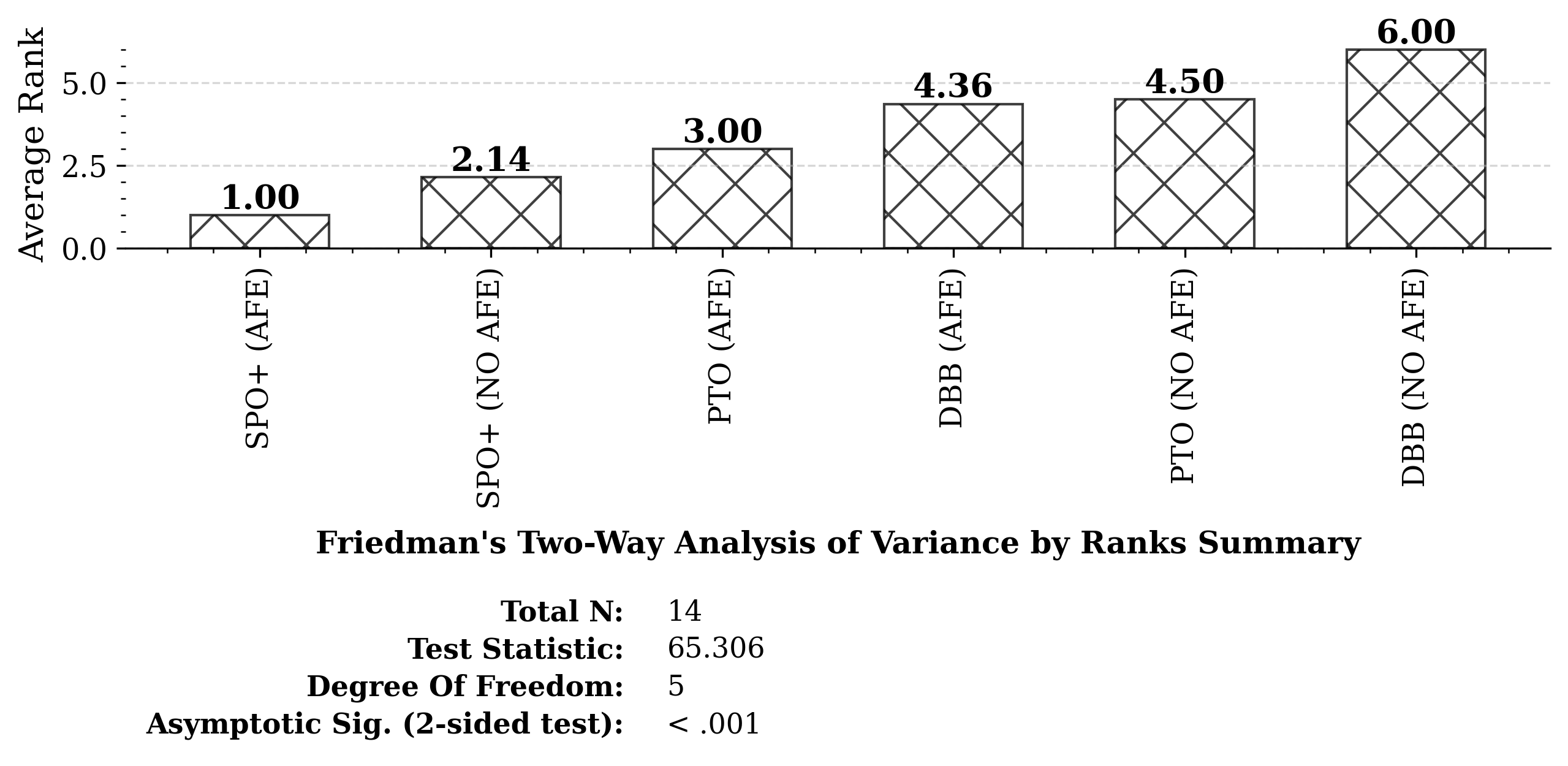}
\caption{Result of comparisons on the test set between the PTO and DFL (SPO$^{+}$ and DBB) methods, with and without AFE, using the Friedman average ranking (lower is better). The significance level is $0.05$.}
\label{Friedman_Ranking_DFL_paper}
\end{figure}



\begin{table}[!t]
\centering
\footnotesize
\caption{\revision {Results of pairwise comparisons on the test set between the PTO and DFL (SPO$^{+}$ and DBB) methods, with and without AFE, using the Wilcoxon signed-rank test. Each row tests the null hypothesis that the distributions of Sample 1 (i.e., Method 1) and Sample 2 (i.e., Method 2) are the same. Asymptotic p-values (2-sided tests) are displayed. The significance level is $0.05$. P-values below the significance level indicate rejection of the null hypothesis. Values reported as < .001 signify p-values less than 0.001, providing strong statistical evidence.\\
\footnotesize{
$^a$ Significance values have been adjusted by the Bonferroni correction for multiple tests.}}}
\begin{tabular}{@{} p{6cm} c c c c @{}} 
\toprule
Method 1 vs Method 2 & Test Statistic & Std. Test Statistic & Sig. & Adj. Sig.$^{\text{a}}$ \\
\midrule
SPO$^{+}$ (AFE) vs SPO$^{+}$ (No AFE) & -1.143 & -1.616 & .106 & 1.000 \\
SPO$^{+}$ (AFE) vs PTO (AFE) & 2.000 & 2.828 & .005 & .070 \\
SPO$^{+}$ (AFE) vs DBB (AFE) & -3.357 & -4.748 & $< .001$ & .000 \\
SPO$^{+}$ (AFE) vs PTO (No AFE) & 3.500 & 4.950 & $< .001$ & .000 \\
SPO$^{+}$ (AFE) vs DBB (No AFE) & -5.000 & -7.071 & $< .001$ & .000 \\
SPO$^{+}$ (No AFE) vs PTO (AFE) & 0.857 & 1.212 & .225 & 1.000 \\
SPO$^{+}$ (No AFE) vs DBB (AFE) & -2.214 & -3.131 & .002 & .026 \\
SPO$^{+}$ (No AFE) vs PTO (No AFE) & 2.357 & 3.334 & $< .001$ & .013 \\
SPO$^{+}$ (No AFE) vs DBB (No AFE) & -3.857 & -5.455 & $< .001$ & .000 \\
PTO (AFE) vs DBB (AFE) & -1.357 & -1.919 & .055 & .824 \\
PTO (AFE) vs PTO (No AFE) & -1.500 & -2.121 & .034 & .508 \\
PTO (AFE) vs DBB (No AFE) & -3.000 & -4.243 & $< .001$ & .000 \\
DBB (AFE) vs PTO (No AFE) & 0.143 & 0.202 & .840 & 1.000 \\
DBB (AFE) vs DBB (No AFE) & -1.643 & -2.323 & .020 & .302 \\
PTO (No AFE) vs DBB (No AFE) & -1.500 & -2.121 & .034 & .508 \\
\bottomrule
\end{tabular}

\label{tab:pairwise_comparisons}
\end{table}
\subsection{Statistical Significance Analysis}
\label{Statistical_Significance}
As shown in \autoref{Friedman_Ranking_DFL_paper}, Friedman's test applied to the fourteen-day test horizon reveals a statistically significant global difference among the six methods ($p < .001$), providing compelling evidence that some of the observed performance variations reflect genuine methodological distinctions in handling BESS optimisation under data scarcity rather than random chance. The mean ranks demonstrate that methods with AFE occupy the top three positions out of the top four ranks, with SPO$^{+}$ achieving the best rank, suggesting the superior decision-making efficacy of DFL approaches when combined with AFE for BESS optimisation problems.

Pairwise comparisons using the Wilcoxon signed‑rank test with Bonferroni correction, as shown in \autoref{tab:pairwise_comparisons}, pinpoint where differences are statistically meaningful. SPO$^{+}$ (AFE) demonstrates statistically significant superiority over multiple competing methods. Most notably, SPO$^{+}$ (AFE) significantly outperforms both DBB variants and the PTO approach without AFE (adjusted $p < .001$), reflecting substantial and consistent reductions in regret across the test period. Intriguingly, differences between SPO$^{+}$ (AFE) and PTO (AFE) or between SPO$^{+}$ (AFE) and its own no‑AFE baseline are not significant after correction (adjusted $p=0.070$ and $1.000$, respectively), suggesting that the superiority of SPO$^{+}$ (AFE) over PTO (AFE) or improvements from AFE alone cannot be firmly established (i.e., are not statistically robust under correction). 

\revision{None of the within‑method comparisons (i.e., PTO (AFE) vs PTO (No AFE), DBB (AFE) vs DBB (No AFE), or SPO$^{+}$ (AFE) vs SPO$^{+}$ (No AFE)) remain significant after Bonferroni adjustment, despite unadjusted $p$‑values below 0.05. Nevertheless, AFE consistently provided substantial practical improvements across all methods, with performance gains of 56.48\%, 19.42\%, and 22.87\% for SPO$^{+}$, PTO, and DBB, respectively, without degrading performance on any test day. The conservative Bonferroni correction reduces the power to detect within-method AFE benefits, and the limited test horizon of fourteen days, while sufficient to detect strong global differences, constrains the statistical power to identify more subtle pairwise differences after stringent multiple comparison corrections. Despite these statistical limitations, the consistent directional improvements and substantial magnitude of performance gains across all methods provide compelling evidence for the practical value of AFE in BESS optimisation problems.}

\section{Summary}
\label{ch6:Conclusion}

This work addresses gaps that challenge effective BESS optimisation, stemming from: A) DFL methods promise to align prediction with downstream objectives \cite{Wilder2019, Mandi2020}; however, they are relatively new and have been tested primarily on synthetic datasets or small-scale problems (i.e., simplified benchmarks) \cite{mandi2023towards, Zhou2024}, highlighting the need to assess their practical viability in real-world applications such as BESS problems; and B) real-world datasets often exhibit greater variability and data scarcity due to practical constraints \cite{grinsztajn2022tree, hollmann2022tabpfn} which can compromise DFL performance and necessitate enhanced feature representations to extract richer information from limited data without the need for domain expertise. 

This work proposes a decision-aware, end-to-end ML framework that directly addresses each gap: first, to overcome data scarcity limitations, the framework leverages domain-specific AFE \cite{Alkhulaifi2025} to extract richer representations without heavy reliance on domain expertise; second, rather than treating forecasting and optimisation as separate tasks, the framework jointly learns to forecast electricity demand and prices while optimising battery operations using a regret-based objective, ensuring prediction errors directly inform decision quality; and third, the work evaluates this framework using a novel dataset collected from a UK household property, providing empirical assessment and evidence of the practical viability of DFL methods in real-world BESS applications.

In this framework, two DFL methods (SPO$^{+}$ and DBB) and a PTO approach are compared under baseline and AFE-enhanced conditions. The results show that AFE improves the performance of all three methods, with particularly pronounced benefits for the DFL approaches (SPO$^{+}$: 56.48\%, DBB: 22.87\%, PTO: 19.42\%). Statistical analysis further confirms significant global differences among methods ($p < .001$), with SPO$^{+}$ (AFE) ranking the best (i.e., lowest regret), demonstrating its superior decision-making efficacy for BESS optimisation under data scarcity. The significance of this work extends beyond the proposed framework and methodological comparisons to practical implications for energy management systems, especially pertinent for real-world BESS deployments, where acquiring large, high-quality datasets is difficult and manual FE is time-consuming and error-prone. \revision{These findings suggest that, for the studied dataset and operating conditions, aligning prediction and optimisation with FE \textit{may} improve economic outcomes (as measured by regret and realised cost). However, this evidence comes from a single real-world case study and therefore should be interpreted as indicative rather than universal; validating the economic benefit across additional sites, seasons, and market conditions remains necessary.} 

\revision{Note that, although this chapter evaluates the proposed AFE--DFL framework on a single real-world dataset, this is an intentional case-study design rather than an attempt to define a universal benchmark. DFL remains an emerging paradigm \cite{Mandi2024}, and much of the literature (see \autoref{ch2:Paper4}) evaluates such methods on synthetic decision problems, with comparatively limited evidence in real operational energy settings where forecasts interact with BESS physical constraints and tariffs \cite{Mandi2024,Wilder2019}. Using a real BESS dataset, therefore tests whether the proposed approach is feasible and beneficial in real-world settings such as data scarcity and constraint sensitivity that toy setups often abstract away (referred to as \textit{toy-level} problems in \cite{Mandi2020, Zhou2024}). Accordingly, the findings can be interpreted as supporting evidence, while broader generalisation requires validation across additional sites, seasons, and operating conditions.} \revision{The limitations of this work are discussed in \autoref{con:Research_Limitations}.}

\cleardoublepage
\chapterwithquote{%
  \parbox{0.9\textwidth}{\raggedright
  "We can only see a short distance ahead, but we can see plenty there that needs to be done" - Alan Turing}%
}{Conclusions, Limitations, and Future Work}{ch:conclusion}


This chapter concludes the thesis by synthesising the research findings and reflecting on their broader significance. \autoref{con:Main_Contributions} summarises the key contributions of this work in relation to the research aim and objectives. \autoref{con:Research_Limitations} acknowledges the limitations while \autoref{con:Future_Work} outlines directions for future research. Finally, \autoref{Potential_Applications_Beyond_ECF} discusses how AutoEnergy (see \autoref{ch:mainchapter5}) and the AFE-DFL framework (see \autoref{ch:mainchapter6}) developed in this thesis could be applied beyond ECF problems.
\newpage 

\section{Main Contributions}
\label{con:Main_Contributions}
The aim of this thesis was to develop ML models for ECF that minimise reliance on domain knowledge and can be applied across diverse energy systems while maximising forecasting accuracy and optimising downstream tasks. \autoref{tab:Conclusion_Chapters_Novelties} summarises the individual contributions of each chapter included in this thesis. The specific objectives set to achieve the overall thesis aim and conclusions relevant to these objectives are discussed below:

\begin{table}[!t]
\centering
\caption{Summary of publications arising from this thesis.}
\scriptsize
\begin{tabular}{>{\centering\arraybackslash}m{1.6cm}|%
                >{\raggedright\arraybackslash}m{2.8cm}|%
                >{\justifying\noindent\arraybackslash}m{7.5cm}}
\toprule
\textbf{Chapter} & \textbf{Title of Article} & \textbf{Novelty} \\
\midrule
\autoref{ch:mainchapter3} & Machine Learning Pipeline for Energy and Environmental Prediction in Cold Storage Facilities & Proposed a comprehensive ML pipeline for ECF capable of one-week-ahead hourly forecasting \revision{and evaluated under data-constrained settings}. The pipeline was validated on two novel UK-based cold storage facility datasets. Systematic evaluation of eight FS techniques with extensive analysis of domain knowledge impact, feature importance, and dataset size implications in ECF applications. \revision{This provides a baseline for later automation of FE.} \\
\hline

Appendix~\ref{ch:mainchapter4} & Exploring Automated Feature Engineering for Energy Consumption Forecasting with AutoML & Developed an ECF-specific AFE prototype to \revision{reduce} the domain expertise needed for FE. \revision{Reported performance gains in the evaluated settings} across four AutoML frameworks (AutoGluon, H2O, TPOT, and FLAML). \\
\hline

\autoref{ch:mainchapter5} & AutoEnergy: An Automated Feature Engineering Algorithm for Energy Consumption Forecasting with AutoML & Introduced \textit{AutoEnergy}, a novel, fully \revision{automated FE} algorithm tailored for ECF that generates interpretable features from timestamps and past consumption values through rule-based transformations, integrating with AutoML to \revision{reduce} human intervention. Evaluated across eighteen diverse real-world energy datasets representing different energy systems, AutoEnergy \revision{showed promising generalisability across the evaluated datasets while typically reducing error and achieving competitive processing time} compared to existing methods, with \revision{integration with the TabPFN foundation model on eligible datasets}. \\
\hline

\autoref{ch:mainchapter6} & Decision-Focused Learning Enhanced by Automated Feature Engineering for Energy Storage Optimisation & Leveraged AutoEnergy to \revision{support} the downstream task and \revision{study} the nascent DFL, validated through a novel BESS dataset in real-world settings. \revision{On the evaluated case study,} results show that incorporating AutoEnergy further improves \revision{DFL decision quality} by achieving 22.9-56.5\% lower operating costs compared to the same models without it. \\

\bottomrule
\end{tabular}
\label{tab:Conclusion_Chapters_Novelties}
\end{table}



\begin{enumerate}
  \revision{
    \item \textbf{RO1: To establish a baseline ML model for ECF and investigate the role of domain knowledge in FE for such forecasting problems.}}\\

\revision{This objective was addressed in \autoref{ch:mainchapter3} by developing and validating an end-to-end ECF pipeline on two real-world datasets collected purposefully for this investigation.  This comprehensive approach ultimately produces a practical pipeline that guides practitioners in efficiently implementing domain knowledge-based features for ECF applications and establishes the empirical foundation necessary for subsequent FE automation efforts. The findings underscore the importance of FE, given that the engineered features have a noticeable impact on model outputs, as evidenced by SHAP analysis.}

\revision{

    \item \textbf{RO2: To develop an AFE method for ECF problems, thereby streamlining ML model development by addressing the most time-consuming and expert-dependent task in the pipeline.}} \\

\revision{This objective was fulfilled through the development of \textit{AutoEnergy} and presented in \autoref{ch:mainchapter5}.AutoEnergy automatically generates interpretable features from timestamps and past consumption values through rule-based transformations, integrating them with AutoML for fully automated ECF modelling while reducing human intervention.} The performance of AutoEnergy was evaluated using eighteen diverse real-world energy consumption datasets spanning residential, commercial, industrial, and grid power domains. Through extensive benchmarking against baseline AutoML without FE and established FE methods, namely TSFresh (with TSEff and TSMin configurations) and FT, AutoEnergy demonstrated significant improvements in both predictive accuracy and computational efficiency. These findings highlight AutoEnergy's potential to improve AutoML performance while reducing reliance on domain expertise for FE, paving the way for fully automated ML pipelines in ECF applications. Satisfying this objective produced an AFE algorithm that researchers can use to automatically generate interpretable features tailored for ECF problems, minimising both time and expertise barriers to accurate ECF modelling while improving AutoML performance.


\revision{
        \item \textbf{RO3: To leverage the proposed AFE algorithm to optimise the downstream tasks beyond predictive performance.}}\\

\revision{This objective was fulfilled through the work in \autoref{ch:mainchapter6}. Traditional two-stage approaches for BESS treat forecasting and optimisation as separate processes, allowing prediction errors to cascade into suboptimal decisions. While DFL methods promise to address this limitation by integrating prediction and optimisation, their practical viability remains unproven due to evaluation being confined to synthetic datasets or small-scale problems, see our discussion and review of DFL in \autoref{ch2:Paper4}. Additionally, real-world applications are further constrained by data scarcity resulting from collection limitations.}

To address these challenges, this work proposed an end-to-end framework that jointly forecasts electricity prices and demand for the next day while optimising BESS operations using a regret-based objective, thereby eliminating the separation between prediction and optimisation. To overcome data limitations, the framework integrates AFE that extracts richer representations while minimising domain expertise requirements. To address the lack of real-world DFL evaluation, effectiveness is demonstrated through evaluation on a novel real-world UK-based BESS dataset, comparing two DFL methods (SPO$^{+}$ and DBB) against a traditional two-stage approach with and without FE. On average, SPO$^{+}$ achieves the lowest regret, representing a 67.16\% reduction compared to the two-stage method. The introduction of FE further amplified these advantages, substantially enhancing performance across all methods (SPO$^{+}$: 56.48\%, DBB: 22.87\%, two-stage: 19.42\%). \revision{For this dataset and these operating conditions, the results indicate that incorporating AFE into DFL may lead to better economic performance, reflected in lower regret and reduced realised cost.}

\end{enumerate}

\revision{It is worth clarifying that, while this thesis considers data scarcity settings, it does not claim to provide a comprehensive robustness analysis of data scarcity across all energy systems. However, \autoref{Ch3_results_Dataset_Size_Implications} includes an additional experiment examining how reducing the training-set size affects forecasting performance (i.e., dataset size implications). This analysis is necessarily limited in scope and does not replace a systematic, multi-site robustness evaluation across different energy systems. Accordingly, claims about handling data scarcity should be interpreted as evidence from the studied cases, with broader generalisation requiring further validation.}


\newpage
\section{\revision{Research Limitations}}
\label{con:Research_Limitations}

\revision{Several limitations emerged throughout the research and should be acknowledged and discussed. These limitations are discussed in relation to the specific objectives set to achieve the overall thesis aim.}

\textbf{RO1: To establish baseline ML models for ECF and investigate the role of domain knowledge in FE for such forecasting problems.}

\revision{While providing empirical evidence towards a robust baseline pipeline for ECF, particularly in FDCS environments, and offering practitioners a reproducible approach for incorporating domain-informed features in similar domains, several limitations should be considered when interpreting and applying the findings. First, the datasets used for validation were predominantly collected during colder months (late October to late January), so the learned relationships may not fully reflect energy behaviour under warmer conditions, different defrost regimes, or altered thermal loads. Second, the generalisability of the resulting models is only partially evidenced: although evaluation across two FDCS sites suggests potential transferability, it does not systematically capture variation in facility scale, equipment configurations, control policies, or geographic and climatic contexts that could materially affect performance and feature relevance.}


\textbf{RO2: To develop an AFE method for ECF problems, thereby streamlining ML model development by addressing the most time-consuming and expert-dependent task in the pipeline.}

\revision{Although the proposed algorithm (AutoEnergy) has the potential to enable domain practitioners (e.g., energy engineers) and the wider ML community working on similar forecasting tasks to automatically generate interpretable features, reducing reliance on specialist domain knowledge and iterative manual feature engineering while improving AutoML performance, several limitations should be acknowledged.} First, the algorithm's FE process is specifically tailored to energy time series data, which may limit its generalisability to other domains without substantial modifications. Second, while AutoEnergy has demonstrated robust performance on datasets of up to approximately 145k samples (e.g., the PJME and PJMW datasets), an evaluation of its scalability on much larger datasets and real-time streaming scenarios remains necessary to confirm its computational efficiency under industrial-scale workloads. 

Third, while the empirical sensitivity analysis provides evidence supporting the one-third sequence-length constraint for nested window features, this parameter remains heuristically determined rather than theoretically \revision{established}. \revision{In addition, the interpretability evidence reported for the AutoML experiment (see \autoref{Analysis_Feature_Statistical_Testing}) is intentionally limited to permutation-based feature importance, as this is the standard library-level explanation output available across AutoGluon's automated ensembles, and more detailed model-specific explanations would require fixing a single learner or exporting models for post-hoc analysis}. \revision{Finally, the evaluation scope is empirically constrained. Accordingly, although five pipelines and eighteen diverse datasets were assessed alongside established benchmarks, it is not feasible to compare AutoEnergy against the full range of FE methods given the rapid pace of AutoML development and the diversity of domain-specific methods.}


\textbf{RO3: To leverage the proposed AFE algorithm to optimise the downstream tasks beyond predictive performance.}

Through fulfilling this objective, the development of AFE was enhanced to not only maximise forecasting accuracy but also be decision-aware, thus delivering tangible operational value in practical ECF applications. However, certain limitations warrant consideration.\revision{The study’s relatively small dataset (55 days) collected from a single UK property during winter months (i.e., from 1 January to 24 February 2025) does impose substantial generalisation limitations. Concretely, the empirical findings are case-study specific: (A) they do not establish that the same cost reductions or regret improvements will hold under different regions, tariffs, market structures, or regulatory settings; and (B) the winter-only period does not represent seasonal changes in consumption, renewable availability, or temperature-driven demand behaviours. This matters because the DFL optimiser is sensitive to the joint distribution of energy demand, price, and operational constraints; shifting these conditions can change both the learned forecasting signal and the resulting schedule.}

Furthermore, hyperparameter optimisation was constrained to a limited search space due to computational resource limitations, as detailed in \autoref{tab:hyperparameters_space}, which may have prevented the discovery of optimal model architectures and learning parameters that could have further improved performance. 
Another limitation worth acknowledging is that the BESS optimisation model uses several assumptions (e.g., treating all operational parameters as deterministic and modelling the system as a single aggregated unit) that may not fully capture the complexities and uncertainties inherent in real-world BESS installations.

Moreover, although the proposed framework and conducted experiments used a real-world dataset, the conclusions drawn, though insightful, \revision{remain data-dependent} and would require validation across diverse geographical locations, temporal periods, and energy system configurations to establish broader generalisability and robustness of the DFL-AFE approach. \revision{Accordingly, any economic benefit claims should be treated as case-study evidence, and may not hold under different assets, tariffs, or operational constraints without further validation.} \revision{Finally, the current framework is evaluated using point forecasts and regret distributions, without producing calibrated predictive uncertainty (e.g., prediction intervals) or using probabilistic forecasters that explicitly model forecast uncertainty, which limits risk-aware decision-making and uncertainty propagation into the optimisation stage.}


\revision{
\section{\revision{Future Work}}
\label{con:Future_Work}

\revision{The following sections outline promising research directions that build upon the work presented in this thesis.}

\subsection{\revision{Data Collection and Seasonal Generalisation}}
Expanding the empirical evidence base is crucial for establishing the robustness of the proposed approaches. Future studies should prioritise data collection across different seasons, particularly warmer months (unlike the two datasets collected for the investigation in \autoref{ch:mainchapter3}), to better understand how varying outdoor conditions and seasonal dynamics affect FDCS environments and ML model performance. Complementary to this, validation efforts should focus on enhancing the universality of these models across various geographical locations and operational contexts, exploring their applicability across a wider spectrum of FDCS settings with different operational scales and regulatory environments.

\subsection{\revision{Algorithm Enhancement and Scalability}}
The AutoEnergy algorithm, proposed in \autoref{ch:mainchapter5}, can be further refined through improved optimisation methods that dynamically adjust the maximum window length (\autoref{Lags_windows}) based on underlying temporal patterns and series characteristics, rather than relying on heuristic determination. \revision{Future work may also explore hybridising AutoEnergy with domain-agnostic feature generators such as genetic programming-based feature synthesis \cite{Guo2005_Genetic_Programming, Gulati2025_Genetic_Programming_limitations} and anomaly or change-point detection signals, particularly for problems where rare events dominate the decision objective.} These enhancements would position AutoEnergy as a more comprehensive step towards end-to-end, fully automated ML models for ECF and similar forecasting problems.

\subsection{\revision{Framework Extension to Integrated Energy Systems}}
Extending the proposed DFL-AFE framework, introduced in \autoref{ch:mainchapter6}, to incorporate multiple renewable generation sources (e.g., solar panels) would provide a more comprehensive assessment of DFL's potential in integrated battery energy storage systems. Additionally, testing the framework across longer forecasting horizons, beyond the current one-day-ahead predictions, would strengthen the evidence for DFL's applicability in real-world BESS operations and support longer-term energy management strategies.

\subsection{\revision{Multi-objective Optimisation and Environmental Considerations}}
While the used regret metric (see \autoref{Evaluation_metrics}), focuses solely on cost minimisation, future work could incorporate multi-objective optimisation frameworks that balance economic objectives with environmental considerations, such as carbon emissions reduction, or operational constraints, such as user comfort requirements. This would enable BESS systems to contribute to broader sustainability goals while maintaining economic viability.

\subsection{\revision{Advanced Learning Paradigms}}
Investigating the proposed AFE-DFL framework further with other learning paradigms offers considerable promise. Ensemble methods that combine multiple DFL approaches, transfer learning techniques \cite{Gunduz2023_Transfer_learning} that leverage knowledge from larger energy datasets, and hybrid approaches combining AFE-DFL with reinforcement learning \cite{Shen2024_Reinforcement_learning}, or meta-learning frameworks may enable more adaptive and robust BESS optimisation strategies. These approaches could facilitate rapid adjustment to changing energy market conditions and emerging operational requirements. Additionally, exploring other DFL approaches, such as noise-contrastive estimation \cite{Mulamba2021}, may yield further improvements in decision quality.

\subsection{\revision{Uncertainty Quantification}}
Future work may investigate risk-aware scheduling formulations that propagate predictive uncertainty into the optimisation stage. This includes developing and reporting explicit predictive uncertainty quantification (e.g., calibrated prediction intervals) and exploring uncertainty-aware forecasting methods that can better account for forecast variability when making operational decisions. Such approaches would enhance the robustness of energy management strategies and improve their reliability in real-world deployments.

\subsection{\revision{Generative Model Integration}}
\label{con:Generative_Model_Integratio}

Future work could explore integrating recent generative time-series models \cite{generative_model_Langevin2023} as probabilistic forecasters within the proposed pipeline, either to provide calibrated predictive distributions (and sampled scenarios) for uncertainty-aware or risk-aware optimisation, or to augment small real-world datasets via conditional generation. For example, pretrained probabilistic foundation models for time series forecasting (e.g., Lag-Llama \cite{generative_model_Lag-Llama_2023}, Chronos \cite{generative_model_Ansari_Chronos}) can produce distributional forecasts that are potentially aligned with downstream decision-making under uncertainty, while diffusion-based predictors (e.g., TimeDiff \cite{generative_model_Shen_TimeDiff}) offer an alternative generative route for scenario generation and robust forecasting. Such integration could complement AutoEnergy by combining interpretable rule-based features with learned generative representations, while also enabling more principled uncertainty reporting beyond point forecasts.

}


\newpage
\section{Potential Applications Beyond Energy Consumption Forecasting}
\label{Potential_Applications_Beyond_ECF}

Beyond ECF problems, the methodologies investigated and developed in this thesis, particularly the AFE method embodied in AutoEnergy \cite{Alkhulaifi2025} (see \autoref{ch:mainchapter5}), are potentially transferable to other time series or energy-like forecasting problems that exhibit similar modelling challenges (i.e., heavy reliance on domain expertise for extracting and capturing multi-scale temporal patterns, dealing with real-world small datasets, ensuring forecasts support downstream decision tasks). For instance, AutoEnergy could be applied to air-quality and pollution prediction in environmental monitoring \cite{Liu2021_air_quality}, where diurnal and seasonal cycles are present, as well as to gas \cite{Yukseltan2021_gas_Demand} and water \cite{GonzlezPerea2019_water_Demand} demand forecasting, where daily and seasonal usage is shaped by weather and operational regimes. It could also potentially be applied to healthcare resource utilisation forecasting problems \cite{ Tello2022_bed_demand, GarciaVicua2023_patient_flow}, such as bed occupancy and patient flows, which may follow daily and weekly patterns that reflect variations in demand and staffing. In addition, it may be applied to traffic flow forecasting \cite{Razali2021_traffic_flow_AI, Sayed2023_traffic_flow_AI}, where rush-hour peaks, weekly commuting cycles, and seasonal variations are observed. 

To some degree, a common thread across these applications is the presence of consumption or usage patterns driven by human behaviour, repeated operational procedures, and external factors, which makes them potential candidates for AutoEnergy’s methodological framework without fundamental algorithmic changes. \revision{It is worth acknowledging that other AFE frameworks may exist in the wider literature, and the claim here is not that the proposed AFE method is novel across all contexts, but that it is purposefully designed and tailored to the ECF problems. Rather, it is a domain-specific AFE method tailored to ECF problems, which is comprehensively empirically validated across diverse energy datasets. On this basis, it may also be applicable to energy-like domains that exhibit similar patterns; however, such transferability is prospective, conditional, and would require empirical validation in each target domain.}

\revision{The rationale for potential transfer beyond ECF is that AutoEnergy (see \autoref{ch:mainchapter5}) mainly encodes generalisable modelling knowledge. In particular, some \textit{energy-like} forecasting problems share recurring cycles (daily, weekly, seasonal), autocorrelation, and multi-scale temporal variation, which makes the same families/categories of engineered representations plausible starting points. However, transferability is conditional: where a target domain is dominated by non-temporal drivers, regime shifts, or rich exogenous signals, the marginal benefit of these rule-based temporal features may be smaller. Therefore, the contribution should be interpreted as a reusable AFE baseline for time series with recurring usage patterns, not as a universal replacement for domain expertise.}

The DFL investigation presented in this thesis \cite{alkhulaifi2025DFL} (see \autoref{ch:mainchapter6}) addresses the broader operational forecasting challenge of ensuring that automated predictive gains translate into better outcomes in real-world PTO settings under data scarcity. For instance, this work could potentially extend to water management \cite{Kavya2023_water_Demand_AI}, where demand forecasts inform pumping schedules and reservoir control, helping to optimise distribution costs and maintain supply reliability during periods of peak usage. Another potential application lies in healthcare, where DFL could support more effective resource allocation \cite{ Kim2024_Bed_Capacity} by linking patient-flow forecasts directly to staffing and capacity planning, thereby helping to minimise wait times and improve care quality within budgetary constraints. Supply chain demand forecasting \cite{Feizabadi2020_ML_supply_chain, Tirkolaee2021_ML_Supply_chain, Zhu2021_supply_chain_ML} offers another potential application, where DFL could link demand predictions to inventory and replenishment decisions, thereby helping to reduce costs and mitigate stockout risks. To some extent, these applications beyond ECF share a PTO structure in which forecasts feed downstream optimisation. 

Future work, therefore, could adapt the proposed DFL framework by formulating domain-specific cost functions and operational constraints and by leveraging AutoEnergy so that forecasting gains carry through to decisions. In other words, while the problem formulation presented in \autoref{ch6:Method}, particularly the mathematical formulation in \autoref{Mathematical_model}, is tailored specifically for BESS applications, the overarching framework methodology is adaptable to other contexts. The core principles of jointly forecasting uncertain parameters while optimising operational decisions remain applicable, though the specific constraints, decision variables, and objective functions would require modification for different systems.


\newpage
\revision{

\section{\revision{Reflection in Light of Recent Advances in AI}}
\label{Reflection_on_recent_AI}

Looking back at this thesis from the vantage point of the last four years, I am genuinely struck by how quickly the centre of gravity in AI has shifted towards foundation models \cite{Hollmann2025}, general-purpose AI \cite{triguero2024general}, and LLM-based tools \cite{zhang2025opportunities}. When I began my PhD journey, it was tempting to assume that ever-larger models would simply make domain-aware FE obsolete. Instead, the work repeatedly pulled me in the opposite direction: the most stubborn failures came from real constraints that recent AI progress does not magically erase, namely limited and messy data, high stakes around interpretability, and the fact that forecasting is only useful insofar as it improves downstream decisions. In that sense, the thesis sits slightly sideways to the mainstream narrative of scale: it argues, through evidence across multiple energy settings and environments, that carefully chosen inductive bias (here, domain-aware temporal features with traceable provenance) can be a pragmatic substitute for data and can make automation more trustworthy rather than merely more opaque.

This perspective also sharpens what I see as the core trade-off in the AutoEnergy algorithm proposed in \autoref{ch:mainchapter5}. Recent advances have produced powerful general tabular learners (including transformer-based approaches such as TabPFN \cite{Hollmann2025}), yet the results in this thesis suggest that domain-specific structure still matters: a general learner can benefit materially from being handed features that encode energy-relevant periodicity, lag structure, and multi-scale dynamics in a human-readable form. At the same time, the thesis makes clear where this approach is brittle. AutoEnergy is intentionally heuristic and predominantly target-and-timestamp driven, so its marginal value shrinks when high-quality exogenous drivers already dominate, and its computational profile still needs stress-testing in genuinely large-scale and streaming regimes. If I were re-running the research today, I would make these boundary conditions explicit from the outset, because the current trend towards \textit{one model for everything} can encourage an uncritical assumption that scale alone will dominate. In practice, this thesis suggests there are still real-world systems (e.g., buildings, wind turbines) where data are limited, and domain structure is decisive, so investing in a robust, automated, theory-informed feature representation remains not just helpful, but necessary.

Finally, the most important shift in how I now frame the thesis is that it is not really about forecasting accuracy as an endpoint; it is about aligning learning with use. The decision-focused work introduced in \autoref{ch:mainchapter5} anticipates a broader movement towards objective alignment, where models are judged by the quality of the decisions they enable rather than by proxy losses alone. The thesis shows that this alignment is not automatic: sophisticated paradigms can underperform in real, data-scarce settings unless the representation is strengthened, which is precisely where automated, interpretable FE proved valuable. My main takeaway, and the clearest direction for future work, is to deliberately combine the strengths of recent AI advances (e.g., pretrained tabular models such as TabPFN) with the safeguards this thesis leans on (traceability, domain constraints, and decision-level evaluation), rather than treating them as competing approaches.

}

\begin{appendices}
\setcounter{section}{0}
\renewcommand\thesection{\Alph{section}}
\renewcommand*\sectionautorefname{Appendix}
\renewcommand*\subsectionautorefname{Appendix}

\counterwithout{figure}{chapter}
\counterwithout{table}{chapter}
\counterwithin{figure}{section}
\counterwithin{table}{section}
\renewcommand\thefigure{\thesection.\arabic{figure}}
\renewcommand\thetable{\thesection.\arabic{table}}

\cleardoublepage
\cleardoublepage

\section{Exploring Automated Feature Engineering for Energy Consumption Forecasting with AutoML}

\label{ch:mainchapter4}


This preliminary \revision{work}, titled "\href{https://doi.org/10.1109/SMC54092.2024.10831959}{Exploring Automated Feature Engineering for Energy Consumption Forecasting with AutoML}", initiates the pursuit of \textbf{RO2} by exploring the automation of FE in ECF, proposing an initial AFE method designed to minimise reliance on domain-specific expertise. The \revision{work} presents a preliminary study on FE automation, demonstrating improved performance when the proposed method is integrated with four different AutoML frameworks. By reducing the most time-consuming and expert-dependent task in the ML pipeline, this work establishes the foundational approach for streamlining ML model development in ECF while paving the way for more sophisticated automation techniques in subsequent research presented in \autoref{ch:mainchapter5}.
\subsection{Introduction and Background}
\label{ch4:Intro}

Amid rising energy demands and costs, enhancing energy efficiency is essential for cutting expenses and meeting environmental goals. Accurately predicting energy consumption can be key to optimising operations and reducing consumption. Machine learning (ML) methods have been utilised to forecast energy consumption by framing it as a regression problem, where algorithms learn the correlation between inputs (e.g., previous values) and outcomes (future values) \cite{zhang2021review}. These methods have found application across various energy domains, including commercial and residential buildings, Heating, ventilation, and air conditioning systems, smart grids and renewable energy \cite{Fan2019Deep, zhang2021review}. However, developing such models involves multiple steps such as engineering features, preprocessing data (e.g., handling missing data), selecting the appropriate model, and optimising hyperparameters, all of which can be time-consuming and require domain knowledge.

Recently, AutoML methods such as H2O \cite{h2o}, TPOT \cite{RN575}, AutoGluon \cite{AutoGluon2020}, and FLAML \cite{wang2021flaml} have emerged as promising methods designed to streamline the ML pipeline. Some of these AutoML methods have already been studied in the domain of energy consumption prediction \cite{Wang2019, RN558, RN549}. While these methods excel in automating preprocessing tasks, such as handling missing values and outliers, as well as in feature selection, model selection, and hyperparameter tuning, they still lack the capacity to generate new, useful, and interpretable features, which is crucial for improving predictions in complex environments such as those observed in energy consumption data.   

Having input features that effectively represent the relationship between input and output data is crucial for both ML and AutoML models. This is particularly important in the context of energy forecasting, where such systems exhibit both linear and nonlinear behaviours, irregular usage patterns, and weather influences \cite{manandhar2023current}. Although the use of input features for energy forecasting varies across different studies in the literature, the most prominent input features used are historical data and weather data \cite{sun2020review}. Historical data reveal consumption patterns, while weather data represents environmental factors, such as outdoor temperature, that influence energy use. Some studies have incorporated additional engineered features by transforming and deriving new attributes from the raw data, such as time-based features, sine and cosine transformations of cyclical features, lag features, and rolling-window statistical features to create a more comprehensive representation of energy temporal patterns \cite{Fan_2017, Moon_2019}. However, manually extracting these features can be time-consuming and requires a combination of energy domain expertise and data science skills. The energy domain experts can identify the key factors that drive energy consumption patterns, while the data scientists are needed to validate these observations through data analysis and to properly extract the relevant features from the raw data. 

Feature extraction techniques such as PCA \cite{Li_2015}, Wavelet Decomposition \cite{Peng_2022}, and Autoencoders \cite{Fan2019Deep} are also utilised in forecasting energy consumption but come with notable limitations. While PCA is effective at reducing data dimensionality, it can make the interpretation of the variables more challenging as the transformed principal components may not have a clear, intuitive relationship to the original data features. Similarly, wavelet decomposition produces coefficients that are more complex and less interpretable than the raw data and interpreting these wavelet features often requires specialised domain knowledge.  Autoencoders, meanwhile, produce abstract features with no clear link to the original data, making interpretation difficult. This interpretability issue is crucial in energy consumption forecasting, where comprehending the factors influencing energy use is key to developing informed policies and solutions.

Some works have been introduced in the literature to streamline the process of FE, such as Deep Feature Synthesis \cite{kanter2015deep}, Feature-Engine \cite{RN589}, Tsfresh \cite{Christ2018}, and Autofeat \cite{RN573}. While these methods can automate some aspects of FE, they still require some expertise in identifying appropriate lags and other important feature transformations. Moreover, methods such as Deep Feature Synthesis have demonstrated promising results in specific domains such as education and e-commerce, particularly in predicting project excitement and repeat buyer behaviour. However, the exploration of these methods in other sectors, such as the energy consumption prediction domain, remains limited. Considering these limitations, this work addresses this gap by presenting an AFE method tailored to enhance AutoML for energy forecasting problems. To assess and validate the impact of the proposed method on AutoML performance, eleven publicly available real-world datasets representing various energy consumption patterns in different domains were utilised. The contribution of this work lies in:

\begin{itemize}[leftmargin=15pt]
\item Improving  AutoML performance in forecasting energy consumption.
\item Reducing the need for domain knowledge needed for FE in energy consumption forecasting problems.
\item Robustness of the proposed method in enhancing AutoML across various energy consumption patterns in different domains (e.g., residential buildings, wind turbines, regional energy consumption) and different dataset sizes.
\item Enhance interpretability, as all newly generated features are explainable.
\end{itemize}

\subsection{The Proposed Automated Feature Engineering Method}
\label{ch4:method}
This section presents a definition of the problem under investigation in \autoref{ch4:problem_definitio}, followed by a description of the proposed method to address it in \autoref{ch4:Proposed_Method}.
\subsubsection{Problem Definition}
\label{ch4:problem_definitio}
Given an energy time series dataset $\mathcal{D}$ composed of $N$ instances, where each instance is represented by a tuple $(t_i, y_i)$, including a timestamp $t_i$ and a target variable $y_i$, we define the dataset as:
\begin{equation}
\mathcal{D} = \left\{ (t_1, y_1), (t_2, y_2), \ldots, (t_N, y_N) \right\}
\end{equation}

\noindent{The objective is to construct an algorithm $\mathcal{A}$ that employs an automatic, comprehensive FE approach to improve the performance of predictive model $\mathcal{M}$ within an AutoML framework.}

\subsubsection{Proposed Method}
\label{ch4:Proposed_Method}

\begin{figure}
\centering
\includegraphics[width=1\linewidth]{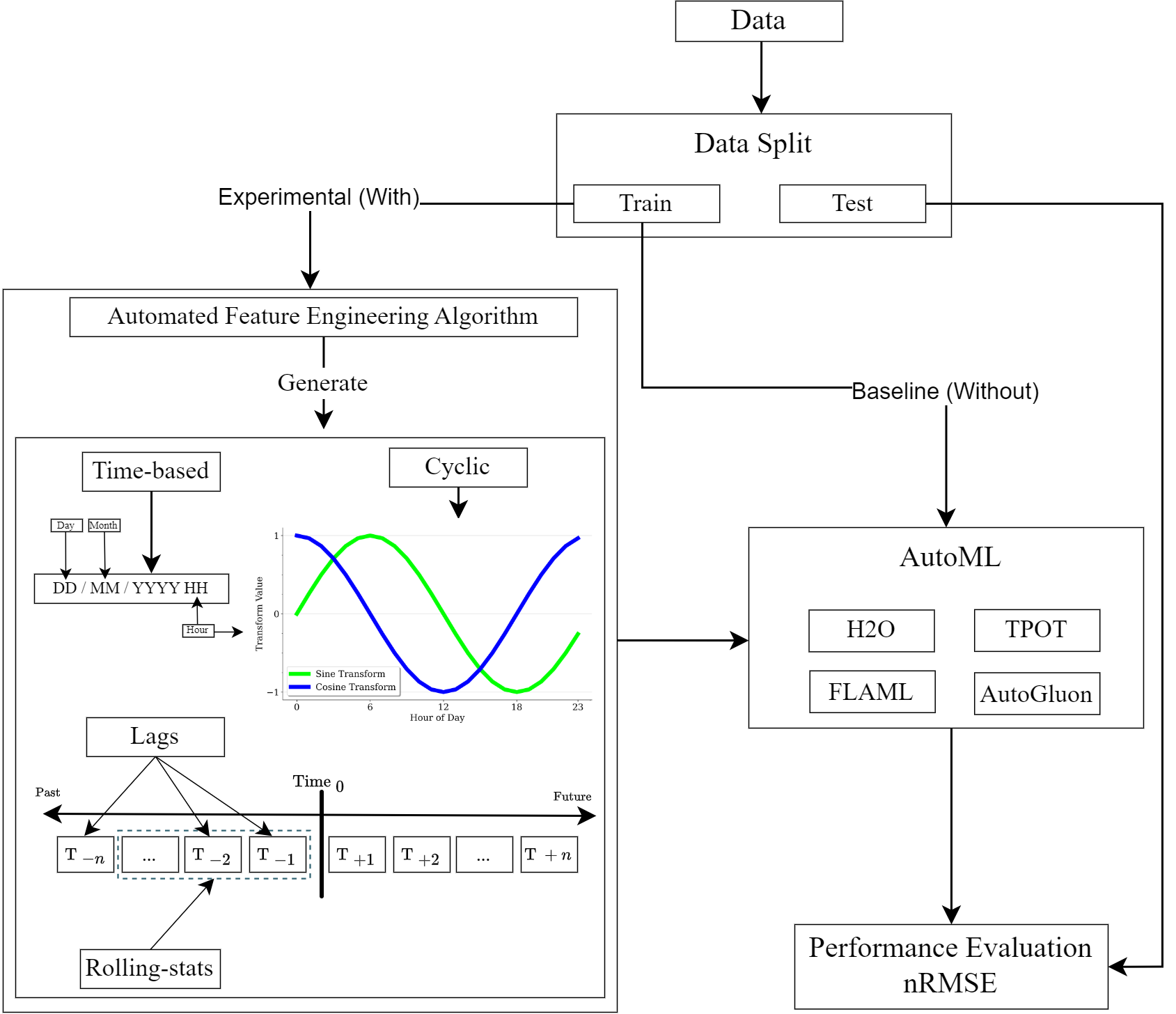}
\caption{Proposed FE algorithm and experimental setup.}
\vspace{-6mm}
\label{ch4:experimental_design_fig}
\end{figure}

The proposed \hyperref[alg:A]{Algorithm 1} executes a series of FE functions $\{\text{EngineeredFeatures}_j\}_{j=1}^{M}$, where these functions process the timestamp $t_i$ and the target variable $y_i$ to generate a subset of features $\mathbf{F}'_{i,j}$. While these generated features are commonly used in the literature, this method introduces a level of automation in their engineering (i.e., their extraction and selection), particularly for the lag and rolling-window statistical features, as explained below and as shown in \autoref{ch4:experimental_design_fig}:

\begin{itemize}[label=-, leftmargin=15pt]

\item Time-based features: these features leverage temporal data to discern patterns influenced by time, such as the hour of the day (e.g., on-peak and off-peak hours), the day of the week, and weekdays versus weekends. These features contain time-related information and can provide insights into patterns based on time intervals.
\item Sine and cosine transformation features: these transformations help in modelling periodic patterns, allowing the algorithm to recognise and leverage cyclic trends that repeat over time, such as daily or weekly cycles in energy consumption.
\item Lag features: these features introduce historical data points as part of the inputs, enabling the model to understand dependencies on past values. This is particularly important for energy forecasting, where past observations can influence future values. However, identifying the optimal lags requires domain knowledge \cite{zhang2021review}, which may not always be available. Thus, this work automates the decision on the temporal extent and the number of lags by acknowledging that most energy systems generally exhibit daily patterns (e.g., occupancy levels and human activities) and weekly patterns (e.g., weekdays and weekends) \cite{Bourdeau2019}. Therefore, the proposed method looks back up to one week. Moreover, it employs Pearson correlation \cite{Khalil2022} to select only the top 10 lags that have the highest correlation with the target variable $y$ and this approach is justified by a) the need to prevent unnecessary expansion of the feature space, b) maintaining computational efficiency and c) maintaining method simplicity, potential generalisability and to prevent overfitting, which could diminish the method's performance with new, unseen data.
\item Rolling-window statistical features: these features include statistical measures such as the maximum, minimum, mean, kurtosis, skewness, and standard deviation, each calculated over a rolling window. This approach encapsulates short-term trends and fluctuations by providing a dynamic view of the data's behaviour over specified time frames. It enables the capture of variability and volatility in energy consumption, offering valuable insights for predictive modelling. The window size for each statistical calculation is also automatically set to the number of optimal lags identified in the previous step, ensuring that each window captures relevant temporal patterns within the data, which enhances the model’s predictive accuracy while maintaining method simplicity and computational efficiency.

\end{itemize}

The comprehensive feature vector $\mathbf{F}'_i$ for each instance $i$ is then constructed by concatenating these feature subsets:
\begin{equation}
\mathbf{F}'_i = \bigoplus_{j=1}^{M} \text{EngineeredFeatures}_j(t_i, y_i), \quad \forall i \in \{1, \ldots, N\}
\end{equation}

where $\bigoplus$ denotes the concatenation operation, combining all feature subsets $\mathbf{F}'_{i,j}$ generated by the functions into a single feature vector for each instance.

The predictive model $\mathcal{M}$ is subsequently trained using these comprehensive feature vectors $\mathbf{F}'_i$ along with their corresponding target variable $y_i$, in an AutoML framework:
\begin{equation}
\mathcal{M} = \text{AutoML}(\{(\mathbf{F}'_i, y_i)\}_{i=1}^{N})
\end{equation}

\begin{algorithm}
\renewcommand{\thealgorithm}{1}
\caption{Automated Feature Engineering for Energy Consumption Forecasting Problems.}
\label{alg:A}
\footnotesize
\begin{algorithmic}[1] 
\Require DataFrame $D$ with columns $t$ (time stamps) and target variable $y$ (e.g., electricity consumption)
\Ensure DataFrame $D'$ with new features

\State $D' \gets D$
\State $Features_{time} \gets F_{time}(D, t)$
\State $Features_{cyclical} \gets F_{cyclical}(Features_{time})$
\State $Features_{lags} \gets F_{lags}(D, y)$
\State $Features_{stats} \gets F_{stats}(D, y)$
\State $D' \gets D$ append $Features_{time}$, $Features_{cyclical}$, $Features_{lags}$, $Features_{stats}$

\Function{$F_{time}$}{$D, t$}
    \State $Features_{time} \gets$ Extract time-based features from $t$ {\tiny{\Comment{(e.g., hour of day)}}}
    \State \Return $Features_{time}$
\EndFunction

\Function{$F_{cyclical}$}{$Features_{time}$}
    \State $Features_{cyclical} \gets$ empty list
    \For{feature $f$ in $Features_{time}$}
        \State $f_{sin}, f_{cos} \gets$ Compute $\sin$ and $\cos$ transformations of $f$
        \State Append $f_{sin}, f_{cos}$ to $Features_{cyclical}$
    \EndFor
    \State \Return $Features_{cyclical}$
\EndFunction

\Function{$F_{lags}$}{$D, y$}
    \State $Features_{lags} \gets$ empty list
    \For{$l$ in predefined lags}
        \State $lag\_feature \gets$ shift $y$ by $l$ time units
        \State Compute and append $lag\_feature$ to $Features_{lags}$ if highly correlated with $y${\tiny{\Comment{use Pearson correlation to identify top 10 lags}}}
    \EndFor
    \State \Return $Features_{lags}$
\EndFunction

\Function{$F_{stats}$}{$D, y, Features_{lags}$}
    \State $Features_{stats} \gets$ empty list
    \For{each $lag$ in $Features_{lags}$}
        \State $window \gets lag${\tiny{\Comment{use lag number as a window size}}}
        \State $stat\_features \gets$ Compute rolling window statistics of $y$ over $window$:
        \State \quad Mean: $\mu = \frac{1}{N}\sum_{i=1}^N y_i$
        \State \quad Standard Deviation: $\sigma = \sqrt{\frac{1}{N-1}\sum_{i=1}^N (y_i - \mu)^2}$
        \State \quad Skewness: $\gamma_1 = \frac{1}{N}\sum_{i=1}^N \left(\frac{y_i - \mu}{\sigma}\right)^3$
        \State \quad Kurtosis: $\gamma_2 = \frac{1}{N}\sum_{i=1}^N \left(\frac{y_i - \mu}{\sigma}\right)^4 - 3$
        \State {\tiny{\Comment{(where $N$ is the number of observations in the window, $y_i$ is the i-th value of the target variable.)}}}
        \State Append $stat\_features$ to $Features_{stats}$
    \EndFor
    \State \Return $Features_{stats}$
\EndFunction

\end{algorithmic}
\end{algorithm}


\subsubsection*{Notations and Definitions:}
\begin{itemize}[label={}]
    \item $\mathcal{D}$: Dataset consisting of $N$ timestamp-target pairs.
    \item $t_i$: Timestamp of the $i$-th instance.
    \item $y_i$: Target variable for the $i$-th instance.
    \item $\mathbf{F}'_{i,j}$: Feature subset extracted from timestamp $t_i$ using the $j$-th FE function.
    \item $\mathbf{F}'_i$: Comprehensive feature vector for the $i$-th instance, derived from concatenating all feature subsets $\mathbf{F}'_{i,j}$.
    \item \hyperref[alg:A]{Algorithm 1}: Algorithm for comprehensive FE.
    \item $\mathcal{M}$: Predictive model trained on the enhanced feature set in an AutoML framework.
\end{itemize}

\noindent{The \hyperref[alg:A]{Algorithm 1} is designed with the following considerations:}
\begin{itemize}[label={--}]
    \item Engineered features should be interpretable, facilitating feature importance analysis and offering insights into the temporal patterns influencing the target variable $y_i$.
    \item The \hyperref[alg:A]{Algorithm 1} should maintain computational efficiency to accommodate the processing of large datasets efficiently.
    \item Features at time $t_i$ must be computed solely from data available up to $t_i$, without utilising future data points, to prevent look-ahead bias.

\end{itemize}

\begin{landscape}
    \renewcommand{\arraystretch}{2} 
    \begin{table}
    \centering
    \caption{Datasets used in this study representing different energy patterns in residential buildings, renewable energy, and regional energy consumption. The multivariate datasets mean they have additional features (e.g., weather).}
    \scriptsize	
    \begin{tabular}{l|l|l|l|l|l|l|l|l|l} 
        \toprule
        \textbf{Dataset} & \textbf{Description} & \textbf{Type} & \textbf{Granularity} & \textbf{Samples} & \textbf{Mean} & \textbf{Std} & \textbf{Min} & \textbf{Max} & \textbf{Ref.} \\
        \midrule
        Appliances & Home appliances energy consumption & Multivariate (27 features) & 10mins & 19,735 & 97.69 & 102.5 & 10.0 & 1,080.0 & \cite{RN559} \\
        TCity & Power consumption of the city of Tetouan, Morocco & Multivariate (5 features) & 10mins & 52,416 & 32,344.9 & 7,130.5 & 13,895.7 & 52,204.4 & \cite{RN604} \\
        WindT & SCADA system data for wind turbines & Multivariate (2 features) & 10mins & 50,530 & 1,307.6 & 1,312.4 & 2.4 & 3,618.7 & \cite{RN602} \\
        AEP & American Electric Power consumption & Univariate & Hourly & 121,273 & 15,499.6 & 2,591.3 & 9,581.0 & 25,695.0 & \cite{MULLA18} \\
        PJME & PJM East Region power consumption & Univariate & Hourly & 144,366 & 32,080.5 & 6,463.8 & 14,544.0 & 62,009.0 & \cite{MULLA18} \\
        PJMW & PJM West Region power consumption & Univariate & Hourly & 143,206 & 5,602.4 & 979.1 & 487.0 & 9,594.0 & \cite{MULLA18} \\
        COMED & Commonwealth Edison - Illinois power consumption & Univariate & Hourly & 66,497 & 3,105.1 & 599.7 & 907.0 & 5,445.0 & \cite{MULLA18} \\
        FE & FirstEnergy (FE) - Ohio power consumption & Univariate & Hourly & 62,874 & 7,792.2 & 1,331.2 & 0.0 & 14,032.0 & \cite{MULLA18} \\
        NI & Northern Illinois Hub power consumption & Univariate & Hourly & 58,450 & 11,701.6 & 2,371.5 & 7,003.0 & 23,631.0 & \cite{MULLA18} \\
        DEOK & Duke Energy Ohio/Kentucky power consumption & Univariate & Hourly & 57,739 & 3,105.1 & 599.7 & 907.0 & 5,445.0 & \cite{MULLA18} \\
        EKPC & East Kentucky Power Cooperative & Univariate & Hourly & 45,334 & 1,464.2 & 378.8 & 514.0 & 3,490.0 & \cite{MULLA18} \\
        \bottomrule
    \end{tabular}
    \label{ch4:tab:datasets}
\end{table}
\end{landscape}
\subsection{Experimental setup}
\label{ch4:experimental_setup}

This section describes the experimental setup used in this study, detailing the datasets employed, benchmarking methods,  evaluation metrics and statistical tests. The steps undertaken in the experimental procedure were as follows:
\begin{enumerate}
\item Step One: each dataset was divided into two parts: 80\% for training the AutoML models, and 20\% for testing and evaluating the models' performance on new, unseen data. As this is an exploratory work, this train-test split validation procedure was chosen due to its simplicity and computational efficiency, particularly since the average number of samples across the datasets is nearly 75,000. It is worth mentioning that the data are not shuffled since the energy consumption data exhibit temporal patterns, and it is crucial to maintain the chronological order of the time stamps. Additionally, when using an AutoML framework, there is typically no need for a separate validation set besides the test set to monitor the training process, as the model selection and hyperparameter tuning processes are automated within the AutoML pipeline.
\item Step Two: this step consists of two different experimental settings (i.e., scenarios) for performance evaluation and comparison. In the first scenario, the AutoML models were trained without the proposed FE to establish a baseline. It is worth noting that the univariate datasets used in this scenario, consisting only of the target variable and timestamp, serve as a baseline scenario to evaluate AutoML's forecasting capabilities with minimal input. This deliberately simplistic setup is designed to mimic the performance of AutoML in conditions that resemble a) a worst-case scenario, b) an absence of domain-specific FE knowledge, and c) a preliminary testing phase where the model must learn with limited data. In the second scenario, the AutoML models were trained with the experimental scenario (i.e., with the proposed FE method).
\item Step Three: All AutoML models, in both baseline and experimental settings, were evaluated on the test set to compare the performance of each model under each scenario.
\end{enumerate}

\subsubsection{Datasets}
Given the scarcity of standardised benchmark datasets for energy forecasting using AutoML, this study opted to utilise publicly available real-world datasets employed in related studies and Kaggle repositories, as shown in \autoref{ch4:tab:datasets}. It leveraged eleven datasets that span diverse domains such as energy consumption in residential buildings (i.e., home appliances), renewable energy sources (i.e., wind turbines), and city/regional energy consumption. These datasets feature both univariate and multivariate forecasting problems, across different sizes and granularities. 

\subsubsection{Benchmarking Methods}
To ensure a comprehensive evaluation of the effectiveness of the proposed method across various AutoML methods, a mix of well-established methods, namely H2O, TPOT, AutoGluon, and FLAM,L was selected. These methods were chosen because they represent the leading open-source AutoML methods that have been effectively used in numerous forecasting problems, thereby allowing for a robust benchmark across diverse state-of-the-art approaches. All models were trained using Python 3.11.

\subsubsection{Evaluation Metrics and Statistical Tests}

The Normalised Root Mean Squared Error (nRMSE) is utilised as an evaluation metric \cite{joseph2022modern} to compare the performance of each AutoML method with and without applying the proposed FE method, as shown in \autoref{ch4:eq:nRMSE}, providing a normalised measure of error magnitude, thereby enabling performance comparison across various datasets with differing $y_i$ scales.
\begin{equation}
\label{ch4:eq:nRMSE}
nRMSE = \frac{\sqrt{\frac{1}{n}\sum_{i=1}^{n}(y_i - \hat{y}_i)^2}}{y_{\max} - y_{\min}}
\end{equation}

where $y_i$ represents the actual values, $\hat{y}_i$ represents the predicted values, $n$ is the number of observations, and $y_{\max}$ and $y_{\min}$ are the maximum and minimum of the actual values, respectively.

In this study, the Wilcoxon signed-rank test \cite{garcia2008extension} is employed for pairwise comparisons with a significance level of \(\alpha = 0.05\). Specifically comparing with versus without the implementation of the proposed FE method (i.e., experimental vr baseline), to provide statistical support for the analysis of results. Additionally, the Friedman test \cite{hodges2011rank} is utilised to assess whether there are statistically significant differences among the performances of the AutoML methods across multiple datasets, while the Friedman ranking is employed to determine the ranking of the AutoML methods based on their average performance across all datasets.

\subsection{Results and Discussion}
\label{ch4:Results_and_Discussion}

This section presents an analysis and discussion of the experiment results. This includes the overall performance of the proposed method across different energy consumption domains using various AutoML methods, as quantified with nRMSE values, followed by Wilcoxon signed-rank test results in \autoref{ch4:sec:Overall_performance}. Ablation analysis of the feature subsets and their contributions to model performance is presented in \autoref{ch4:sec:Ablation}.

\begin{table}[t]
\centering
\footnotesize
\caption{Comparison of nRMSEs (lower is better) of AutoML methods across different datasets with vs. without using the proposed FE method. The boldface is the lowest (i.e., best) nRMSE value.}
\label{tab:nRMSE}
\begin{tabular}{l|l l|l l|l l|l l}

\toprule
\textbf{Dataset} & \multicolumn{2}{c|}{\textbf{H2O}} & \multicolumn{2}{c|}{\textbf{TPOT}} & \multicolumn{2}{c|}{\textbf{AutoGluon}} & \multicolumn{2}{c}{\textbf{FLAML}} \\
 & with & without & with & without & with & without & with & without \\
\midrule
Appliances & \textbf{0.0725} & 0.2398 & 0.0777 & 0.1054 & 0.0876 & 0.1109 & 0.0770 & 0.1507 \\
TCity & \textbf{0.0118} & 0.2274 & 0.0149 & 0.2064 & 0.0122 & 0.2331 & 0.0131 & 0.2418 \\
WindT & \textbf{0.0442} & 0.1383 & 0.0595 & 0.1419 & 0.0608 & 0.1445 & 0.0497 & 0.1395 \\
AEP & 0.0107 & 0.2637 & 0.0173 & 0.2120 & \textbf{0.0101} & 0.2421 & 0.0104 & 0.2649 \\
PJME & 0.0072 & 0.2832 & 0.0142 & 0.1769 & \textbf{0.0066} & 0.1797 & 0.0082 & 0.2894 \\
PJMW & 0.0105 & 0.2829 & 0.0159 & 0.1769 & \textbf{0.0104} & 0.1797 & 0.0114 & 0.2894 \\
COMED & 0.0157 & 0.2154 & \textbf{0.0153} & 0.1450 & 0.0154 & 0.1547 & 0.0155 & 0.2101 \\
FE & \textbf{0.0097} & 0.2109 & 0.0159 & 0.2403 & 0.0107 & 0.1699 & 0.0118 & 0.2502 \\
NI & 0.0098 & 0.2594 & 0.0115 & 0.1636 & \textbf{0.0079} & 0.1604 & 0.0088 & 0.2543 \\
DEOK & 0.0156 & 0.2155 & \textbf{0.0153} & 0.1450 & 0.0154 & 0.1547 & 0.0155 & 0.2101 \\
EKPC & 0.0154 & 0.2111 & 0.0172 & 0.1513 & \textbf{0.0138} & 0.1436 & 0.0141 & 0.2099 \\
\bottomrule 
\end{tabular}
\end{table}


\subsubsection{Overall Performance}
\label{ch4:sec:Overall_performance}
The results, as shown in \autoref{tab:nRMSE}, demonstrate that the proposed algorithm successfully achieves its primary objective of automatically improving the performance of AutoML methods for forecasting energy consumption in various domains, while simultaneously reducing the need for domain knowledge required for FE. Among the eleven datasets, which represent various energy consumption environments, AutoGluon emerged as the top-performing AutoML framework in five instances, establishing itself as the most consistently high-performing AutoML framework across these datasets. The analysis further revealed that the greatest improvements were observed in regional energy consumption datasets, where the proposed FE method led to substantial reductions in nRMSE. Furthermore, the domain of renewable energy, exemplified by wind turbine data, along with the sector of residential buildings, represented by the energy consumption data of home appliances, also saw improvements from the proposed method, albeit to a lesser degree, indicating its broad applicability across different energy-related domains. Although the results are self-explanatory, the Wilcoxon test, as shown in \autoref{tab:WilcoxonResults}, reveals statistically significant improvement in nRMSE scores across all examined AutoML methods when the proposed AFE method was applied for each AutoML framework. Additionally, each AutoML framework indicates that the proposed method consistently enhanced predictive performance, without any instances where the baseline scenario (i.e., without the proposed method) outperformed the experimental scenario (i.e., with the proposed method). These findings provide statistical evidence that the proposed AFE technique consistently and significantly improves the predictive accuracy of AutoML models across diverse energy datasets, with no cases of performance degradation observed.

\begin{table}
\centering
\scriptsize
\caption{Results of pairwise comparisons of each AutoML using the Wilcoxon Test.}
\begin{tabular}{c|cccc}
\toprule
\textbf{} & \textbf{H2O} & \textbf{TPOT} & \textbf{AutoGluon} & \textbf{FLAML} \\
\midrule
\textbf{Z value$^a$} & -2.934 & -2.936 & -2.936 & -2.936 \\
\textbf{P-value} & 0.003 & 0.003 & 0.003 & 0.003 \\
\bottomrule
\end{tabular} \\
\vspace{1pt} 
\footnotesize{$a$. Based on negative ranks.}
\label{tab:WilcoxonResults}
\end{table}

\begin{figure}[H]
\centering
\includegraphics[width=1\linewidth]{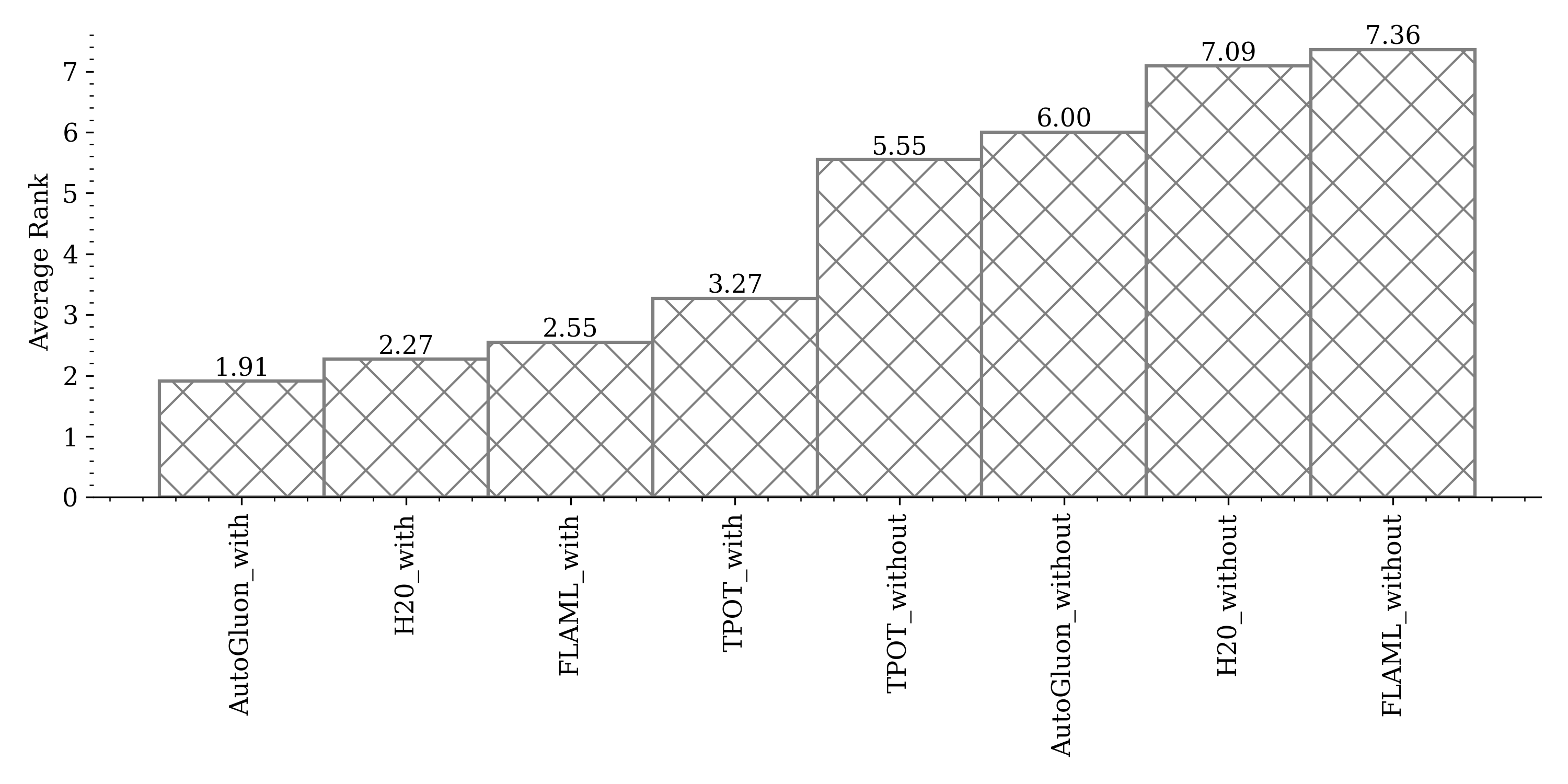}
\caption{Average Friedman Ranks of all AutoML methods. N=11, P $<$ .001, Chi-Square = 64.636.}
\label{ch4:Friedman_Ranking}
\end{figure}

\vspace{-7mm}
\subsubsection{Ablation Analysis of Features Contribution}
\label{ch4:sec:Ablation}
As the proposed FE algorithm generates various input features, the effect of these inputs on AutoML performance was analysed. AutoGluon was selected for this experiment due to its overall good performance, as demonstrated in the previous section and supported by the average Friedman rankings shown in \autoref{ch4:Friedman_Ranking}. The result of this experiment is shown in \autoref{tab:Ablation} where the extracted input features were categorised into three classes: A) All: including all features; B) Time-based and cyclical: which includes features such as the hour of the day, day of the week, and their sine and cosine transformations; C) Lags and rolling statistics: representing past values of the target variable and statistics calculated over a moving window, respectively. The findings show that the use of all features generated by the proposed method results in the best performance with the lowest nRMSE values for predicting energy consumption with AutoML. Using the lags and rolling stats features provides the second-best set of input features and even outperforms the use of all features for the WindT dataset, indicating that such features contain valuable information from which AutoML models can learn. Meanwhile, using only time-based and cyclical features yields the worst performance with the highest nRMSE values compared to other categories of feature sets generated by the proposed method.

\begin{table}[H]
\centering
\scriptsize
\caption{Comparison of nRMSEs (lower is better) using various input features across different datasets.}
\begin{tabular}{l|c|c|c}
\toprule
\textbf{Dataset} & \multicolumn{3}{c}{\textbf{AutoGluon}} \\
                 & All & Time-based \& cyclic & Lags \& rolling-stats \\

\midrule
Appliances & \textbf{0.0876} & 0.1090 & 0.0929 \\
TCity & \textbf{0.0122} & 0.1540 & 0.0124 \\
WindT & 0.0608 & 0.1250 & \textbf{0.0386} \\
AEP & \textbf{0.0101} & 0.1471 & 0.0122 \\
PJME & \textbf{0.0066} & 0.1190 & 0.0081 \\
PJMW & \textbf{0.0104} & 0.1351 & 0.0120 \\
COMED & \textbf{0.0154} & 0.1076 & 0.0162 \\
FE & \textbf{0.0107} & 0.1271 & 0.0113 \\
NI &\textbf{0.0079} & 0.1077 & 0.0092 \\
DEOK & \textbf{0.0154} & 0.1076 & 0.0162 \\
EKPC & \textbf{0.0138} & 0.1273 & 0.0158 \\
\bottomrule
\end{tabular}
\label{tab:Ablation}
\end{table}
\subsection{Summary}
\label{ch4:Conclusion}
This exploratory work presented an AFE method tailored for energy forecasting problems that aimed to improve the performance of AutoML methods and reduce the need for domain knowledge for FE. The proposed method was validated across eleven datasets from various energy domains, including residential buildings, renewable energy, and regional energy consumption. The method was tested with various state-of-the-art AutoML methods. The results demonstrated a noticeable reduction in prediction errors across the examined datasets. The Wilcoxon test indicated a statistically significant improvement in performance across all the examined AutoML frameworks when the proposed AFE method was employed. This study highlights the potential of AFE to improve the performance of AutoML for energy consumption forecasting problems, addressing the limitations of existing AutoML methods (i.e., lack of generating new useful features) and reducing the reliance on domain-specific expertise. Future work will involve more extensive testing of the proposed methods across additional datasets from various energy consumption domains, including industrial and manufacturing settings.
\cleardoublepage

\section{Complementary Results from Chapter~\ref{ch:mainchapter5}}

\label{Appendix_A}
The following metrics are commonly used to evaluate the predictive performance of regression models:

\begin{align}
\text{MAE} &= \frac{1}{n} \sum_{i=1}^{n} \left| y_i - \hat{y}_i \right| \\[6pt]
\text{RMSE} &= \sqrt{\frac{1}{n} \sum_{i=1}^{n} \left( y_i - \hat{y}_i \right)^2} \\[6pt]
\text{MAPE} &= \frac{100\%}{n} \sum_{i=1}^{n} \left| \frac{y_i - \hat{y}_i}{y_i} \right| \\[6pt]
R^2 &= 1 - \frac{\sum_{i=1}^{n} \left( y_i - \hat{y}_i \right)^2}{\sum_{i=1}^{n} \left( y_i - \bar{y} \right)^2}
\end{align}

\noindent
where $y_i$ denotes the actual value, $\hat{y}_i$ denotes the predicted value, $\bar{y}$ is the mean of the actual values, and $n$ is the total number of observations.

\noindent
MAE (Mean Absolute Error) measures the average magnitude of prediction errors without considering their direction. RMSE (Root Mean Squared Error) penalises larger errors more heavily, making it sensitive to outliers. MAPE (Mean Absolute Percentage Error) expresses errors as a percentage of actual values, providing scale-independent interpretability; however is undefined when any $y_i = 0$. Coefficient of Determination ($R^2$) indicates the proportion of variance in the dependent variable explained by the model, with values closer to 1 representing a better fit.


\subsection{Results Across All Test Sets with AutoGluon}
\label{Appendix_A_AutGluon}

\autoref{tab:rmse_mae} and \autoref{tab:mape_r2} present comprehensive evaluation results using four additional metrics (RMSE, MAE, MAPE, and R²) across all 18 datasets, providing multiple perspectives on FE methods' performance with AutoGluon. These results are complementary to those presented in \autoref{tab:nRMSE_Time}.

\begin{landscape}
\begin{table}
\centering
\footnotesize
\caption{RMSE and MAE results for various FE methods using AutoGluon across all test sets. Lower is better. Bold indicates best results. These results are complementary to those presented in \autoref{tab:nRMSE_Time}.}
\renewcommand{\arraystretch}{1.5}
\begin{tabular}{lcccccccccc}
\toprule
& \multicolumn{10}{c}{\textbf{FE Methods}} \\
\cmidrule(lr){2-11}
\textbf{Dataset} & \multicolumn{2}{c|}{\textbf{AutoEnergy}} & \multicolumn{2}{c|}{\textbf{FT}} & \multicolumn{2}{c|}{\textbf{TSMin}} & \multicolumn{2}{c|}{\textbf{TSEff}} & \multicolumn{2}{c}{\textbf{No.Feat.}} \\
& RMSE & MAE & RMSE & MAE & RMSE & MAE & RMSE & MAE & RMSE & MAE \\
\midrule
AEP & \textbf{125.88} & \textbf{90.310} & 148.90 & 112.09 & 2515.3 & 1962.9 & 2633.5 & 2054.9 & 2445.8 & 1959.3 \\
Appliances & 72.205 & 26.727 & \textbf{68.366} & \textbf{23.436} & 151.98 & 78.951 & 126.89 & 53.078 & 100.58 & 69.336 \\
CAISO\_Elec & \textbf{573.73} & \textbf{368.86} & 818.03 & 592.44 & 8785.4 & 7376.6 & 3782.1 & 2943.9 & 7171.6 & 5677.0 \\
COMED & \textbf{64.451} & \textbf{28.250} & 64.819 & 31.482 & 658.33 & 534.51 & 676.95 & 541.09 & 816.74 & 664.80 \\
DEOK & \textbf{65.096} & \textbf{28.533} & 65.468 & 31.797 & 664.91 & 539.85 & 683.72 & 546.50 & 824.91 & 671.45 \\
EKPC & \textbf{35.211} & \textbf{26.018} & 40.448 & 30.771 & 505.42 & 411.83 & 578.16 & 482.25 & 391.87 & 314.65 \\
FDCS\_1 & \textbf{243.44} & \textbf{167.72} & 258.66 & 184.84 & 535.56 & 461.58 & 281.64 & 183.11 & 766.88 & 583.88 \\
FDCS\_2 & \textbf{345.89} & \textbf{246.55} & 373.22 & 283.55 & 573.15 & 476.24 & 526.69 & 422.81 & 758.16 & 668.23 \\
FE & \textbf{75.202} & \textbf{48.521} & 92.156 & 61.690 & 1326.0 & 1062.4 & 1362.6 & 1099.0 & 1167.6 & 906.03 \\
NI & \textbf{101.35} & \textbf{61.563} & 119.97 & 84.023 & 2503.8 & 1972.6 & 2448.3 & 1955.9 & 2998.8 & 2367.1 \\
PJME & \textbf{220.78} & \textbf{148.42} & 253.55 & 188.43 & 7858.0 & 5770.4 & 7392.1 & 5491.2 & 13728 & 11465 \\
PJMW & \textbf{59.313} & \textbf{44.096} & 69.084 & 52.694 & 1132.3 & 892.57 & 1101.2 & 860.49 & 995.97 & 794.60 \\
Solar\_Home & \textbf{0.20367} & \textbf{0.14167} & 0.21669 & 0.15633 & 0.52050 & 0.44938 & 0.42167 & 0.35878 & 0.37987 & 0.30742 \\
Steel & 6.5323 & 3.3761 & 6.5576 & 3.7719 & \textbf{5.5918} & \textbf{3.0670} & 7.0883 & 4.2280 & 109.38 & 76.673 \\
TCity & \textbf{360.15} & \textbf{263.27} & 2704.0 & 2233.1 & 7269.4 & 5907.7 & 5161.6 & 4140.5 & 8099.1 & 6903.9 \\
UNICON & \textbf{0.93407} & \textbf{0.68319} & 0.96273 & 0.70705 & 7.0677 & 5.2662 & 4.8874 & 3.6194 & 6.5745 & 5.4347 \\
Victoria & \textbf{48.557} & \textbf{31.940} & 53.593 & 36.121 & 1415.5 & 1141.7 & 553.51 & 423.68 & 912.61 & 707.86 \\
WindT & \textbf{221.48} & \textbf{112.34} & 335.26 & 218.04 & 593.74 & 447.32 & 390.85 & 261.25 & 1483.6 & 1322.4 \\
\midrule
Mean & \textbf{145.54} & \textbf{94.280} & 304.04 & 231.60 & 2027.5 & 1613.4 & 1539.2 & 1192.4 & 2376.1 & 1952.9 \\
\bottomrule
\end{tabular}
\label{tab:rmse_mae}
\end{table}
\end{landscape}

\begin{landscape}
\begin{table}
\centering
\footnotesize
\caption{MAPE(\%) and $R^2$ Results for various FE methods using AutoGluon across all test sets. Lower MAPE and higher $R^2$ are better. Bold indicates best results. These results are complementary to those presented in \autoref{tab:nRMSE_Time}.}
\renewcommand{\arraystretch}{1.5}
\begin{tabular}{lcccccccccc}
\toprule
& \multicolumn{10}{c}{\textbf{FE Methods}} \\
\cmidrule(lr){2-11}
\textbf{Dataset} & \multicolumn{2}{c|}{\textbf{AutoEnergy}} & \multicolumn{2}{c|}{\textbf{FT}} & \multicolumn{2}{c|}{\textbf{TSMin}} & \multicolumn{2}{c|}{\textbf{TSEff}} & \multicolumn{2}{c}{\textbf{No.Feat.}} \\
& MAPE\% & R2 & MAPE\% & R2 & MAPE\% & R2 & MAPE\% & R2 & MAPE\% & R2 \\
\midrule
AEP & \textbf{0.62000\%} & \textbf{0.99730} & 0.77000\% & 0.99630 & 12.940\% & 0.040100 & 13.440\% & -0.052300 & 13.660\% & -0.0014000 \\
Appliances & 18.030\% & 0.37090 & \textbf{14.350\%} & \textbf{0.43610} & 82.030\% & -1.7868 & 46.070\% & -0.94280 & 90.740\% & -0.22070 \\
CAISO\_Elec & \textbf{1.9900\%} & \textbf{0.99010} & 3.0200\% & 0.97990 & 32.960\% & -1.3214 & 14.770\% & 0.56980 & 31.570\% & -0.54690 \\
COMED & \textbf{1.0500\%} & \textbf{0.98920} & 1.1500\% & 0.98910 & 18.270\% & -0.15890 & 17.920\% & -0.22540 & 21.640\% & -0.72930 \\
DEOK & \textbf{1.0605\%} & \textbf{0.97931} & 1.1615\% & 0.97921 & 18.453\% & -0.15731 & 18.099\% & -0.22315 & 21.856\% & -0.72201 \\
EKPC & \textbf{1.7400\%} & \textbf{0.99180} & 2.0600\% & 0.98920 & 30.440\% & -0.72100 & 36.180\% & -1.2520 & 21.660\% & -0.010600 \\
FDCS\_1 & 21.760\% & \textbf{0.82470} & \textbf{20.810\%} & 0.80210 & 137.90\% & 0.15150 & 25.350\% & 0.76530 & 168.05\% & -0.73980 \\
FDCS\_2 & \textbf{27.620\%} & \textbf{0.60140} & 32.510\% & 0.53590 & 64.640\% & -0.094500 & 52.570\% & 0.075700 & 98.190\% & -0.91520 \\
FE & \textbf{0.63000\%} & \textbf{0.99660} & 0.83000\% & 0.99490 & 14.310\% & -0.052800 & 14.820\% & -0.11170 & 11.640\% & 0.17650 \\
NI & \textbf{0.52000\%} & \textbf{0.99820} & 0.71000\% & 0.99740 & 17.430\% & -0.14990 & 17.560\% & -0.099500 & 20.330\% & -0.61260 \\
PJME & \textbf{0.47000\%} & \textbf{0.99880} & 0.61000\% & 0.99850 & 17.390\% & -0.44900 & 16.380\% & -0.28230 & 41.430\% & -3.4767 \\
PJMW & \textbf{0.80000\%} & \textbf{0.99640} & 0.97000\% & 0.99520 & 15.810\% & -0.18400 & 15.040\% & -0.11970 & 14.520\% & -0.0035000 \\
Solar\_Home & \textbf{25.740\%} & \textbf{0.59270} & 28.950\% & 0.53890 & 108.63\% & -1.6604 & 86.690\% & -0.74600 & 60.000\% & -0.41700 \\
Steel & 6.0600\% & 0.99730 & 6.4200\% & 0.99730 & \textbf{5.3300\%} & \textbf{0.99800} & 8.7300\% & 0.99690 & 220.76\% & 0.25290 \\
TCity & \textbf{0.93000\%} & \textbf{0.99660} & 7.1100\% & 0.80770 & 18.630\% & -0.38970 & 13.660\% & 0.29930 & 24.050\% & -0.72510 \\
UNICON & \textbf{1.7300\%} & \textbf{0.96560} & 1.7900\% & 0.96350 & 12.200\% & -0.96770 & 8.5500\% & 0.059100 & 13.670\% & -0.70260 \\
Victoria & \textbf{0.70000\%} & \textbf{0.99680} & 0.78000\% & 0.99610 & 22.790\% & -1.7180 & 8.7500\% & 0.58440 & 15.040\% & -0.12980 \\
WindT & \textbf{127.32\%} & \textbf{0.97280} & 514.93\% & 0.93760 & 145.91\% & 0.80430 & 144.05\% & 0.91520 & 7235.3\% & -0.22200 \\
\midrule
Mean & \textbf{13.264\%} & \textbf{0.90369} & 35.496\% & 0.88582 & 43.104\% & -0.43439 & 31.025\% & 0.011589 & 451.33\% & -0.54184 \\
\bottomrule
\end{tabular}
\label{tab:mape_r2}
\end{table}
\end{landscape}


\subsection{Results Across All Test Sets with TabPFN}
\label{Appendix_A_TabPFN}

\autoref{tab:rmse_mae_methods} and \autoref{tab:mape_r2_methods} present a comprehensive evaluation using four additional metrics (RMSE, MAE, MAPE, and R²) across all 18 datasets, providing multiple perspectives on FE methods' performance with TabPFN. These results are complementary to those presented in \autoref{tab:TabPFN_AutoEnergy_Improvement}.


\begin{landscape}
\begin{table}
\centering
\footnotesize
\caption{RMSE and MAE results for various FE methods using TabPFN across all test sets. Bold indicates the best per row and metric. These results are complementary to those presented in \autoref{tab:TabPFN_AutoEnergy_Improvement}.}
\renewcommand{\arraystretch}{1.5}
\begin{tabular}{lcccccccccc}
\toprule
& \multicolumn{10}{c}{\textbf{Methods}} \\
\cmidrule(lr){2-11}
\textbf{Dataset} & \multicolumn{2}{c|}{\textbf{TabPFN}} & \multicolumn{2}{c|}{\textbf{TabPFN\_AutoEnergy}} & \multicolumn{2}{c|}{\textbf{TabPFN\_FT}} & \multicolumn{2}{c|}{\textbf{TabPFN\_TSEff}} & \multicolumn{2}{c}{\textbf{TabPFN\_TSMin}} \\
& RMSE & MAE & RMSE & MAE & RMSE & MAE & RMSE & MAE & RMSE & MAE \\
\midrule
Appliances & 117.18 & 102.10 & \textbf{58.891} & 24.478 & 59.070 & \textbf{24.179} & 95.739 & 74.172 & 124.23 & 111.26 \\
FDCS\_1 & 444.96 & 321.52 & \textbf{206.34} & \textbf{127.91} & 223.44 & 153.03 & 282.83 & 200.28 & 523.57 & 414.20 \\
FDCS\_2 & 443.46 & 343.62 & \textbf{332.63} & \textbf{238.43} & 376.95 & 277.53 & 446.51 & 348.85 & 477.77 & 373.33 \\
Steel & 4.2997 & 1.8133 & 4.3478 & 1.9157 & 4.3778 & 2.1459 & 7.2560 & 4.1050 & \textbf{4.0600} & \textbf{1.7699} \\
TCity & 6936.3 & 5547.4 & \textbf{473.57} & \textbf{191.57} & 545.77 & 379.33 & 1807.3 & 1343.5 & 5925.3 & 4732.0 \\
UNICON & 5.0920 & 4.1715 & \textbf{0.89409} & \textbf{0.62719} & 0.96541 & 0.67900 & 4.6337 & 3.7580 & 4.4883 & 3.7482 \\
Victoria & 709.85 & 588.43 & 74.830 & \textbf{28.601} & \textbf{69.195} & 39.349 & 471.69 & 364.28 & 689.46 & 548.88 \\
WindT & 645.94 & 346.10 & \textbf{303.05} & \textbf{150.93} & 351.15 & 153.10 & 574.87 & 405.29 & 457.21 & 334.79 \\
\midrule
Mean & 1163.4 & 906.89 & \textbf{181.82} & \textbf{95.557} & 203.86 & 128.67 & 461.36 & 343.02 & 1025.8 & 814.99 \\
\bottomrule
\end{tabular}
\label{tab:rmse_mae_methods}
\end{table}
\end{landscape}


\begin{landscape}
\begin{table}
\centering
\footnotesize
\caption{MAPE(\%) and $R^2$ results for various FE methods using TabPFN across all test sets. Bold indicates the best per row and metric. These results are complementary to those presented in \autoref{tab:TabPFN_AutoEnergy_Improvement}.}
\renewcommand{\arraystretch}{1.5}
\begin{tabular}{lcccccccccc}
\toprule
& \multicolumn{10}{c}{\textbf{Methods}} \\
\cmidrule(lr){2-11}
\textbf{Dataset} & \multicolumn{2}{c|}{\textbf{TabPFN}} & \multicolumn{2}{c|}{\textbf{TabPFN\_AutoEnergy}} & \multicolumn{2}{c|}{\textbf{TabPFN\_FT}} & \multicolumn{2}{c|}{\textbf{TabPFN\_TSEff}} & \multicolumn{2}{c}{\textbf{TabPFN\_TSMin}} \\
& MAPE\% & R2 & MAPE\% & R2 & MAPE\% & R2 & MAPE\% & R2 & MAPE\% & R2 \\
\midrule
Appliances & 146.00\% & -0.57350 & 19.610\% & \textbf{0.60250} & \textbf{18.830\%} & 0.60010 & 97.080\% & -0.050400 & 161.76\% & -0.76870 \\
FDCS\_1 & 88.250\% & 0.50360 & \textbf{17.600\%} & \textbf{0.89330} & 30.150\% & 0.87480 & 43.360\% & 0.79940 & 179.56\% & 0.31270 \\
FDCS\_2 & 46.260\% & 0.34480 & \textbf{28.400\%} & \textbf{0.63130} & 37.090\% & 0.52660 & 46.650\% & 0.33570 & 52.540\% & 0.23940 \\
Steel & \textbf{1.7000\%} & 0.99880 & 2.1500\% & 0.99880 & 2.6400\% & 0.99880 & 5.4200\% & 0.99670 & 1.7900\% & \textbf{0.99900} \\
TCity & 19.460\% & -0.21430 & \textbf{0.70000\%} & \textbf{0.99430} & 1.4100\% & 0.99250 & 4.2800\% & 0.91760 & 17.220\% & 0.11390 \\
UNICON & 10.700\% & -0.021300 & \textbf{1.5800\%} & \textbf{0.96850} & 1.7100\% & 0.96330 & 9.4400\% & 0.15420 & 9.6000\% & 0.20650 \\
Victoria & 12.420\% & 0.34320 & \textbf{0.60000\%} & 0.99270 & 0.84000\% & \textbf{0.99380} & 7.4000\% & 0.71000 & 11.260\% & 0.38040 \\
WindT & 247.27\% & 0.77000 & \textbf{91.090\%} & \textbf{0.94940} & 156.86\% & 0.93200 & 234.74\% & 0.81780 & 165.61\% & 0.88480 \\
\midrule
Mean & 71.507\% & 0.26891 & \textbf{20.216\%} & \textbf{0.87885} & 31.191\% & 0.86024 & 56.046\% & 0.58513 & 74.917\% & 0.29600 \\
\bottomrule
\end{tabular}
\label{tab:mape_r2_methods}
\end{table}
\end{landscape}

\cleardoublepage

\section{Analysis of the Top Ten Features Generated by AutoEnergy in Chapter~\ref{ch:mainchapter5}}

\label{Appendix_B}

\begin{tcolorbox}[title=Feature Importance in AutoGluon]
\footnotesize
Feature importance scores are calculated using permutation importance, where:
\begin{itemize}
    \item Scores represent the performance drop when a feature's values are randomly shuffled.
    \item Higher scores indicate greater feature importance.
\end{itemize}

\footnotesize
The output includes the following columns:
\begin{itemize}
    \item \textbf{Importance}: Estimated feature importance score, representing the performance drop when the feature's values are shuffled.
    \item \textbf{Std Dev}: Standard deviation of the feature importance score; if NaN, insufficient shuffle sets were used to calculate variance.
    \item \textbf{p-value}: P-value for a statistical t-test of the null hypothesis (importance = 0) versus the alternative (importance > 0); lower values suggest the feature is confidently useful.
    \item \textbf{n}: Number of shuffles performed to estimate the importance score.
    \item \textbf{p99 High}: Upper end of the 99\% confidence interval for the true feature importance score.
    \item \textbf{p99 Low}: Lower end of the 99\% confidence interval for the true feature importance score.
\end{itemize}

For more information, see:
\begin{itemize}
   \item AutoGluon Tabular Feature Importance: \url{https://auto.gluon.ai/dev/api/autogluon.tabular.TabularPredictor.feature_importance.html}
   \item AutoGluon Time Series Feature Importance: \url{https://auto.gluon.ai/dev/api/autogluon.timeseries.TimeSeriesPredictor.feature_importance.html}
\end{itemize}
\end{tcolorbox}

\begin{table}[!t]
\footnotesize
\centering
\caption{Feature Importance Analysis for Appliances Dataset}
\label{tab:feature_importance_Appliances}
\resizebox{\textwidth}{!}{
\begin{tabular}{lccccccc}
\toprule
\textbf{Feature} & \textbf{Importance} & \textbf{Std Dev} & \textbf{p-value} & \textbf{n} & \textbf{p99 High} & \textbf{p99 Low} \\
\midrule
y\_lag\_1 & 0.023 & 0.003 & 0.000001 & 7 & 0.028 & 0.019 \\
y\_window\_2\_min & 0.012 & 0.003 & 0.000021 & 7 & 0.016 & 0.008 \\
y\_lag\_2 & 0.008 & 0.002 & 0.000057 & 7 & 0.011 & 0.004 \\
y\_window\_2\_max & 0.007 & 0.003 & 0.000324 & 7 & 0.012 & 0.003 \\
y\_window\_2\_mean & 0.007 & 0.003 & 0.000396 & 7 & 0.011 & 0.003 \\
y\_window\_3\_min & 0.005 & 0.003 & 0.002797 & 7 & 0.009 & 0.001 \\
lights & 0.004 & 0.002 & 0.000618 & 7 & 0.007 & 0.002 \\
hour\_sin & 0.004 & 0.003 & 0.002231 & 7 & 0.008 & 0.001 \\
y\_window\_4\_min & 0.004 & 0.003 & 0.007820 & 7 & 0.009 & -0.000 \\
y\_window\_2\_std & 0.003 & 0.001 & 0.000155 & 7 & 0.005 & 0.002 \\
\bottomrule
\end{tabular}}
\end{table}

\begin{table}[!t]
\footnotesize
\centering
\caption{Feature Importance Analysis for TCity Dataset}
\label{tab:feature_importance_TCity}
\resizebox{\textwidth}{!}{
\begin{tabular}{lccccccc}
\toprule
\textbf{Feature} & \textbf{Importance} & \textbf{Std Dev} & \textbf{p-value} & \textbf{n} & \textbf{p99 High} & \textbf{p99 Low} \\
\midrule
y\_lag\_1 & 6472.919 & 61.023 & 0.000001 & 10 & 6535.631 & 6410.207 \\
y\_window\_2\_mean & 274.030 & 4.320 & 0.000001 & 10 & 278.470 & 269.590 \\
y\_window\_2\_min & 133.148 & 3.328 & 0.000001 & 10 & 136.568 & 129.728 \\
hour & 84.677 & 2.237 & 0.000001 & 10 & 86.977 & 82.378 \\
y\_window\_2\_max & 76.294 & 2.970 & 0.000001 & 10 & 79.347 & 73.242 \\
hour\_sin & 66.290 & 3.312 & 0.000001 & 10 & 69.693 & 62.886 \\
y\_window\_3\_max & 46.340 & 1.916 & 0.000001 & 10 & 48.308 & 44.371 \\
hour\_cos & 45.151 & 2.785 & 0.000001 & 10 & 48.013 & 42.289 \\
y\_window\_3\_mean & 28.397 & 1.299 & 0.000001 & 10 & 29.733 & 27.062 \\
y\_window\_7\_std & 27.916 & 3.361 & 0.000001 & 10 & 31.369 & 24.462 \\
\bottomrule
\end{tabular}}
\end{table}

\begin{table}[!t]
\footnotesize
\centering
\caption{Feature Importance Analysis for WindT Dataset}
\label{tab:feature_importance_WindT}
\resizebox{\textwidth}{!}{
\begin{tabular}{lccccccc}
\toprule
\textbf{Feature} & \textbf{Importance} & \textbf{Std Dev} & \textbf{p-value} & \textbf{n} & \textbf{p99 High} & \textbf{p99 Low} \\
\midrule
Wind Speed & 1089.044 & 14.551 & 0.000001 & 4 & 1131.540 & 1046.547 \\
y\_lag\_1 & 221.907 & 4.591 & 0.000001 & 4 & 235.314 & 208.500 \\
y\_window\_2\_mean & 39.566 & 0.603 & 0.000001 & 4 & 41.327 & 37.804 \\
y\_window\_2\_min & 28.250 & 0.786 & 0.000003 & 4 & 30.546 & 25.955 \\
y\_window\_2\_max & 11.408 & 0.761 & 0.000041 & 4 & 13.632 & 9.185 \\
month\_sin & 8.698 & 0.838 & 0.000122 & 4 & 11.146 & 6.250 \\
y\_window\_11\_max & 5.441 & 0.347 & 0.000036 & 4 & 6.456 & 4.426 \\
y\_window\_3\_min & 5.165 & 0.537 & 0.000153 & 4 & 6.733 & 3.597 \\
y\_window\_4\_max & 5.084 & 0.273 & 0.000021 & 4 & 5.881 & 4.287 \\
y\_window\_9\_max & 3.779 & 0.240 & 0.000035 & 4 & 4.480 & 3.077 \\
\bottomrule
\end{tabular}}
\end{table}

\begin{table}[!t]
\footnotesize
\centering
\caption{Feature Importance Analysis for AEP Dataset}
\label{tab:feature_importance_AEP}
\resizebox{\textwidth}{!}{
\begin{tabular}{lccccccc}
\toprule
\textbf{Feature} & \textbf{Importance} & \textbf{Std Dev} & \textbf{p-value} & \textbf{n} & \textbf{p99 High} & \textbf{p99 Low} \\
\midrule
y\_lag\_1 & 2959.685 & 61.399 & 0.000001 & 4 & 3138.997 & 2780.374 \\
hour\_cos & 164.392 & 7.553 & 0.000013 & 4 & 186.449 & 142.335 \\
y\_window\_2\_mean & 91.869 & 7.072 & 0.000063 & 4 & 112.522 & 71.216 \\
hour & 86.483 & 2.501 & 0.000003 & 4 & 93.786 & 79.180 \\
y\_window\_2\_min & 84.227 & 6.214 & 0.000055 & 4 & 102.376 & 66.078 \\
hour\_sin & 80.137 & 5.265 & 0.000039 & 4 & 95.513 & 64.761 \\
y\_lag\_2 & 58.332 & 4.071 & 0.000047 & 4 & 70.221 & 46.442 \\
y\_window\_2\_std & 50.792 & 3.088 & 0.000031 & 4 & 59.810 & 41.774 \\
month\_cos & 36.956 & 2.788 & 0.000059 & 4 & 45.098 & 28.813 \\
y\_window\_2\_max & 30.679 & 3.258 & 0.000163 & 4 & 40.194 & 21.165 \\
\bottomrule
\end{tabular}}
\end{table}

\begin{table}[!t]
\footnotesize
\centering
\caption{Feature Importance Analysis for PJME Dataset}
\label{tab:feature_importance_PJME}
\resizebox{\textwidth}{!}{
\begin{tabular}{lccccccc}
\toprule
\textbf{Feature} & \textbf{Importance} & \textbf{Std Dev} & \textbf{p-value} & \textbf{n} & \textbf{p99 High} & \textbf{p99 Low} \\
\midrule
y\_lag\_1 & 9328.038 & 160.553 & 0.000001 & 5 & 9658.618 & 8997.458 \\
y\_lag\_2 & 1072.594 & 35.865 & 0.000001 & 5 & 1146.442 & 998.747 \\
hour\_cos & 627.283 & 34.718 & 0.000001 & 5 & 698.767 & 555.798 \\
y\_window\_2\_std & 260.277 & 19.365 & 0.000004 & 5 & 300.150 & 220.404 \\
hour & 206.776 & 13.390 & 0.000002 & 5 & 234.347 & 179.206 \\
hour\_sin & 187.717 & 14.653 & 0.000004 & 5 & 217.887 & 157.546 \\
y\_window\_3\_skew & 84.794 & 9.722 & 0.000020 & 5 & 104.811 & 64.777 \\
month\_cos & 81.218 & 7.545 & 0.000009 & 5 & 96.754 & 65.682 \\
y\_window\_10\_max & 62.058 & 6.857 & 0.000018 & 5 & 76.176 & 47.941 \\
y\_window\_2\_mean & 58.304 & 9.382 & 0.000078 & 5 & 77.622 & 38.985 \\
\bottomrule
\end{tabular}}
\end{table}

\begin{table}[!t]
\footnotesize
\centering
\caption{Feature Importance Analysis for PJMW Dataset}
\label{tab:feature_importance_PJMW}
\resizebox{\textwidth}{!}{
\begin{tabular}{lccccccc}
\toprule
\textbf{Feature} & \textbf{Importance} & \textbf{Std Dev} & \textbf{p-value} & \textbf{n} & \textbf{p99 High} & \textbf{p99 Low} \\
\midrule
y\_lag\_1 & 1178.745 & 3.675 & 0.000001 & 4 & 1189.479 & 1168.011 \\
hour\_cos & 87.534 & 2.280 & 0.000002 & 4 & 94.192 & 80.876 \\
y\_window\_2\_mean & 27.700 & 0.239 & 0.000001 & 4 & 28.397 & 27.002 \\
hour\_sin & 25.967 & 0.836 & 0.000005 & 4 & 28.408 & 23.527 \\
hour & 25.166 & 0.589 & 0.000002 & 4 & 26.886 & 23.445 \\
y\_window\_2\_max & 19.485 & 0.249 & 0.000001 & 4 & 20.211 & 18.759 \\
y\_window\_2\_min & 19.347 & 0.722 & 0.000007 & 4 & 21.455 & 17.239 \\
y\_window\_2\_std & 18.926 & 0.585 & 0.000004 & 4 & 20.633 & 17.219 \\
y\_lag\_2 & 17.020 & 0.483 & 0.000003 & 4 & 18.432 & 15.608 \\
month\_cos & 13.070 & 1.041 & 0.000069 & 4 & 16.111 & 10.029 \\
\bottomrule
\end{tabular}}
\end{table}

\begin{table}[!t]
\footnotesize
\centering
\caption{Feature Importance Analysis for COMED Dataset}
\label{tab:feature_importance_COMED}
\resizebox{\textwidth}{!}{
\begin{tabular}{lccccccc}
\toprule
\textbf{Feature} & \textbf{Importance} & \textbf{Std Dev} & \textbf{p-value} & \textbf{n} & \textbf{p99 High} & \textbf{p99 Low} \\
\midrule
y\_lag\_1 & 666.802 & 7.442 & 0.000001 & 7 & 677.230 & 656.374 \\
hour\_cos & 39.205 & 2.043 & 0.000001 & 7 & 42.067 & 36.342 \\
y\_window\_2\_mean & 26.871 & 0.558 & 0.000001 & 7 & 27.654 & 26.089 \\
y\_window\_2\_min & 17.563 & 0.665 & 0.000001 & 7 & 18.495 & 16.631 \\
y\_lag\_2 & 10.928 & 0.733 & 0.000001 & 7 & 11.955 & 9.901 \\
hour & 10.744 & 0.609 & 0.000001 & 7 & 11.597 & 9.890 \\
y\_window\_2\_std & 9.699 & 0.741 & 0.000001 & 7 & 10.737 & 8.661 \\
y\_window\_2\_max & 8.487 & 0.348 & 0.000001 & 7 & 8.974 & 7.999 \\
hour\_sin & 6.370 & 0.289 & 0.000001 & 7 & 6.775 & 5.966 \\
month\_cos & 3.861 & 0.269 & 0.000001 & 7 & 4.237 & 3.484 \\
\bottomrule
\end{tabular}}
\end{table}

\begin{table}[!t]
\footnotesize
\centering
\caption{Feature Importance Analysis for FE Dataset}
\label{tab:feature_importance_FE}
\resizebox{\textwidth}{!}{
\begin{tabular}{lccccccc}
\toprule
\textbf{Feature} & \textbf{Importance} & \textbf{Std Dev} & \textbf{p-value} & \textbf{n} & \textbf{p99 High} & \textbf{p99 Low} \\
\midrule
y\_lag\_1 & 1696.252 & 17.912 & 0.000001 & 4 & 1748.564 & 1643.940 \\
hour\_cos & 115.522 & 7.444 & 0.000037 & 4 & 137.263 & 93.781 \\
y\_lag\_2 & 66.634 & 4.003 & 0.000030 & 4 & 78.326 & 54.942 \\
hour & 55.829 & 9.527 & 0.000667 & 4 & 83.652 & 28.005 \\
hour\_sin & 38.892 & 3.179 & 0.000075 & 4 & 48.176 & 29.608 \\
month\_cos & 20.839 & 1.478 & 0.000049 & 4 & 25.157 & 16.522 \\
y\_window\_2\_std & 17.458 & 4.735 & 0.002578 & 4 & 31.286 & 3.630 \\
y\_window\_2\_mean & 10.947 & 3.461 & 0.003995 & 4 & 21.055 & 0.838 \\
y\_window\_2\_min & 9.882 & 3.466 & 0.005346 & 4 & 20.004 & -0.239 \\
y\_lag\_3 & 8.213 & 0.435 & 0.000020 & 4 & 9.483 & 6.943 \\
\bottomrule
\end{tabular}}
\end{table}

\begin{table}[!t]
\footnotesize
\centering
\caption{Feature Importance Analysis for NI Dataset}
\label{tab:feature_importance_NI}
\resizebox{\textwidth}{!}{
\begin{tabular}{lccccccc}
\toprule
\textbf{Feature} & \textbf{Importance} & \textbf{Std Dev} & \textbf{p-value} & \textbf{n} & \textbf{p99 High} & \textbf{p99 Low} \\
\midrule
y\_lag\_1 & 2666.381 & 53.292 & 0.000001 & 4 & 2822.018 & 2510.743 \\
hour & 259.923 & 7.318 & 0.000003 & 4 & 281.295 & 238.550 \\
hour\_cos & 155.461 & 7.185 & 0.000014 & 4 & 176.445 & 134.477 \\
y\_window\_2\_max & 112.643 & 10.584 & 0.000113 & 4 & 143.551 & 81.734 \\
y\_window\_2\_mean & 85.701 & 8.234 & 0.000121 & 4 & 109.749 & 61.652 \\
y\_window\_2\_std & 84.605 & 5.606 & 0.000040 & 4 & 100.978 & 68.232 \\
y\_window\_2\_min & 65.611 & 7.589 & 0.000211 & 4 & 87.776 & 43.447 \\
hour\_sin & 38.306 & 3.626 & 0.000116 & 4 & 48.897 & 27.716 \\
month\_cos & 29.564 & 1.229 & 0.000010 & 4 & 33.153 & 25.975 \\
y\_window\_9\_max & 20.477 & 3.537 & 0.000691 & 4 & 30.806 & 10.149 \\
\bottomrule
\end{tabular}}
\end{table}

\begin{table}[!t]
\footnotesize
\centering
\caption{Feature Importance Analysis for DEOK Dataset}
\label{tab:feature_importance_DEOK}
\resizebox{\textwidth}{!}{
\begin{tabular}{lccccccc}
\toprule
\textbf{Feature} & \textbf{Importance} & \textbf{Std Dev} & \textbf{p-value} & \textbf{n} & \textbf{p99 High} & \textbf{p99 Low} \\
\midrule
y\_lag\_1 & 666.802 & 7.442 & 0.000001 & 7 & 677.230 & 656.374 \\
hour\_cos & 39.205 & 2.043 & 0.000001 & 7 & 42.067 & 36.342 \\
y\_window\_2\_mean & 26.871 & 0.558 & 0.000001 & 7 & 27.654 & 26.089 \\
y\_window\_2\_min & 17.563 & 0.665 & 0.000001 & 7 & 18.495 & 16.631 \\
y\_lag\_2 & 10.928 & 0.733 & 0.000001 & 7 & 11.955 & 9.901 \\
hour & 10.744 & 0.609 & 0.000001 & 7 & 11.597 & 9.890 \\
y\_window\_2\_std & 9.699 & 0.741 & 0.000001 & 7 & 10.737 & 8.661 \\
y\_window\_2\_max & 8.487 & 0.348 & 0.000001 & 7 & 8.974 & 7.999 \\
hour\_sin & 6.370 & 0.289 & 0.000001 & 7 & 6.775 & 5.966 \\
month\_cos & 3.861 & 0.269 & 0.000001 & 7 & 4.237 & 3.484 \\
\bottomrule
\end{tabular}}
\end{table}

\begin{table}[!t]
\footnotesize
\centering
\caption{Feature Importance Analysis for EKPC Dataset}
\label{tab:feature_importance_EKPC}
\resizebox{\textwidth}{!}{
\begin{tabular}{lccccccc}
\toprule
\textbf{Feature} & \textbf{Importance} & \textbf{Std Dev} & \textbf{p-value} & \textbf{n} & \textbf{p99 High} & \textbf{p99 Low} \\
\midrule
y\_lag\_1 & 437.837 & 3.513 & 0.000001 & 10 & 441.447 & 434.227 \\
hour\_cos & 18.492 & 0.615 & 0.000001 & 10 & 19.124 & 17.859 \\
y\_window\_2\_min & 17.231 & 0.517 & 0.000001 & 10 & 17.762 & 16.700 \\
hour\_sin & 14.479 & 0.296 & 0.000001 & 10 & 14.783 & 14.175 \\
y\_window\_2\_mean & 14.187 & 0.393 & 0.000001 & 10 & 14.590 & 13.783 \\
y\_window\_2\_max & 10.753 & 0.341 & 0.000001 & 10 & 11.103 & 10.403 \\
hour & 9.994 & 0.244 & 0.000001 & 10 & 10.246 & 9.743 \\
y\_lag\_3 & 9.345 & 0.530 & 0.000001 & 10 & 9.890 & 8.801 \\
month\_cos & 5.666 & 0.222 & 0.000001 & 10 & 5.894 & 5.438 \\
y\_lag\_10 & 3.101 & 0.125 & 0.000001 & 10 & 3.230 & 2.973 \\
\bottomrule
\end{tabular}}
\end{table}

\begin{table}[!t]
\footnotesize
\centering
\caption{Feature Importance Analysis for Steel Dataset}
\label{tab:feature_importance_Steel}
\resizebox{\textwidth}{!}{
\begin{tabular}{lccccccc}
\toprule
\textbf{Feature} & \textbf{Importance} & \textbf{Std Dev} & \textbf{p-value} & \textbf{n} & \textbf{p99 High} & \textbf{p99 Low} \\
\midrule
Lagging\_Current\_Reactive.Power\_kVarh & 138.335 & 1.550 & 0.000001 & 10 & 139.928 & 136.742 \\
Lagging\_Current\_Power\_Factor & 71.301 & 0.840 & 0.000001 & 10 & 72.164 & 70.438 \\
Leading\_Current\_Power\_Factor & 26.245 & 0.690 & 0.000001 & 10 & 26.954 & 25.536 \\
y\_lag\_1 & 7.810 & 0.187 & 0.000001 & 10 & 8.003 & 7.618 \\
Leading\_Current\_Reactive\_Power\_kVarh & 2.586 & 0.196 & 0.000001 & 10 & 2.787 & 2.385 \\
y\_window\_2\_max & 0.748 & 0.036 & 0.000001 & 10 & 0.784 & 0.711 \\
hour\_cos & 0.491 & 0.073 & 0.000001 & 10 & 0.566 & 0.416 \\
y\_window\_3\_mean & 0.467 & 0.037 & 0.000001 & 10 & 0.504 & 0.429 \\
hour\_sin & 0.226 & 0.022 & 0.000001 & 10 & 0.250 & 0.203 \\
y\_window\_5\_min & 0.192 & 0.015 & 0.000001 & 10 & 0.208 & 0.176 \\
\bottomrule
\end{tabular}}
\end{table}

\begin{table}[!t]
\footnotesize
\centering
\caption{Feature Importance Analysis for UNICON Dataset}
\label{tab:feature_importance_UNICON}
\resizebox{\textwidth}{!}{
\begin{tabular}{lccccccc}
\toprule
\textbf{Feature} & \textbf{Importance} & \textbf{Std Dev} & \textbf{p-value} & \textbf{n} & \textbf{p99 High} & \textbf{p99 Low} \\
\midrule
y\_lag\_1 & 2.608 & 0.046 & 0.000001 & 10 & 2.655 & 2.560 \\
hour\_cos & 0.365 & 0.015 & 0.000001 & 10 & 0.380 & 0.349 \\
y\_window\_2\_mean & 0.231 & 0.009 & 0.000001 & 10 & 0.241 & 0.221 \\
hour\_sin & 0.155 & 0.010 & 0.000001 & 10 & 0.165 & 0.144 \\
y\_window\_2\_min & 0.152 & 0.007 & 0.000001 & 10 & 0.160 & 0.145 \\
hour & 0.115 & 0.012 & 0.000001 & 10 & 0.128 & 0.103 \\
y\_window\_2\_max & 0.095 & 0.006 & 0.000001 & 10 & 0.101 & 0.089 \\
y\_lag\_25 & 0.071 & 0.010 & 0.000001 & 10 & 0.081 & 0.061 \\
y\_lag\_24 & 0.068 & 0.011 & 0.000001 & 10 & 0.079 & 0.057 \\
day\_of\_month & 0.027 & 0.007 & 0.000001 & 10 & 0.034 & 0.020 \\
\bottomrule
\end{tabular}}
\end{table}

\begin{table}[!t]
\footnotesize
\centering
\caption{Feature Importance Analysis for Victoria Dataset}
\label{tab:feature_importance_Victoria}
\resizebox{\textwidth}{!}{
\begin{tabular}{lccccccc}
\toprule
\textbf{Feature} & \textbf{Importance} & \textbf{Std Dev} & \textbf{p-value} & \textbf{n} & \textbf{p99 High} & \textbf{p99 Low} \\
\midrule
y\_lag\_1 & 805.529 & 6.196 & 0.000010 & 3 & 841.032 & 770.026 \\
y\_window\_2\_max & 80.172 & 1.050 & 0.000029 & 3 & 86.187 & 74.156 \\
y\_window\_2\_mean & 66.534 & 1.053 & 0.000042 & 3 & 72.567 & 60.501 \\
y\_window\_2\_min & 55.862 & 0.858 & 0.000039 & 3 & 60.781 & 50.943 \\
hour & 22.948 & 0.249 & 0.000020 & 3 & 24.378 & 21.519 \\
hour\_sin & 15.737 & 0.878 & 0.000518 & 3 & 20.766 & 10.708 \\
hour\_cos & 15.583 & 0.205 & 0.000029 & 3 & 16.757 & 14.410 \\
Temp & 10.602 & 0.571 & 0.000483 & 3 & 13.873 & 7.330 \\
y\_window\_3\_max & 9.137 & 0.727 & 0.001052 & 3 & 13.303 & 4.972 \\
y\_window\_3\_mean & 6.739 & 0.561 & 0.001151 & 3 & 9.954 & 3.524 \\
\bottomrule
\end{tabular}}
\end{table}

\begin{table}[!t]
\footnotesize
\centering
\caption{Feature Importance Analysis for CAISO\_Elec Dataset}
\label{tab:feature_importance_CAISO_Elec}
\resizebox{\textwidth}{!}{
\begin{tabular}{lccccccc}
\toprule
\textbf{Feature} & \textbf{Importance} & \textbf{Std Dev} & \textbf{p-value} & \textbf{n} & \textbf{p99 High} & \textbf{p99 Low} \\
\midrule
y\_lag\_1 & 6180.657 & 53.102 & 0.000001 & 10 & 6235.229 & 6126.085 \\
y\_lag\_2 & 973.309 & 11.782 & 0.000001 & 10 & 985.418 & 961.201 \\
hour\_sin & 673.672 & 9.721 & 0.000001 & 10 & 683.662 & 663.682 \\
y\_lag\_23 & 504.967 & 8.934 & 0.000001 & 10 & 514.148 & 495.786 \\
hour\_cos & 404.239 & 10.045 & 0.000001 & 10 & 414.562 & 393.915 \\
y\_window\_2\_std & 114.964 & 3.910 & 0.000001 & 10 & 118.982 & 110.946 \\
month\_cos & 86.968 & 5.851 & 0.000001 & 10 & 92.981 & 80.955 \\
y\_window\_2\_mean & 78.341 & 3.736 & 0.000001 & 10 & 82.181 & 74.502 \\
y\_window\_2\_max & 75.825 & 3.540 & 0.000001 & 10 & 79.462 & 72.187 \\
y\_window\_2\_min & 60.110 & 3.626 & 0.000001 & 10 & 63.836 & 56.384 \\
\bottomrule
\end{tabular}}
\end{table}

\begin{table}[!t]
\footnotesize
\centering
\caption{Feature Importance Analysis for Solar\_Home Dataset}
\label{tab:feature_importance_Solar_Home}
\resizebox{\textwidth}{!}{
\begin{tabular}{lccccccc}
\toprule
\textbf{Feature} & \textbf{Importance} & \textbf{Std Dev} & \textbf{p-value} & \textbf{n} & \textbf{p99 High} & \textbf{p99 Low} \\
\midrule
y\_lag\_1 & 0.068 & 0.003 & 0.000001 & 10 & 0.071 & 0.065 \\
hour\_sin & 0.008 & 0.001 & 0.000001 & 10 & 0.009 & 0.007 \\
hour & 0.006 & 0.001 & 0.000001 & 10 & 0.007 & 0.006 \\
hour\_cos & 0.004 & 0.000 & 0.000001 & 10 & 0.004 & 0.003 \\
y\_lag\_2 & 0.002 & 0.000 & 0.000001 & 10 & 0.003 & 0.002 \\
y\_window\_11\_mean & 0.002 & 0.000 & 0.000001 & 10 & 0.002 & 0.002 \\
y\_window\_2\_mean & 0.002 & 0.000 & 0.000001 & 10 & 0.002 & 0.001 \\
y\_window\_4\_std & 0.001 & 0.000 & 0.000001 & 10 & 0.001 & 0.001 \\
y\_window\_9\_max & 0.001 & 0.000 & 0.000002 & 10 & 0.001 & 0.001 \\
y\_window\_2\_min & 0.001 & 0.000 & 0.000001 & 10 & 0.001 & 0.001 \\
\bottomrule
\end{tabular}}
\end{table}

\begin{table}[!t]
\footnotesize
\centering
\caption{Feature Importance Analysis for FDCS\_1 Dataset}
\label{tab:feature_importance_FDCS_1}
\resizebox{\textwidth}{!}{
\begin{tabular}{lccccccc}
\toprule
\textbf{Feature} & \textbf{Importance} & \textbf{Std Dev} & \textbf{p-value} & \textbf{n} & \textbf{p99 High} & \textbf{p99 Low} \\
\midrule
y\_lag\_24 & 176.012 & 13.971 & 0.000001 & 10 & 190.369 & 161.655 \\
hour\_cos & 52.557 & 4.317 & 0.000001 & 10 & 56.994 & 48.120 \\
y\_lag\_48 & 16.092 & 2.042 & 0.000001 & 10 & 18.190 & 13.994 \\
hour\_sin & 11.471 & 1.721 & 0.000001 & 10 & 13.239 & 9.702 \\
hour & 9.078 & 1.740 & 0.000001 & 10 & 10.867 & 7.290 \\
y\_lag\_72 & 8.721 & 1.203 & 0.000001 & 10 & 9.958 & 7.484 \\
Outdoor Temperature (C) & 6.528 & 2.036 & 0.000002 & 10 & 8.621 & 4.436 \\
Specific Humidity  (g/kg) & 5.478 & 2.153 & 0.000011 & 10 & 7.691 & 3.265 \\
Outdoor Wet Bulb Temperature (C) & 5.220 & 1.173 & 0.000001 & 10 & 6.425 & 4.015 \\
y\_window\_6\_max & 4.195 & 0.806 & 0.000001 & 10 & 5.023 & 3.366 \\
\bottomrule
\end{tabular}}
\end{table}

\begin{table}[!t]
\footnotesize
\centering
\caption{Feature Importance Analysis for FDCS\_2 Dataset}
\label{tab:feature_importance_FDCS_2}
\resizebox{\textwidth}{!}{
\begin{tabular}{lccccccc}
\toprule
\textbf{Feature} & \textbf{Importance} & \textbf{Std Dev} & \textbf{p-value} & \textbf{n} & \textbf{p99 High} & \textbf{p99 Low} \\
\midrule
y\_lag\_3 & 111.814 & 7.866 & 0.000001 & 10 & 119.898 & 103.730 \\
y\_lag\_18 & 47.902 & 4.638 & 0.000001 & 10 & 52.668 & 43.136 \\
y\_lag\_19 & 26.304 & 4.133 & 0.000001 & 10 & 30.551 & 22.057 \\
hour\_cos & 21.691 & 5.233 & 0.000001 & 10 & 27.068 & 16.313 \\
hour\_sin & 16.210 & 4.318 & 0.000001 & 10 & 20.647 & 11.772 \\
y\_lag\_20 & 14.978 & 1.293 & 0.000001 & 10 & 16.308 & 13.649 \\
y\_window\_3\_max & 14.296 & 3.039 & 0.000001 & 10 & 17.419 & 11.173 \\
y\_window\_3\_mean & 14.191 & 2.771 & 0.000001 & 10 & 17.038 & 11.343 \\
y\_window\_5\_mean & 10.158 & 2.125 & 0.000001 & 10 & 12.342 & 7.974 \\
y\_lag\_21 & 9.835 & 1.472 & 0.000001 & 10 & 11.348 & 8.322 \\
\bottomrule
\end{tabular}}
\end{table}

\cleardoublepage

\section{Results of Ten Independent Runs Across Test Days in Chapter~\ref{ch:mainchapter6}}

\label{6_Appendix_A}

To ensure the robustness and reliability of the experimental results, models were trained and evaluated across ten independent runs with different random seeds and presented in the following tables. This practice mitigates variance introduced by stochastic training procedures (e.g., such as random weight initialisation) and enables statistically meaningful comparisons by reporting mean performance metrics together with their standard deviations.


\begin{table}[!b]
\centering
{\footnotesize
\caption{Daily test regrets for Run 1.}
\setlength{\extrarowheight}{1pt}
\begin{tabularx}{\textwidth}{ @{} >{\centering\arraybackslash}m{2.8cm} >{\centering\arraybackslash}X >{\centering\arraybackslash}X !{\vrule width 0.8pt} >{\centering\arraybackslash}X >{\centering\arraybackslash}X !{\vrule width 0.8pt} >{\centering\arraybackslash}X >{\centering\arraybackslash}X @{} }
\toprule
\multirow{4}{*}{Date} & \multicolumn{6}{c}{Test Set Regrets — Run 1} \\
\cmidrule(lr){2-7}
 & \multicolumn{2}{c!{\vrule width 0.8pt}}{PTO} & \multicolumn{2}{c!{\vrule width 0.8pt}}{SPO$^{+}$} & \multicolumn{2}{c}{DBB} \\
\cmidrule(lr){2-3} \cmidrule(lr){4-5} \cmidrule(lr){6-7}
 & AFE & No AFE & AFE & No AFE & AFE & No AFE \\
\midrule
11 Feb & 0.0460 & 0.1005 & 0.0022 & 0.0292 & 0.0252 & 0.1617 \\
12 Feb & 0.1518 & 0.1661 & 0.0252 & 0.0643 & 0.0901 & 0.2267 \\
13 Feb & 0.2054 & 0.2000 & 0.0050 & 0.0918 & 0.1022 & 0.2584 \\
14 Feb & 0.2161 & 0.2286 & 0.0378 & 0.1566 & 0.0963 & 0.3694 \\
15 Feb & 0.0456 & 0.1574 & 0.0191 & 0.0502 & 0.0703 & 0.1765 \\
16 Feb & 0.0284 & 0.0628 & 0.0093 & 0.0304 & 0.0434 & 0.1480 \\
17 Feb & 0.0761 & 0.1353 & 0.0202 & 0.0472 & 0.0875 & 0.2857 \\
18 Feb & 0.1735 & 0.2851 & 0.0535 & 0.1758 & 0.0855 & 0.4257 \\
19 Feb & 0.1674 & 0.2131 & 0.0126 & 0.1128 & 0.0707 & 0.3851 \\
20 Feb & 0.0999 & 0.1801 & 0.0134 & 0.0511 & 0.0357 & 0.2622 \\
21 Feb & 0.3149 & 0.2348 & 0.0066 & 0.0655 & 0.1477 & 0.6182 \\
22 Feb & 0.5036 & 0.4319 & 0.1550 & 0.1490 & 0.2509 & 0.9141 \\
23 Feb & 0.1597 & 0.4811 & 0.2323 & 0.2930 & 0.3333 & 0.4447 \\
24 Feb & 0.7369 & 0.6280 & 0.2310 & 0.3056 & 0.3390 & 1.2273 \\
\bottomrule
\end{tabularx}

\label{tab:test_regret_run_1}
}
\end{table}


\begin{table}[!t]
\centering
{\footnotesize
\caption{Daily test regrets for Run 2.}
\setlength{\extrarowheight}{1pt}
\begin{tabularx}{\textwidth}{ @{} >{\centering\arraybackslash}m{2.8cm} >{\centering\arraybackslash}X >{\centering\arraybackslash}X !{\vrule width 0.8pt} >{\centering\arraybackslash}X >{\centering\arraybackslash}X !{\vrule width 0.8pt} >{\centering\arraybackslash}X >{\centering\arraybackslash}X @{} }
\toprule
\multirow{4}{*}{Date} & \multicolumn{6}{c}{Test Set Regrets — Run 2} \\
\cmidrule(lr){2-7}
 & \multicolumn{2}{c!{\vrule width 0.8pt}}{PTO} & \multicolumn{2}{c!{\vrule width 0.8pt}}{SPO$^{+}$} & \multicolumn{2}{c}{DBB} \\
\cmidrule(lr){2-3} \cmidrule(lr){4-5} \cmidrule(lr){6-7}
 & AFE & No AFE & AFE & No AFE & AFE & No AFE \\
\midrule
11 Feb & 0.0351 & 0.0647 & 0.0148 & 0.0437 & 0.0455 & 0.1620 \\
12 Feb & 0.0886 & 0.1312 & 0.0544 & 0.0808 & 0.0551 & 0.2263 \\
13 Feb & 0.1096 & 0.1917 & 0.0118 & 0.0727 & 0.0576 & 0.2750 \\
14 Feb & 0.1404 & 0.2171 & 0.1067 & 0.1041 & 0.1247 & 0.3127 \\
15 Feb & 0.0548 & 0.0931 & 0.0168 & 0.0329 & 0.0732 & 0.1849 \\
16 Feb & 0.0364 & 0.0575 & 0.0141 & 0.0628 & 0.0683 & 0.1626 \\
17 Feb & 0.0697 & 0.1040 & 0.0679 & 0.0721 & 0.0521 & 0.2282 \\
18 Feb & 0.1337 & 0.1048 & 0.0445 & 0.1193 & 0.0419 & 0.3194 \\
19 Feb & 0.1391 & 0.1435 & 0.0415 & 0.0487 & 0.0416 & 0.2164 \\
20 Feb & 0.0662 & 0.1265 & 0.0604 & 0.0915 & 0.0852 & 0.2246 \\
21 Feb & 0.0920 & 0.3488 & 0.0554 & 0.0457 & 0.1348 & 0.5514 \\
22 Feb & 0.2629 & 0.5926 & 0.2325 & 0.3339 & 0.4857 & 0.8166 \\
23 Feb & 0.2188 & 0.1189 & 0.1790 & 0.4527 & 0.2858 & 0.5197 \\
24 Feb & 0.4366 & 0.8192 & 0.3598 & 0.2898 & 0.7148 & 1.0354 \\
\bottomrule
\end{tabularx}

\label{tab:test_regret_run_2}
}
\end{table}


\begin{table}[!t]
\centering
{\footnotesize
\caption{Daily test regrets for Run 3.}
\setlength{\extrarowheight}{1pt}
\begin{tabularx}{\textwidth}{ @{} >{\centering\arraybackslash}m{2.8cm} >{\centering\arraybackslash}X >{\centering\arraybackslash}X !{\vrule width 0.8pt} >{\centering\arraybackslash}X >{\centering\arraybackslash}X !{\vrule width 0.8pt} >{\centering\arraybackslash}X >{\centering\arraybackslash}X @{} }
\toprule
\multirow{4}{*}{Date} & \multicolumn{6}{c}{Test Set Regrets — Run 3} \\
\cmidrule(lr){2-7}
 & \multicolumn{2}{c!{\vrule width 0.8pt}}{PTO} & \multicolumn{2}{c!{\vrule width 0.8pt}}{SPO$^{+}$} & \multicolumn{2}{c}{DBB} \\
\cmidrule(lr){2-3} \cmidrule(lr){4-5} \cmidrule(lr){6-7}
 & AFE & No AFE & AFE & No AFE & AFE & No AFE \\
\midrule
11 Feb & 0.0541 & 0.0877 & 0.0044 & 0.0453 & 0.0755 & 0.1527 \\
12 Feb & 0.0876 & 0.1597 & 0.0329 & 0.0371 & 0.1857 & 0.2540 \\
13 Feb & 0.1369 & 0.1321 & 0.0166 & 0.0954 & 0.1669 & 0.2749 \\
14 Feb & 0.1310 & 0.2120 & 0.0346 & 0.1521 & 0.2073 & 0.3706 \\
15 Feb & 0.0501 & 0.0950 & 0.0093 & 0.0386 & 0.1226 & 0.1336 \\
16 Feb & 0.0468 & 0.0751 & 0.0222 & 0.0253 & 0.0963 & 0.1188 \\
17 Feb & 0.0297 & 0.2254 & 0.0181 & 0.0068 & 0.1493 & 0.2511 \\
18 Feb & 0.0757 & 0.2214 & 0.0548 & 0.0572 & 0.1438 & 0.4026 \\
19 Feb & 0.0985 & 0.1056 & 0.0199 & 0.0580 & 0.1310 & 0.2910 \\
20 Feb & 0.0846 & 0.1829 & 0.0092 & 0.0360 & 0.1461 & 0.2275 \\
21 Feb & 0.1985 & 0.3051 & 0.0101 & 0.1144 & 0.3393 & 0.6889 \\
22 Feb & 0.3787 & 0.4844 & 0.1947 & 0.3232 & 0.6863 & 0.9105 \\
23 Feb & 0.2842 & 0.4195 & 0.2073 & 0.2800 & 0.2723 & 0.4835 \\
24 Feb & 0.4929 & 0.6812 & 0.2672 & 0.5244 & 1.0021 & 1.1669 \\
\bottomrule
\end{tabularx}

\label{tab:test_regret_run_3}
}
\end{table}


\begin{table}[!t]
\centering
{\footnotesize
\caption{Daily test regrets for Run 4.}
\setlength{\extrarowheight}{1pt}
\begin{tabularx}{\textwidth}{ @{} >{\centering\arraybackslash}m{2.8cm} >{\centering\arraybackslash}X >{\centering\arraybackslash}X !{\vrule width 0.8pt} >{\centering\arraybackslash}X >{\centering\arraybackslash}X !{\vrule width 0.8pt} >{\centering\arraybackslash}X >{\centering\arraybackslash}X @{} }
\toprule
\multirow{4}{*}{Date} & \multicolumn{6}{c}{Test Set Regrets — Run 4} \\
\cmidrule(lr){2-7}
 & \multicolumn{2}{c!{\vrule width 0.8pt}}{PTO} & \multicolumn{2}{c!{\vrule width 0.8pt}}{SPO$^{+}$} & \multicolumn{2}{c}{DBB} \\
\cmidrule(lr){2-3} \cmidrule(lr){4-5} \cmidrule(lr){6-7}
 & AFE & No AFE & AFE & No AFE & AFE & No AFE \\
\midrule
11 Feb & 0.1763 & 0.0823 & 0.0065 & 0.0494 & 0.1317 & 0.0730 \\
12 Feb & 0.2184 & 0.1498 & 0.0334 & 0.0785 & 0.1565 & 0.2161 \\
13 Feb & 0.2814 & 0.1939 & 0.0286 & 0.0912 & 0.2282 & 0.1393 \\
14 Feb & 0.3800 & 0.2134 & 0.0848 & 0.1494 & 0.2801 & 0.2521 \\
15 Feb & 0.2370 & 0.1235 & 0.0344 & 0.0561 & 0.1751 & 0.0969 \\
16 Feb & 0.1498 & 0.0829 & 0.0133 & 0.0307 & 0.0460 & 0.0852 \\
17 Feb & 0.2398 & 0.1047 & 0.0504 & 0.0805 & 0.1843 & 0.1184 \\
18 Feb & 0.2501 & 0.1697 & 0.0531 & 0.0718 & 0.1094 & 0.2380 \\
19 Feb & 0.2627 & 0.1616 & 0.0244 & 0.0555 & 0.1667 & 0.1599 \\
20 Feb & 0.2411 & 0.1409 & 0.0492 & 0.1082 & 0.1393 & 0.1579 \\
21 Feb & 0.5240 & 0.2744 & 0.0229 & 0.1753 & 0.3515 & 0.3383 \\
22 Feb & 1.0862 & 0.4876 & 0.0366 & 0.3711 & 0.9279 & 0.5470 \\
23 Feb & 0.5422 & 0.3496 & 0.3582 & 0.2341 & 0.1308 & 0.1880 \\
24 Feb & 1.1804 & 0.6144 & 0.0183 & 0.7131 & 1.1578 & 0.9066 \\
\bottomrule
\end{tabularx}

\label{tab:test_regret_run_4}
}
\end{table}


\begin{table}[!t]
\centering
{\footnotesize
\caption{Daily test regrets for Run 5.}
\setlength{\extrarowheight}{1pt}
\begin{tabularx}{\textwidth}{ @{} >{\centering\arraybackslash}m{2.8cm} >{\centering\arraybackslash}X >{\centering\arraybackslash}X !{\vrule width 0.8pt} >{\centering\arraybackslash}X >{\centering\arraybackslash}X !{\vrule width 0.8pt} >{\centering\arraybackslash}X >{\centering\arraybackslash}X @{} }
\toprule
\multirow{4}{*}{Date} & \multicolumn{6}{c}{Test Set Regrets — Run 5} \\
\cmidrule(lr){2-7}
 & \multicolumn{2}{c!{\vrule width 0.8pt}}{PTO} & \multicolumn{2}{c!{\vrule width 0.8pt}}{SPO$^{+}$} & \multicolumn{2}{c}{DBB} \\
\cmidrule(lr){2-3} \cmidrule(lr){4-5} \cmidrule(lr){6-7}
 & AFE & No AFE & AFE & No AFE & AFE & No AFE \\
\midrule
11 Feb & 0.0426 & 0.0965 & 0.0117 & 0.1344 & 0.1952 & 0.0534 \\
12 Feb & 0.0691 & 0.2341 & 0.0186 & 0.0891 & 0.2858 & 0.0611 \\
13 Feb & 0.0901 & 0.2732 & 0.0477 & 0.0827 & 0.3764 & 0.0994 \\
14 Feb & 0.1358 & 0.3106 & 0.0394 & 0.2744 & 0.3711 & 0.1486 \\
15 Feb & 0.0583 & 0.1310 & 0.0221 & 0.0604 & 0.2316 & 0.0657 \\
16 Feb & 0.0615 & 0.0886 & 0.0210 & 0.0407 & 0.1257 & 0.0225 \\
17 Feb & 0.0343 & 0.1737 & 0.0629 & 0.0710 & 0.2508 & 0.0930 \\
18 Feb & 0.0544 & 0.3498 & 0.0438 & 0.0786 & 0.3129 & 0.0659 \\
19 Feb & 0.0616 & 0.3066 & 0.0436 & 0.0357 & 0.2701 & 0.0621 \\
20 Feb & 0.0787 & 0.1858 & 0.0199 & 0.0832 & 0.3175 & 0.0933 \\
21 Feb & 0.1494 & 0.4373 & 0.0873 & 0.0638 & 0.6248 & 0.2712 \\
22 Feb & 0.3453 & 0.7395 & 0.1990 & 0.0827 & 1.0908 & 0.4603 \\
23 Feb & 0.2395 & 0.6477 & 0.3209 & 0.3372 & 0.7048 & 0.3129 \\
24 Feb & 0.6137 & 0.9798 & 0.1601 & 0.2619 & 1.2803 & 0.7476 \\
\bottomrule
\end{tabularx}

\label{tab:test_regret_run_5}
}
\end{table}


\begin{table}[!t]
\centering
{\footnotesize
\caption{Daily test regrets for Run 6.}
\setlength{\extrarowheight}{1pt}
\begin{tabularx}{\textwidth}{ @{} >{\centering\arraybackslash}m{2.8cm} >{\centering\arraybackslash}X >{\centering\arraybackslash}X !{\vrule width 0.8pt} >{\centering\arraybackslash}X >{\centering\arraybackslash}X !{\vrule width 0.8pt} >{\centering\arraybackslash}X >{\centering\arraybackslash}X @{} }
\toprule
\multirow{4}{*}{Date} & \multicolumn{6}{c}{Test Set Regrets — Run 6} \\
\cmidrule(lr){2-7}
 & \multicolumn{2}{c!{\vrule width 0.8pt}}{PTO} & \multicolumn{2}{c!{\vrule width 0.8pt}}{SPO$^{+}$} & \multicolumn{2}{c}{DBB} \\
\cmidrule(lr){2-3} \cmidrule(lr){4-5} \cmidrule(lr){6-7}
 & AFE & No AFE & AFE & No AFE & AFE & No AFE \\
\midrule
11 Feb & 0.0593 & 0.0884 & 0.0041 & 0.0138 & 0.0888 & 0.1592 \\
12 Feb & 0.0949 & 0.2681 & 0.0455 & 0.0927 & 0.1472 & 0.1502 \\
13 Feb & 0.1030 & 0.2631 & 0.0308 & 0.0800 & 0.2264 & 0.2567 \\
14 Feb & 0.1240 & 0.2599 & 0.0680 & 0.1366 & 0.1880 & 0.2247 \\
15 Feb & 0.0802 & 0.1017 & 0.0266 & 0.0579 & 0.0996 & 0.1471 \\
16 Feb & 0.0691 & 0.0620 & 0.0214 & 0.0416 & 0.0753 & 0.1231 \\
17 Feb & 0.1229 & 0.2083 & 0.0034 & 0.0717 & 0.1035 & 0.2083 \\
18 Feb & 0.1298 & 0.2551 & 0.0290 & 0.0565 & 0.2034 & 0.2709 \\
19 Feb & 0.0803 & 0.2826 & 0.0259 & 0.0663 & 0.1998 & 0.2156 \\
20 Feb & 0.1131 & 0.1471 & 0.0170 & 0.0211 & 0.1439 & 0.2179 \\
21 Feb & 0.1565 & 0.4190 & 0.0304 & 0.0434 & 0.2256 & 0.3996 \\
22 Feb & 0.4466 & 0.7152 & 0.1465 & 0.1474 & 0.4251 & 0.6713 \\
23 Feb & 0.4235 & 0.4190 & 0.2202 & 0.3071 & 0.4484 & 0.6310 \\
24 Feb & 0.6327 & 1.0741 & 0.1850 & 0.1499 & 0.5910 & 0.7309 \\
\bottomrule
\end{tabularx}

\label{tab:test_regret_run_6}
}
\end{table}


\begin{table}[!t]
\centering
{\footnotesize
\caption{Daily test regrets for Run 7.}
\setlength{\extrarowheight}{1pt}
\begin{tabularx}{\textwidth}{ @{} >{\centering\arraybackslash}m{2.8cm} >{\centering\arraybackslash}X >{\centering\arraybackslash}X !{\vrule width 0.8pt} >{\centering\arraybackslash}X >{\centering\arraybackslash}X !{\vrule width 0.8pt} >{\centering\arraybackslash}X >{\centering\arraybackslash}X @{} }
\toprule
\multirow{4}{*}{Date} & \multicolumn{6}{c}{Test Set Regrets — Run 7} \\
\cmidrule(lr){2-7}
 & \multicolumn{2}{c!{\vrule width 0.8pt}}{PTO} & \multicolumn{2}{c!{\vrule width 0.8pt}}{SPO$^{+}$} & \multicolumn{2}{c}{DBB} \\
\cmidrule(lr){2-3} \cmidrule(lr){4-5} \cmidrule(lr){6-7}
 & AFE & No AFE & AFE & No AFE & AFE & No AFE \\
\midrule
11 Feb & 0.0716 & 0.0491 & 0.0051 & 0.1292 & 0.0520 & 0.1019 \\
12 Feb & 0.2049 & 0.1306 & 0.0172 & 0.1614 & 0.1301 & 0.1913 \\
13 Feb & 0.2482 & 0.2323 & 0.0026 & 0.1848 & 0.1023 & 0.2722 \\
14 Feb & 0.2851 & 0.2162 & 0.0121 & 0.2314 & 0.2267 & 0.2332 \\
15 Feb & 0.0683 & 0.0737 & 0.0090 & 0.1520 & 0.0913 & 0.1130 \\
16 Feb & 0.0466 & 0.0659 & 0.0156 & 0.1236 & 0.0551 & 0.0957 \\
17 Feb & 0.1677 & 0.0915 & 0.0486 & 0.1377 & 0.0882 & 0.2388 \\
18 Feb & 0.2554 & 0.1866 & 0.0549 & 0.2292 & 0.1734 & 0.3287 \\
19 Feb & 0.2652 & 0.1927 & 0.0219 & 0.2044 & 0.1697 & 0.2919 \\
20 Feb & 0.1718 & 0.1233 & 0.0351 & 0.1404 & 0.0948 & 0.1634 \\
21 Feb & 0.3835 & 0.3276 & 0.0081 & 0.3090 & 0.1878 & 0.3576 \\
22 Feb & 0.5363 & 0.5205 & 0.0013 & 0.4702 & 0.3418 & 0.8135 \\
23 Feb & 0.2500 & 0.2480 & 0.4078 & 0.5966 & 0.3434 & 0.6278 \\
24 Feb & 1.0286 & 0.8276 & 0.0034 & 0.5737 & 0.5705 & 0.8065 \\
\bottomrule
\end{tabularx}

\label{tab:test_regret_run_7}
}
\end{table}


\begin{table}[!t]
\centering
{\footnotesize
\caption{Daily test regrets for Run 8.}
\setlength{\extrarowheight}{1pt}
\begin{tabularx}{\textwidth}{ @{} >{\centering\arraybackslash}m{2.8cm} >{\centering\arraybackslash}X >{\centering\arraybackslash}X !{\vrule width 0.8pt} >{\centering\arraybackslash}X >{\centering\arraybackslash}X !{\vrule width 0.8pt} >{\centering\arraybackslash}X >{\centering\arraybackslash}X @{} }
\toprule
\multirow{4}{*}{Date} & \multicolumn{6}{c}{Test Set Regrets — Run 8} \\
\cmidrule(lr){2-7}
 & \multicolumn{2}{c!{\vrule width 0.8pt}}{PTO} & \multicolumn{2}{c!{\vrule width 0.8pt}}{SPO$^{+}$} & \multicolumn{2}{c}{DBB} \\
\cmidrule(lr){2-3} \cmidrule(lr){4-5} \cmidrule(lr){6-7}
 & AFE & No AFE & AFE & No AFE & AFE & No AFE \\
\midrule
11 Feb & 0.0305 & 0.0459 & 0.0050 & 0.0405 & 0.0898 & 0.0975 \\
12 Feb & 0.0408 & 0.1060 & 0.0168 & 0.0785 & 0.1520 & 0.1342 \\
13 Feb & 0.0617 & 0.1135 & 0.0080 & 0.1805 & 0.1569 & 0.1610 \\
14 Feb & 0.0864 & 0.2115 & 0.0072 & 0.1535 & 0.2392 & 0.1871 \\
15 Feb & 0.0437 & 0.0520 & 0.0025 & 0.0267 & 0.1386 & 0.0967 \\
16 Feb & 0.0337 & 0.0506 & 0.0170 & 0.0400 & 0.0902 & 0.0779 \\
17 Feb & 0.0667 & 0.0903 & 0.0148 & 0.0892 & 0.1977 & 0.1118 \\
18 Feb & 0.0746 & 0.0564 & 0.0678 & 0.1609 & 0.1506 & 0.1537 \\
19 Feb & 0.0599 & 0.0592 & 0.0234 & 0.1947 & 0.1049 & 0.0662 \\
20 Feb & 0.0558 & 0.0991 & 0.0102 & 0.0820 & 0.1902 & 0.1685 \\
21 Feb & 0.1448 & 0.1802 & 0.0293 & 0.0922 & 0.3862 & 0.2653 \\
22 Feb & 0.2212 & 0.4995 & 0.1276 & 0.2505 & 0.6550 & 0.6502 \\
23 Feb & 0.3143 & 0.2166 & 0.2378 & 0.2806 & 0.6567 & 0.4280 \\
24 Feb & 0.3886 & 0.6921 & 0.1764 & 0.4467 & 0.9219 & 0.8343 \\
\bottomrule
\end{tabularx}

\label{tab:test_regret_run_8}
}
\end{table}


\begin{table}[!t]
\centering
{\footnotesize
\caption{Daily test regrets for Run 9.}
\setlength{\extrarowheight}{1pt}
\begin{tabularx}{\textwidth}{ @{} >{\centering\arraybackslash}m{2.8cm} >{\centering\arraybackslash}X >{\centering\arraybackslash}X !{\vrule width 0.8pt} >{\centering\arraybackslash}X >{\centering\arraybackslash}X !{\vrule width 0.8pt} >{\centering\arraybackslash}X >{\centering\arraybackslash}X @{} }
\toprule
\multirow{4}{*}{Date} & \multicolumn{6}{c}{Test Set Regrets — Run 9} \\
\cmidrule(lr){2-7}
 & \multicolumn{2}{c!{\vrule width 0.8pt}}{PTO} & \multicolumn{2}{c!{\vrule width 0.8pt}}{SPO$^{+}$} & \multicolumn{2}{c}{DBB} \\
\cmidrule(lr){2-3} \cmidrule(lr){4-5} \cmidrule(lr){6-7}
 & AFE & No AFE & AFE & No AFE & AFE & No AFE \\
\midrule
11 Feb & 0.0501 & 0.0789 & 0.0205 & 0.0521 & 0.0518 & 0.1518 \\
12 Feb & 0.1817 & 0.1710 & 0.0914 & 0.0695 & 0.1261 & 0.2286 \\
13 Feb & 0.1752 & 0.1990 & 0.0511 & 0.1262 & 0.1794 & 0.2573 \\
14 Feb & 0.2021 & 0.2713 & 0.0723 & 0.1508 & 0.1832 & 0.2796 \\
15 Feb & 0.0889 & 0.0958 & 0.0275 & 0.0302 & 0.0567 & 0.1645 \\
16 Feb & 0.0728 & 0.0667 & 0.0205 & 0.0352 & 0.0552 & 0.0978 \\
17 Feb & 0.0776 & 0.1496 & 0.1100 & 0.1716 & 0.1742 & 0.1665 \\
18 Feb & 0.1697 & 0.1664 & 0.1305 & 0.0513 & 0.2399 & 0.3013 \\
19 Feb & 0.1831 & 0.1577 & 0.0986 & 0.0266 & 0.2006 & 0.2788 \\
20 Feb & 0.1043 & 0.1874 & 0.1136 & 0.1738 & 0.1005 & 0.2011 \\
21 Feb & 0.3591 & 0.3571 & 0.0765 & 0.2709 & 0.4189 & 0.4476 \\
22 Feb & 0.5028 & 0.6358 & 0.1699 & 0.5511 & 0.4658 & 0.7541 \\
23 Feb & 0.2339 & 0.2043 & 0.3113 & 0.3017 & 0.4783 & 0.5219 \\
24 Feb & 0.8042 & 0.9195 & 0.3985 & 0.6728 & 0.7704 & 0.9855 \\
\bottomrule
\end{tabularx}

\label{tab:test_regret_run_9}
}
\end{table}


\begin{table}[!t]
\centering
{\footnotesize
\caption{Daily test regrets for Run 10.}
\setlength{\extrarowheight}{1pt}
\begin{tabularx}{\textwidth}{ @{} >{\centering\arraybackslash}m{2.8cm} >{\centering\arraybackslash}X >{\centering\arraybackslash}X !{\vrule width 0.8pt} >{\centering\arraybackslash}X >{\centering\arraybackslash}X !{\vrule width 0.8pt} >{\centering\arraybackslash}X >{\centering\arraybackslash}X @{} }
\toprule
\multirow{4}{*}{Date} & \multicolumn{6}{c}{Test Set Regrets — Run 10} \\
\cmidrule(lr){2-7}
 & \multicolumn{2}{c!{\vrule width 0.8pt}}{PTO} & \multicolumn{2}{c!{\vrule width 0.8pt}}{SPO$^{+}$} & \multicolumn{2}{c}{DBB} \\
\cmidrule(lr){2-3} \cmidrule(lr){4-5} \cmidrule(lr){6-7}
 & AFE & No AFE & AFE & No AFE & AFE & No AFE \\
\midrule
11 Feb & 0.0547 & 0.0858 & 0.0051 & 0.0734 & 0.0763 & 0.1337 \\
12 Feb & 0.0959 & 0.1523 & 0.0339 & 0.0928 & 0.1110 & 0.1692 \\
13 Feb & 0.1616 & 0.1696 & 0.0026 & 0.2173 & 0.0892 & 0.2493 \\
14 Feb & 0.1555 & 0.2224 & 0.0066 & 0.2267 & 0.1844 & 0.3002 \\
15 Feb & 0.0733 & 0.1070 & 0.0057 & 0.0661 & 0.0741 & 0.1558 \\
16 Feb & 0.0536 & 0.0685 & 0.0147 & 0.0650 & 0.0528 & 0.1110 \\
17 Feb & 0.1008 & 0.1008 & 0.0098 & 0.1325 & 0.0890 & 0.1166 \\
18 Feb & 0.1780 & 0.1440 & 0.0546 & 0.1411 & 0.0642 & 0.2043 \\
19 Feb & 0.1648 & 0.1614 & 0.0219 & 0.0881 & 0.0651 & 0.1893 \\
20 Feb & 0.1284 & 0.1565 & 0.0163 & 0.1223 & 0.1125 & 0.1457 \\
21 Feb & 0.1722 & 0.3764 & 0.0029 & 0.2148 & 0.2243 & 0.4058 \\
22 Feb & 0.3200 & 0.6637 & 0.0725 & 0.5195 & 0.4594 & 0.6788 \\
23 Feb & 0.3280 & 0.2579 & 0.3160 & 0.2836 & 0.2421 & 0.3912 \\
24 Feb & 0.4409 & 0.7854 & 0.0352 & 0.4050 & 0.7503 & 1.1105 \\
\bottomrule
\end{tabularx}

\label{tab:test_regret_run_10}
}
\end{table}

\cleardoublepage

\section{Optimal BESS Scheduling Strategy Results Across Test Days in Chapter~\ref{ch:mainchapter6}}

\label{6_Appendix_B}

\begin{figure}
\centering
\includegraphics[width=1\linewidth]{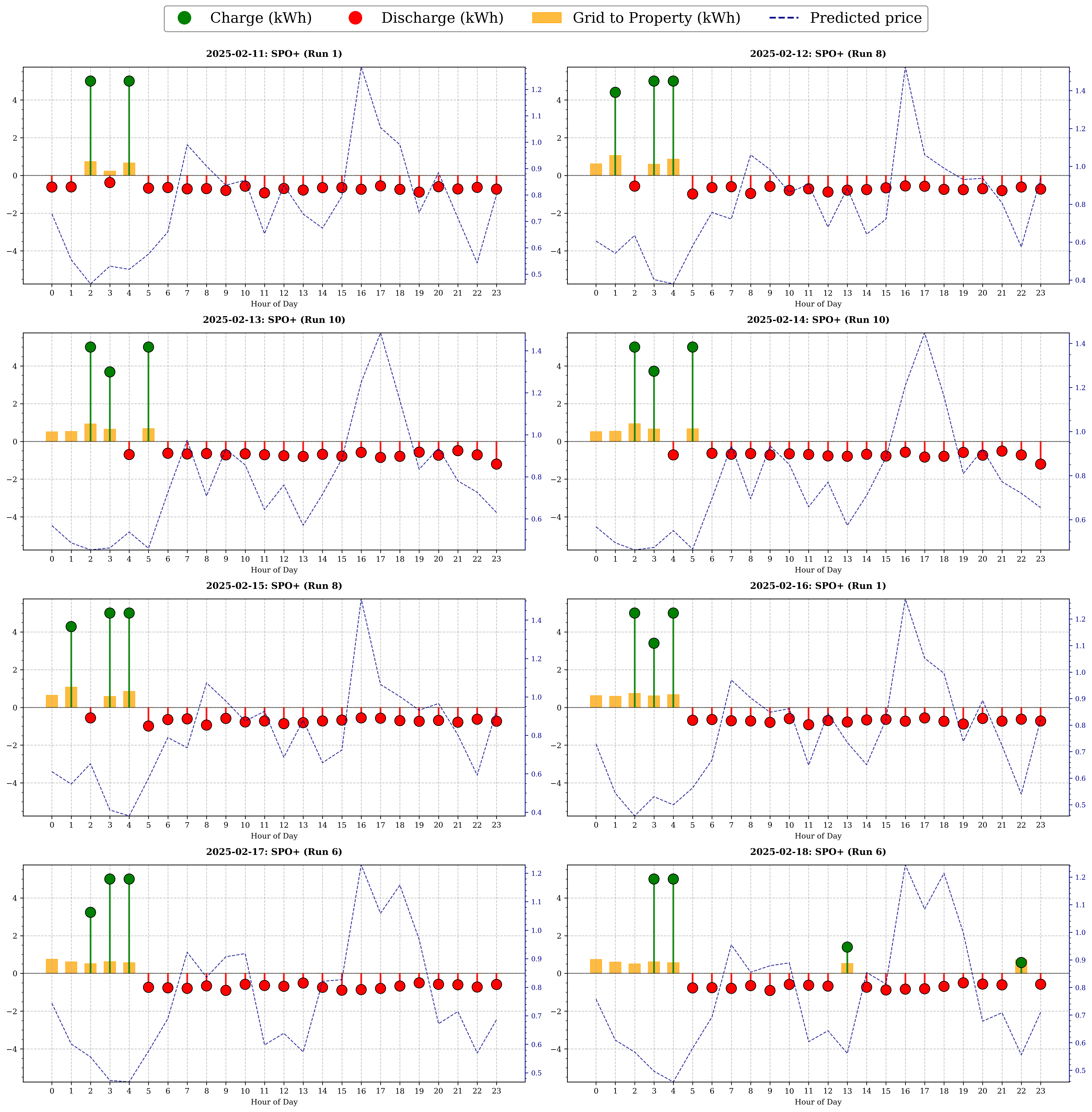}
\caption{Daily BESS optimisation results showing best-performing methods per day. Green circles indicate battery charging, red circles show discharging, and orange background bars represent direct grid supply. The dashed blue line shows predicted electricity prices (right y-axis).}
\label{Scheduling_Strategy_1}
\end{figure}
\begin{figure}[!t]
\centering
\includegraphics[width=1\linewidth]{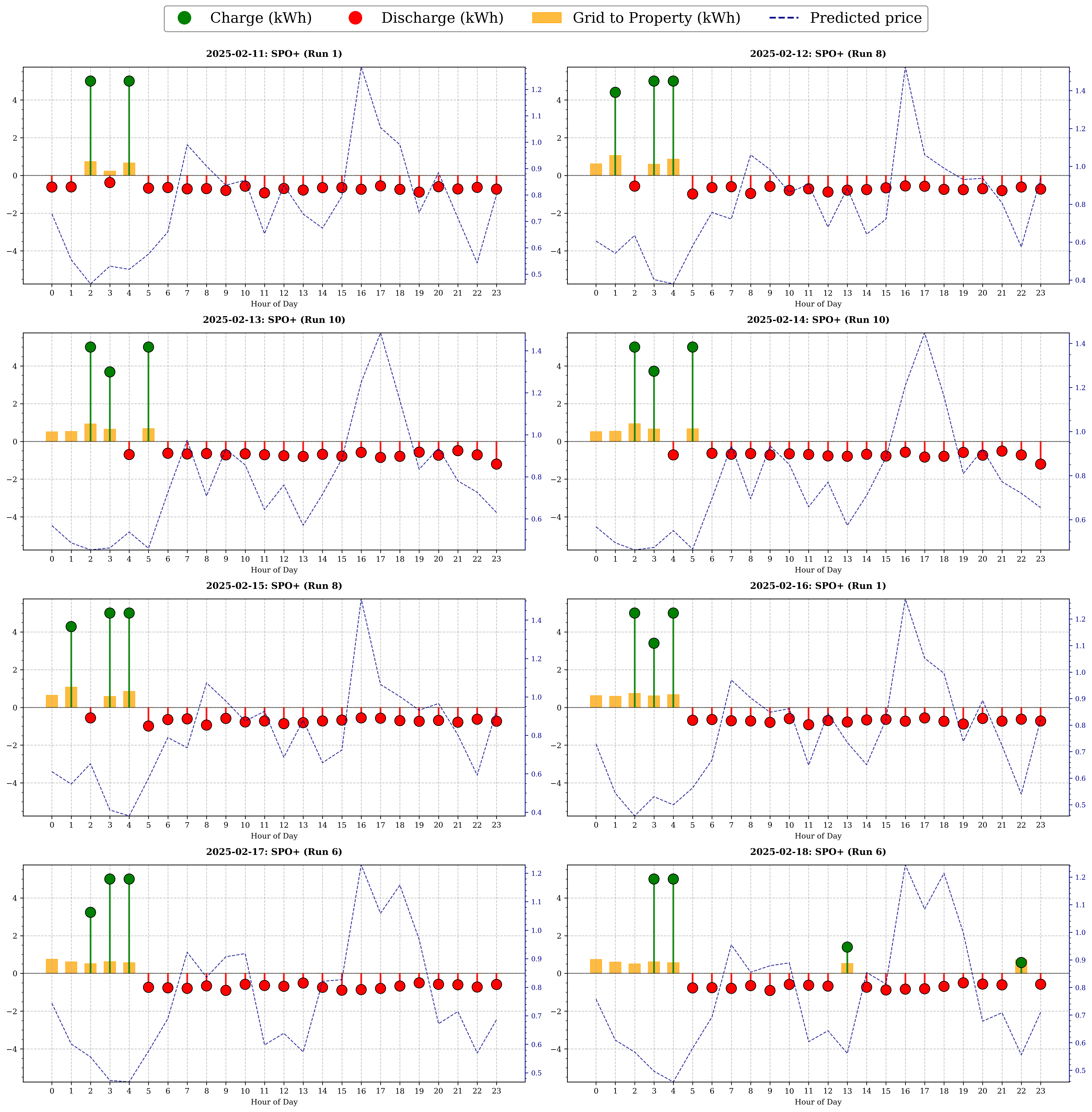}
\caption{Daily BESS optimisation results showing best-performing methods per day. Green circles indicate battery charging, red circles show discharging, and orange background bars represent direct grid supply. The dashed blue line shows predicted electricity prices (right y-axis).}
\label{Scheduling_Strategy_2}
\end{figure}

The following figures present daily BESS optimisation results from multiple experimental runs over all test set days, with each subplot showing the best-performing method for that day based on the lowest regret values. As shown in \autoref{Scheduling_Strategy_1},  and \autoref{Scheduling_Strategy_2}, the visualisation uses a lollipop chart design where battery energy flows are represented by coloured circles connected to stems: green circles above the zero line indicate charging periods (energy stored), while red circles below represent discharging periods (energy released to meet property demand). Orange background bars show direct grid-to-property energy supply where needed. Predicted electricity prices are overlaid as dashed blue lines referenced to the secondary y-axis. Each subplot title identifies the test set date, the best-performing method, and the corresponding experimental run. The horizontal time axis spans 24 hours (00-23), enabling assessment of temporal scheduling patterns in response to price signals.

\cleardoublepage
\end{appendices}


\makeatletter
\renewcommand\@biblabel[1]{[#1]\hspace{0.5em}}
\makeatother

\clearpage

\addcontentsline{toc}{chapter}{Bibliography}
\renewcommand\bibname{Bibliography}

\bibliographystyle{myunsrtnat}

{\small
\bibliography{bibliography/bibliography}
}

\end{document}